\documentclass[a4paper,12pt,authoryear,online,oneside]{Classes/PhDThesisPSnPDF}
\input{imports}

\title{Neural World Models for Computer Vision}

\author{Anthony Hu}

\dept{Department of Engineering}

\university{University of Cambridge}
\crest{\includegraphics[width=0.2\textwidth]{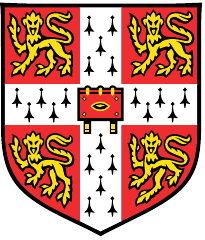}}

\supervisorrole{\textbf{Supervisor: }}

\supervisorlinewidth{0.35\textwidth}

\advisorrole{\textbf{Advisor: }}

\advisorlinewidth{0.35\textwidth}

\degreetitle{Doctor of Philosophy}

\college{Wolfson College}

\degreedate{August 2022} 

\subject{LaTeX} \keywords{{LaTeX} {PhD Thesis} {Engineering} {University of
Cambridge}}

\ifdefineAbstract
\fi

\ifdefineChapter
\fi

\begin{document}

\frontmatter

\maketitle


\begin{dedication} 

I would like to dedicate this thesis to my parents.


\end{dedication}


\begin{declaration}

I hereby declare that except where specific reference is made to the work of 
others, the contents of this thesis are original and have not been 
submitted in whole or in part for consideration for any other degree or 
qualification in this, or any other university. This thesis is my own 
work and contains nothing which is the outcome of work done in collaboration 
with others, except as specified in the text and Acknowledgements. This 
thesis contains fewer than 65,000 words including appendices, 
bibliography, footnotes, tables and equations and has fewer than 150 figures.


\end{declaration}

\begin{acknowledgements}
I would like to acknowledge Roberto Cipolla for supervising me as a PhD student and Toshiba Europe for sponsoring me. I'm thankful for the wonderful experience I've had in Cambridge. 
I'm grateful to Alex Kendall for his guidance on research, his incredible wisdom, and his constant support. Thank you for providing direction so early in my PhD, for helping me grow as an independent researcher, and for all the opportunities at Wayve. Thanks to my colleagues at Wayve who inspired many ideas in my research and for creating such a vibrant atmosphere, especially to Corina, Fergal, Gianluca, Hudson, Jamie, Jeff, Nicolas, Nikhil, Sofía, Vijay, and Zak.

I would like to thank Joseph, Miguel, Oumayma, Quentin, Thomas and Thiziri for their help and advice as fellow research students.
I wanted to show my appreciation to Juba for his unfaltering support throughout my thesis. Thank you for giving me strength during the lows, and celebrating with me during the highs. I'm thankful to my childhood friends Alexandre, Edwin, Estelle and Mélanie for brightening this journey with their cheerful mood.

Thanks to my cousins Jenny, Mélodie, Sheila, Shirlin and Yang Ling for their steady presence, their positivity and their encouragements. I'm grateful to my grandmothers who are the embodiment of selflessness. Thank you for taking such good care of me, for ensuring I did not lack anything, and for always keeping me well fed.

I would like to thank my sister for all the joy she has brought to me during my thesis. Thank you for your support and for being an attentive listening ear these past years. Every time I would come back home to Drancy I would feel invigorated. I'm grateful to my brother who has always been my greatest role model. Thank you for believing in me, more than anyone else. You've taught me how to strive for excellence, how to continuously learn new skills, and how to persevere. Thanks to both of you for being two of the most stable pillars in my life. I'm incredibly lucky to have you.

And finally, I would like to thank my parents for all the opportunities I've had in my life. Thank you for providing a nurturing environment for me to grow and pursue my interests. I've learned by example the value of hard work from you, and this will be a lesson I will carry on forever. I'm grateful for your unconditional support in all my endeavours. Thank you for your love and for raising me, without you, none of this would have been possible.







\end{acknowledgements}
\begin{abstract}
Humans navigate in their environment by learning a mental model of the world through passive observation and active interaction.
Their world model allows them to anticipate what might happen next and act accordingly with respect to an underlying objective. Such world models hold strong promises for planning in complex environments like in autonomous driving. A human driver, or a self-driving system, perceives their surroundings with their eyes or their cameras. They infer an internal representation of the world which should: (i) have spatial memory (e.g. occlusions), (ii) fill partially observable or noisy inputs (e.g. when blinded by sunlight), and (iii) be able to reason about unobservable events probabilistically (e.g. predict different possible futures). They are embodied intelligent agents that can predict, plan, and act in the physical world through their world model. In this thesis we present a general framework to train a world model and a policy, parameterised by deep neural networks, from camera observations and expert demonstrations. We leverage important computer vision concepts such as geometry, semantics, and motion to scale world models to complex urban driving scenes.

In our framework, we derive the probabilistic model of this active inference setting where the goal is to infer the latent dynamics that explain the observations and actions of the active agent. We optimise the lower bound of the log evidence by ensuring the model predicts accurate reconstructions as well as plausible actions and transitions. 

First, we propose a model that predicts important quantities in computer vision: depth, semantic segmentation, and optical flow. We then use 3D geometry as an inductive bias to operate in the bird's-eye view space. We present for the first time a model that can predict probabilistic future trajectories of dynamic agents in bird's-eye view from \ang{360} surround monocular cameras only. Finally, we demonstrate the benefits of learning a world model in closed-loop driving. Our model can jointly predict static scene, dynamic scene, and ego-behaviour in an urban driving environment.

\end{abstract}


\tableofcontents

\listoffigures

\listoftables


\nomenclature[a-0]{$a$}{A scalar}
\nomenclature[a-1]{$\va$}{A vector}
\nomenclature[a-2]{$\mA$}{A matrix}
\nomenclature[a-4]{$\mI_n$}{Identity matrix with $n$ rows and $n$ columns}
\nomenclature[a-5]{$\mI$}{Identity matrix with dimension implied by context}
\nomenclature[a-6]{$\mathrm{diag}(\va)$}{A square, diagonal matrix with diagonal entries given by $\va$}
\nomenclature[a-7]{$\ra$}{A scalar random variable}
\nomenclature[a-8]{$\rva$}{A tensor-valued random variable}

\nomenclature[b-1]{$\R$}{The set of real numbers}
\nomenclature[b-2]{$\{0,1\}$}{The set containing $0$ and $1$}
\nomenclature[b-3]{$\{0,1,\dots,n\}$}{The set of all integers between $0$ and $n$}
\nomenclature[b-4]{$[a, b]$}{The real interval including $a$ and $b$}
\nomenclature[b-5]{$(a, b]$}{The real interval excluding $a$ but including $b$}

\nomenclature[c-4]{$\rva_{1:t}$}{List of random tensors $(\rva_1,\dots,\rva_t)$}
\nomenclature[c-5]{$\rva_{<t}$}{List of random tensors $(\rva_1,\dots,\rva_{t-1})$}

\nomenclature[d-0]{$P(\rx)$}{A probability distribution over a discrete variable}
\nomenclature[d-1]{$p(\rx)$}{A probability distribution over a continuous
variable, or over a variable whose type has not been specified}
\nomenclature[d-2]{$\rx \sim P$}{Random variable $\rx$ has distribution $P$}
\nomenclature[d-3]{$\E_{\rx \sim P}[f(\rx)] \text{ or } \E[f(\rx)]$}{Expectation of $f(\rx)$ with respect to $P(\rx)$}
\nomenclature[d-4]{$\Var(f(\rx))$}{Variance of $f(\rx)$ under $P(\rx)$}
\nomenclature[d-5]{$\Cov(f(\rx), g(\rx))$}{Covariance of $f(\rx)$ and $g(\rx)$ under $P(\rx)$}
\nomenclature[d-6]{$H(\rx)$}{Shannon entropy of the random variable $\rx$}
\nomenclature[d-7]{$\KL(P \Vert Q)$}{Kullback-Leibler divergence of $P$ from $Q$}
\nomenclature[d-8]{$\mathcal{N}(\rvx;\vmu,\mSigma)$}{Gaussian distribution over $\rvx$ with mean $\vmu$ and covariance $\mSigma$}

\nomenclature[e-0]{$f(\vx;\vtheta$)}{A function of $\vx$ parameterised by $\vtheta$. (To lighten the notation, $f(\vx)$ can also be used, the dependence over $\vtheta$ being implicit)}
\nomenclature[e-1]{$\log x$}{Natural logarithm of $x$}
\nomenclature[e-2]{$\lVert \vx \lVert_p $}{$\normlp$ norm of $\vx$}
\nomenclature[e-3]{$\lVert \vx \lVert$}{$\normltwo$ norm of $\vx$}
\nomenclature[e-4]{$\indicator_\mathrm{condition}$}{is $1$ if the condition is true, $0$ otherwise}

\nomenclature[z-bev]{BeV}{Bird's-eye View}
\nomenclature[z-rl]{RL}{Reinforcement Learning}
\nomenclature[z-kl]{KL}{Kullback-Leibler}
\nomenclature[z-cnn]{CNN}{Convolutional Neural Network}
\nomenclature[z-cnn]{RNN}{Recurrent Neural Network}

\printnomenclature

\mainmatter

\setlength{\parskip}{0.5em}

\chapter{Introduction}

\graphicspath{{Chapter1/Figures/}}

Mental models represent the way humans understand the external world through their perceptual senses \citep{craik43,johnson-laird83}. These \newterm{world models} \citep{sutton90,thrun90} are an internal representation of the surrounding world, the interactions of the various parts in it, and the consequences of their actions on it. World models are hypothesised to play a major role in reasoning and decision-making \citep{johnson-laird91,staggers93,jones11}.

Humans are believed to develop their internal model of the world by constantly predicting the future, noticing inconsistencies, and updating their world model to explain their sensory inputs (e.g. visual system) \citep{barlow1989unsupervised,friston10}. They resolve uncertainty by minimising prediction error \citep{rao99}. In particular if we consider visual navigation,  their world model can be seen as a unified representation of the spatial environment enabling navigation and goal reaching in complex environments \citep{madl15,epstein17,park18}. 


\newterm{Computer vision} is the field of artificial intelligence that seeks to understand the visual world with computers, and take actions or make recommendations. Its goal is to automate the capabilities of the human visual system \citep{ballard82}. In the past decade, progress in computer vision has been driven by deep neural networks trained on static datasets with a large amount of annotated data, assumed to be independent and identically distributed. Two prominent examples are AlexNet \citep{krizhevsky12} and ResNet \citep{he16} trained on ImageNet \citep{imagenet09} for image classification. 

However, the independent and identically distributed assumption no longer holds when considering an active visual agent that interacts with their environment. Examples include drones \citep{jung18,padhy18,foehn20} or autonomous driving \citep{pomerleau1988alvinn,bojarski_end_2016,hawke2020urban}.
What the agent observes through their visual senses depends on their actions and their underlying objective. Moreover, the class of active agents we consider here can physically interact with the world. This is the field of \newterm{embodied intelligence} \citep{cangelosi15} and aims at understanding intelligent behaviour in the physical world. In this thesis, we consider the novel problem of learning a world model with deep neural networks from visual observations to control an \newterm{autonomous vehicle}. In particular, we explore how inductive biases from computer vision can scale training large neural world models on complex urban driving scenes.

The remainder of the introduction is organised as follows. We begin by discussing the motivations behind solving autonomous driving. Then, we give a brief history of autonomous driving: how it emerged as a research field, and where it stands today. We continue with how we approach the problem of autonomous driving through learning world models with deep neural networks, and finally, we conclude with the contributions of this thesis.

\section{Motivations}
Solving autonomous driving has the potential to bring considerable benefits to society. It could save millions of lives and transform mobility as we know it. 

Every year, 1.35 million people die from road accidents, tens of millions are injured. Road accidents is the leading cause of death for the population aged between 5 and 29 \citep{who-road-safety-2018}. Distraction, inebriation, and fatigue are the most common causes of road fatalities. They are all linked to the human fallible attention, perceptional senses, motor skills, and reaction time.

\paragraph{Distraction.} 25\% of car crashes are attributed to driver distraction, mainly due to the usage of mobile phones \citep{LIPOVAC2017132}. Using a mobile phone while driving results in physical distraction (one hand could not be available to perform driving maneuvers), visual distraction (looking away from the road), and cognitive distraction (as the driver has to focus on more than one task at a time). Telephone calls increase the likelihood of a car crash by 4, and texting by a factor of 23 \citep{farmer10}. 

\paragraph{Inebriation.} Driving while inebriated led to 22\% of road deaths \citep{alcohol-related-fatalities} by impairing attention, perception, and motor skills \citep{alonso15}. Furthermore, consumption of alcohol creates a feeling of overconfidence, even though it adversely affects the judgement of speed and distances and increases reaction time \citep{hernandez07}. 

\paragraph{Fatigue.} Sleep deprivation, or the task itself of driving being mentally demanding, can lead to fatigue. Sleepiness accounts for 20\% of road accidents \citep{shekari17}, and can be explained by a decrease in attention, reaction time, or in the most extreme case the impossibility to react by falling in a complete state of sleep \citep{vanlaar08}. 

\newterm{Autonomous vehicles} (AV) are not prone to distraction, inebriation or fatigue. They should be able to perform at optimal performance at all time, bypassing these human biological limitations. Autonomous driving could also enhance the driving experience and redefine cities.

In the United Kingdom, 71\% of the workers commute by car with an average commute time of around 1 hour \citep{lyons08}. Such a commute is taxing as the driver needs to pay visual attention to the road and maneuver the car to reach their destination. According to a study by the \citet{commute-wellbeing14}, workers with long commutes are less productive, more tired, and have lower personal well-being. Self-driving cars could allow people to work in their vehicle to make the commute productive, or to simply rest and relax.

Electric vehicles are better suited for autonomous driving because of the additional electricity requirements for sensing and compute. Controlling the vehicle is also more efficient as the signal is electrical and not mechanical \citep{baxter18}. In itself, transitioning to electric vehicles will reduce carbon emission in more than 95\% of world regions, even in the pessimistic case where power production policies are not changed to more renewable options \citep{knobloch20}. Autonomous electric vehicles could have an even greater impact on the environment by greatly reducing city traffic congestion \citep{OVERTOOM2020195}. Parking spaces could also be significantly reduced and replaced with vegetation.

Autonomous driving has the potential to save millions of lives, reduce global carbon emission, decongestion cities, and give back free time to people. The technological challenge ahead in terms of perceptual sensing, prediction of complex dynamics, and planning in dense and uncertain environments makes it one of the most exciting research problem of the 21st century.

\clearpage
\section{A Brief History of Autonomous Driving}

\subsection{The Inception}
In 1979, Sadayuki Tsugawa and his team from the Mechanical Engineering Laboratory in Tokyo, Japan, created the first autonomous system that could lane follow. Their system could drive on various road conditions at 30km/h with two on-board cameras and a road pattern recognition module
\citep{Tsugawa1979AnAW}. The vehicle was actuated both longitudinally (throttle and brake) and laterally (steering wheel) through a table look-up using the features from the pattern recognition module. 

A decade later Ernst Dickmanns developed an autonomous van that could operate in German highways \citep{DICKMANNS1987221}. Their system tracked road lane boundaries with multiple image frames using contour correlation, curvature models and perspective geometry. These features were the input to Kalman filters that estimated vehicle states and road curvature parameters. The model actuated longitudinal and lateral control and could drive at high speed on well structured highway roads.

In 1989 at Carnegie Mellon University, Dean Pomerleau developed ALVINN \citep{pomerleau1988alvinn}: the first end-to-end neural network that learned to lane follow from data. The inputs of the network were a $30\times32$ grayscale image and a $8\times32$ observation from a laser range finder. The neural network itself was small: one fully-connected hidden layer with 29 neurons, followed by one fully-connected output layer with 46 neurons. But even such a simple architecture proved to be able to control lateral direction for lane following by training the network from simulated road images. This pioneering work established two initial results: (i) it was possible to lane follow with a neural network trained end-to-end without any feature engineering, and (ii) domain transfer was possible from a network entirely trained in simulation and deployed in the real world.

The \newterm{Defense Advanced Research Projects Agency} (DARPA) announced a Grand Challenge for autonomous driving in February 2003 that initiated the self-driving cars modern era. The challenge was created to keep away soldiers from harm in dangerous war zones by deploying unmanned military vehicles. A prize of 1 million dollars would be awarded to the team that successfully drove 142 miles in a desert in California. In March 2004, fifteen teams competed in the challenge, but none could complete it. The Red Team from Carnegie Mellon went the furthest with 7.4 miles driven before veering off course in a mountainous region \citep{darpa04}. One year later, in 2005, the second edition of the Grand Challenge involved passing through three narrow tunnels, more than 100 sharp left and right turns in a total of 132 miles in the Mojave desert, California. This time, five different teams crossed the finish lane, with Stanley, from the Stanford Racing Team, finishing first. 

While the two previous challenges took place in the desert, without any interaction with other vehicles, the 2007 DARPA Urban Challenge tested the ability of an autonomous system to navigate in a city environment. Previous efforts in autonomous driving was focused on reaching a destination in a structured setting without any dynamic interaction, but this time, the challenge was testing the ability of a self-driving car to make intelligent decision based on the actions of other vehicles. Each vehicle was tasked to drive a total of 60 miles avoiding static obstacles, passing cars, making U-turns, performing parking maneuvers, and merging into traffic. One particularly challenging task was to negotiate a four-way stop sign intersection requiring to keep precedence of other human-controlled and autonomous vehicles. The winning team was Tartan Racing, a collaborative effort between Carnegie Mellon and General Motors \citep{darpa-urban-07}. 

The DARPA challenges launched the autonomous driving research era as we know it. In 2009, Google created the Google Self-Driving Car Project, which arguably is a legacy of the DARPA challenges. The project was led by Sebastian Thrun and Anthony Levandowski, both in Team Stanford that finished second at the Urban Challenge. Michael Montermelo and Dmitri Dolgov, responsible of planning and optimisation in Team Stanford were recruited, along with Chris Urmson, Bryan Salesky and Dave Ferguson respectively director of technology, software lead, and planning lead in the winning team Tartan Racing. The Google Self-Driving Car Project then became Waymo in 2016 \citep{wakabayashi16}, one of the leading autonomous driving companies. 

Most key members of the Google Self-Driving Car Project decided to start their own venture. Dave Ferguson co-founded in 2016 Nuro, which specialises in the delivery of goods \citep{bhuiyan16}. Argo AI was also created in 2016 by Bryan Saleski, former software lead in the Tartan Racing team. Aurora, founded by Chris Urmson in 2017, is building a combination of software and hardware named the Aurora Driver that can be mounted into existing cars to make them autonomous \citep{schiffer20}.

Outside from the Google Self-Driving Car Project, but among the DARPA Challenge veterans, Kyle Vogt from the MIT team co-founded Cruise that develops fully driverless vehicles for ride hailing \citep{marshall17}. Jesse Levinson (former localisation and mapping lead in Team Stanford) co-founded Zoox in 2014 to also build robot taxis \citep{vance18}. The DARPA Challenges additionally drew attention to how unreliable sensors were at the time, and some key modern sensors were developed thanks to it. An example is the Velodyne Lidar (Light Detection and Ranging) \citep{velodyne17} that enabled real-time surround lidar sensing. External to the DARPA Challenges, a prominent actor in autonomous driving is Tesla, an electric car manufacturer \citep{tesla22}. They are building a full self-driving system called Tesla Autopilot, which is compatible on all the new Tesla vehicles.


\subsection{Levels of Autonomy}
In order to classify the different existing autonomous driving systems, the Society of Automotive Engineers (SAE) have defined five levels of autonomy \citep{sae21}. Levels 1 to 3 requires the human driver to be able to take over at any moment, while levels 4 and above are fully autonomous but differ in scopes.

\begin{itemize}
    \item Level 1 autonomy (driver assistance) provides support in steering \emph{or} brake/acceleration, e.g. lane centering \emph{or} cruise control.
    \item Level 2 autonomy (partial automation) is an assistance in steering \emph{and} brake/acceleration, e.g. lane centering \emph{and} cruise control at the same time.
    \item In level 3 autonomy (conditional automation), it is possible to completely hand over control to the self-driving system in certain situations. For example in highway driving, or in traffic jams.
    \item In level 4 autonomy (high automation), the system can drive autonomously without the human driver ever taking control, but only in certain geographical areas. Examples of use are local driverless taxis. The pedals and steering wheel may or may not be present in the vehicle.
    \item Level 5 autonomy (full automation) describes a system that can drive autonomously anywhere, at anytime.
\end{itemize}

Solving autonomous driving would mean reaching level 5 autonomy, but at the time of writing we are still far from it. The most advanced autonomous systems are level 4 with Cruise operating in San Francisco, California \citep{cruise22} and Waymo in Phoenix, Arizona \citep{waymo-one22}. We are now going to describe the autonomous driving stack used by most of the companies mentioned so far.

\subsection{Traditional Autonomous Driving Stack}
\label{intro:subsection:traditional-av}

\begin{figure}
  \centering
  \includegraphics[width=\linewidth]{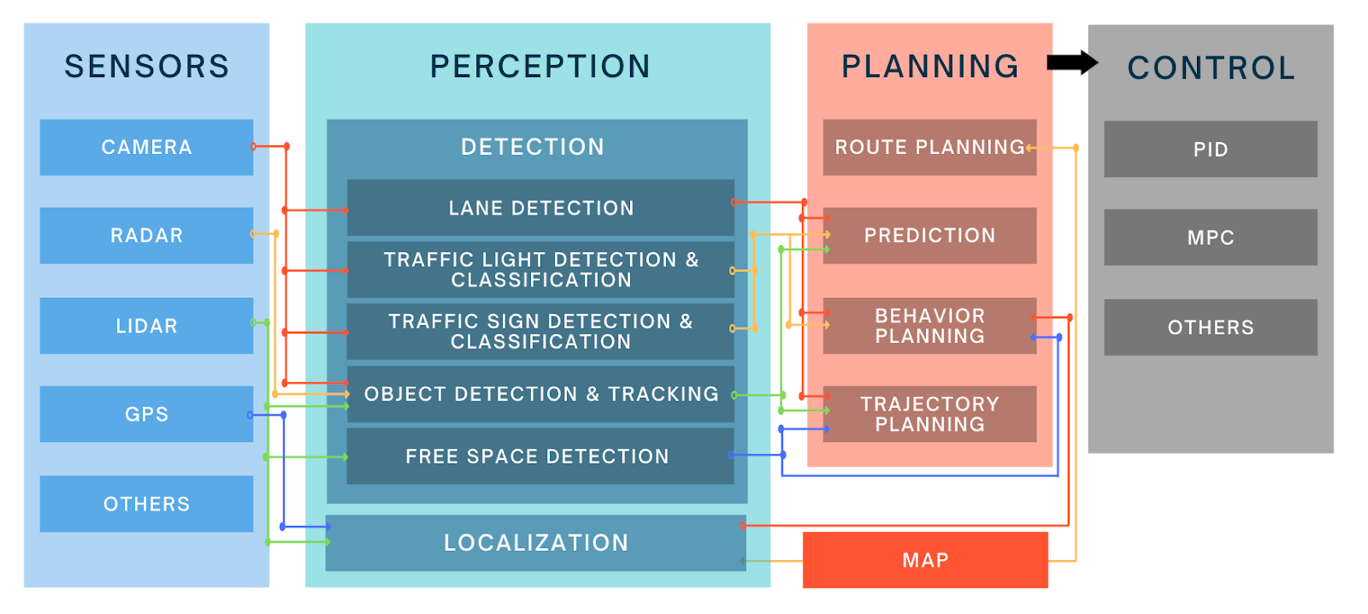}%
  \captionsetup{singlelinecheck=off}
  \caption[Traditional autonomous driving stack.]{Traditional autonomous driving stack. Figure from \citet{wayve21}.
  \begin{enumerate}[itemsep=0.5mm, parsep=0pt]
    \item The inputs are sensor data from cameras, radars, lidars as well as GPS (global positioning system) or IMU (inertial measurement unit).
    \item The perception module detects the static scene (road, sidewalk, static obstacles, traffic lights) as well as dynamic objects (pedestrians, vehicles) from sensor data. It is also responsible of localisation to register the ego-vehicle in the high-definition map (HD map).
    \item The planning module predicts the future trajectory of dynamic objects using the outputs from the perception module and the HD map. It is also responsible for generating a safe trajectory for the ego-vehicle.
    \item Finally, a low-level control module outputs vehicle control to actuate the car in order to follow the path as described by the planning module.
  \end{enumerate}
  }
  \label{intro-fig:av-stack}
\end{figure}

Most autonomous driving companies \citep{ponyai21,waymo22,mobileye21,motional21,apollo20} share the same core autonomous driving stack as illustrated in \Cref{intro-fig:av-stack}. It contains a set of sensors and an high-definition map (HD map), a perception module, a planning module and a control module. We further discuss the sensors, the HD map, and the limitations of this approach.

\paragraph{Sensor data.} Cameras provide the highest resolution information and is comparable to the visual senses humans rely on to drive. Radars can detect distance and velocity of objects through the Doppler effect. Lidars return a 3D point cloud of the environment by sending light beams and estimating distances by calculating the time it gets for the beam to be reflected back. Radars and lidars are useful to detect static or dynamic obstacles. GPS gives a coarse positioning of the ego-vehicle, and IMU returns orientation and angular velocity as well as proper acceleration (acceleration of a body in its own instantaneous rest frame).

\paragraph{HD maps.} High-definition maps are accurate 3D mapping of a geographical area \citep{liu_wang_zhang_2020}. They include a precise description of the roads, road connectivity, static obstacles, traffic signs, traffic lights, all at the centimeter level accuracy.

\paragraph{Limitations.} HD maps need to be updated over time and restrict the geographical area where an autonomous vehicle system can operate. This autonomous driving system is extremely reliant on precise 3D localisation in order to know exactly where the ego-vehicle is. It is modular and errors that propagate downstream cannot be corrected. There are also many hand-designed interfaces between the perception and planning modules, and in the planning module as decision-making is based on rules.

\clearpage
\section{End-to-End Autonomous Driving}
Another radically different approach for autonomous driving is to optimise a single end-to-end neural network. In this section, we begin by reviewing the recent breakthroughs in end-to-end deep learning, we then compare the end-to-end approach with the traditional modular AV stack. Finally, within the framework of end-to-end autonomous driving, we discuss more specifically how learning a world model could help advance autonomous driving.

\subsection{End-to-End Deep Learning}
In the past decade, end-to-end deep learning has made tremendous breakthroughs in multiple different scientific fields. AlphaGo \citep{SilverHuangEtAl16nature} beat the world champion in the game of Go with a deep reinforcement learning network. The research community thought this feat was still decades away due to the sheer combinatorial complexity of Go: there are \SI{2e170} legal positions which is more than the number of atoms in the observable universe estimated at $10^{80}$.

Google DeepMind and OpenAI developed superhuman models on two of the most complex strategy games: respectively AlphaStar on Starcraft II \citep{vinyals-alphastar19} and OpenAI Five in Dota 2 \citep{openaifive19}. These two games require long-term planning because actions taken now will often only reap rewards much later in the game. As a comparison, Chess and Go end after 40 and 150 moves on average, while Starcraft II and Dota 2 necessitate more than tens of thousands of actions. Another challenge is partial observability, and the fact that actions are executed in real-time in a large continuous space. Lastly, Starcraft II and Dota 2 involve game theory as there is no single best strategy. Any strategy can be countered by another and thus requires a constant rethinking of strategic knowledge.

GPT-3 \citep{brown20} is a large language model trained on a massive amount of unsupervised text data to predict the next word. It can generate prose so realistic that it is extremely hard to distinguish from human writing. Moreover, it can perform new tasks in a zero-shot manner without ever having been explicitly trained to do so. A prompt is given to the model as a text input, and GPT-3 simply predicts the next words that follow this prompt to solve the task. Examples include language translation, question answering, and code writing.

End-to-end deep learning has also radically changed the field of protein folding with AlphaFold \citep{alphafold-jumper21} that solved a 50-year grand challenge in biology. AlphaFold can predict the 3D structure of a protein from its sequence of amino acids at the atomic level of accuracy. The 3D structure of a protein determines its function, and being able to predict it could have huge impact on understanding diseases and discovering new drugs.

\subsection{Modular vs. End-to-End} 
The vast majority of autonomous driving companies adopted a modular approach as described in \Cref{intro:subsection:traditional-av}. The end task of controlling the vehicle is subdivided into a discrete pipeline of tasks including perception, prediction, planning, and control. This results in high latency and cascading errors in the downstream modules if uncertainty is not propagated.  Crucially, in a modular approach, decision-making in the planning module is hand-coded and rule-based. The underlying assumption is that it is possible to list all the possible driving scenarios and engineer rules to handle of all them. Another radically different approach is to learn the driving policy from human driving, which is the unique approach taken by Wayve.

Drawing inspiration from the end-to-end deep neural networks breakthroughs, Wayve trained the first deep reinforcement learning model to lane follow on quiet countryside roads \citep{kendall2019learning}. The model was initialised with random weights and optimised a reward equal to the distance travelled by the agent without an intervention from the safety driver. With only a dozen of optimisation steps, the model was able to lane follow. A year later, Wayve demonstrated it was possible to scale the end-to-end approach to complex urban driving scenes with conditional imitation learning 
\citep{hawke2020urban}. They trained a policy to imitate expert human driving with behaviour cloning. The policy was conditioned on a route map that indicates the destination of the autonomous vehicle.

\begin{figure}
  \centering
  \includegraphics[width=\linewidth]{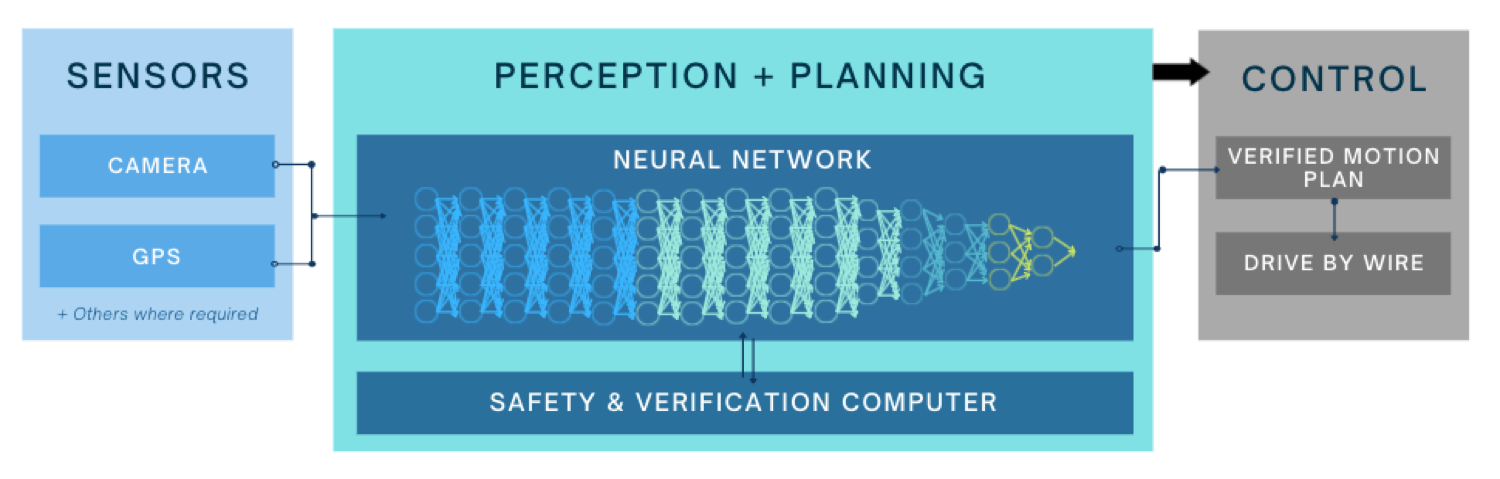}%
  \captionsetup{singlelinecheck=off}
  \caption[End-to-end neural network for autonomous driving.]{End-to-end neural network for autonomous driving. Figure from \citet{wayve21}.
  \begin{enumerate}[itemsep=0.5mm, parsep=0pt]
    \item Sensors are reduced to cameras and GPS. There is no longer a reliance on an HD map.
    \item Perception and planning is done jointly through a deep neural network.
    \item A control interface executes the plan from the neural network.
  \end{enumerate}
  }
  \label{intro-fig:end-to-end-net}
\end{figure}

A modular approach implies hand-engineering the perception features that are needed for the prediction task, and hand-coding the rules to generate a plan for the autonomous vehicle. This supposes it is possible to enumerate all the possible perception features and driving situations. In this thesis, we favour an end-to-end approach to learn a driving policy. We build a camera-first framework to train an end-to-end deep neural network on expert human driving data (\Cref{intro-fig:end-to-end-net}).

\subsection{World Models and Autonomous Driving}
Driving involves dealing with uncertainty. Many situations require the driver to take action in ambiguous scenarios. Coming up with a safe course of actions entails reasoning over uncertainty. There exists two types of uncertainty: epistemic uncertainty and aleatoric uncertainty \citep{fox11}. 

Epistemic uncertainty, derived from Greek \emph{episteme} (knowledge), captures our lack of knowledge or information. It is uncertainty that we could, in principle, resolve with more observations or knowledge.
It is represented as a single case (true or false) and is attributed to missing information or expertise. Examples of epistemic uncertainty include occlusions, where an object is no longer visible from the observations, or over-exposition due to sun light. In both cases, there is uncertainty but there is only a single outcome. There is or there is not a pedestrian occluded by a vehicle. There is or there is not a stop sign that is hard to see due to light conditions. 

Aleatoric uncertainty, derived from Latin \emph{alea} (chance), captures the stochastic nature of the real world. It is uncertainty that we could not resolve even if we wanted to.
It is represented by a class of possible outcomes and is focused on assessing an event's likelihood. This means that if we were able to run multiple times an experiment, different outcomes would be observed. An example is predicting the future which is inherently stochastic. Many different futures are possible and each have a different likelihood to occur. Modelling aleatoric uncertainty means being able to measure how likely different futures are.

Understanding how uncertain we are is essential to make safe decisions, especially in a safety-critical system such as an autonomous vehicle where a mistake could be fatal. Driving comes with an enormous set of challenges. One is dealing with imperfect information. The observations obtained from the sensors are not perfect. They may contain occlusions, or over and under exposition making it hard to distinguish objects and shapes. Another challenge is predicting the future. How to leverage past experience to infer which futures are plausible, and how likely they are. In this thesis, we examine how learning a world model that models both aleatoric and epistemic uncertainty leads to better driving performance. Similarly to Bayesian inference, we do not differentiate between epistemic and aleatoric uncertainties \citep{lawrence20} and model them with probability distributions.

\clearpage
\section{Contributions}
Learning a mental model of the world consists of reducing uncertainty by minimising prediction error. The world model should be able to reason about epistemic uncertainty (due to imperfect knowledge or information), and aleatoric uncertainty (due to the stochastic nature of the world). We parameterise the world model with deep neural networks and train it on a large corpus of driving data. We show that our model (i) has spatial memory, (ii) can fill in missing or partially observable information, and (iii) can predict and reason about future events. The thesis is organised as follows. 

\begin{itemize}
    \item In \textbf{\Cref{chapter:computer-vision}}, we review world models: how they were introduced in the context of reinforcement learning, and how they relate to video prediction. Then, we discuss important concepts from computer vision that will help scale training of large world models to urban driving scenes: geometry, semantics, and motion.
    \item \textbf{\Cref{chapter:generative-models}} formally introduces the probabilistic framework to infer a world model and a policy from observations and expert demonstrations. The remaining chapters will be instantiation of this framework in different settings.
    \item \textbf{\Cref{chapter:video-scene-understanding}} presents a model that predicts future depth, semantic segmentation, and optical flow for video scene understanding. This chapter focuses on the different network architectures to learn a representation from high resolution video. This work demonstrates for the first time diverse and plausible future predictions from high-dimensional video inputs. It was published at the European Conference on Computer Vision 2020 as "Probabilistic Future Prediction for Video Scene Understanding" \citet{hu2020probabilistic}.
    \item In \textbf{\Cref{chapter:instance-prediction}}, we propose a model that predicts future trajectories of dynamic agents in bird's-eye view from surround monocular cameras. We leverage 3D geometry as an inductive bias to learn a unified bird's-eye view representation from multiple camera views. Our predictions are consistent across cameras, and exhibit multi-modal behaviour of dynamic agents. This chapter studies the importance of learning a bird's-eye view representation for autonomous driving. It was published at the International Conference on Computer Vision 2021 as "FIERY: Future Instance Prediction in Bird's-Eye View From Surround Monocular Cameras" \citet{hu2021fiery}.
    \item And finally, in \textbf{\Cref{chapter:imitation-learning}}, we present an active agent that learns a world model and a policy to control a vehicle in a driving simulator. Our model jointly predicts static scene, dynamic scene, and ego-behaviour. This chapter demonstrates the benefits of learning a world model to improve performance in closed-loop driving. It was published at Neural Information Processing Systems as "Model-Based Imitation Learning for Urban Driving" \citet{hu2022mile}.
\end{itemize}

\begin{figure}[h]
  \centering
  \includegraphics[width=\linewidth]{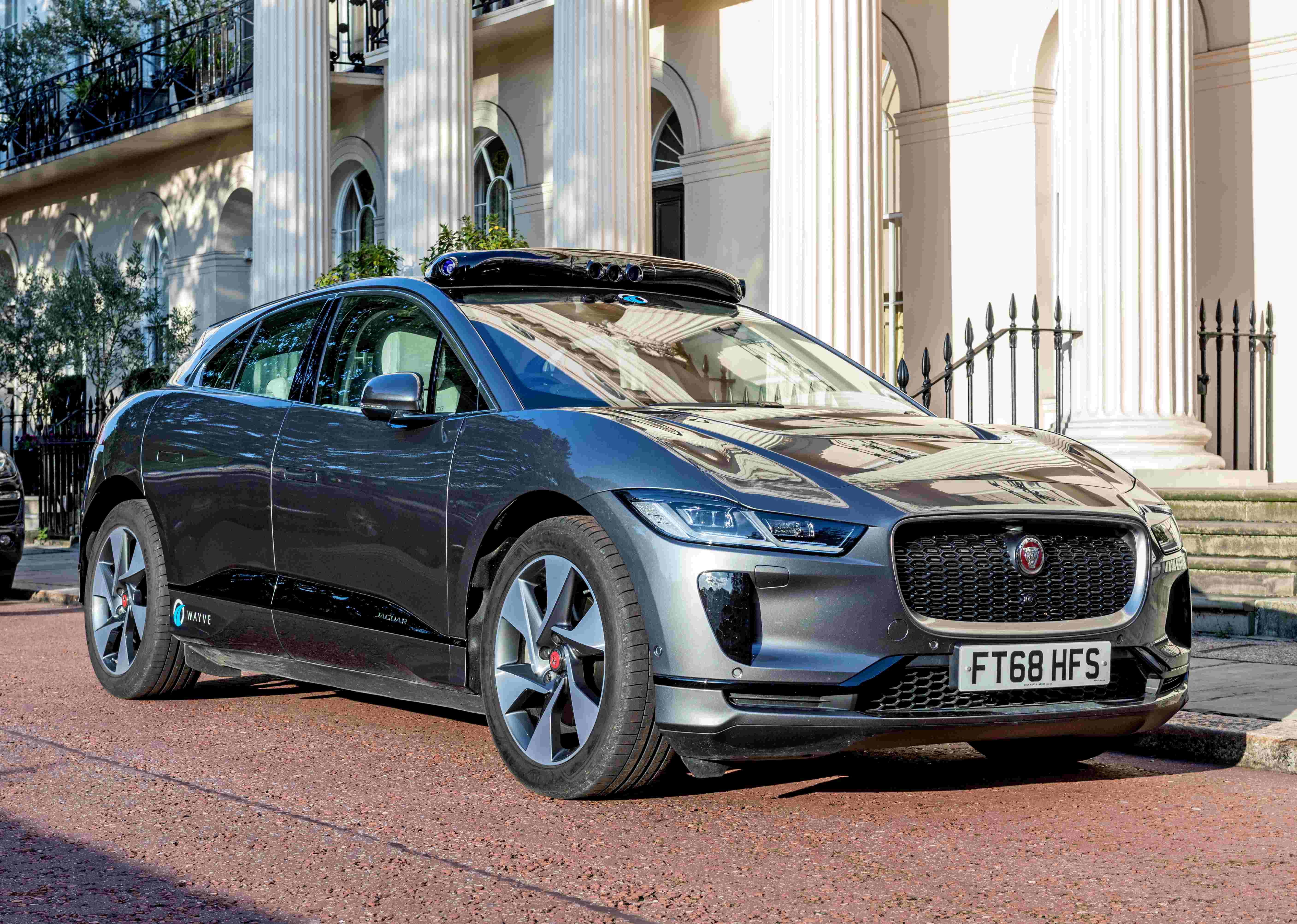}%
  \caption{Wayve's autonomous vehicle.}
  \label{intro-fig:wayve-vehicle}
\end{figure}

The content of this work is deployed on a real autonomous driving system at Wayve (\Cref{intro-fig:wayve-vehicle}). Advancing embodied intelligence by learning world models in complex environments is the purpose of this thesis.

\chapter{World Models and Computer Vision}
\label{chapter:computer-vision}
\graphicspath{{Chapter2/Figures/}}

The term \emph{world model} \citep{sutton90,thrun90} was originally introduced in the context of reinforcement learning \citep{minsky61}. \citet{sutton90,thrun90} defined a world model as the mapping between an agent's state and action to their future state and reward. They parametrised the world model with a neural network \citep{rosenblatt62,minsky69} and optimised the weights of the network through backpropagation \citep{lecun85,rumelhart86}. The optimised loss was the error between the predicted future state and reward, and the ground truth future state and reward from the environment. 

Using a world model in the context of reinforcement learning is called model-based reinforcement learning. The learned world model becomes a substitute of the real environment and can be used for either planning (using the world model to look ahead and select a trajectory that maximises future rewards) \citep{thrun90}, or learning (generate more training data with the world model) \citep{sutton90}. 

Current model-based reinforcement learning algorithms operate on either vector states (e.g.: 3D position, velocity) \citep{thrun90,deisenroth11}, grid worlds \citep{sutton-dyna90,SilverHuangEtAl16nature}, or low resolution images in game environments \citep{ha18,kaizer19,hafner2021dreamerv2}. In this thesis, we are interested in training world models from high-resolution visual inputs in complex urban driving scenes. Further, we assume that we do not have access to a ground truth reward, and that we cannot interact with the environment. 
\newpage
We thus use the more general definition of world models by \citet{friston21}. A world model is a \newterm{generative model} \citep{dayan95,leroux11}, which is a probabilistic description of how causes (e.g. latent states) lead to consequences (e.g. data or sensations). Our considered world models do not predict rewards, but only future states conditioned on current state and action.
We train world models from an offline corpus of high-resolution video and expert demonstrations. 

We will see that scaling world models to high-resolution images and complex driving scenes requires leveraging inductive biases from computer vision: \emph{geometry}, \emph{semantics}, and \emph{motion}. \emph{Geometry} refers to understanding the 3D structure of a scene, \emph{semantics} to segmenting the scene in meaningful semantic classes, and finally, \emph{motion} relates to reasoning about scene dynamics and predicting how objects will move in the future. 

In this chapter, we first review world models in the context of reinforcement learning with visual observations (\Cref{cv-section:world-models-rl}). Then, we study video prediction (\Cref{cv-section:video-prediction}), which consists of predicting future video frames from past image observations. Video prediction is closely related to world modelling as the problem can be framed as inferring future latent states that can be decoded to future image frames. Finally, we review important concepts in computer vision that will help scale our world models to complex urban driving scenes: geometry (\Cref{cv-section:geometry}), semantics (\Cref{cv-section:semantics}), and motion (\Cref{{cv-section:temporal-representation}}).
\clearpage
\section{World Models in Reinforcement Learning}
\label{cv-section:world-models-rl}

Learning a world model (or dynamics model) and using this model to train a policy has been explored as early as in the 80s. \citet{werbos89,nguyen89} trained world models with feed-forward neural networks and \citet{schmidhuber90,schmidhuber91} with recurrent neural networks. More recently, PILCO \citep{deisenroth11} modelled the world with Gaussian Processes and used the model to generate trajectories in order to optimise a controller to swing a pendulum. \citet{depeweg17} parameterised their world model with Bayesian neural networks instead of Gaussian Processes. These methods have demonstrated promising results when the states are known, well-defined, and relatively low-dimensional. We now consider modelling dynamics from higher dimensional image inputs.

Learning the dynamics of a system from video observations alone is a challenging but important problem \citep{bounou21}. To get around the difficult problem of learning dynamics from high-dimensional images, researchers have explored using neural networks to first learn a compressed representation of the images. \citet{wahlstrom15,wahlstrom2-15} used the bottleneck features of an image autoencoder to train a dynamics model and a policy to control a pendulum. Learning dynamics from a compressed latent state has been shown to enable RL algorithms to be much more data-efficient \citep{finn-autoencoders16,watter15}. 

\begin{figure}[h]
    \centering
    \includegraphics[height=0.12\textheight]{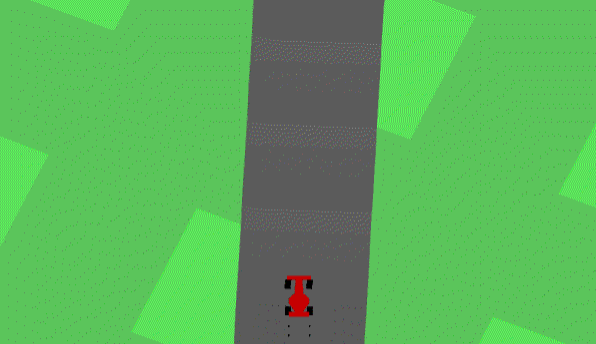}
    \caption[The Car Racing environment.]{The Car Racing environment \citep{carracing16}.}
\label{cv-fig:car-racing}
\end{figure}

\begin{figure}[h]
    \centering
    \begin{subfigure}[b]{0.24\textwidth}
        \centering
        \includegraphics[height=0.12\textheight]{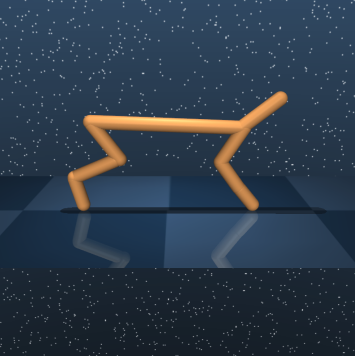}
        \caption{Cheetah}
    \end{subfigure}
    \begin{subfigure}[b]{0.24\textwidth}
        \centering
        \includegraphics[height=0.12\textheight]{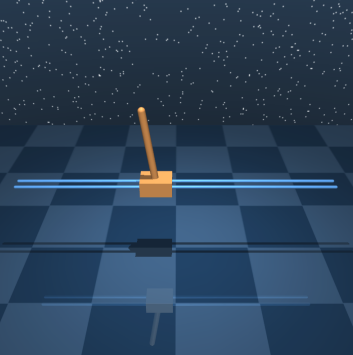}
        \caption{Cartpole}
    \end{subfigure}
    \begin{subfigure}[b]{0.24\textwidth}
        \centering
        \includegraphics[height=0.12\textheight]{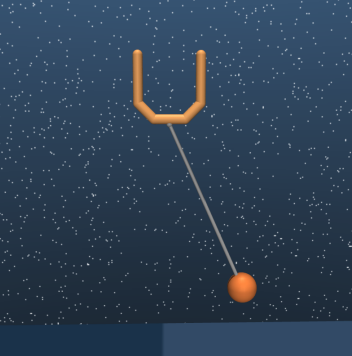}
        \caption{Cup}
    \end{subfigure}
    \begin{subfigure}[b]{0.24\textwidth}
        \centering
        \includegraphics[height=0.12\textheight]{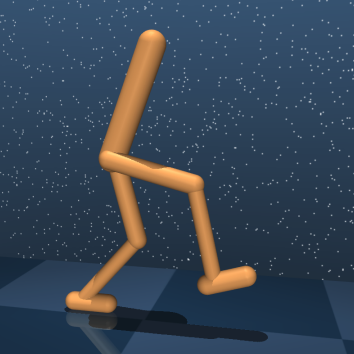}
        \caption{Walker}
    \end{subfigure}
    \caption[Examples of control tasks in the DeepMind Control Suite.]{Examples of control tasks in the DeepMind Control Suite. \citep{tassa18}}
\label{cv-fig:deepmind-control-suite}
\end{figure}

\begin{figure}[h]
    \centering
    \begin{subfigure}[b]{0.32\textwidth}
        \centering
        \includegraphics[height=0.12\textheight]{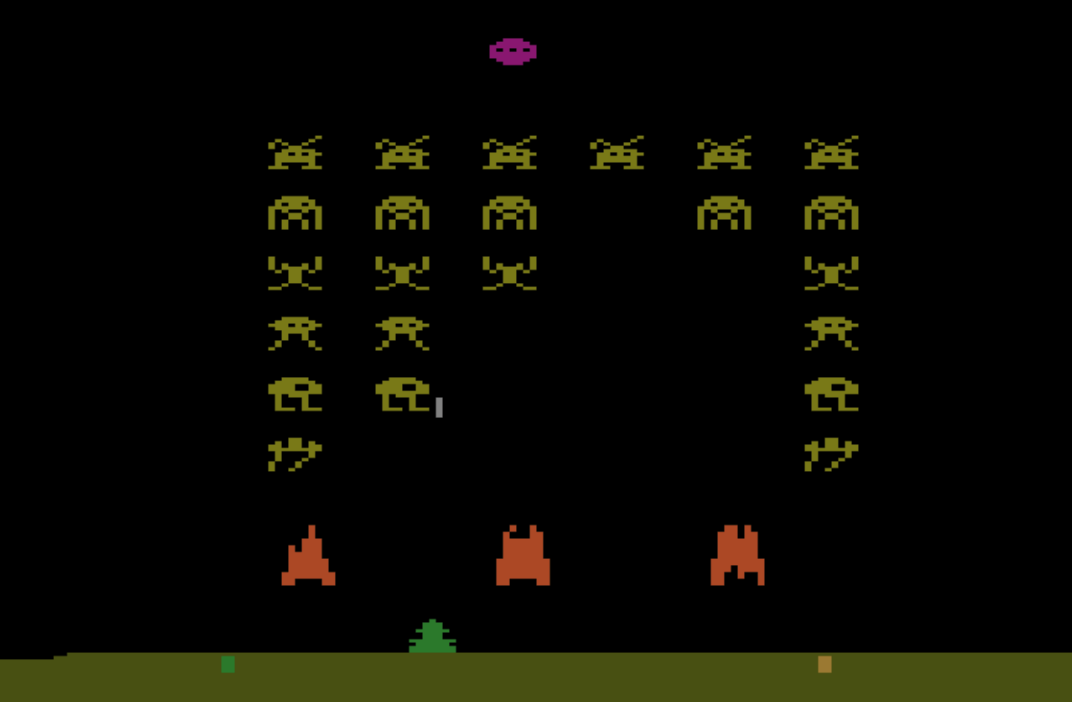}
        \caption{\textsc{Space Invaders}}
    \end{subfigure}
    \begin{subfigure}[b]{0.32\textwidth}
        \centering
        \includegraphics[height=0.12\textheight]{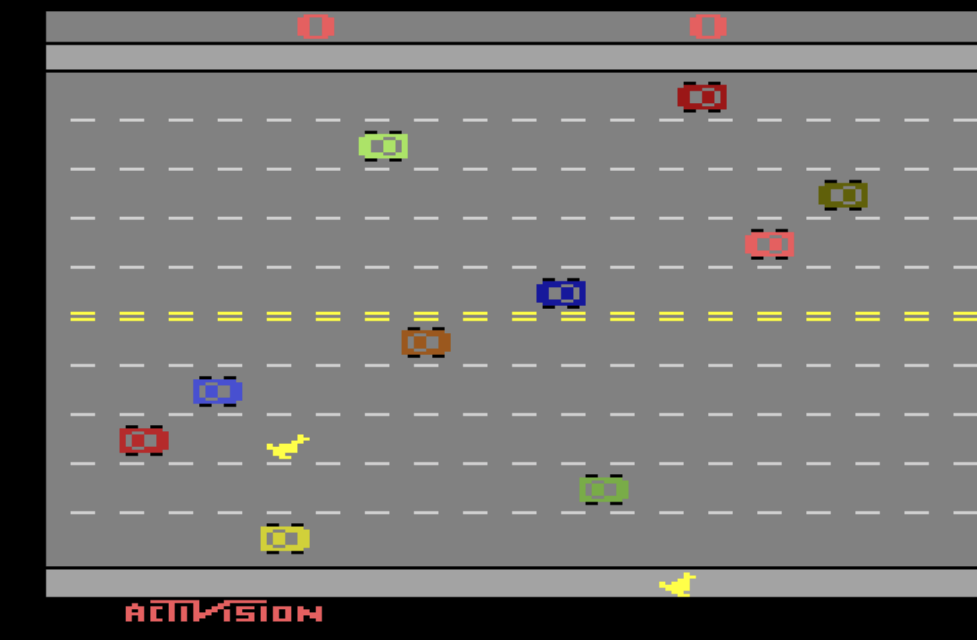}
        \caption{\textsc{Freeway}}
    \end{subfigure}
    \begin{subfigure}[b]{0.32\textwidth}
        \centering
        \includegraphics[height=0.12\textheight]{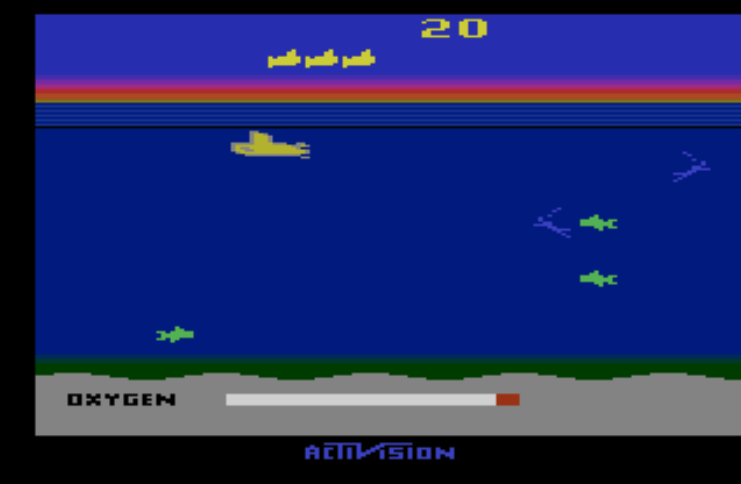}
        \caption{\textsc{Seaquest}}
    \end{subfigure}
    \caption[Examples of Atari games.]{Examples of Atari games \citep{bellemare-arcade13}.}
\label{cv-fig:atari}
\vspace{-10pt}
\end{figure}

Recent works have shown impressive performance by learning latent dynamics from pixel observations and either using the learned representation to optimise a policy \citep{ha18}, or planning  with the world model \citep{hafner2019planet,hafner2021dreamerv2}. The environments considered are: CarRacing \citep{ha18} in \Cref{cv-fig:car-racing}, DeepMind Control Suite \citep{hafner2019planet} in \Cref{cv-fig:deepmind-control-suite}, and Atari \citep{hafner2021dreamerv2} in \Cref{cv-fig:atari}.

\begin{figure}
\centering
\includegraphics[width=0.8\textwidth]{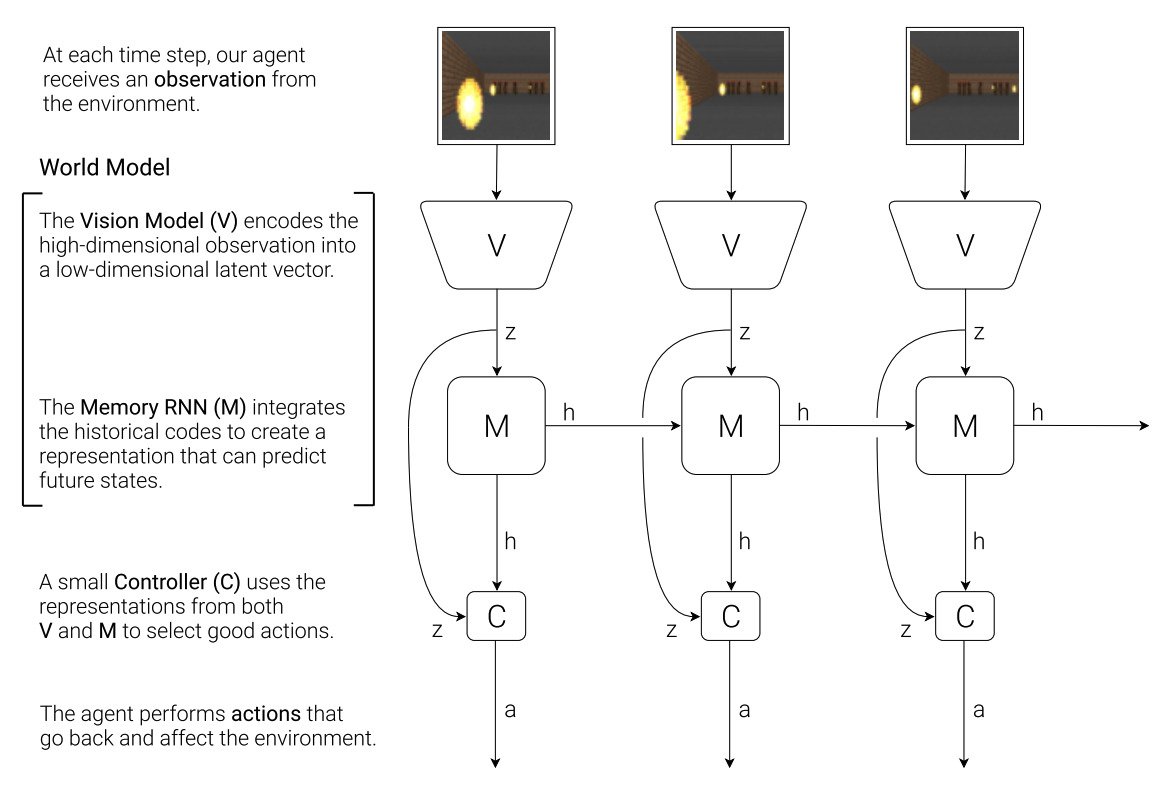}
\caption[An example of world model architecture.]{An example of world model architecture, figure from \citet{ha-detailed18}}
\label{cv-fig:ha-world-model}
\end{figure}

To get a better understanding of how world models are used in the reinforcement learning framework, we will now describe the method of \citet{ha18}, which is one of the first work to solve game-like environments by learning latent dynamics. The architecture of their model is given in \Cref{cv-fig:ha-world-model}. It comprises a Vision model (a variational-autoencoder \citep{kingma14} that encodes the input image $\vo$ to a representation $\vz$), a Memory RNN model, which creates a temporal representation $\vh$, and finally a Controller model that uses both $\vz$ and $\vh$ to predict action $\va$.

The world model (Vision model + Memory RNN model) is a large neural network trained in an unsupervised way to reconstruct image frames. The controller is optimised using evolution strategy. More concretely, the training procedure is the following.

\begin{enumerate}
    \item Collect 10,000 episodes with a random policy.
    \item Train the Vision model to encode each frame $\vo\in \R^{3\times64\times64}$ into a latent vector $\vz\in \R^{32}$.
    \item Train the Memory RNN model to predict the next latent vector $p(\vz_{t+1}|\vz_t, \va_t, \vh_t)$, with $\vh_t \in \R^{256}$.
    \item Define the Controller as $\va_t = \mW_c[\vz_t, \vh_t] + \vb_c$.
    \item Estimate the parameters $\mW_c$ and $\vb_c$ with Covariance-Matrix Adaptation Evolution Strategy (CMA-ES) \citep{hansen16} to maximise the expected cumulative reward.
\end{enumerate}

This approach side-steps the RL credit assignment problem (knowing which actions taken in the past led to positive or negative regards) that makes large RL models hard to train. The evolution strategy can focus on the credit assignment problem on a much smaller search space, while the world model fully captures the representation of the environment. In reinforcement learning, smaller neural network architectures are often preferred over larger and more expressive ones because of the credit assignment problem. The idea is to train a world model that would encompass all the complexity of the scenes, and then train a smaller decision model that focuses on the credit assignment problem.
\clearpage
\section{Video Prediction}
\label{cv-section:video-prediction}

We reviewed world models in the context of reinforcement learning. We saw that it was much more efficient to apply RL algorithms on a learned latent space than directly in pixel space, which is very high-dimensional. The environments considered were simulated, and rather low-resolution ($64\times64$). In this section, we will discuss video prediction, which consists of predicting future image frames from past image observations \citep{srivastava15,oh15}. Video prediction is closely related to visual world models (world models that operate on image observations), as they both aim to predict future observations.

Contrarily to the previous section, we will review video prediction methods that are applied to natural real-world scenes \citep{finn16,franceschi20}, and are higher resolution \citep{clark19}. The learned representation through video prediction training can be subsequently be the input to a policy to solve a downstream task \citep{srivastava15,lotter17}.

\subsection{Video Pixel Prediction}
Videos have the convenient property that consecutive frames are temporally consistent. This property enables a learning framework of predicting past or future frames, which is entirely self-supervised as the targets are the RGB frames. Despite this desirable property, learning a predictive model from videos remains extremely challenging because of the high dimensionality of the data, the stochasticity and complexity of the scenes, and the partial observability of the environment \citep{oprea20}.

\subsubsection{First Video Prediction Models}
\begin{figure}[h]
    \centering
    \begin{subfigure}[b]{0.25\textwidth}
        \centering
        \includegraphics[height=0.12\textheight]{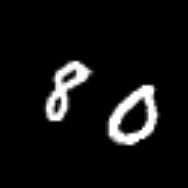}
    \end{subfigure}
    \begin{subfigure}[b]{0.25\textwidth}
        \centering
        \includegraphics[height=0.12\textheight]{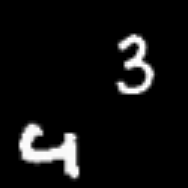}
    \end{subfigure}
    \begin{subfigure}[b]{0.25\textwidth}
        \centering
        \includegraphics[height=0.12\textheight]{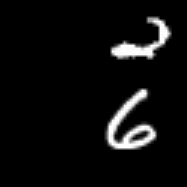}
    \end{subfigure}
    \caption[Examples of frames in the Moving MNIST dataset.]{Examples of frames in the Moving MNIST dataset \citep{srivastava15}.}
\label{cv-fig:moving-mnist}
\end{figure}

\paragraph{Moving MNIST.}
\citet{srivastava15} presented one of the first work on video pixel prediction. They trained an LSTM-based model to predict future frames on a new dataset they introduced: Moving MNIST dataset. This dataset contains videos of two $28\times 28$ digits moving in a $64\times 64$ black background as shown in \Cref{cv-fig:moving-mnist}.  

\paragraph{Action-conditional prediction on Atari.}
\citet{oh15} tackled the problem of action-conditional future frame prediction in Atari \citep{bellemare-arcade13}. The environments encountered in Atari are more diverse (tens of objects that can influence each other) and depend on the action of the active agent that interacts with the environment. They showed that a trained controller (Deep Q-Network \citep{mnih13}) could still operate well even by replacing the ground truth frames with predicted frames from their model. 

The video prediction models presented so far were all deterministic and predicted a single future. The future is however stochastic. We now present the three main families of probabilistic future prediction models: generative adversarial networks, autoregressive models, and variational models.

\subsubsection{Generative Adversarial Networks}
\citet{srivastava15} and \citet{oh15} both used an $L_2$ loss for their predicted images reconstruction. However, when multiple futures are possible, the network will be encouraged to predict the mean future since the $L_2$ loss assumes the data is drawn from a Gaussian distribution. This results in blurry image predictions. \citet{mathieu16} introduced an adversarial loss \citep{goodfellow14} to better model multimodal distributions. If the model outputs the average future frames, it will not fool the discriminator, therefore, the generator has to select one the mode (conditioned on the noise) to successfully deceive it. 

\subsubsection{Autoregressive Models} \citet{kalchbrenner16} proposed a predictive model from another class of generative models: autoregressive. Their probabilistic video model, named Video Pixel Network (VPN), estimated the joint probability distribution of raw pixel values in a video. VPN was the first model able to predict future frames without any artifacts (e.g. blurring) on Moving MNIST, as well as generating detailed samples on the Robotic Pushing dataset \citep{finn16}. One downside is that generating images is computationally expensive as it involves generating each pixel one after the other due to the factorisation of the joint probability distribution of pixels.

\subsubsection{Variationally Trained Models} The last family of probabilistic video prediction is variational \citep{babaeizadeh18,denton18,franceschi20}. They are called variational models because these networks optimise the variational lower bound of the log evidence of the observed data. Variational models infer the latent dynamics from the image observations. Predictions are made in the latent space, and are decoded to image observations.
Two distributions parametrised by neural networks are optimised jointly: the prior and the posterior distributions. The prior network estimates the next latent state given the past observations up to time $t-1$. The posterior network has additionally access to the ground truth image at time $t$. During training, the prior is encouraged to match the posterior distribution with a mode-covering Kullback-Leibler divergence loss.  

\subsection{Transfer Learning from Video Prediction}
A video prediction model learns an internal representation that captures both the scene content and the dynamics \citep{clark19}. Therefore, such a learned representation could transfer to other downstream tasks \citep{srivastava15,lotter17}.

\citet{srivastava15} trained a video prediction model on the UCF101 dataset \citep{soomro12}, which consists of realistic action videos. They showed that by using the learned representation, they improved classification accuracy on the action recognition task. \citet{lotter17} trained a model to predict future frames from the front-facing camera of a vehicle. They subsequently used the learned representation to predict the steering angle of the vehicle. This representation proved to be a powerful signal as being able to predict the future requires an implicit knowledge of how the objects of the scene are allowed to move.

\subsection{Scaling World Models to Autonomous Driving}
We have seen that predicting future video frames was a difficult problem due to data complexity \citep{clark19}, and modelling the stochasticity of the future \citep{babaeizadeh18,denton18}. In our case however, our end goal is to control an autonomous vehicle. Is it necessary to reconstruct the image input at the pixel-level accuracy? Intuitively, the learned representation must contain information about the following.
\begin{itemize}
    \item \textbf{Static scene}. To navigate in a 3D space, an active agent needs to understand semantics (what is around me?) and geometry (how far are the obstacles?).
    \item \textbf{Dynamic scene}. They must interact and reason about the trajectories of dynamic agents (motion).
\end{itemize}

We will now review these three concepts in computer vision: geometry, semantics, and motion. We will then see in the remaining of the thesis how these concepts can be applied to scale training world models on high-resolution images of urban driving scenes.

\clearpage
\section{Geometry}
\label{cv-section:geometry}

Navigating in the real world requires knowing how far the obstacles are \citep{reichardt94}. That relative distance can be estimated in several ways. (i) Using images from two synchronised cameras separated by a known distance, (ii) with a moving camera (structure from motion), or (iii) if the scene is static, using images with different luminosity conditions. We are going to focus on methods (i) and (ii) as the scenes we are interested in are dynamic.

Initial learned methods \citep{saxena09, eigen14} for depth estimation from monocular images were based on supervised learning that assumed a ground truth depth map was available. This ground truth depth could be acquired using radars or lidars by measuring the time an emitted beam takes to bounce on an obstacle. Radars have a longer range than lidars (200-300m vs. 50-100m) as the radio waves (mm) are less absorbed than light waves (nm) when bouncing on obstacles. However, the short wavelength of lidars enables the beams to be modulated faster, resulting in a higher temporal and spatial resolution compared to radars \citep{liang2018}. 

\begin{figure}[h]
\centering
\includegraphics[width=\textwidth]{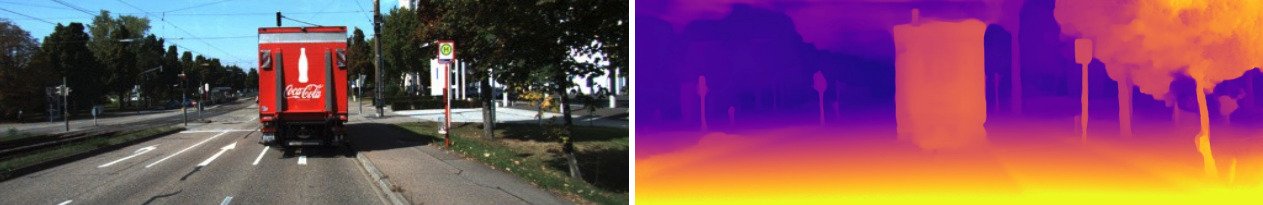}
\caption[Example of predicted dense depth map.]{Example of predicted dense depth map. Brighter colours indicate pixels that are closer to the camera. Figure from \citet{guizilini20}.}
\label{cv-fig:depth}
\end{figure}

More recent methods do not rely on ground depth depth labels and are completely self-supervised. \citet{godard17} introduced a self-supervised method to estimate depth using synchronised stereo images (method (i)). By reconstructing one image from the other view, the model effectively learns disparity and depth can then be calculated if the distance between the stereo cameras is known. \citet{zhou17} proposed a self-supervised method from monocular video data, where the learning signal comes from the temporal consistency of consecutive frames (method (ii)). \Cref{cv-fig:depth} shows an example of predicted depth.

\subsection{Supervised Methods}
Supervised monocular depth estimation can be formalised as finding a function $f$ that maps an input image $I$ to its per-pixel depth map $\hat{\vd} = f(I)$. The optimisation process iteratively refines the function $f$ using the prediction error from the ground truth depth $\vd$.

\citet{saxena09} proposed a model that builds patches of images and estimated their 3D location and orientation, followed by a Markov Random Field that combines predictions. The model has difficulties on thin structures, and as predictions are made locally, lacks the global context to generate realistic outputs. 

\citet{eigen14} used a two-scale deep network taking raw pixel values as input to produce depth values. Contrary to \citet{saxena09}, their model does not rely on hand-crafted features but directly learns a representation from pixels. Several works have then built upon this approach using techniques such as Conditional Random Fields \citep{li15}, using more robust loss functions \citep{laina16}, or incorporating scene priors \citep{wang15}.

\subsection{Self-Supervised Stereo Methods}
Supervised methods require a large quantity of high quality annotated data. That can be challenging to acquire in a range of environments as laser measurements can be imprecise in natural scenes due to motion and reflections. \citet{godard17} treats depth estimation as an image reconstruction problem using pairs of images acquired from cameras separated by a known distance. Their method does not aim at directly predicting the depth map, but instead tries to a find a dense correspondence field $d^r$ such that when it is applied to the left image $I^l$, the right image can be reconstructed $\tilde{I}^r = I^l(d^r)$. By assuming that the images are rectified (i.e. the left and right images lie on a common image plane), then $d^r$ is the image disparity. If additionally, the baseline distance $b$ between the two cameras and the focal length $f$ are known, the depth $\hat{d}^r$ can be obtained with the following formula: $\hat{d}^r = \frac{bf}{d^r}$.



\paragraph{Reconstruction loss.} The network learns to generate images by sampling from the opposite stereo image using the fully-differentiable image sampler from Spatial Transformer Network \citep{jaderberg15}. The photometric error $pe(I, \tilde{I})$ between the ground truth image $I$ and reconstructed image $\tilde{I}$  is a combination of $L_1$ and Structural Similarity (SSIM) loss: 
\begin{equation}
pe(I, \tilde{I}) = \frac{1}{N} \sum_{i, j} \left(\alpha \frac{1 - \text{SSIM}(I_{ij}, \tilde{I}_{ij})}{2} + (1 - \alpha) \Vert I_{ij} -  \tilde{I}_{ij}\Vert_1 \right)
\end{equation}
with $N$ the number of pixels and $\alpha$ a scalar in $[0,1]$.

\subsection{Self-Supervised Monocular Methods}
It is also possible to estimate depth from monocular videos, where the supervision comes from consecutive temporal frames \citep{zhou17,godard19,guizilini20}. In addition to predicting the depth map, ego-motion also has to be inferred. Camera pose prediction is only needed during training to constrain the depth estimation network. Contrarily to the stereo methods, only a single monocular camera is required which makes this approach widely applicable.

\begin{figure}[h]
  \centering
  \includegraphics[width=.5\linewidth]{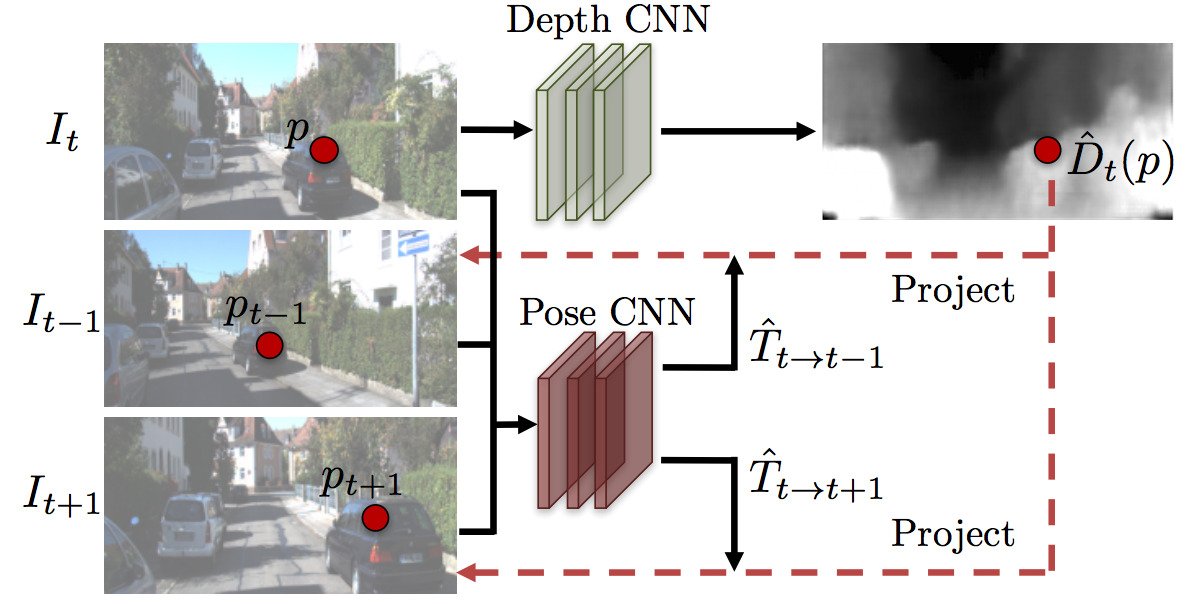}
  \caption[Self-supervised monocular depth prediction model.]{Self-supervised monocular depth prediction model, figure from \citet{zhou17}}
  \label{cv-fig:self-supervised-mono}
\end{figure}

\citet{zhou17} trained a depth estimation network with a separate pose estimation network under the assumption that the scene is static. This assumptions means that appearance change is mostly due to camera motion. The supervision comes from novel view synthesis: generating a new image of the scene from a different camera pose. Let us denote by $(I_1, I_2, ..., I_T)$ a sequence of images, with one of the frame $I_t$ being the target view, and the other frames being the source views. The reprojection loss is:

\begin{equation}
\mathcal{L}_{p} = \sum_{s \neq t} pe(I_t, \hat{I}_{s\rightarrow t})
\end{equation}
with $\hat{I}_{s\rightarrow t}$ the synthesised view of $I_t$ from source image $I_s$ using the predicted depth $\hat{D}_t$ and the predicted $4\times4$ camera transformation $\hat{T}_{t\rightarrow s}$. As illustrated in Figure \ref{cv-fig:self-supervised-mono}, we can sample pixels from the source image $I_s$ to reconstruct $I_t$ . 
Let us denote by $p_t$ the coordinate of a pixel in the target image $I_t$ in homogenous coordinate. Given the camera intrinsic matrix $K$ and the mapping $\varphi$ from image plane to camera coordinate,
the corresponding pixel in the source image is provided by:

\begin{equation}
p_s \sim K\varphi^{-1}[\hat{T}_{t\rightarrow s}\varphi(K^{-1}p_t, \hat{D_t}(p_t))]
\end{equation}

Since the projected coordinates $p_s$ are continuous values, the Spatial Transformer Network \citep{jaderberg15} sampling mechanism is used to select the 4 neighbouring pixels to populate the reconstructed image $\hat{I}_{s\rightarrow t}$.

\subsubsection{Minimum Reprojection Loss}
If the depth of a pixel is correctly predicted on the target image, but that pixel either (i) disappears from the field of view of the source image due to camera motion, or (ii) is occluded, then the reprojection loss will incur a large error. The standard approach is to sum the reprojection losses from each source image. \citet{godard19} instead proposes a minimum reprojection los: 

\begin{equation}
    \mathcal{L}_{p} = \sum_p \min_{s \neq t} pe(I_t(p), \hat{I}_{s\rightarrow t}(p))
\end{equation}

\begin{figure}[h]
  \centering
  \includegraphics[width=.5\linewidth]{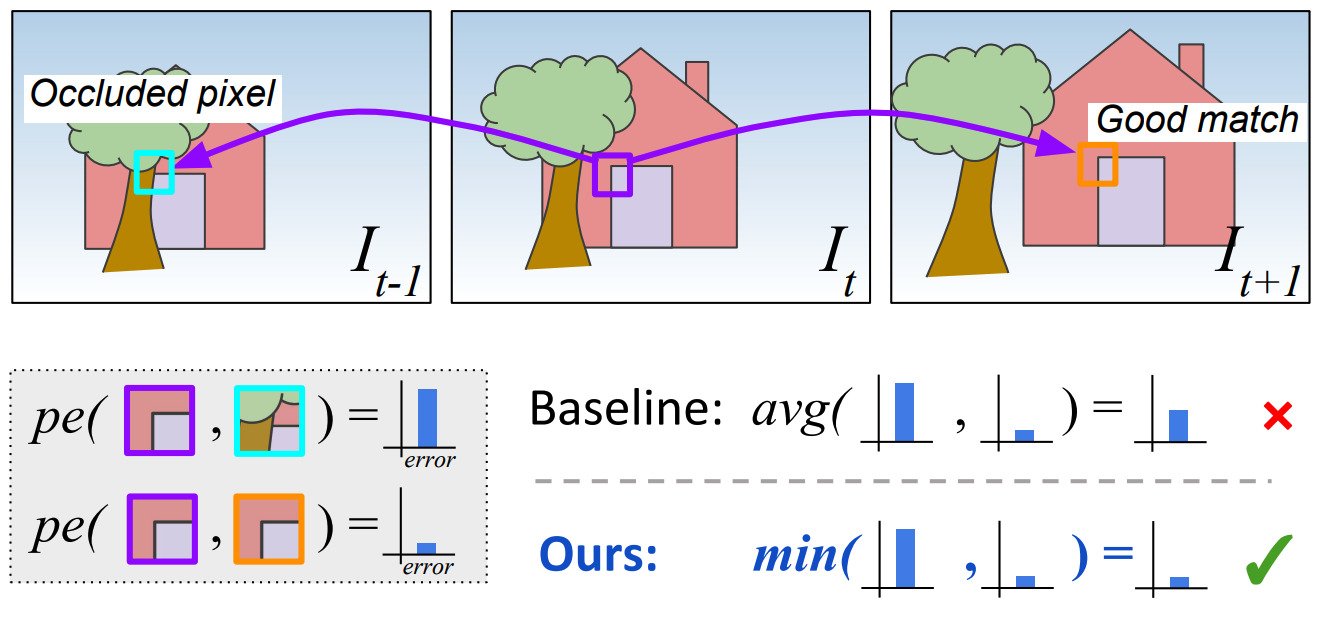}
  \caption[Per-pixel min reprojection loss.]{Per-pixel min reprojection loss, figure from \citet{godard19}}
  \label{cv-fig:min-reprojection}
\end{figure}

As shown in \Cref{cv-fig:min-reprojection}, this reformulation prevents pixels that are occluded or out-of-field of view to be considered and results in sharper depth prediction in occlusion boundaries and reduces artifacts around image borders.

\clearpage
\section{Semantics}
\label{cv-section:semantics}
Depth prediction allows the understanding of the geometry of a scene by estimating distances. Semantic segmentation tackles the problem of understanding the semantics: \textbf{what is in the scene?}

\subsection{Semantic Segmentation}
Semantic segmentation refers to assigning a class label to every pixel in an image. It can also be seen as clustering parts of the image into different classes \citep{brice70}.

\begin{figure}[h]
\centering
\includegraphics[width=\textwidth]{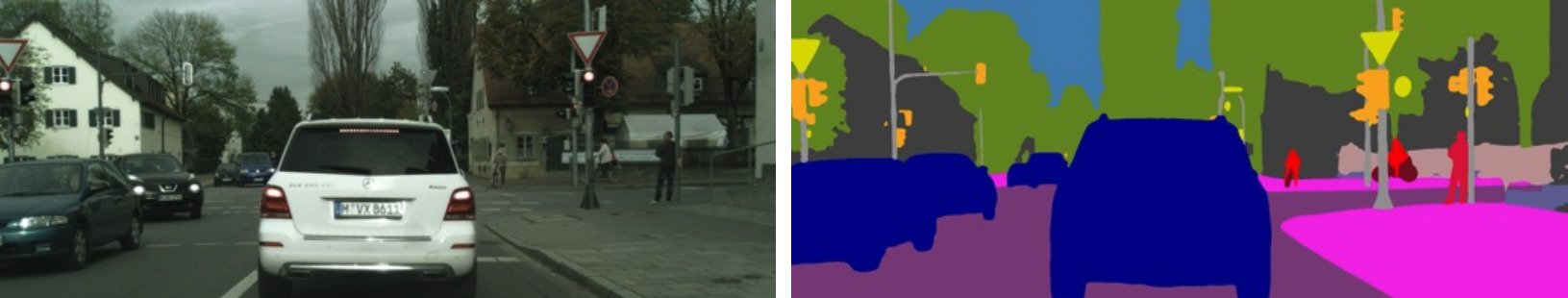}
\caption[Example of predicted semantic segmentation.]{Example of predicted semantic segmentation. Figure from \citet{kendall18}.}
\label{cv-fig:semantic-segmentation}
\end{figure}

As context helps disambiguating between classes, semantic segmentation models must learn the spatial relationship between objects. Recent approaches use an encoder \citep{long15,badrinarayanan15}. An encoder is a convolutional neural network (CNN) that spatially downsamples the image to a bottleneck feature. A decoder then upsamples the encoded feature to output a dense class prediction. \Cref{cv-fig:semantic-segmentation} shows an example of predicted semantic segmentation.

\begin{figure}[h]
    \centering
    \begin{subfigure}[b]{0.23\textwidth}
        \centering
        \includegraphics[height=0.12\textheight]{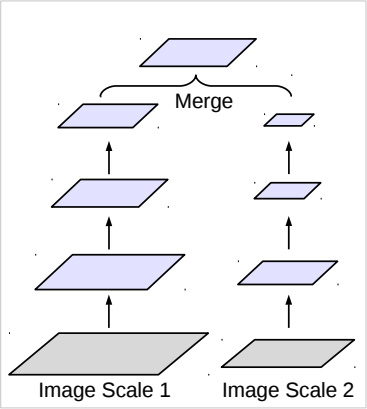}
        \caption{\tiny{Image Pyramid.}}
    \end{subfigure}
    \begin{subfigure}[b]{0.23\textwidth}
        \centering
        \includegraphics[height=0.12\textheight]{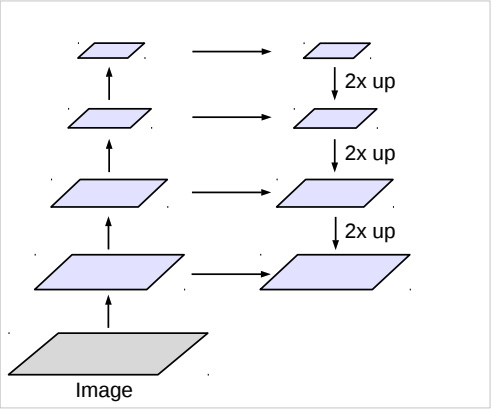}
        \caption{\tiny{Encoder-Decoder.}}
    \end{subfigure}
    \begin{subfigure}[b]{0.23\textwidth}
        \centering
        \includegraphics[height=0.12\textheight]{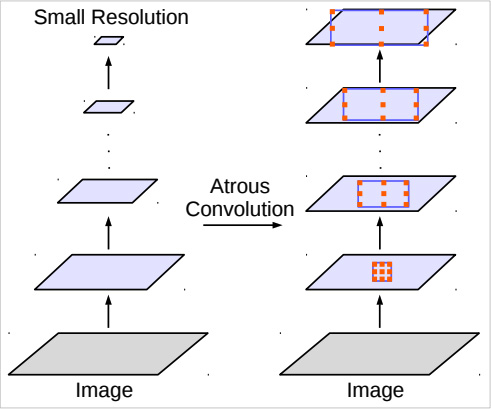}
        \caption{\tiny{Atrous Convolutions.}}
    \end{subfigure}
    \begin{subfigure}[b]{0.23\textwidth}
        \centering
        \includegraphics[height=0.12\textheight]{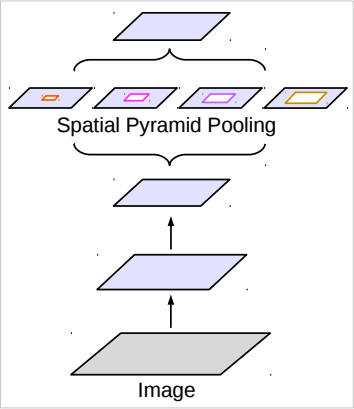}
        \caption{\tiny{Spatial Pyramid Pooling.}}
    \end{subfigure}
    \caption[Different architectures to capture multi-scale context in images.]{Different architectures to capture multi-scale context in images, figure from \citet{chen17}}
    \label{cv-fig:semantic-architecture}
\end{figure}

Objects can appear in different scales in the image. There exist multiple approaches to deal with multi-scale as shown in \Cref{cv-fig:semantic-architecture}:
\begin{enumerate}[label=(\alph*)]
	\item \textbf{Image Pyramid}. Encoding the image at different image scales and merging the resulting feature maps \citep{farabet13}.
	\item \textbf{Encoder-Decoder}. This architecture is the most popular, and is also often called U-Net \citep{ronneberger15}. It uses skip-connections with intermediate features to inject higher resolution context in the decoded features.
	\item \textbf{Atrous Convolutions} (also called dilated convolutions). This approach uses dilated convolutions instead of strided convolutions. The spatial receptive field is thus unchanged but comes at the cost of additional compute and memory requirements \citep{chen-deeplab17}.
	\item \textbf{Spatial Pyramid Pooling}. This method adds a step of spatially pooling the encoded feature at different resolution to add global context to the bottleneck feature \citep{chen17}.
\end{enumerate}

\subsection{Instance Segmentation}
\begin{figure}[h]
\centering
\includegraphics[width=\textwidth]{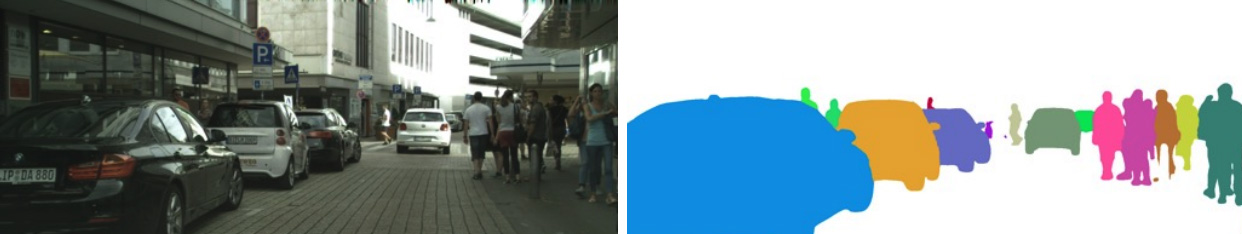}
\caption[Example of predicted instance segmentation.]{Example of predicted instance segmentation. Each detected object is coloured differently. Figure from \citet{kendall18}.}
\label{cv-fig:instance-segmentation}
\end{figure}

Instance segmentation refers to segmenting countable objects in an image and assigning them a unique identifier \citep{tu03}, as shown in \Cref{cv-fig:instance-segmentation}.

There exist two main approaches for instance segmentation: region-proposal based \citep{he17,chen17b,liu18} and embedding based \citep{brabandere17, kong18,fathi17}.  Region-proposal methods first predict bounding boxes where an object might be present. Overlapping bounding boxes are processed with non-maximum suppression and the object in the bounding box is subsequently segmented with a mask head. Embedding-based methods predict a dense pixel-wise embedding feature and use a clustering method to obtain individual instances. 

Even though region-proposal methods tend to perform better, they have several shortcomings. (i) Two objects may share the same bounding box and in that situation, the mask head does not know which object to pick in that particular box, (ii) a pixel can belong to two separate objects as each prediction is done independently,  and (iii) the number of detected objects is limited by the fixed number of proposals of the network. The embedding-based methods rely on a clustering method to output the individual instances and this may cause an instance to be split in two, or false positives depending on the sensitivity of the clustering algorithm.

\citet{kirillov2019panoptic} introduced the concept of \textbf{Panoptic Segmentation} that unifies semantic segmentation and instance segmentation. An image is divided into \emph{stuff} (uncountable) and \emph{things} (countable objects). Both stuff and things are classified through semantic segmentation, with things being additionally assigned a unique identifier.









\subsection{Video Instance Segmentation}
All the tasks mentioned so far operate on a single image. The same instance detected in two consecutive images is not tracked. To overcome this shortcoming, \citet{yang19} proposed the task of \textbf{Video Instance Segmentation} where the goal is to not only segment instances in each frame, but also track detected instances over time. We present the two main approaches for video instance segmentation: region-proposal based methods and embedding-based methods.

\subsubsection{Region-Proposal Methods}
This family of methods is an extension of the region-proposal methods used in instance segmentation. An example is Track R-CNN \citep{voigtlaender19}, which extends Mask R-CNN \citep{he17} in two ways. They incorporate 3D convolutions to integrate temporal information, and add an association head that produces an association vector for each detection, inspired from person re-identification \citep{beyer17,tang17}. Each detected instance has an association vector, allowing tracking by comparing the Euclidean distance between the lists of detected vectors in two consecutive frames and performing optimal assignment.

\subsubsection{Embedding-Based Methods}
Embedding-based methods map video-pixels to a high-dimensional embedding space. This space encourages video-pixels of the same instance to be close together and distinct from other instances. Segmentation and tracking is done through clustering in the embedding space.

\citet{hu19} introduced a spatio-temporal embedding loss with three competing forces, similarly to \citet{brabandere17}. The attraction force (\Cref{loss1}) encourages the video-pixels embedding of a given instance to be close to its embedding mean. The repulsion force (\Cref{loss2}) incites the embedding mean of a given instance to be far from all others instances. And finally, the regularisation force (\Cref{loss3}) prevents the embedding to diverge from the origin.

Let $\vo$ denote the RGB video frames. A neural network $f$ outputs a spatio-temporal embedding $\vy = f(\vo)$.
Let us denote by $K$ the number of instances, and by $S_k$ the set of all video-pixels of instance $k \in \{1,\dots,K \}$. For all $i \in S_k$, we denote by $\vy_i$ the embedding for pixel $i$ and by $\vmu_k$ the mean embedding of instance $k$: $\vmu_k = \frac{1}{|S_k|} \sum_{i \in S_k} \vy_i$. The embedding loss is given by:

\begin{align}
    \label{loss1}
	\mathcal{L}_{\text{a}} ={}& \frac{1}{K} \sum_{k=1}^K \frac{1}{|S_k|} \sum_{i \in S_k} \max(0, \Vert \vmu_k - \vy_i \Vert_2 - \rho_a)^2\\
	\label{loss2}
	\mathcal{L}_{\text{r}} ={}& \frac{1}{K(K-1)} \sum_{k_1 \neq k_2} \max(0, 2\rho_r - \Vert \vmu_{k_1} - \vmu_{k_2} \Vert_2)^2\\
	\label{loss3}
	\mathcal{L}_{\text{reg}} ={}& \frac{1}{K} \sum_{k=1}^K \Vert \vmu_k\Vert_2
\end{align}
where: 
\begin{itemize}
\item $\rho_a$ defines the attraction radius, constraining the embedding to be within a distance $\rho_a$ to its mean.
\item $\rho_r$ is the repulsion radius, constraining the mean embedding of two different instances to be at least $2\rho_r$ apart. Therefore, if we set $\rho_r > 2\rho_a$, a pixel embedding of an instance $k$ will be closer to all the pixel embeddings $i \in S_k$ of instance $k$, than to the pixel embeddings of any other instance.
\end{itemize}

\begin{figure}[h]
\begin{center}
\includegraphics[width=\linewidth]{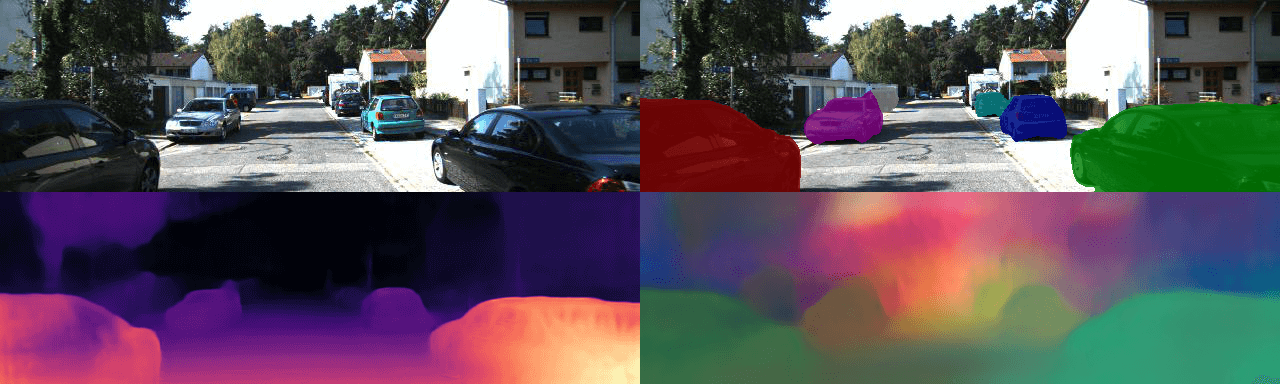}
\end{center}
\caption[An illustration of the embedding-based video instance segmentation model.]{An illustration of the embedding-based video instance segmentation model. Clockwise from top left: input image, predicted multi-object instance segmentation, visualisation of the high-dimensional embedding and predicted monocular depth. Figure from \citet{hu19}.}
\label{cv-fig:embedding}
\vspace{-10pt}
\end{figure}

During inference, each pixel of the considered frame is assigned to an instance by randomly picking an unassigned pixel and aggregating close-by pixels with the mean shift algorithm \citep{comaniciu2002mean} until convergence. In the ideal case, with a test loss of zero, this will result in perfect instance segmentation. \Cref{cv-fig:embedding} shows an example of prediction.

\begin{figure}[h]
    \centering
    \begin{subfigure}[b]{\textwidth}
        \includegraphics[width=\linewidth]{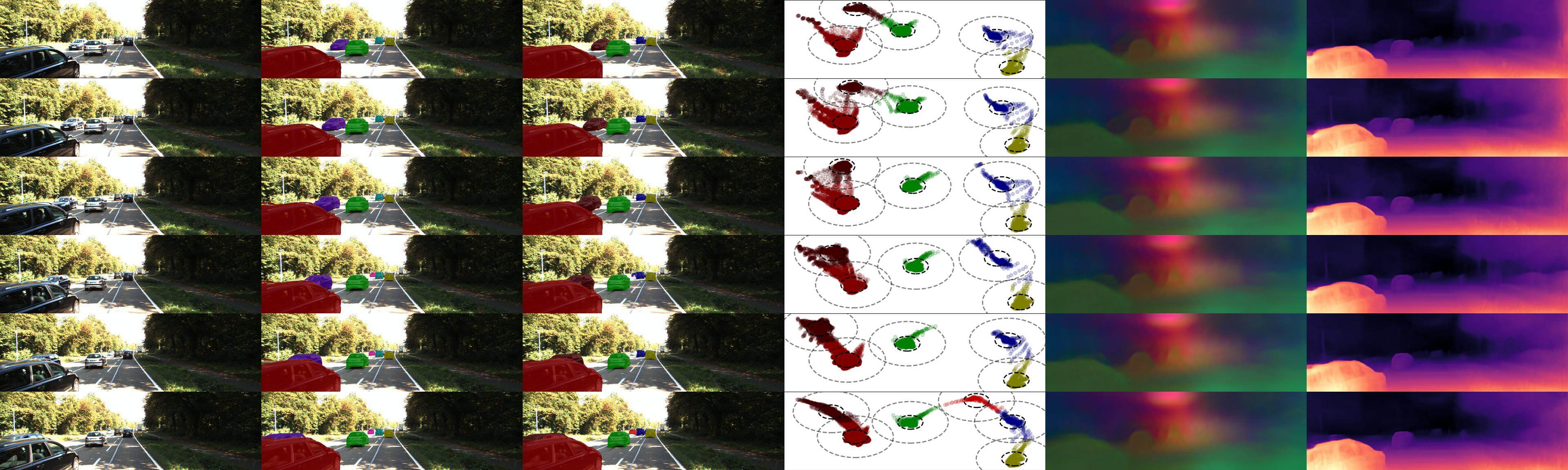}
        \caption{\textbf{Partial occlusion.} The brown car is correctly segmented even when being partially occluded by the red car, as the embedding contains past temporal context and is aware of the motion of the brown car.}
        \label{cv-fig:partial-occlusion}
    \end{subfigure}
    \begin{subfigure}[b]{\textwidth}
        \includegraphics[width=\linewidth]{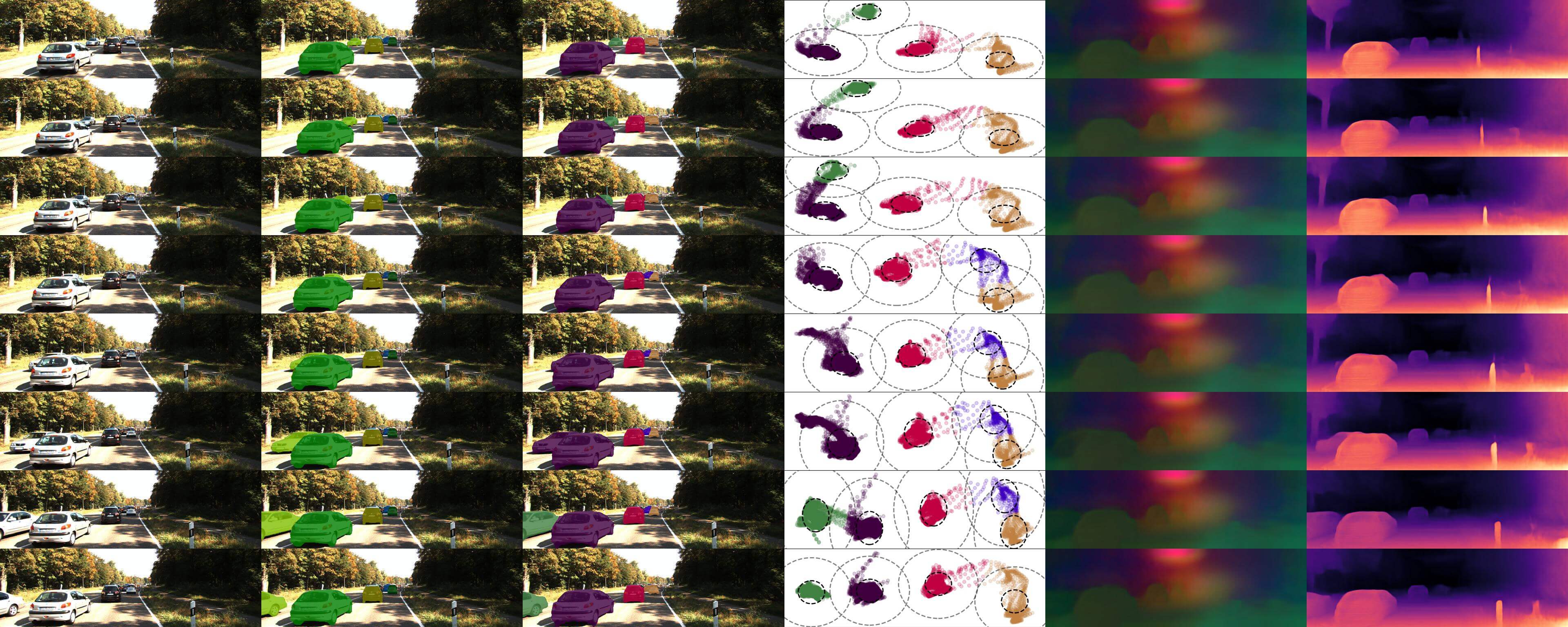}
        \caption{\textbf{Total occlusion.} The green car is correctly tracked, even though it was completely occluded by another car.}
        \label{cv-fig:total-occlusion}
    \end{subfigure}
    \caption[Predicted instance segmentation through partial and total occlusions.]{Predicted instance segmentation through partial and total occlusions. From left to right: RGB input image, ground truth instance segmentation, predicted instance segmentation, embedding visualised in 2D, embedding visualised in RGB and predicted monocular depth. Figure from \citet{hu19}.}
\end{figure}

\paragraph{Qualitative examples.} Their model can consistently segment instances over time on the following challenging scenarios: tracking through partial (\Cref{cv-fig:partial-occlusion}) and full occlusion (\Cref{cv-fig:total-occlusion}).

Just like semantic segmentation and instance segmentation got combined into panoptic segmentation, joint semantic segmentation and video instance segmentation is called \textbf{Video Panoptic Segmentation} \citep{kim2020vps}.
\clearpage
\section{Motion}
\label{cv-section:temporal-representation}
We have reviewed geometry and semantics that were set to answer the questions: `what is around me?' and `how far is it from me?'. The final concept left to examine is motion. There is so much more we can infer from a video compared to a still image. Motion naturally present in a video enables us to grasp scene dynamics or reason on occlusions for example.

Images are traditionally processed with 2D convolutional neural networks. A natural extension to model video is to use \newterm{3D convolutional neural networks} where the kernel filter operates in both space and time. Sequences (e.g. audio, text) are typically modelled with recurrent neural networks. Another popular approach to model video, which is essentially a sequence of images, is therefore to use \newterm{convolutional recurrent networks}. Finally, \newterm{transformers} have become ubiquitous in natural language processing (NLP), and are becoming increasingly popular in computer vision. Transformers were originally applied on tokens of text, then extended to tokens of image patches. Therefore a natural extension is to apply this method to videos, by modelling a sequence of temporal image patches. In this section we are going to review the three most popular methods to process video data: 3D convolutions, spatial recurrent networks, and transformers.

\subsection{3D Convolutions}
The first spatio-temporal models made their appearance in video action recognition, where the goal is to classify the action in a video. Compared to image classification, the challenge is to understand motion as extracting individual features on each frame and averaging the features would completely disregard causality and dynamics.

Classical action recognition models extended hand-crafted features from image recognition methods to videos with local spatio-temporal features. Examples include spatio-temporal interest points (STIPs) \citep{laptev03}, the temporal extension of histogram of oriented gradients (HOG3D) \citep{dalal05,klaser08}, 3D scale-invariant feature transform (SIFT-3D) \citep{scovanner07}, or histogram of optical flow (3D-HOF) \citep{laptev04}. 

\citet{karpathy14} pioneered action recognition with a data-driven learning framework, by collecting a new video action recognition dataset named Sports-1M. This dataset consists of one million Youtube videos associated with their action label (487 sport classes). It is two orders of magnitude larger than UCF101 \citep{soomro12}, the previous benchmark dataset for action recognition, which contains about 13,000 videos and 101 categories. They investigated different methods to combine temporal information with 3D convolutional neural networks.

\subsubsection{Time Information Fusion}
\begin{figure}
\centering
\includegraphics[width=\textwidth]{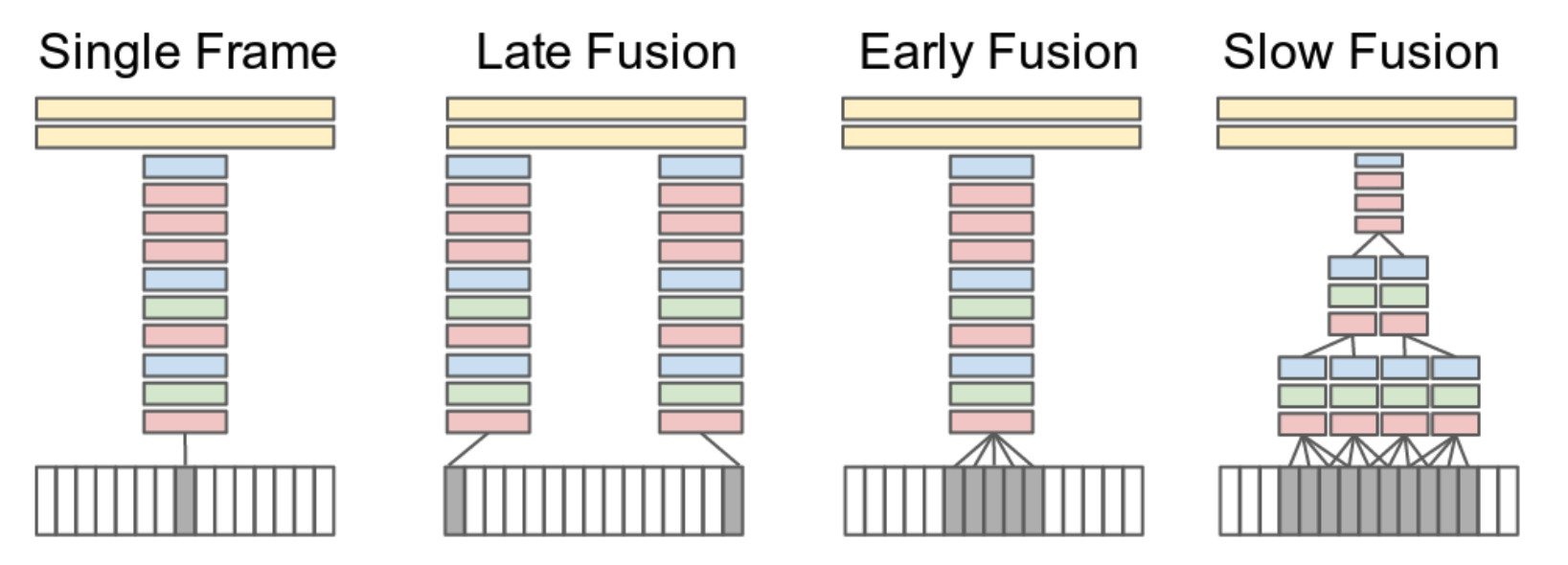}
\caption[Different temporal fusion methods for video using 3D convolutions.]{Different temporal fusion methods for video using 3D convolutions, figure from \citet{karpathy14}}
\label{cv-fig:3d-architectures}
\end{figure}

\citet{karpathy14} investigated three different approaches to fuse temporal information (Figure \ref{cv-fig:3d-architectures}).
\begin{enumerate}
    \item \textbf{Late Fusion}. This method merges the high-level 2D convolutional maps extracted from individual frames with 3D convolutions. 
	\item \textbf{Early Fusion}. This approach combines information directly at the pixel level by replacing the early 2D convolutions with 3D convolutions. 
	\item \textbf{Slow Fusion}. This method is a balanced mix between Early and Late Fusion: higher level layers get progressively access to more context in both space and time.
\end{enumerate}

Their spatio-temporal model displayed a significant performance increase compared to hand-crafted features baselines, but surprisingly, only a marginal improvement compared to a single-image baseline. This may hint that their current approach to model temporal information might not be adapted to effectively learn dynamics. Next we are going to review several methods that improved upon \citet{karpathy14}.

\subsubsection{Two-Stream Convolutional Network}
\citet{simonyan14} hypothesised that learning both spatial and temporal information from video frames was too difficult for the network, they therefore created a two-stream architecture. A spatial stream performs action recognition on still images, and a temporal stream learns to recognise action from motion with optical flow. The two streams would then be fused before the action classification layer. Their architecture is inspired by the two-stream hypothesis \citep{goodale92}: the human visual cortex contains two pathways, the ventral stream (object recognition) and dorsal stream (motion). One advantage of decoupling spatial and temporal networks is that the spatial network can benefit from a transfer of knowledge with networks trained on large scale image recognition datasets such as ImageNet \citep{imagenet09}.

\subsubsection{Fully 3D Convolutional Networks}
\citet{tran15} argued that \citet{karpathy14} could not successfully learn spatio-temporal information as their use of 3D convolution was sparse. They proposed a fully 3D convolutional network that models appearance and motion simultaneously throughout the whole video. One key characteristic of their model is that it creates a spatio-temporal hierarchical representation of data.

Due to the lack of labeled data and the high-dimensionality of the parametrisation of 3D convolutional networks, most action recognition networks remained shallow (up to 8 layers). More recently, \citet{carreira17} showed that very deep classification neural networks trained on ImageNet (Inception \citep{ioffe15}, VGG-16 \citep{simonyan15}, ResNet \citep{he16}) could be inflated to 3D and that the pretrained weights were a valuable initialisation. The inflation from 2D to 3D kernels transforms a $k\times k$ kernel into a $t\times k\times k$ kernel and initialises the weights of the new 3D kernel by dividing the weights from the original 3D kernel by $t$, such that the activation of the 3D convolutional network applied to the same ImageNet image repeated $t$ times remains identical. 

\subsubsection{Factorised 3D Convolutions}
\citet{tran18} demonstrated that factorising 3D convolutions into separate spatial and temporal convolutions facilitated optimisation. They decomposed a 3D convolution as the succession of a 2D convolution in space and a 1D convolution in time, which they named (2+1)D convolution. This approach effectively doubles the number of non-linearities of the network, allowing it to learn more complex functions. Further, they found that (2+1)D convolution were easier to train than 3D convolutions where appearance and dynamics are entangled. 

\subsection{Spatial Recurrent Neural Network}

Inspired by temporal models from speech recognition and natural language processing, \citet{ng15} introduced a video architecture with recurrent units. Their model first extracted convolutional features from each video frame with a CNN and then used a recurrent module formed of Long-Short Term Memory units (LSTM) \citep{hochreiter97} on the flattened features.

However, as all spatial information is decimated when flattening the image feature descriptors, learning precise spatial motion is difficult. Further, this approach is not invariant to the dimension of the inputs, and the number of network parameters scales linearly with the input size. Another way to represent temporal information in video is with the Spatial Recurrent Neural Network (Spatial RNN).

A recurrent network that operates on convolutional feature maps implies a significant increase in the number of parameters and memory. Spatial RNNs introduce sparsity and locality in the RNN units by replacing the fully-connected operations with convolutions. As a consequence, the number of parameters and memory usage is greatly reduced.

Let us consider a convolutional feature map of dimension $C\times H\times W$, and denote by $H$ the number of hidden units. Applying a recurrent unit directly to this feature map would require input-to-hidden parameters of size $\mathcal{O}(H\times C\times H\times W)$, which could potentially be really large. Furthermore, fully-connected layers do not take advantage of the properties of convolutional maps. The latter are composed of patterns with strong local correlation that are repeated at different spatial locations. In addition, videos have smooth temporal variation over time, therefore the changes only occur locally which is perfectly adapted for convolutions (that model local spatial neighbourhood). 

A spatial RNN replaces the fully connected layers with convolutions and results in hidden activations that preserve the spatial topology. The number of parameters is reduced to $\mathcal{O}(H\times C\times k\times k)$ with $k$ the kernel size, resulting in a subsequent decrease in parameters if $k\times k$ is smaller than $H\times W$.

\subsection{Transformers}
Transformers are self-attention based architectures and have become the de-facto standard for natural language processing \citep{vaswani17}. The usual practice is to pre-train on a large corpus of text data and fine-tune on a given task \citep{devlin-bert19}. Due to the computational efficiency and scalability of transformers, models of unprecedented size (more than 100B of parameters) have been trained \citep{brown20,lepikhin21}. What is impressive is that, with models and datasets both increasing in size, performance does not seem to saturate. 

Text is a sequence of words. Similarly, video can be seen as a sequence of images. Given that there is a wealth of video data available, would it be possible to transfer the success of transformers in natural language processing to video understanding? And if not, what are the current challenges that prevent this from happening? In this section, we will review how the transformer works, then we will highlight the challenges of scaling this architecture to videos, and we will examine recent approaches of transformers applied to images and videos. 

\subsubsection{Self-Attention}
Transformers operate on sequences and rely on self-attention to learn a representation. Self-attention indicates how each element in the sequence relates to the others. 
Let us consider a sequence $(\vx_1, \dots, \vx_n)$ with $\vx_i \in \mathbb{R}^p$. A self-attention module outputs a sequence of features $(\vz_1,\dots, \vz_n)$, with $\vz_i \in \mathbb{R}^d$, such that each feature map $\vz_i$ `attends' to all the inputs $(\vx_1, \dots, \vx_n)$. 

More specifically, the module contains 3 matrices: a query matrix $\mW^q \in \mathbb{R}^{p\times d_k}$, a key matrix $\mW^k \in \mathbb{R}^{p\times d_k}$, and a value matrix $\mW^v \in \mathbb{R}^{p\times d}$, with $d_k$ the dimension of the query and key vectors \citep{alammar18}. Let us denote by $\mX \in \mathbb{R}^{n\times p}$ the inputs. The self-attention module computes the queries $\mQ=\mX\mW^q \in \mathbb{R}^{n\times d_k}$, the keys $\mK=\mX\mW^k \in \mathbb{R}^{n\times d_k}$, and the values $\mV=\mX\mW^v \in \mathbb{R}^{n\times d}$. Each output vector $\vz_i$ is a linear combination of the row vectors of $\mV$. The linear combination is obtained through self-attention using the queries $\mQ$ and keys $\mK$. The outputs $\mZ \in \mathbb{R}^{n\times d}$ is obtained with:



\begin{equation}
	\mZ = \operatorname{softmax}\left(\frac{\mQ\mK^T}{\sqrt{d_k}}\right)\mV
\end{equation}
The attention module can be extended to multiple heads with $K$ different matrices $\mW^q, \mW^k, \mW^v$ initialised differently to yield $(\mZ_1, \dots, \mZ_K)$ outputs. These outputs are then concatenated and multiplied by a weight matrix $\mW^0 \in \R^{Kd\times d_0}$ to combine information from all the heads, with $d_0$ the dimension of the resulting feature after the self-attention module.

Naively applying self-attention to an image would result in computing an attention matrix that is quadratic in the number of pixels. If we denote an image by $\vo \in \R^{3\times H\times W}$, then we could construct a sequence $(\vx_1,\dots,\vx_n)$ with $\vx_i \in \R^3$ the channels of a pixel and $n=H\times W$ the number of pixels. For a video the attention matrix would be even larger with $n=T \times H \times W$ the number of video pixels assuming there are $T$ image frames. For high resolution videos, calculating such an attention matrix would be prohibitive. As an example, if the image resolution is $H \times W = 600\times 960$ and there are $T=10$ video frames, then the attention matrix would contain $n^2=(10\times 600 \times 960)^2 \approx 10^{13}$ elements. We will now see the techniques used by recent transformer architectures to scale to images and videos.

\subsubsection{Image Transformers}
\begin{figure}
\centering
\includegraphics[width=\textwidth]{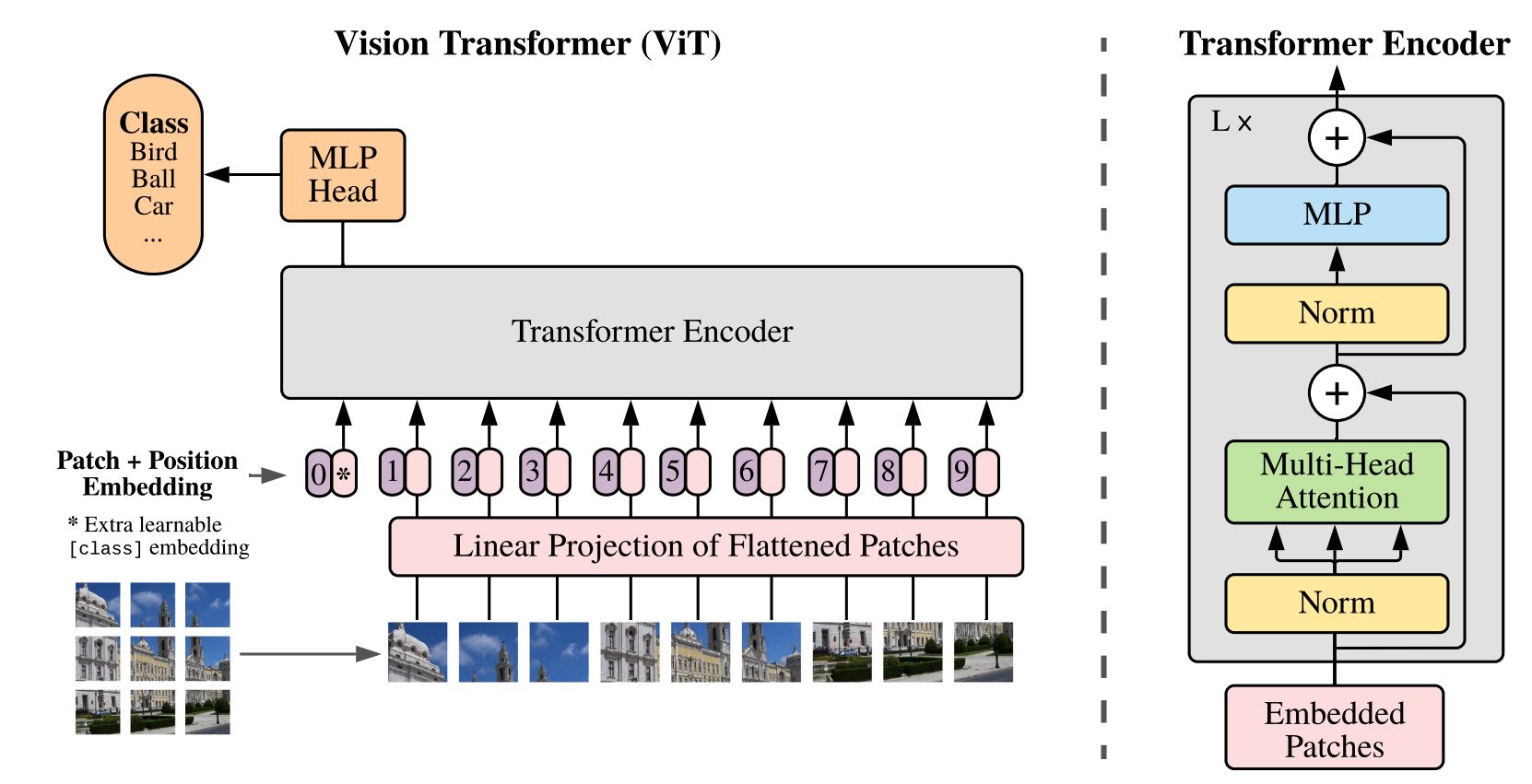}
\caption[Vision Transformer architecture.]{Vision Transformer architecture, figure from \citet{dovositskiy-vit21}}
\label{cv-fig:vision-transformers}
\end{figure}

\citet{dovositskiy-vit21} scaled transformers to mid-resolution images on ImageNet with an architecture called Vision Transformers. Their core idea was to split an image of size $H \times W$ in $p \times p$ small image patches. Each image patch would be flattened and linearly embedded into a 1D vector. Those vectors along with a positional encoding would be the inputs of a transformer architecture. \Cref{cv-fig:vision-transformers} depicts the Vision Transformer. On ImageNet, the image resolution is roughly $H\times W = 224\times 224$, and they have selected a patch size of $p \times p = 16 \times 16$. They would thus construct a sequence of size $HW/p^2 = 196$ as opposed to the naive sequence of $HW = 50,176$ pixels. An image patch is treated like tokens (words) in natural language processing. Except for the construction of the tokens, there was no modification on the transformer architectures from NLP. Vision Transformers could therefore make use of the network optimisations and progress made in the entirely different field of NLP. 

When trained on medium-scale datasets (e.g. ImageNet which contains around 1.3M images), the performance of Vision Transformers was a few percentages below convolutional networks of the same size. This was to be expected because vision transformers lack the inductive biases that convolutional layers have, such as equivariance and locality. However, when pre-trained on larger datasets (14M-300M data points, e.g. Imagenet 21k \citep{imagenet09} or JFT \citep{sun17}) and finetuned, the picture changed and vision transformers reached or beat the state-of-the-art on multiple image recognition benchmarks. These results indicated that large-scale training trumps inductive bias in image recognition, and that the paradigm of pre-training on large datasets and finetuning on the task at hand from NLP was transferable to computer vision.

\subsubsection{Video Transformers}
\begin{figure}
\centering
\includegraphics[width=\textwidth]{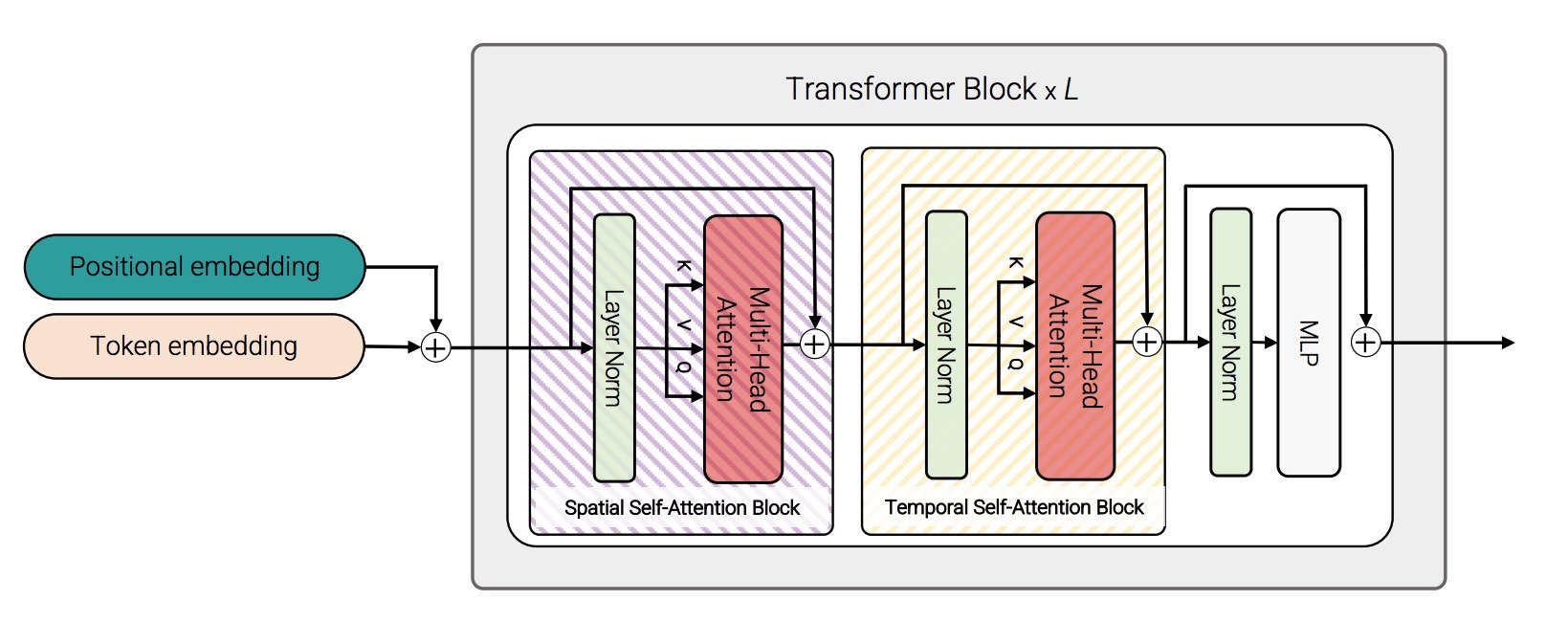}
\caption[Factorised self-attention for video inputs.]{Factorised self-attention for video inputs, figure from \citet{arnab-vivit21}}
\label{cv-fig:factorised-transformer}
\end{figure}

Similarly to the Vision Transformer, the Video Vision Transformer \citep{arnab-vivit21} created patches from video frames, linearly embedded them into tokens as inputs to self-attention modules for video action recognition. However, by adding time as an additional dimension, the size of the sequence rapidly grew. If we take the previous image resolution of $H\times W = 224\times 224$, patch size $p \times p = 16 \times 16$, and $T=10$ video frames, then the sequence size is equal to $THW/p^2 = 1960$. This kind of dimensionality is not easy to handle in standard GPU hardware. \citet{bartasius21,arnab-vivit21} proposed to factorise the space and time attentions, by first attending to tokens spatially, then in time as shown in \Cref{cv-fig:factorised-transformer}.

More precisely, each token would first attend all the other spatial tokens in the same time index. Then, all tokens would attend to all other time tokens at the same spatial index. If we denote by $n_s=HW/p^2$ the number of spatial tokens, the standard spatio-temporal attention has a complexity of $\mathcal{O}(T^2n_s^2)$. Factorising self-attention in space and time reduces complexity to $\mathcal{O}(T^2 +n_s^2)$. Although efficient, the downside of this approach is that temporal attention only occurs within the same spatial position. This means objects that move fast and span several spatial positions would not be properly modelled. 

\clearpage
\section{Summary}
We reviewed world models in the context of reinforcement learning. In particular, we emphasised how our problem setting differed from reinforcement learning. We do not assume access to a reward function, or any interaction with an online environment. Our objective is to train world models on offline video observations and expert demonstrations of urban driving scenes. 

Then, we studied different methods for video prediction. Differently from video prediction, we want to not only learn a model of the world, but also an optimal policy for a downstream task (controlling an autonomous vehicle). We have seen the difficulty of learning latent dynamics in video prediction due to the complexity of natural scenes and the stochasticity of the future. Finally, we reviewed the concepts of geometry and semantics through the tasks of dense pixel prediction with depth estimation and panoptic segmentation. We also explored different neural network architectures to model motion from video with 3D convolutions, spatial recurrent units, and transformers. 

In the next chapter (\Cref{chapter:generative-models}), we will  formally define the probabilistic generative framework of this thesis. In this framework, the objective is to infer a world model and the expert policy from video observations and expert demonstrations. In the following chapters (\Cref{chapter:video-scene-understanding,chapter:instance-prediction,chapter:imitation-learning}), we will show how we can leverage computer vision to scale world modelling to complex urban driving scenes.

\chapter{Probabilistic World Models}
\label{chapter:generative-models}
\graphicspath{{Chapter3/Figures/}}

In this chapter, we formally define the probabilistic generative model in order to jointly learn a world model, and an optimal policy from video observations and expert actions. Probabilistic generative modelling is a general framework that links causes (latent states) to consequences (observed data, or sensations). A formulation of such a framework was introduced by \citet{friston10} as the free-energy principle. Through the lens of neuroscience, information theory, and optimal control theory, the free-energy framework states that the same quantity is optimised over the course of learning: surprise, or prediction error. In other words, the underlying principle behind learning is the minimisation of prediction error.

The free-energy principle states that any adaptive system (e.g. a biological agent) resist a natural tendency to disorder \citep{ashby1947principles,Prigogine1985}. They reduce surprises (as defined in the information theoretical sense: the negative log-probability of an outcome) over time to ensure their states remain within physiological bounds \citep{friston10}. 

A biological agent cannot estimate surprises, as it would involve knowing all the hidden states of the world causing sensory inputs.
This is where the free-energy comes in: the free-energy is an upper-bound on the surprise, and therefore minimising it implicitly implies minimising the surprise. The free-energy can be estimated because it depends on quantity the biological agent has access to, namely its sensory states and a recognition density (a probabilistic representation of what caused a particular sensation). The free-energy can be interpreted as the difference between an organism's prediction of its sensory inputs (embedded in its model of the world), and its actual sensations \citep{friston12}.

Free-energy was originally introduced in statistical physics in order to ease the surprisal estimation to a tractable optimisation problem \citep{feynman72} by introducing a variational distribution. It has now also been used in many machine learning and statistics problems \citep{Hinton93keepingneural,MacKay94freeenergy,radford98}
\section{Free-Energy}
We now formally define the free-energy. Let $\rvo$ be a vector of observed data or sensations, and $\rvs$ the latent state that generated those sensations. The objective of the free-energy principle is to minimise the surprise, defined as the negative log-likelihood of the observed data $-\log p(\rvo)$. In machine learning, this is equivalent to maximising the log evidence $\log p(\rvo)$. However, because we do not observe the latent state $\rvs$, we cannot evaluate the surprise as it would require integrating over all possible latent states:

\begin{equation}
    -\log p(\rvo) = -\int_{\rvs} \log p(\rvs, \rvo) \diff \rvs
\end{equation}

To overcome this problem, we introduce a variational distribution $q(\rvs|\rvo)$. The \newterm{Kullback-Leibler} (KL) divergence of the variational distribution $q(\rvs|\rvo)$ from the posterior distribution $p(\rvs|\rvo)$ is:

\begin{align}
    \KL(q(\rvs|\rvo)~||~p(\rvs|\rvo)) &= \E_{q(\rvs|\rvo)} \left[ \log \frac{q(\rvs|\rvo)}{p(\rvs|\rvo)} \right] \notag \\
    &= \E_{q(\rvs|\rvo)} \left[ \log \frac{q(\rvs|\rvo)p(\rvo)}{p(\rvs)p(\rvo|\rvs)} \right] \label{framework-eq:first-bayes} \\
    &= \log p(\rvo) - \E_{q(\rvs|\rvo)}[\log p(\rvo|\rvs)] + \KL(q(\rvs|\rvo)~||~p(\rvs)) \label{framework-eq:split}
\end{align}
where \Cref{framework-eq:first-bayes} follows from Bayes rules, and \Cref{framework-eq:split} is obtained by splitting the logarithms.

By definition of the KL divergence, $\KL(q(\rvs|\rvo)~||~p(\rvs|\rvo)) \geq 0$, therefore we obtain the following inequality:

\begin{equation}
    -\log p(\rvo) \leq  \underbrace{-\E_{q(\rvs|\rvo)}[\log p(\rvo|\rvs)] + \KL(q(\rvs|\rvo)~||~p(\rvs))}_{\text{Free-energy}}
    \label{framework-eq:free-energy-inequality}
\end{equation}

The free-energy $F$ is defined as:
\begin{equation}
    F \eqdef \underbrace{-\E_{q(\rvs|\rvo)}[\log p(\rvo|\rvs)]}_{\text{accuracy}} + \underbrace{\KL(q(\rvs|\rvo)~||~p(\rvs))}_{\text{complexity}}
\end{equation}
It contains an accuracy term that specifies how good the predicted sensory inputs are given the model of the world, and a complexity term that controls the variational distribution (also named recognition density).

Additionally, the inequality in \Cref{framework-eq:free-energy-inequality} becomes an equality when\\ $\KL(q(\rvs|\rvo)~||~p(\rvs|\rvo)) = 0$, that is to say when $q(\rvs|\rvo)=p(\rvs|\rvo)$. In other words, the surprisal is exactly equal to the free-energy when the variational distribution is equal to the posterior distribution.

\paragraph{Connection with Bayesian inference.} The free-energy principle can be linked to Bayesian inference. Minimising the free-energy implies minimising the KL divergence between the prior distribution (the prior beliefs of the agent) and the variational recognition density (that has access to the actual observed data or sensations). This corresponds to Bayesian inference on unknown latent states of the world causing sensory data \citep{Knill2004TheBB}. 

Models that minimise the free-energy provide an accurate explanation of the data under complexity constraints. This can be understood as Bayesian model selection, selecting the model that best explains the data under complexity constraints. The complexity is the divergence between the variational distribution and the prior beliefs about hidden states (or the effective degrees of freedom necessary to explain the data). This interpretation can be linked to the fact that biological agents build parsimonious models to explain their world \citep{dayan95}.

\paragraph{Generative and inference model} The generative model $p$ models the causal relationship between latent states $\rvs$ and observed data $\rvo$:

\begin{equation}
    p(\rvs, \rvo) = p(\rvs) p(\rvo|\rvs)
\end{equation}
Let us denote the parameters of the generative model by $\theta$. 

The inference model $q(\rvs|\rvo)$ updates the beliefs about the model of the world with observations. It is parametrised with parameters $\phi$. Therefore, minimising the free-energy consists of finding the parameters of the generative model $\theta^{\star}$ and the parameters of the inference model $\phi^{\star}$ such that:

\begin{equation}
    (\theta^{\star},\phi^{\star}) = \argmin_{(\theta,\phi)} F(\theta, \phi)
\end{equation}

In this thesis, the generative and inference models are deep neural networks with weights $\theta$ and $\phi$. The weights of the neural networks are updated with gradient descent on the free-energy, which is the loss function.
\clearpage
\section{Active Inference}
We have assumed in the previous section that the agent was a passive observer of the world that could not interact with the environment (as in video prediction). Its sole goal was to recover a model of the world to understand the dynamics of the world. In this section, we are going to describe the free-energy framework in its most general form where the agent can execute actions and interact with its environment. This framework is called \newterm{active inference} \citep{friston_2013}.

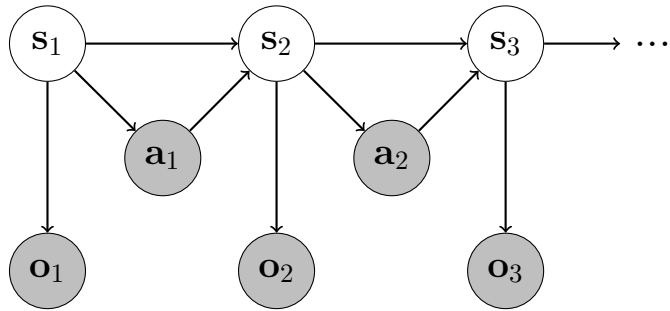
\begin{figure}[h] 
\centering 
\begin{tikzpicture} 
    \node[draw=black,circle,inner sep=0pt,minimum size =10mm] (s1) {\large$\rvs_1$};
    \node[draw=black,circle,inner sep=0pt,right=2cm of s1,minimum size =10mm] (s2) {\large$\rvs_2$};
    \node[draw=black,circle,inner sep=0pt,right=2cm of s2,minimum size =10mm] (s3) {\large$\rvs_3$};
    
    \node[draw=black,circle,below=2cm of s1,inner sep=0pt,minimum size =10mm,fill=lightgray] (o1) {\large$\rvo_1$};
    \node[draw=black,circle,below=2cm of s2,inner sep=0pt,minimum size =10mm,fill=lightgray] (o2) {\large$\rvo_2$};
    \node[draw=black,circle,below=2cm of s3,inner sep=0pt,minimum size =10mm,fill=lightgray] (o3) {\large$\rvo_3$}; 
    
    \node[draw=black,circle,below right=0.8cm and 0.8cm of s1,inner sep=0pt,minimum size =10mm,fill=lightgray] (a1) {\large$\rva_1$};
    \node[draw=black,circle,below right=0.8cm and 0.8cm of s2,inner sep=0pt,minimum size =10mm,fill=lightgray] (a2) {\large$\rva_2$};  
    \node[draw=white,circle,right=1cm of s3] (s4){\Large $...$};
    \path[->,thick]
        (s1) edge (o1)
        (s2) edge (o2)
        (s3) edge (o3)
        (s1) edge (s2)
        (s2) edge (s3)
        (s1) edge (a1)
        (a1) edge (s2)
        (s2) edge (a2)
        (a2) edge (s3)
        (s3) edge (s4);
\end{tikzpicture}
\caption[Graphical model representing the conditional dependence between states of the generative model in active inference.]{Graphical model representing the conditional dependence between states of the generative model in active inference. Observed states are in gray.} \label{framework-fig:probabilistic-model} 
\end{figure}

More formally, given a sequence of observations and actions $(\rvo_{1:T}, \rva_{1:T})$, we would like to infer a policy that exhibits intelligent behaviour. We assume there are latent states $\rvs_{1:T}$ that governs both the observed data and the actions taken by the agent. A graphical model illustrating this framework is depicted in \Cref{framework-fig:probabilistic-model}.

This framework is extremely general and makes very little assumptions. We do not need to:
\begin{enumerate}
    \item know the true dynamics of the environment.
    \item have access to a reward function.
    \item interact with the environment (in order to e.g. gather more training data, explore the environment, or use black-box optimisation such as evolution algorithms \citep{kai18}).
\end{enumerate}

Let us now define the generative model $p$ (parametrised by $\theta$) and the inference model $q$ (parametrised by $\phi$) in order to have an expression of the free-energy $F(\theta, \phi)$ we can minimise. Given the graphical model, the full distribution factorises as:

\begin{equation}
    \label{framework-eq:generative-model}
    \begin{split}
    p(\rvo_{1:T}, \rva_{1:T}, \rvs_{1:T}) =\prod_{t=1}^T p(\rvs_t|\rvs_{t-1},\rva_{t-1})p(\rvo_t|\rvs_t)p(\rva_t|\rvs_t)
    \end{split}
\end{equation}

with $p(\rvs_1)$ the initial distribution.

We introduce a variational distribution $q_S$ defined and factorised as follows:

\begin{equation}
    \label{framework-eq:inference-model}
    q_{S} \triangleq q(\rvs_{1:T}|\rvo_{1:T}, \rva_{1:T}) = \prod_{t=1}^T q(\rvs_{t}|\rvo_{\le t},\rva_{<t})
\end{equation}

We can then compute a lower bound on the log evidence (or upper bound on the surprisal -- both formulations are equivalent and only differ by a minus sign):
\begin{align*}
    \log p(\rvo_{1:T},\rva_{1:T})
    \geq&~\sum_{t=1}^T \E_{\rvs_{1:t} \sim q(\rvs_{1:t}| \rvo_{\leq}, \rva_{<t})} \Big[ \log p(\rvo_t|\rvs_t) + \log p(\rva_t|\rvs_t) \Big]\\
    & - \sum_{t=1}^T \E_{\rvs_{1:t-1} \sim q(\rvs_{1:t-1}|\rvo_{<t},\rva_{<t-1})} \Big[ \KL \big( q(\rvs_t|\rvo_{\le t}, \rva_{<t})~||~p(\rvs_t|\rvs_{t-1})\big) \Big]
\end{align*}

\begin{proof}
The Kullback-Leibler divergence of the variational distribution $q_S$ from the posterior distribution $p(\rvs_{1:T}|\rvo_{1:T},\rva_{1:T})$ is:

\begin{align}
   &\KL\big(q(\rvs_{1:T}|\rvo_{1:T},\rva_{1:T}) ~||~ p(\rvs_{1:T}|\rvo_{1:T},\rva_{1:T}) \big) \notag\\
    =& \E_{\rvs_{1:T} \sim q_{S}} \left[ \log \frac{q(\rvs_{1:T}|\rvo_{1:T},\rva_{1:T})}{p(\rvs_{1:T}|\rvo_{1:T},\rva_{1:T})}\right] \notag\\
    =& \E_{\rvs_{1:T} \sim q_{S}} \left[ \log \frac{q(\rvs_{1:T}|\rvo_{1:T},\rva_{1:T})p(\rvo_{1:T},\rva_{1:T})}{p(\rvs_{1:T}) p(\rvo_{1:T},\rva_{1:T}|\rvs_{1:T})} \right] \notag\\
    =& \log p(\rvo_{1:T},\rva_{1:T}) - \E_{\rvs_{1:T} \sim q_{S}} \big[ \log p(\rvo_{1:T},\rva_{1:T}|\rvs_{1:T}) \big] \notag \\
    &+ \KL \big(q(\rvs_{1:T}|\rvo_{1:T},\rva_{1:T})~||~p(\rvs_{1:T})\big) \notag
\end{align}

Since $\KL\big(q(\rvs_{1:T}|\rvo_{1:T},\rva_{1:T}) ~||~ p(\rvs_{1:T}|\rvo_{1:T},\rva_{1:T}) \big) \geq 0$, we obtain the following evidence lower bound:
\begin{align}
    \log p(\rvo_{1:T},\rva_{1:T}) \geq &\E_{ \rvs_{1:T} \sim q_{S}} \big[ \log p(\rvo_{1:T},\rva_{1:T}|\rvs_{1:T}) \big] \notag \\
    &- \KL \big(q(\rvs_{1:T}|\rvo_{1:T},\rva_{1:T})~||~p(\rvs_{1:T})\big) \label{framework-eq:lower-bound}
\end{align}

Let us now calculate the two terms of this lower bound separately. On the one hand:
\begin{align}
   \E_{\rvs_{1:T} \sim q_{S}} \big[ \log p(\rvo_{1:T},\rva_{1:T}|\rvs_{1:T}) \big]
    &=~ \E_{\rvs_{1:T} \sim q_{S}} \left[ \log \prod_{t=1}^T p(\rvo_t|\rvs_t) p(\rva_t|\rvs_t) \right] \label{framework-eq:expectation1}\\
    &=~\sum_{t=1}^T \E_{\rvs_{1:t} \sim q( \rvs_{1:t}| o_{\le t}, a_{<t})} \big[ \log p(\rvo_t|\rvs_t) + \log p(\rva_t|\rvs_t) \big] \label{framework-eq:last-expectation}
\end{align}
where \Cref{framework-eq:expectation1} follows from the factorisation defined in \Cref{framework-eq:generative-model}, and \Cref{framework-eq:last-expectation} was obtained by integrating over remaining latent variables $\rvs_{t+1:T}$.

On the other hand:
\begin{align}
    &\KL \big(q(\rvs_{1:T}|\rvo_{1:T},\rva_{1:T})~||~p(\rvs_{1:T})\big) \notag\\
    =&~\E_{\rvs_{1:T} \sim q_{S}} \left[\log \frac{q(\rvs_{1:T}| \rvo_{1:T}, \rva_{1:T})}{p(\rvs_{1:T})} \right] \notag\\
    =&~\int_{\rvs_{1:T}} q( \rvs_{1:T}| \rvo_{1:T}, \rva_{1:T}) \log \frac{q(\rvs_{1:T}| \rvo_{1:T}, \rva_{1:T})}{p(\rvs_{1:T})} \diff \rvs_{1:T} \notag\\
    =&~\int_{\rvs_{1:T}} q(\rvs_{1:T}| \rvo_{1:T}, \rva_{1:T}) \log\left[ \prod_{t=1}^T \frac{q(\rvs_t| \rvo_{\le t}, \rva_{<t})}{p(\rvs_t | \rvs_{t-1})} \right] \diff \rvs_{1:T} \label{framework-eq:kl-step-1}
\end{align}
where \Cref{framework-eq:kl-step-1} results from the factorisations defined in \Cref{framework-eq:generative-model} and \Cref{framework-eq:inference-model}

Thus:
\begin{align}
    &\KL \big(q(\rvs_{1:T}|\rvo_{1:T},\rva_{1:T})~||~p(\rvs_{1:T})\big) \notag\\
    =&~\int_{\rvs_{1:T}} \left(\prod_{t=1}^T q(\rvs_{t}|\rvo_{\le t}, \rva_{<t}) \right)\left(  \sum_{t=1}^T \log \frac{q(\rvs_t| \rvo_{\le t}, \rva_{<t})}{p(\rvs_t | \rvs_{t-1})} \right)  \diff \rvs_{1:T}\notag \\
    =&~\int_{\rvs_{1:T}} \left(\prod_{t=1}^T q(\rvs_{t}| \rvo_{\le t}, \rva_{<t}) \right)\left( \log \frac{q(\rvs_1| \rvo_1)}{p(\rvs_1)} + \sum_{t=2}^T \log \frac{q(\rvs_t| \rvo_{\le t}, \rva_{<t})}{p(\rvs_t | \rvs_{t-1})} \right)  \diff \rvs_{1:T} \notag \\
    =&~\E_{\rvs_1 \sim q(\rvs_1|\rvo_1)} \left[ \log \frac{q(\rvs_1|\rvo_1)}{p(\rvs_1)}\right] \notag \\
    &+~\int_{\rvs_{1:T}} \left( \prod_{t=1}^T q(\rvs_{t}| \rvo_{\le t}, \rva_{<t}) \right) \left(\sum_{t=2}^T \log \frac{q(\rvs_t| \rvo_{\le t}, \rva_{<t})}{p(\rvs_t | \rvs_{t-1})} \right)  \diff \rvs_{1:T} \label{framework-eq:kl-first-term} \\
    =&~\KL \big( q(\rvs_1|\rvo_1)~||~p(\rvs_1)\big) \notag \\
    &+~\int_{\rvs_{1:T}} \left( \prod_{t=1}^T q(\rvs_{t}| \rvo_{\le t}, \rva_{<t}) \right) \left(\log \frac{q(\rvs_2| \rvo_{1:2}, \rva_1)}{p(\rvs_2 | \rvs_1)} +\sum_{t=3}^T \log \frac{q(\rvs_t| \rvo_{\le t}, \rva_{<t})}{p(\rvs_t | \rvs_{t-1})} \right)  \diff \rvs_{1:T} \notag\\
    =&~\KL \big( q(\rvs_1|\rvo_1)~||~p(\rvs_1)\big) + ~\E_{\rvs_1 \sim q(\rvs_1|\rvo_1)} \Big[ \KL \big( q(\rvs_2|\rvo_{1:2}, \rva_1)~||~p(\rvs_2|\rvs_1)\big) \Big]\notag \\
    &+~\int_{\rvs_{1:T}} \left( \prod_{t=1}^T q(\rvs_{t}| \rvo_{\le t}, \rva_{<t}) \right) \left(\sum_{t=3}^T \log \frac{q(\rvs_t| \rvo_{\le t}, \rva_{<t})}{p(\rvs_t | \rvs_{t-1})} \right)  \diff \rvs_{1:T} \label{framework-eq:kl-second-term}
\end{align}
where \Cref{framework-eq:kl-first-term} and \Cref{framework-eq:kl-second-term} were obtained by splitting the integral in two and integrating over remaining latent variables.
By recursively applying this process on the sum of logarithms indexed by $t$, we get:
\begin{align}
    &\KL \big(q(\rvs_{1:T}|\rvo_{1:T},\rva_{1:T})~||~p(\rvs_{1:T})\big) \notag\\
    =&~ \sum_{t=1}^T \E_{\rvs_{1:t-1} \sim q(\rvs_{1:t-1}|\rvo_{<t},\rva_{<t-1})} \Big[ \KL \big( q(\rvs_t|\rvo_{\le t}, \rva_{<t})~||~p(\rvs_t|\rvs_{t-1})\big) \Big] \label{framework-eq:last-kl}
\end{align}

Finally, we inject \Cref{framework-eq:last-expectation} and \Cref{framework-eq:last-kl} in \Cref{framework-eq:lower-bound} to obtain the desired lower bound:

\begin{align*}
    \log p(\rvo_{1:T},\rva_{1:T})
    \geq&~\sum_{t=1}^T \E_{\rvs_{1:t} \sim q(\rvs_{1:t}| \rvo_{\leq}, \rva_{<t})} \Big[ \log p(\rvo_t|\rvs_t) + \log p(\rva_t|\rvs_t) \Big]\\
    & - \sum_{t=1}^T \E_{\rvs_{1:t-1} \sim q(\rvs_{1:t-1}|\rvo_{<t},\rva_{<t-1})} \Big[ \KL \big( q(\rvs_t|\rvo_{\le t}, \rva_{<t})~||~p(\rvs_t|\rvs_{t-1})\big) \Big] \quad\qedhere
\end{align*}

\end{proof}

Therefore, we have found the formulation of the free-energy in the active inference framework. It is equal to the quantity above multiplied by -1 to switch from lower bound of the log evidence, to the upper bound of the surprisal.

\begin{align*}
    F(\theta,\phi)
    =&~-\sum_{t=1}^T \E_{\rvs_{1:t} \sim q(\rvs_{1:t}| \rvo_{\leq}, \rva_{<t})} \Big[ \log p(\rvo_t|\rvs_t) + \log p(\rva_t|\rvs_t) \Big]\\
    & + \sum_{t=1}^T \E_{\rvs_{1:t-1} \sim q(\rvs_{1:t-1}|\rvo_{<t},\rva_{<t-1})} \Big[ \KL \big( q(\rvs_t|\rvo_{\le t}, \rva_{<t})~||~p(\rvs_t|\rvs_{t-1})\big) \Big]
\end{align*}

In order to have an example of a concrete formulation of this loss function, let us now specify the distributions of the generative model:
\begin{align}
    \label{framework-eq:full-model}
        \rvs_1 &\sim \mathcal{N}(\rvzero, \mI)\\
        \rvs_{t+1} |\rvs_t, \rva_t& \sim \mathcal{N}(\mu_{\theta}(\rvs_{t},\rva_t), \sigma_{\theta}(\rvs_{t},\rva_t)\mI) \\
        \rvo_t | \rvs_t &\sim \mathcal{N}(g_{\theta}(\rvs_t), \mI) \\
        \rva_t | \rvs_t &\sim \mathcal{N}(\pi_{\theta}(\rvs_t), \mI)
\end{align}
where the notation $\sigma_{\theta}(\rvs_{t},\rva_t)\mI=\mathrm{diag}(\sigma_{\theta}(\rvs_{t},\rva_t))$.

Because both the observation $\rvo_t$ and action $\rva_t$ are Gaussian distributions, the resulting loss from the log probability term is the mean squared error. We additionally model the variational distribution as $q(\rvs_t|\rvo_{\le t}, \rva_{<t}) \sim \mathcal{N}(\mu_{\phi}(\rvo_{\le t}, \rva_{<t}), \sigma_{\phi}(\rvo_{\le t}, \rva_{<t})\mI)$ so that the KL divergence $\KL \big( q(\rvs_t|\rvo_{\le t}, \rva_{<t})~||~p(\rvs_t|\rvs_{t-1})\big)$ has a closed-form expression.

To the best of our knowledge, this is the first time the free-energy principle is applied with deep neural networks to infer a world model and a policy from offline observed data and actions.

\paragraph{How to Make Variational Inference Work in Practice}

One known issue of variational inference is that the parameters solution found has a bias. The solution is biased away from the maximum of the likelihood towards regions where the bound between the free-energy and the likelihood is the tightest \citep{turnersahani2011a}. This tendency leads to severe underfitting, as the solution found correspond to a region where the free-energy is tight with the likelihood, but which may be far from the maximum. An example of such underfitting is "mode collapse" when the posterior and prior distributions match exactly, but do not contain any information about the underlying data \citep{Lucas2019UnderstandingPC,he-collapse19}.



A pratical method to counter this issue is to reweight the KL loss with a coefficient $\beta$ to balance the contribution of the reconstruction loss and the KL loss:

\begin{align*}
    \log p(\rvo_{1:T},\rva_{1:T})
    \geq&~\sum_{t=1}^T \E_{\rvs_{1:t} \sim q(\rvs_{1:t}| \rvo_{\leq}, \rva_{<t})} \Big[ \log p(\rvo_t|\rvs_t) + \log p(\rva_t|\rvs_t) \Big]\\
    & - \beta \sum_{t=1}^T \E_{\rvs_{1:t-1} \sim q(\rvs_{1:t-1}|\rvo_{<t},\rva_{<t-1})} \Big[ \KL \big( q(\rvs_t|\rvo_{\le t}, \rva_{<t})~||~p(\rvs_t|\rvs_{t-1})\big) \Big]
\end{align*}

Another solution is to use KL annealing \citep{sonderby16,huang2018} by setting $\beta$ to a value close to $0$ at the beginning of training so that the posterior integrates information from the inputs, and slowly increase $\beta$ to $1.0$ over the course of training. 

\clearpage
\section{Summary}

We presented a general framework to infer a world model and a policy with deep neural networks from camera observations and expert demonstrations. This framework makes very few assumptions about the observation and action space, or on the latent space, which makes it broadly applicable. The following chapters are going to introduce increasingly more complex world models by leveraging computer vision. 

In \Cref{chapter:video-scene-understanding}, we are going to present a model that infers latent dynamics from video driving data. However, instead of reconstruction the RGB images, the latent states will be trained to reconstruct computer vision quantities including depth, semantic segmentation, and optical flow. These quantities are related to the three important concepts we have reviewed previously: geometry, semantics, and motion.

In \Cref{chapter:instance-prediction}, we will transition from reconstructing in the perspective image space, to the bird's-eye view space. The bird's-eye view space is well-suited for autonomous driving since driving happens in the 2D ground plane.

And finally in \Cref{chapter:imitation-learning}, we will demonstrate the effectiveness of the world model and learned driving policy in closed-loop in a driving simulator.

\chapter{Video Scene Understanding}
\label{chapter:video-scene-understanding}

\graphicspath{{Chapter4/Figures/}}

In this chapter, we train a world model and a driving policy on urban driving scenes using the framework we have introduced in the previous chapter. The learned latent states can reconstruct important computer vision quantities: geometry, semantics and motion. Our model learns a representation from RGB video with a spatio-temporal convolutional module. This learned representation can be explicitly decoded to future depth, semantic segmentation and optical flow, in addition to being an input to a learnt driving policy.
To model the stochasticity of the future, we minimise the divergence between the present/prior distribution (what could happen given what we have seen) and the future/posterior distribution (what we observe actually happens). During inference, diverse futures are generated by sampling from the present distribution. The content of this chapter was published in the proceedings of the 16th European Conference on Computer Vision, ECCV 2020 as \citet{hu2020probabilistic}.

\section{Introduction}

Building predictive cognitive models of the world is often regarded as the essence of intelligence.
It is one of the first skills that we develop as infants.
We use these models to enhance our capability at learning more complex tasks, such as navigation or manipulating objects \citep{piaget}.

Unlike in humans, developing prediction models for autonomous vehicles to anticipate the future remains hugely challenging.
Road agents have to make reliable decisions based on forward simulation to understand how relevant parts of the scene will evolve.
There are various reasons why modelling the future is incredibly difficult: natural-scene data is rich in details, most of which are irrelevant for the driving task, dynamic agents have complex temporal dynamics, often controlled by unobservable variables, and the future is inherently uncertain, as multiple futures might arise from a unique and deterministic past.


Current approaches to autonomous driving individually model each dynamic agent by producing hand-crafted behaviours, such as trajectory forecasting, to feed into a decision making module \citep{intentnet18}. 
This largely assumes independence between agents and fails to model multi-agent interaction.  
Most works that holistically reason about the temporal scene are limited to simple, often simulated environments or use low dimensional input images that do not have the visual complexity of real world driving scenes \citep{oh15}.
Some approaches tackle this problem by making simplifying assumptions to the motion model or the stochasticity of the world \citep{NIPS2014_5444,intentnet18}.
Others avoid explicitly predicting the future scene but rather rely on an implicit representation or Q-function (in the case of model-free reinforcement learning) in order to choose an action \citep{model_free_rl_ex1,model_free_rl_ex2,kendall2019learning}.

Real world future scenarios are difficult to model because of the stochasticity and the partial observability of the world.
Our work addresses this by encoding the future state into a low-dimensional \textit{future distribution}. We then allow the model to have a privileged view of the future through the future distribution at training time. As we cannot use the future at test time, we train a \emph{present distribution} (using only the current state) to match the future distribution through a Kullback-Leibler divergence loss. We can then sample from the present distribution during inference, when we do not have access to the future.
We observe that this paradigm allows the model to learn accurate and diverse probabilistic future prediction outputs.

In order to predict the future we need to first encode video into a spatio-temporal representation.
Unlike advances in 2D convolutional architectures \citep{inception14,he16}, learning spatio-temporal features is more challenging due to the higher dimensionality of video data and the complexity of modelling dynamics.
State-of-the-art architectures \citep{chen18,tran18} decompose 3D filters into spatial and temporal convolutions in order to learn more efficiently.
The model we propose further breaks down convolutions into many space-time combinations and context aggregation modules, stacking them together in a more complex hierarchical representation.
We show that the learnt representation is able to jointly predict ego-motion and motion of other dynamic agents.
Ultimately we use this motion-aware and future-aware representation to improve an autonomous vehicle control policy.

Our main contributions are threefold.
Firstly, we present a novel deep learning framework for future video prediction.
Secondly, we demonstrate that our probabilistic model is able to generate visually diverse and plausible futures.
Thirdly, we show our future prediction representation substantially improves a learned autonomous driving policy.

\clearpage
\section{Related Work}

Our proposed model in this chapter is at the intersection of learning scene representation from video, probabilistic modelling of the ambiguity inherent in real-world driving data, and using the learnt representation for control.

\paragraph{Temporal representations.} 
Current state-of-the-art temporal representations from video use recurrent neural networks \citep{shi15,siam2017convolutional}, separable 3D convolutions \citep{Ioannou2016,SunJYS15,XieGDTH16,hara17,tran18}, or 3D Inception modules \citep{carreira17,chen18}. 
In particular, the separable 3D Inception (S3D) architecture \citep{chen18}, which improves on the Inception 3D module (I3D) introduced by  \citet{carreira17}, shows the best trade-off between model complexity and speed, both at training and inference time.
Adding optical flow as a complementary input modality has been consistently shown to improve performance \citep{Feichtenhofer16,simonyan14,simonyan15,Bilen2016DynamicIN}, in particular using flow for representation warping to align features over time \citep{gadde2017semantic,zhu2017deep}.
We propose a new spatio-temporal architecture that can learn hierarchically more complex features with a novel 3D convolutional structure incorporating both local and global space and time context. 

\paragraph{Visual prediction.}
Most works for learning dynamics from video fall under the framework of model-based reinforcement learning \citep{finn16a,finn17,Ebert18,kaizer19} or unsupervised feature learning \citep{srivastava15,denton17}, both regressing directly in pixel space \citep{mathieu16,ranzato14,Jayaraman19} or in a learned feature space \citep{Jaderberg16,finn16}.
For the purpose of creating good representations for driving scenes, directly predicting in the high-dimensional space of image pixels is unnecessary, as some details about the appearance of the world are irrelevant for planning and control.
Our approach is similar to that of \citet{luc17} which trains a model to predict future semantic segmentation using pseudo-ground truth labels generated from a teacher model. However, our model predicts a more complete scene representation with depth, segmentation, and flow and is probabilistic in order to model the uncertainty of the future.

\paragraph{Multi-modality of future prediction.} 
Modelling uncertainty is important given the stochastic nature of real-world data \citep{kendall2017uncertainties}.
\citet{desire17}, \citet{Bhattacharyya18} and \citet{precog19} forecast the behaviour of other dynamic agents in the scene in a probabilistic multi-modal way. 
We distinguish ourselves from this line of work as their approach does not consider the task of video forecasting, but rather trajectory forecasting, and they do not study how useful the representations learnt are for robot control.
\citet{Kurutach2018LearningPR} propose generating multi-modal futures with adversarial training, however spatio-temporal discriminator networks are known to suffer from mode collapse \citep{goodfellow2016tutorial}.

Our variational approach is similar to \citet{kohl18}, although their application domain does not involve modelling dynamics.
Furthermore, while \citet{kohl18} use multi-modal training data, i.e. multiple output labels are provided for a given input, we learn directly from real-world driving data, where we can only observe one future reality, and show that we generate diverse and plausible futures. Most importantly, previous variational video generation methods \citep{lee18,denton18} were restricted to single-frame image generation, low resolution ($64\times64$) datasets that are either simulated (Moving MNIST \citep{srivastava15}) or with static scenes and limited dynamics (KTH actions \citep{schuldt04}, Robot Pushing dataset \citep{ebert17}). Our new framework for future prediction generates entire video sequences on complex real-world urban driving data with ego-motion and complex interactions.

\paragraph{Learning a control policy.}
The representation learned from dynamics models could be used to generate imagined experience to train a policy in a model-based reinforcement learning setting \citep{ha18,hafner2019planet} or to run shooting methods for planning \citep{Chua18}.
Instead we follow the approaches of \citet{bojarski_end_2016}, \citet{codevilla2018end} and \citet{amini2018variational} and learn a policy which predicts longitudinal and lateral control of an autonomous vehicle using Conditional Imitation Learning, as this approach has been shown to be immediately transferable to the real world. 

\clearpage

\section{Model Description}
In this chapter, we consider a special case of the framework from \Cref{chapter:generative-models}. 
From $k$ past observations $\rvo_{1:k}$, we would like to infer the latent state $\rvs_k$ that is predictive of the current action $\rva_k$, as well as future computer vision representations (depth, segmentation and optical flow) $\rvy_{k+1:k+H}$ for $H$ timesteps in the future. We consider only a single latent state $\rvs_k$, and not a sequence of $T$ latent states $\rvs_{1:T}$ as in the general case.

\begin{figure}
  \centering
  \includegraphics[width=\linewidth]{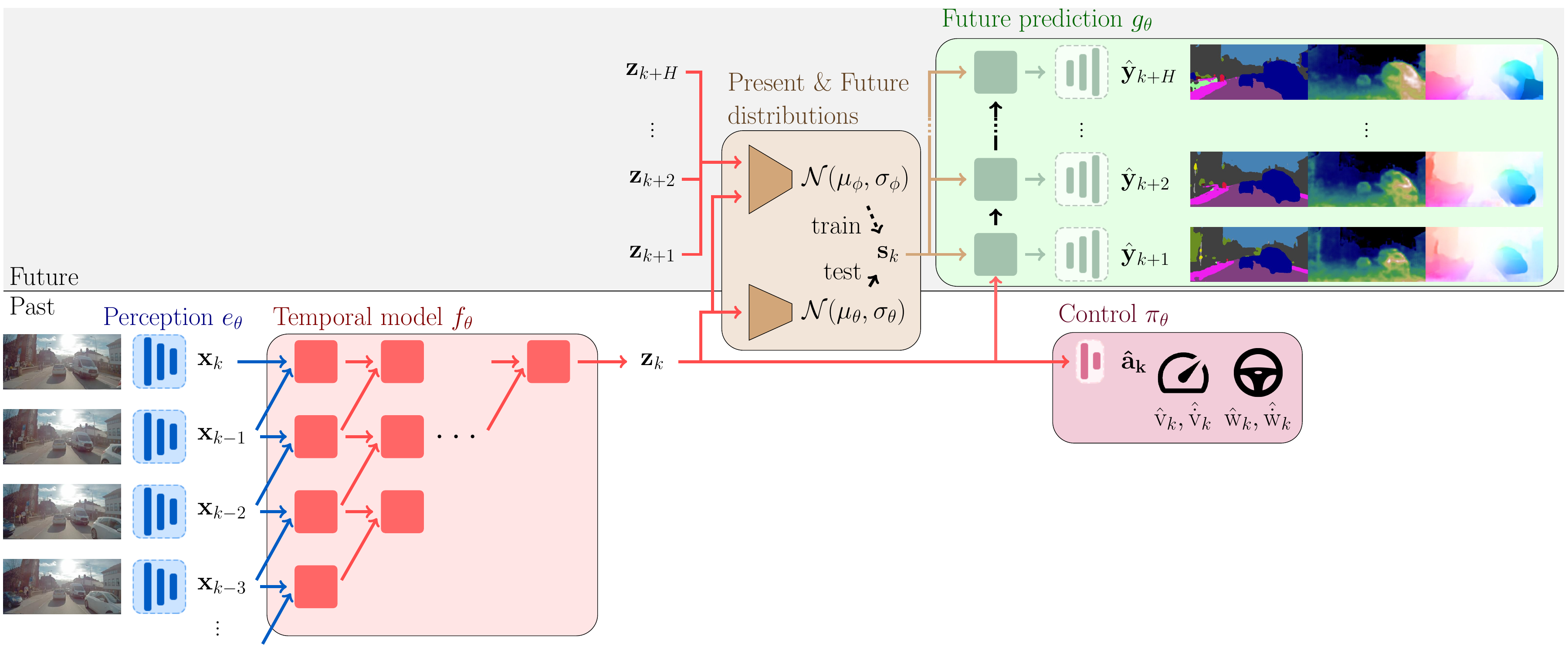}%
  \captionsetup{singlelinecheck=off}
  \caption[Architecture of our probabilistic model.]{Architecture of our probabilistic model.
  \begin{enumerate}[label=(\roman*), itemsep=0.5mm, parsep=0pt]
    \item The \textbf{Perception module} $e_{\theta}$ embeds input images $\rvo_{1:k}$ as scene representation features $\rvx_{1:k}=e_{\theta}(\rvo_{1:k})$.
    \item The \textbf{Temporal model} $f_{\theta}$ builds on these scene features to produce a spatio-temporal representation, $\rvz_k=f_{\theta}(\rvx_{1:k})$, with our proposed \emph{Temporal Block} modules.
    \item The \textbf{Present distribution} $\mathcal{N}(\mu_{\theta}(\rvz_k), \sigma^2_{\theta}(\rvz_k)\mI)$ is encouraged to match the \textbf{Future distribution} $\mathcal{N}(\mu_{\phi}(\{\rvz_{k+j}\}_{j\in J}), \sigma^2_{\phi}(\{\rvz_{k+j}\}_{j\in J}) \mI)$ through a Kullback-Leibler divergence loss.
    \item Using $\rvz_k$ and a sample $\rvs_k$ from the future distribution at training time, or the present distribution at inference time, the \textbf{Future prediction module} $g_{\theta}$ predicts future video scene representation (depth, segmentation and optical flow) $\hat{\rvy}_{k+1:k+H} = g_{\theta}(\rvz_k, \rvs_k)$ . 
    \item Lastly, a \textbf{Control policy} $\pi_{\theta}$ outputs vehicle control $\hat{\rva}_k=\pi_{\theta}(\rvz_k)$ (we do not include $\rvs_k$ as an input to the control policy).
  \end{enumerate}
   
   }
  \label{video-fig:architecture}
\end{figure}

Our model infers a latent state to jointly predict future scene representation (depth, semantic segmentation, optical flow) and train a driving policy. \Cref{video-subsection:probabilistic-modelling} defines the probabilistic framework with the generative model and inference model, and specifies the evidence lower bound we want to maximise. Subsequently, the remaining sections details each part of the network that work in concert to optimise this lower bound. The model contains five components:
\begin{itemize}
    \item \textbf{Perception} (\Cref{video-subsection:perception}), an image scene understanding module;
    \item \textbf{Temporal model} (\Cref{video-subsection:temporal-model}), which learns a spatio-temporal representation;
    \item \textbf{Present and future distributions} (\Cref{video-subsection:distributions}), the probabilistic prior and posterior distributions;
    \item \textbf{Future prediction} (\Cref{video-subsection:future-prediction}), which predicts future video scene representation;
    \item \textbf{Control} (\Cref{video-subsection:control}), which trains a driving policy using expert driving demonstrations.
\end{itemize}

\Cref{video-fig:architecture} gives an overview of the model, and further details are described in \Cref{video-appendix:architecture}.

\subsection{Probabilistic Modelling}
\label{video-subsection:probabilistic-modelling}
Conditioned on $k$ frames of image observations $\rvo_{1:k}$, we want to infer the latent state $\rvs_k$, which is predictive of future scene representation $\rvy_{k+1:k+H} = (\rvd_{k+1:k+H},\rvl_{k+1:k+H},  \rvf_{k+1:k+H})$ (depth, semantic segmentation, and optical flow) for $H$ timesteps in the future, as well as of the current action $\rva_k$. 

The observations $\rvo_{1:k}$ are first embedded with a perception encoder $e_{\theta}$ (\Cref{video-subsection:perception}) as $\rvx_{1:k}=e_{\theta}(\rvo_{1:k})$, and then mapped to a spatio-temporal representation $\rvz_{k} = f_{\theta}(\rvx_{1:k})$ (\Cref{video-subsection:temporal-model}). We parametrise the latent state distribution as a diagonal Gaussian $p(\rvs_k|\rvo_{1:k}) \sim \mathcal{N}(\mu_{\theta}(\rvz_k), \sigma_{\theta}(\rvz_k)\mI)$. For $t=1,\dots,H$, let $\rvd_{k+t}$ be the future depth, $\rvl_{k+t}$ be the future semantic segmentation, and $\rvf_{k+t}$ be the future optical flow. The full generative model, parametrised by $\theta$, is defined as:
\begin{align}
    \label{video-eq:full-model}
        \rvz_k | \rvo_{1:k} &\sim \delta(f_{\theta}(e_{\theta}(\rvo_{1:k}))) \\
        \rvs_k |\rvz_k&\sim \mathcal{N}(\mu_{\theta}(\rvz_k), \sigma_{\theta}(\rvz_k)\mI)\\
         \rvo_{k+t} |\rvz_k, \rvs_k &\sim \mathcal{N}(o_{t,\theta}(\rvz_k,\rvs_k),\mI) \\
        \rvd_{k+t} |\rvz_k, \rvs_k &\sim \mathcal{N}(d_{t,\theta}(\rvz_k,\rvs_k),\mI) \\
        \rvl_{k+t} |\rvz_k, \rvs_k &\sim \mathrm{Categorical}(l_{t,\theta}(\rvz_k,\rvs_k)) \\
        \rvf_{k+t}  |\rvz_k, \rvs_k&\sim \mathcal{N}(m_{t,\theta}(\rvz_k,\rvs_k),\mI) \\
        \rva_k |\rvz_k, \rvs_k&\sim \mathcal{N}(\pi_{\theta}(\rvz_k,\rvs_k), \mI)
\end{align}
with $\delta$ the Dirac delta function, $(\mu_{\theta},\sigma_{\theta})$ a network parametrising the latent state distribution, $\{o_{t,\theta},d_{t,\theta},l_{t,\theta},m_{t,\theta}\}_{t=1,\dots,H}$ parametrising the image, depth, segmentation and flow, and $\pi_{\theta}$ the driving policy. Thus, the joint probability distribution writes as:
\begin{equation}
    \label{video-eq:generative-model}
    \begin{split}
    p(\rvz_k,\rvs_k, \rva_k, \rvo_{k+1:k+H},\rvy_{k+1:k+H}|\rvo_{1:k}) =p(\rvz_k|\rvo_{1:k})p(\rvs_k|\rvz_k) p(\rva_k|\rvz_k,\rvs_k) \prod_{t=1}^H p(\rvo_{k+t},\rvy_{k+t}|\rvz_k,\rvs_k)
    \end{split}
\end{equation}
with $p(\rvz_k|\rvo_{1:k})=\delta(\rvz_k - f_{\theta}(e_{\theta}(\rvo_{1:k})))$ and $p(\rvo_{k+t},\rvy_{k+t}|\rvz_k,\rvs_k)=p(\rvo_{k+t}|\rvz_k,\rvs_k)p(\rvy_{k+t}|\rvz_k,\rvs_k)$.

In order to infer the latent states $(\rvz_k, \rvs_k)$ from observed data, we introduce a variational distribution $q_{Z,S}$ (parametrised by $\phi$) defined and factorised as:
\begin{equation}
    \label{video-eq:inference-model}
    q_{Z,S} \triangleq q(\rvz_k,\rvs_k|\rvo_{1:k+H}) = q(\rvz_k|\rvo_{1:k+H})q(\rvs_k|\rvz_k,\rvo_{k+1:k+H})
\end{equation}
where $q(\rvz_k|\rvo_{1:k+H})=p(\rvz_k|\rvo_{1:k})$ the Dirac delta function defined above since $\rvz_k$ is deterministic. By applying Jensen's inequality, we obtain a lower bound on the log evidence:
\begin{alignat}{2}
&  && \log  p(\rva_k, \rvo_{k+1:k+H},\rvy_{k+1:k+H}|\rvo_{1:k}) \notag \\
& \ge && ~\mathcal{L}(\rva_k, \rvo_{k+1:k+H},\rvy_{k+1:k+H}; \rvo_{1:k},\theta, \phi) \notag \\
 &\triangleq&&  \E_{\rvz_k, \rvs_k \sim q_{Z,S}}\bigg[\underbrace{\log p(\rva_k|\rvz_k,\rvs_k)}_{\text{action}} ~+~ \sum_{t=1}^H\Big(\underbrace{\log p(\rvo_{k+t}| \rvz_k,\rvs_k)}_{\text{image prediction}} ~+~ \underbrace{\log p(\rvy_{k+t}|\rvz_k,\rvs_k)}_{\text{label prediction}}\Big)\bigg]  \notag \\
& && - \underbrace{\KL\Big( q(\rvs_k | \rvz_k,\rvo_{k+1:k+H}) ~||~ p(\rvs_k | \rvz_k) \Big)}_{\text{posterior and prior matching}}
\end{alignat}
Please refer to \Cref{video-appendix:lower-bound} for the full derivation. The probability distribution of $p(\rva_k|\rvz_k,\rvs_k)$ being a diagonal Gaussian, the resulting loss is the mean-squared-error. The same logic applies to $\rvo_{k+t}, \rvd_{k+t}$ and $\rvf_{k+t}$. The semantic segmentation $p(\rvl_{k+t}|\rvz_k,\rvs_k)$ being a categorical distribution, the resulting loss is the cross-entropy. During training, the expectation over the variational distribution is efficiently approximated using a single sample from the distribution and using the reparametrisation trick \citep{kingma14}.

\subsection{Perception}
\label{video-subsection:perception}
The perception component of our system contains two modules: the encoder of a scene understanding model that was trained on single image frames to reconstruct depth and semantic segmentation  \citep{kendall18}, and the encoder of an optical flow network \citep{sun_pwc-net_2018}, trained to predict optical flow. The combined perception features $\rvx_{i} = e_{\theta}(\rvo_i) \in \mathbb{R}^{C\times H \times W}$, for $i=1,\dots, k+H$, form the input to the temporal model. These models are also used as a teacher to distill the information from the future, giving pseudo-ground truth labels for depth, segmentation and flow $\{\rvd_{k+t}, \rvl_{k+t}, \rvf_{k+t}\}$, for $t=1,\dots,H$. See \Cref{video-section:training_data} for more details on the teacher model.

\subsection{Temporal Model}
\label{video-subsection:temporal-model}
Learning a temporal representation from video is extremely challenging because of the high dimensionality of the data, the stochasticity and complexity of natural scenes, and the partial observability of the environment. To train 3D convolutional filters from a sequence of raw RGB images, a large amount of data, memory and compute is required. We instead learn spatio-temporal features with a temporal model that operates on perception encodings, which constitute a more compact representation compared to RGB images.

The temporal model $f_{\theta}$ takes a history of perception features $(\rvx_{1:k})$ with temporal context $k$ and encodes it into a spatio-temporal feature $\rvz_k$.
\begin{equation}
    \rvz_k = f_{\theta}(\rvx_{1:k}) 
\end{equation}

\paragraph{Temporal Block}
We propose a spatio-temporal module, named {\emph{Temporal Block}}, to learn hierarchically more complex temporal features as follows:

\begin{itemize}[noitemsep,topsep=0pt,parsep=2pt,partopsep=2pt]
    \item \textbf{Decomposing the filters}: instead of systematically using full 3D filters $(k_t, k_s, k_s)$, with $k_t$ the time kernel dimension and $k_s$ the spatial kernel dimension, we apply four parallel 3D convolutions with kernel sizes: $(1, k_s, k_s)$ (spatial features), $(k_t, 1, k_s)$ (horizontal motion), $(k_t, k_s, 1)$ (vertical motion), and $(k_t, k_s, k_s)$ (complete motion). All convolutions are preceded by a $(1, 1, 1)$ convolution to compress the channel dimension.
    \item \textbf{Global spatio-temporal context}: in order to learn contextual features, we additionally use three spatio-temporal average pooling layers at: full spatial size $(k_t, H, W)$ ($H$ and $W$ are respectively the height and width of the perception features $x_t$), half size $(k_t, \frac{H}{2}, \frac{W}{2})$ and quarter size  $(k_t, \frac{H}{4}, \frac{W}{4})$. The pooled features are then bilinear upsampled to the original spatial dimension $(H, W)$ go through a $(1, 1, 1)$ convolution. 
\end{itemize}

\begin{figure}
    \centering
    \includegraphics[width=.6\textwidth]{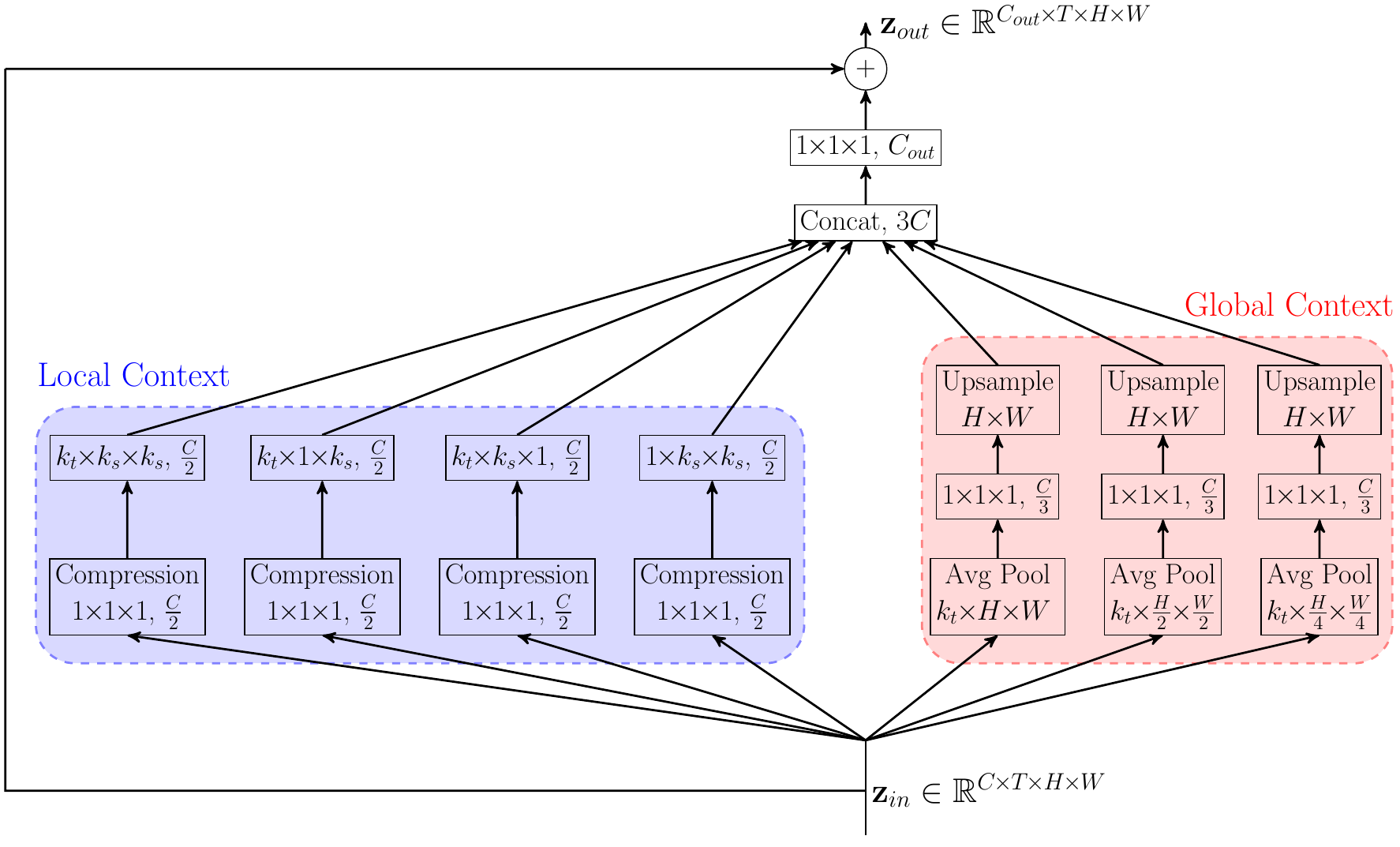}
    \caption[A Temporal Block, our proposed spatio-temporal module.]{A Temporal Block, our proposed spatio-temporal module. From a four-dimensional input $\rvz_{in} \in \mathbb{R}^{C\times T \times H \times W}$, our module learns both local and global spatio-temporal features. The local head learns all possible configurations of 3D convolutions with filters: $(1, k_s, k_s)$ (spatial features), $(k_t, 1, k_s)$ (horizontal motion), $(k_t, k_s, 1)$ (vertical motion), and $(k_t, k_s, k_s)$ (complete motion). The global head learns global spatio-temporal features with a 3D average pooling at full, half and quarter size, followed by a $(1, 1, 1)$ convolution and upsampling to the original spatial dimension $H\times W$. The local and global features are then concatenated and combined in a final $(1, 1, 1)$ 3D convolution.}
    \label{video-fig:temporal_block}
\end{figure}

\Cref{video-fig:temporal_block} illustrates the architecture of the Temporal Block. By stacking multiple temporal blocks, the network learns a representation that incorporates increasingly more temporal, spatial and global context. We also increase the number of channels by a constant $\alpha$ after each temporal block, as after each block, the network has to represent the content of the $k_t$ previous features.


\subsection{Present and Future Distributions}
\label{video-subsection:distributions}
From a unique past, even though many different futures are possible, we only observe one future. Consequently, modelling multi-modal futures from deterministic video training data is  challenging. We adopt a conditional variational approach and model two probability distributions: a {\emph{present distribution}} (or prior) $p(\rvs_k|\rvz_k)$, that represents what could happen given the past context, and a {\emph{future distribution}} (or posterior) $q(\rvs_k|\rvo_{1:k+H})$, that represents what actually happened in that particular observation. This allows us to learn a multi-modal distribution from the input data while conditioning the model to learn from the specific observed future from within this distribution.

The present and the future distributions are diagonal Gaussian, and can therefore be fully characterised by their mean and standard deviation. We parameterise both distributions with a neural network, respectively $(\mu_{\theta}, \sigma_{\theta})$ and $(\mu_{\phi}, \sigma_{\phi})$.

\paragraph{Present distribution} The input is $\rvz_k \in \R^{C_f \times H \times W}$, which represents the past context of the last $k$ frames ($k$ is the time receptive field of our temporal model). The present network contains two downsampling convolutional layers, an average pooling layer and a fully connected layer to map the features to the desired latent dimension $L$. The output of the network is the parametrisation of the present distribution: $(\mu_{\theta}(\rvz_k), \sigma_{\theta}(\rvz_k)) \in \mathbb{R}^L \times \mathbb{R}^L$.

\paragraph{Future distribution} This distribution is not only conditioned by the past $\rvz_k$, but also by the future corresponding to the training sequence. Since we are predicting $H$ steps in the future, the input of the future distribution has to contain information about future frames $(k+1, ..., k+H)$. This is achieved using the learned dynamics features $\{\rvz_{k + j}\}_{j \in J}$, with $J$ the set of indices such that $\{z_{k + j}\}_{j \in J}$ covers all future frames $(k+1, ..., k+H)$, as well as $\rvz_k$. Formally, if we want to cover $H$ frames with features that have a receptive field of $k$, then: \\ $J = \{nk ~ |~ 0 \le n \le \lfloor H / k \rfloor \} \cup \{H\}$.
The architecture of the future network is similar to the present network: for each input dynamics feature $\rvz_{k+j}$, with $j \in J$, we apply two downsampling convolutional layers and an average pooling layer. The resulting features are concatenated, and a fully-connected layer outputs the parametrisation of the future distribution: $(\mu_{\phi}(\{\rvz_{k+j}\}_{j\in J}), \sigma_{\phi}(\{\rvz_{k+j}\}_{j\in J}) \in \mathbb{R}^L \times \mathbb{R}^L$.

We encourage the present distribution $p(\rvs_k|\rvz_k)$ to match the future distribution $q(\rvs_k|\rvo_{1:k+H})$ with a KL loss. As the future is multimodal, different futures might arise from a unique past context $\rvz_k$. Each of these futures will be captured by the future distribution $q(\rvs_k|\rvo_{1:k+H})$ that will pull the present distribution $p(\rvs_k|\rvz_k)$ towards it. Since our training data is extremely diverse, it naturally contains multimodalities. Even if the past context (sequence of images $(\rvo_1, \dots, \rvo_k)$) from two different training sequences will never be the same, the dynamics network will have learned a more abstract spatio-temporal representation that ignores irrelevant details of the scene (such as vehicle colour, weather, road material etc.) to match similar past context to a similar $\rvz_k$. In this process, the present distribution will learn to cover all the possible modes contained in the future.

Note that we can predict multimodal futures from a Gaussian latent because we push these Gaussian distributions through a non-linear decoder (similar to e.g. VAEs \citep{kingma14} or GANs \citep{goodfellow14})

\subsection{Future Prediction}
\label{video-subsection:future-prediction}
A future prediction model $g_{\theta}=\{d_{t,\theta},l_{t,\theta},m_{t,\theta}\}_{t=1,\dots,H}$ unrolls the temporal feature $\rvz_k$ and sample $\rvs_k$ into future scene representation predictions. The future prediction model is a convolutional recurrent network  which creates future features that become the inputs of individual decoders to decode these features to predicted depth $\hat{\rvd}_{k+t}$, segmentation $\hat{\rvl}_{k+t}$, and flow $\hat{\rvf}_{k+t}$ values in the pixel space, for $t=1,\dots, H$. 
\begin{align*}
\hat{\rvd}_{k+t} &=d_{t,\theta}(\rvz_k,\rvs_k)\\
    \hat{\rvl}_{k+t} &=l_{t,\theta}(\rvz_k,\rvs_k)\\
    \hat{\rvf}_{k+t} &=m_{t,\theta}(\rvz_k,\rvs_k)
\end{align*}

We do not decode to the pixel space for efficiency, as matching priors with posteriors in the RGB space would be slower than in the abstract scene representation space defined by depth, semantic segmentation and optical flow. 

The structure of the convolutional recurrent network is the following: a convolutional GRU \citep{ballas16} followed by three spatial residual layers, repeated $D$ times, similarly to \citet{clark19}.  Its initial hidden state is $\rvz_k$, the temporal feature. During training, we sample from the future distribution a vector 
\\$\rvs_k \sim \mathcal{N}(\mu_{\phi}(\{\rvz_{k+j}\}_{j\in J}), \sigma^2_{\phi}(\{\rvz_{k+j}\}_{j\in J}) \mI)$ that conditions the predicted future perception outputs (depth, semantic segmentation, optical flow) on the observed future. As we want our prediction to be consistent in both space and time, we broadcast spatially $\rvs_k \in \R^L$ to $\R^{L\times H \times W}$, and use the same sample throughout the future generation as an input to the GRU to condition the future.

During inference, we sample a vector $\rvs_k$ from the present distribution $\rvs_k \sim \mathcal{N}(\mu_{\theta}(\rvz_k), \sigma^2_{\theta}(\rvz_k) \mI)$, where each sample corresponds to a different future. 

\subsection{Control}
\label{video-subsection:control}
From this rich spatio-temporal representation $\rvz_k$ explicitly trained to predict the future, we train a control model $\pi_{\theta}$ to output a four dimensional vector consisting of estimated speed, acceleration, steering angle and angular velocity $(\hat{\rv_k}, \hat{\dot{\rv_k}}, \hat{\rw_k}, \hat{\dot{\rw_k}})$.
$\pi_{\theta}$ compresses $\rvz_k$ with strided convolutional layers, then stacks several fully connected layers, compressing at each stage, to regress the four dimensional output.

\subsection{Losses}
\label{video-subsection:losses}
The losses optimised by the neural network are slightly different from those defined by the evidence lower bound. These modifications, as defined below, were integrated because they improved performance.
\paragraph{Future Prediction}
The future prediction loss is the weighted sum of future depth, segmentation and optical flow losses. Let the segmentation loss at the future timestep $k+t$ be $L_{l,k+t}$. We use a top-k cross-entropy loss \citep{Wu2016BridgingCA} between the network output $\hat{\rvl}_{k+t}$ and the pseudo-ground truth label $\rvl_{k+t}$. The segmentation loss $L_l$ is computed by summing these individual terms over the future horizon $H$ with a weighted discount term $0 < \gamma_f< 1$:
\begin{equation}
    L_l = \sum_{t=1}^{H} \gamma_f^{t-1} L_{l,k+t}
\end{equation}
For depth, $L_{d,k+t}$ is the scale-invariant depth loss \cite{li_megadepth_2018} between $\hat{d}_{k+t}$ and $d_{k+t}$, For flow, we use a Huber loss between $\hat{f}_{k+t}$ and $f_{k+t}$. Similarly $L_d$ and $L_f$ are the discounted sum. We weight the summed losses by factors $\lambda_d,\lambda_s,  \lambda_f$ to get the future prediction loss $L_{\text{future-pred}}$.
\begin{equation}
    L_{\text{future-pred}} =  \lambda_d L_{d} +\lambda_l L_{l} +  \lambda_f L_{f}
\end{equation}

\paragraph{Control}
We use imitation learning, regressing to the expert's true control actions to generate a \emph{control loss} $L_c$. 
For both speed and steering, we have access to the expert actions $(\rv_k, \rw_k, \dots, \rv_{k+N_c}, \rw_{k+N_c})$ $N_c$ timesteps in the future.

We compare to the linear extrapolation of the generated policy's speed/steering for future $N_c$ timesteps:
\begin{align}
    L_c = \sum_{i=0}^{N_c-1} \gamma_c^i  & \left(  \left(  \rv_{k+i} - \left(\hat{\rv}_k + \Delta_i \hat{\dot{\rv}}_k\right) \right)^2 \right. \nonumber \\ 
    & \hphantom{a.} \left. + \left(  \rw_{k+i} - \left(\hat{\rw}_k + \Delta_i\hat{\dot{\rw}}_k\right) \right)^2 \right)
\end{align}
where $0 < \gamma_c < 1$ is the control discount factor penalising less speed and steering errors further into the future, and $\Delta_i$ is the time interval between two control timesteps.
\paragraph{Total Loss}
If we denote KL matching term as the probabilistic loss:

$$L_{\text{probabilistic}} = \KL\Big( q(\rvs_k | \rvz_k,\rvo_{k+1:k+H}) ~||~ p(\rvs_k | \rvz_k) \Big)$$

The final loss $L$ can be decomposed into the probabilistic loss $L_{\text{probabilistic}}$, the future prediction loss $L_{\text{future-pred}}$, and the control loss $L_c$.
\begin{equation}
    L = \E_{\rvz_k, \rvs_k \sim q_{Z,S}}[\lambda_p L_{\text{probabilistic}} + \lambda_{fp}{L_\text{future-pred}} + \lambda_c L_c]
\end{equation}
In all experiments we use $\gamma_f=0.6$,  $\lambda_d =1.0$, $\lambda_l = 1.0$, $\lambda_f = 0.5$, $\lambda_p=0.005$, $\lambda_{fp}=1$, $\gamma_c=0.7$, $\lambda_c=1.0$, found from hyperparameter tuning.

\clearpage
\section{Experimental Setting}

We have collected driving data in a densely populated, urban environment, representative of most European cities using multiple drivers over the span of six months.
For the purpose of this work, only the front-facing camera images and the measurements of the speed and steering have been used to train our model, all sampled at $5$Hz.

\subsection{Training Data}
\label{video-section:training_data}
\paragraph{Label generation} We first pretrain the scene understanding encoder on a number of heterogeneous datasets to predict depth and semantic segmentation: CityScapes \citep{cityscapes16}, Mapillary Vistas \citep{neuhold_mapillary_2017}, ApolloScape \citep{huang2018apolloscape} and Berkeley Deep Drive \citep{yu2018bdd100k}. The optical flow network is a pretrained PWC-Net from \citep{sun_pwc-net_2018}.
The decoders of these networks are used for generating pseudo-ground truth segmentation and depth labels.
\paragraph{Model training}
The network is trained using 30 hours of driving data from the urban driving dataset we collected and described above.
We address the inherent dataset bias by sampling data uniformly across lateral and longitudinal dimensions.
First, the data is split into a histogram of bins by steering, and subsequently by speed.

\subsection{Metrics}
\label{video-subsection:metrics}
We report standard metrics for measuring the quality of depth, segmentation and flow: respectively scale-invariant logarithmic error, intersection-over-union (IoU), and average end-point error.
For ease of comparison, additionally to individual metrics, we report a unified perception metric $M_{\text{perception}}$ defined as improvement of segmentation, depth and flow metrics with respect to the \emph{Repeat Frame} baseline (repeats the perception outputs of the current frame):
\begin{equation}
    M_{\text{perception}} = \frac{1}{3} (\text{depth}_{\text{\% decrease}} + \text{seg}_{\text{\% increase}} +  \text{flow}_{\text{\% decrease}})
\end{equation}

To measure control performance, we report mean absolute error of speed and steering outputs.

\clearpage
\section{Results}
\subsection{Future Prediction}
We compare our model to the \emph{Repeat frame} baseline (repeating the perception outputs of the current frame at time $k$ for each future frame $k+t$ with $t=1,\dots,H$) in \Cref{video-table:future-prediction}, observing a significant increase in performance.

\begin{table}
\centering
\begin{small}
    \caption[Perception metrics for two seconds future prediction on the collected urban driving data.]{Perception metrics for two seconds future prediction on the collected urban driving data. We measure the prediction accuracy of depth with scale-invariant logarithmic error, optical flow with average end-point error, and semantic segmentation with mean IoU. $M_{\text{perception}}$ indicates overall performance.}
    \begin{tabular}{lcccc}
    \toprule
    {\textbf{\shortstack{Model}}} & $M_{\text{perception}} (\uparrow$) & Depth ($\downarrow$) & Flow ($\downarrow$) & Seg. ($\uparrow$)\\
    \midrule
    Repeat frame & 0.0\%  & 1.467 & 5.707& 0.356\\
    \textbf{Ours} & \textbf{20.0\%} & \textbf{0.970} & \textbf{4.857} & \textbf{0.396}\\
    \bottomrule
    \end{tabular}
    \label{video-table:future-prediction}
\end{small}
\end{table}

\begin{figure}
    \centering
    \includegraphics[width=\textwidth]{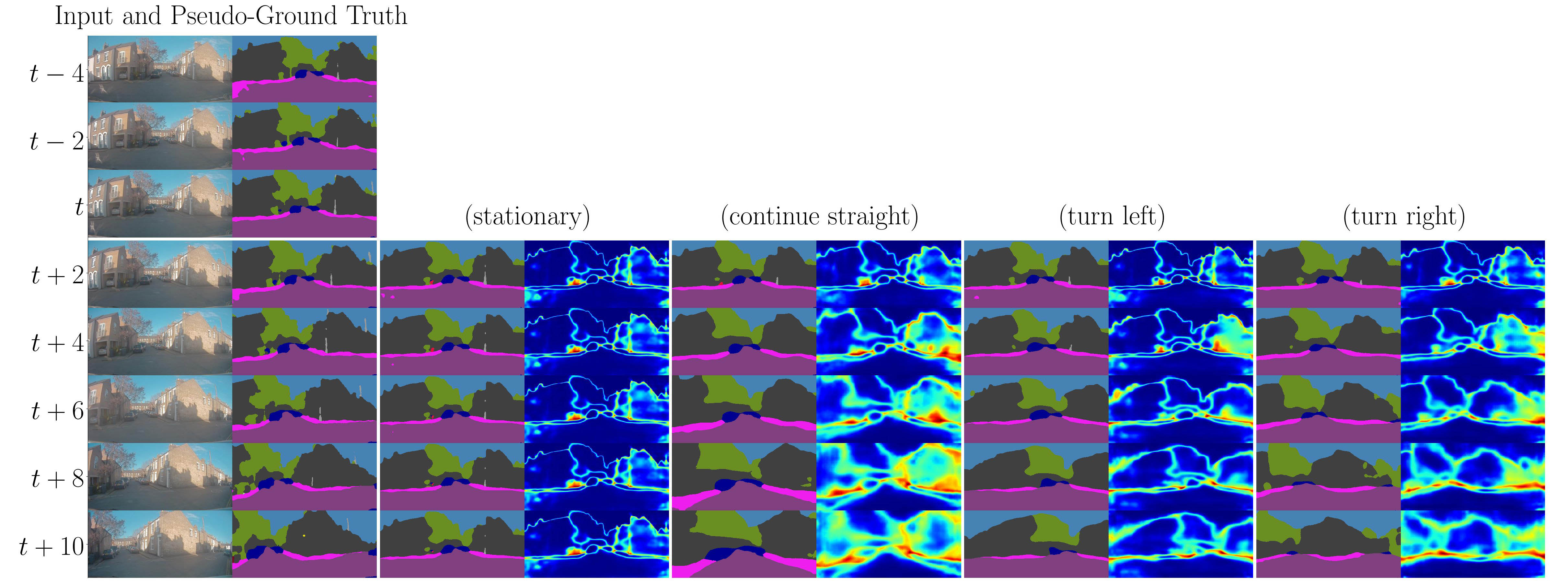}
    \caption[Predicted futures from our model while driving through an urban intersection.]{Predicted futures from our model while driving through an urban intersection. From left, we show the actual past and future video sequence and labelled semantic segmentation. Using four different samples, $\rvs_k$, we observe the model imagining different driving manoeuvres at an intersection: being stationary, driving straight, taking a left or a right turn. We show both predicted semantic segmentation and entropy (uncertainty) for each future. This example demonstrates that our model is able to learn a probabilistic embedding, capable of predicting multi-modal and plausible futures.}
    \label{video-fig:scenarios1}
\end{figure}

\begin{figure}
    \centering
    \includegraphics[width=\textwidth]{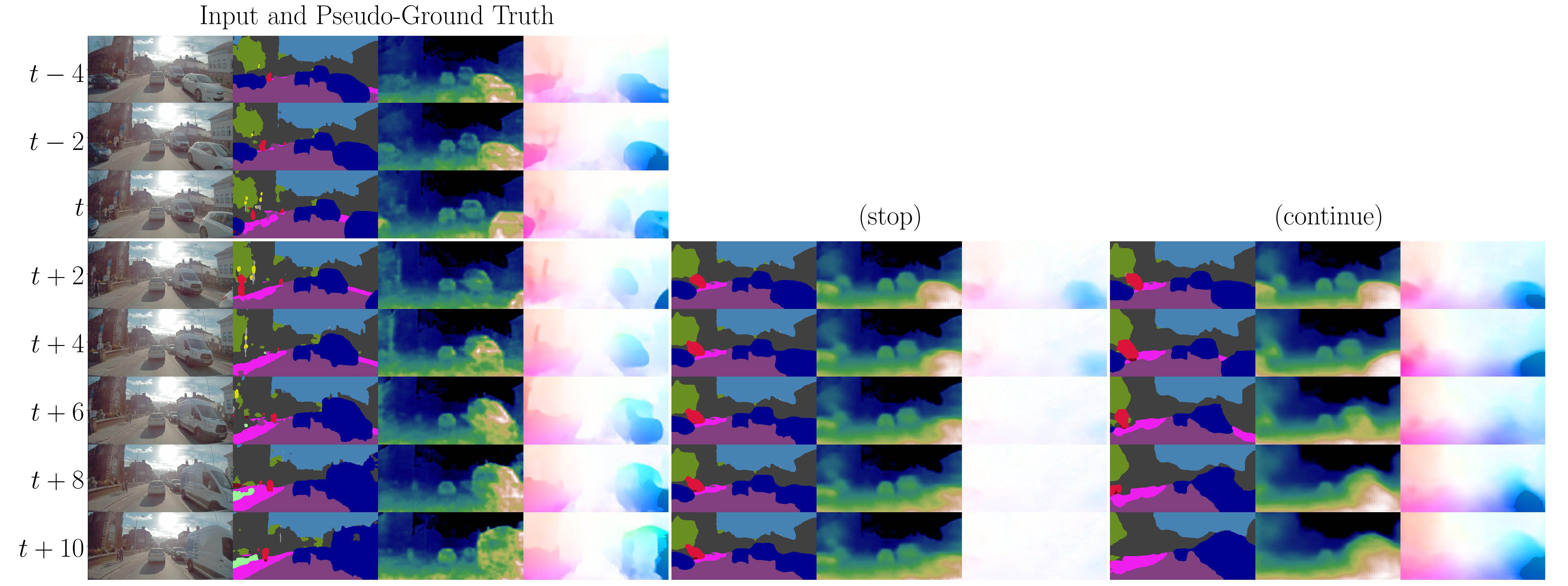}
    \caption[Predicted futures from our model while driving through a busy urban scene.]{Predicted futures from our model while driving through a busy urban scene. From left, we show actual past and future video sequence and labelled depth, semantic segmentation, and optical flow. Using two different samples, $\rvs_k$, we observe the model imagining either stopping in traffic or continuing in motion. This illustrates our model's efficacy at jointly predicting holistic future behaviour of our own vehicle and other dynamic agents in the scene across all modalities. Each colour in the optical flow image represents the vector direction, and the intensity its magnitude.}
    \label{video-fig:scenarios2}
\end{figure}

Perhaps the most striking result of the model is observing that our model can predict diverse and plausible futures from a single sequence of past frames at $5$Hz, corresponding to one second of past context and two seconds of future prediction.
In \Cref{video-fig:scenarios1} and \Cref{video-fig:scenarios2} we show qualitative examples of our video scene understanding future prediction in real-world urban driving scenes.
We sample from the present distribution, $\rvs_k \sim \mathcal{N}(\mu_{\theta}(\rvz_k), \sigma^2_{\theta}(\rvz_k)\mI)$, to demonstrate multi-modality.

Further, our framework can automatically infer which scenes are unusual or unexpected and where the model is uncertain of the future, by computing the differential entropy of the present distribution. Simple scenes (e.g. one-way streets) will tend to have a low entropy, corresponding to an almost deterministic future. Any latent code sampled from the present distribution will correspond to the same future. Conversely, complex scenes (e.g. intersections, roundabouts) will be associated with a high-entropy. Different samples from the present distribution will correspond to different futures, effectively modelling the stochasticity of the future.\footnote{In the accompanying \href{https://wayve.ai/blog/predicting-the-future}{blog post}, we illustrate how diverse the predicted future becomes with varying levels of entropy in an intersection scenario and an urban traffic scenario.}

\begin{figure}[h]
    \centering
    \begin{subfigure}[b]{\textwidth}
        \centering
        \includegraphics[width=0.7\linewidth]{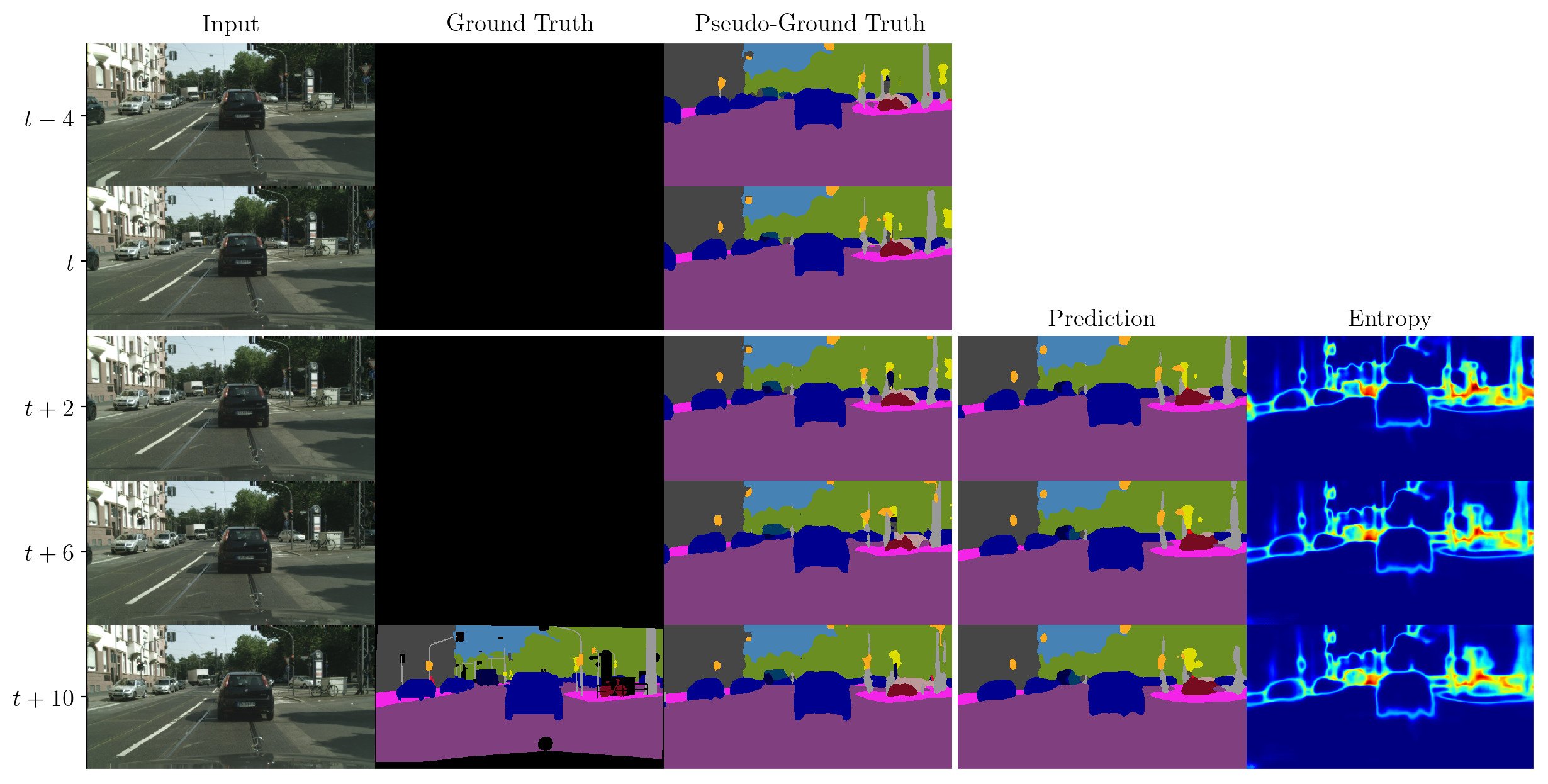}
        \caption{Our model can correctly predict future segmentation of small classes such as poles or traffic lights.}
    \end{subfigure}
    \vskip\baselineskip
    \begin{subfigure}[b]{\textwidth}
        \centering
        \includegraphics[width=0.7\linewidth]{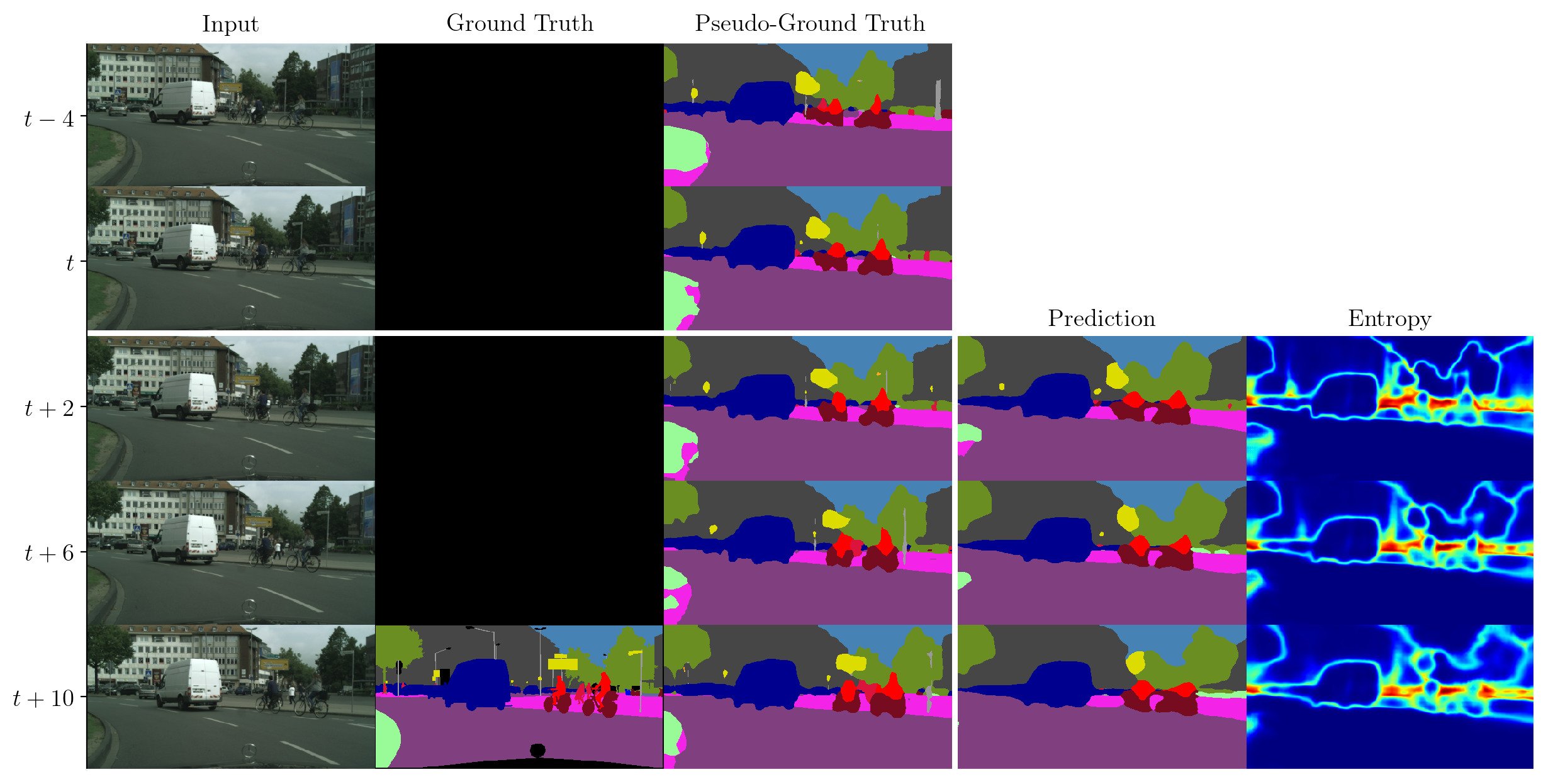}
        \caption{Dynamic agents, i.e. cars and cyclists, are also accurately predicted.}
    \end{subfigure}
    \vskip\baselineskip
    \begin{subfigure}[b]{\textwidth}
        \centering
        \includegraphics[width=0.7\linewidth]{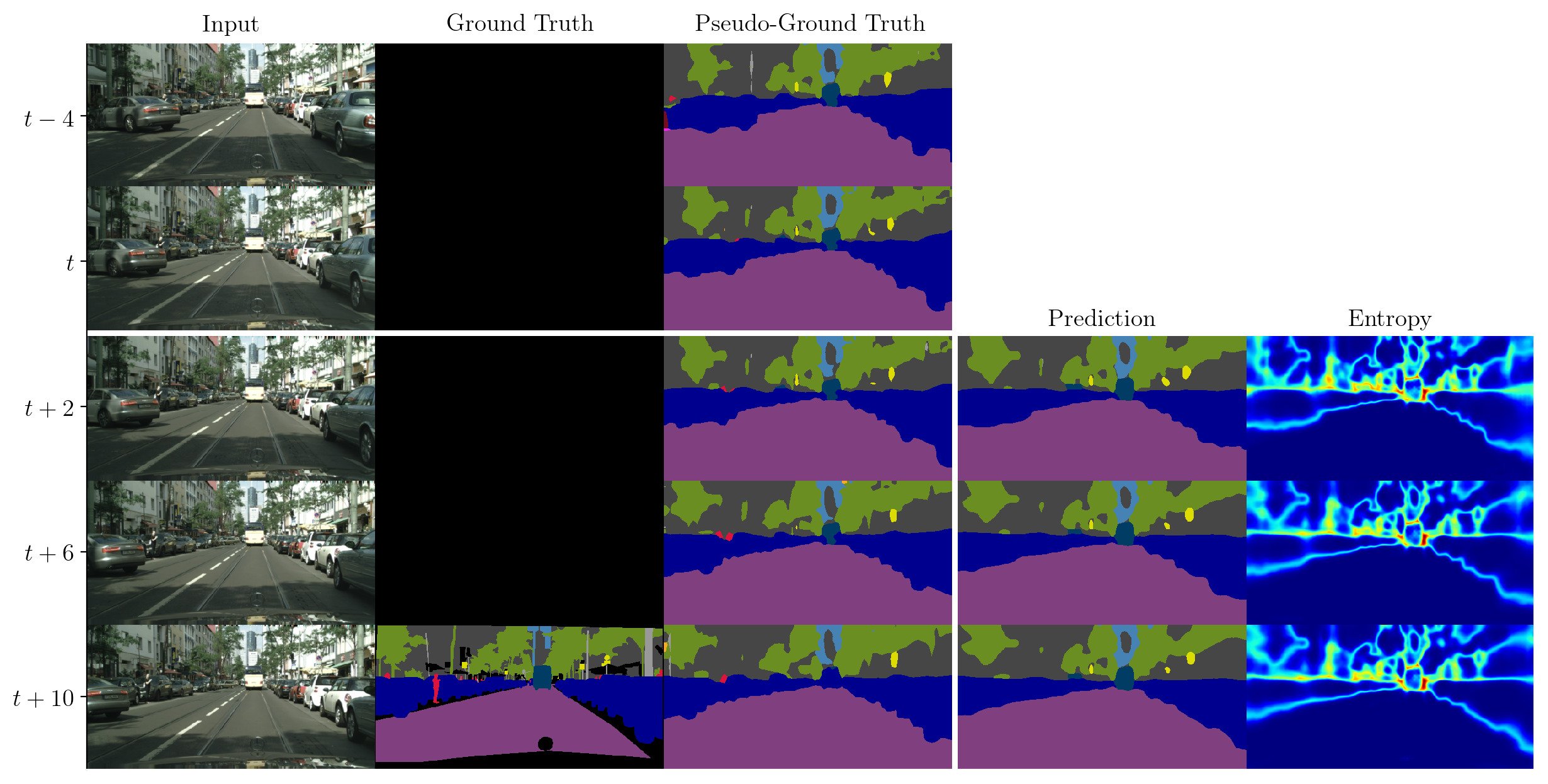}
        \caption{In this example, the bus is correctly segmented, without any class bleeding contrary to the pseudo-ground truth segmentation, showing that our model can reason in a holistic way.}
    \end{subfigure}
    \caption[Future prediction on the Cityscapes dataset.]{Future prediction on the Cityscapes dataset, for 10 frames in the future at 17Hz and $256\times 512$ resolution.}
    \label{video-fig:cityscapes}
\end{figure}

Finally, to allow reproducibility, we evaluate our future prediction framework on Cityscapes \citep{cityscapes16} and report future semantic segmentation performance in \Cref{video-table:cityscapes}. We compare our predictions, at resolution $256\times512$, to the ground truth segmentation at 5 and 10 frames in the future. Qualitative examples on Cityscapes can be found in \Cref{video-fig:cityscapes}.

\begin{table}
\centering
\caption{Future semantic segmentation performance on Cityscapes at $t=5$ and $t=10$ frames in the future using mean forecast (corresponding to respectively \SI{0.29}{\s} and \SI{0.59}{\s} at \SI{17}{\Hz}).}
\begin{tabular}{lcc}
\toprule
{\textbf{Model}} & $\text{IoU}_{t=5}$ ($\uparrow$) & $\text{IoU}_{t=10}$  ($\uparrow$) \\
\midrule
\citet{nabavi18} & - & 0.274\\
\citet{chiu19} & - & 0.408\\
\textbf{Ours} & \textbf{0.464} & \textbf{0.416}\\
\bottomrule
\end{tabular}
\label{video-table:cityscapes}
\end{table}

\paragraph{Limitations.} The interpretation of the futures from \Cref{video-fig:scenarios1} and \Cref{video-fig:scenarios2} were obtained from manual inspection (continue straight, turn left, turn right etc.). It was however also possible to obtain predicted futures that did no seem to correspond to anything interpretable. That was notably the case when the Gaussian sample $\rvs_k$ was far from the mean of the distribution, or when the underlying scene was complex.

Improvements to this visualisation method would be to (i) automatically cluster the predicted futures instead of relying on manual inspection and (ii) understanding the structure of the learned latent space with random walks from the mean.

\subsection{Driving Policy}
We study the influence of the learned temporal representation on driving performance. Our baseline is the control policy learned from a single frame (\emph{One frame})

First we compare to this baseline a model that was trained to directly optimise control, without being supervised with future scene prediction (\emph{Ours w/o future pred.}). It shows only a slight improvement over the static baseline, hinting that it is difficult to learn an effective temporal representation by only using control error as a learning signal.

\begin{table}[h]
\centering
\caption[Evaluation of the driving policy.]{Evaluation of the driving policy. The policy is learned from temporal features explicitly trained to predict the future. We observe a significant performance improvement over non-temporal and non-future-aware baselines.}
\begin{tabular}{lccc}
\toprule
{\textbf{\color{darkgray}{\shortstack{Model}}}}&  Steering ($\downarrow$) & Speed ($\downarrow$) \\
\midrule
One frame & 0.049 & 0.048\\
Ours w/o future pred. & 0.043 & 0.039\\
\textbf{Ours}  & \textbf{0.033} & \textbf{0.026}\\
\bottomrule
\end{tabular}
\label{video-table:perception-and-control}
\end{table}

Our probabilistic model achieves the best performance with a 33\% steering and 46\% speed improvement over the \emph{One frame} baseline.

\clearpage
\section{Summary}

We proposed a deep learning model capable of probabilistic future prediction of static scene and other dynamic agents from camera video. We modelled important quantities from computer vision in the learned latent state with depth (geometry), semantic segmentation (semantics), and optical flow (motion). We showed our model could predict diverse and plausible futures by sampling latent states from the world model. 

This initial work has left a lot of future directions to explore. One is to leverage known priors and structure in the latent representation. So far, the latent state was reasoning in the perspective image space. However, driving occurs in the 2D ground plane. In the next chapter (\Cref{chapter:instance-prediction}), we are going to show the benefit of leveraging this inductive bias by reasoning in the bird's-eye view space.

Another direction to explore is to jointly predict states and actions arbitrarily far in the future. In this work we have restricted the future prediction to a fixed time horizon, and only predict the action of the current timestep, not the future actions. We would also like to evaluate our model in closed-loop. The driving policy was evaluated in an offline manner by comparing the predicted actions with the expert actions. This means the errors from our model are reset over time as it was the expert who generated the data. In closed-loop evaluation, the errors compound over time. It is the responsibility of the model to take actions, which will generate new observations by interacting with the environment. In \Cref{chapter:imitation-learning}, we will jointly predict states and actions and evaluate our model in closed-loop.


\chapter{Bird's-Eye View Future Prediction}
\label{chapter:instance-prediction}

\graphicspath{{Chapter5/Figures/}}

Driving requires interacting with road agents and predicting their future behaviour in order to navigate safely. In this chapter, we present FIERY: a probabilistic future prediction model in bird's-eye view from monocular cameras. Our model predicts future instance segmentation and motion of dynamic agents that can be transformed into non-parametric future trajectories. Contrary to the previous chapter, we only predict future scene representations and not vehicle control. The purpose of the method presented here is to demonstrate it is possible to accurately model the future dynamic scene in bird's-eye view from cameras only. In the following chapter, we will show that integrating this bird's-eye view modelling as an inductive bias results in a large performance increase in closed-loop driving.

Our approach combines the perception, sensor fusion and prediction components of a traditional autonomous driving stack by estimating bird's-eye view prediction directly from surround RGB monocular camera inputs. FIERY learns to model the inherent stochastic nature of the future solely from camera driving data in an end-to-end manner, without relying on HD maps, and predicts multimodal future trajectories.
We show that our model outperforms previous prediction baselines on the NuScenes and Lyft datasets. The content of this chapter was published in the proceedings of the 18th International Conference on Computer Vision, ICCV 2021 as \citet{hu2021fiery}.
    
\section{Introduction}

\begin{figure}
\centering
    \begin{subfigure}[b]{0.55\textwidth}
    \centering
    \includegraphics[width=\linewidth]{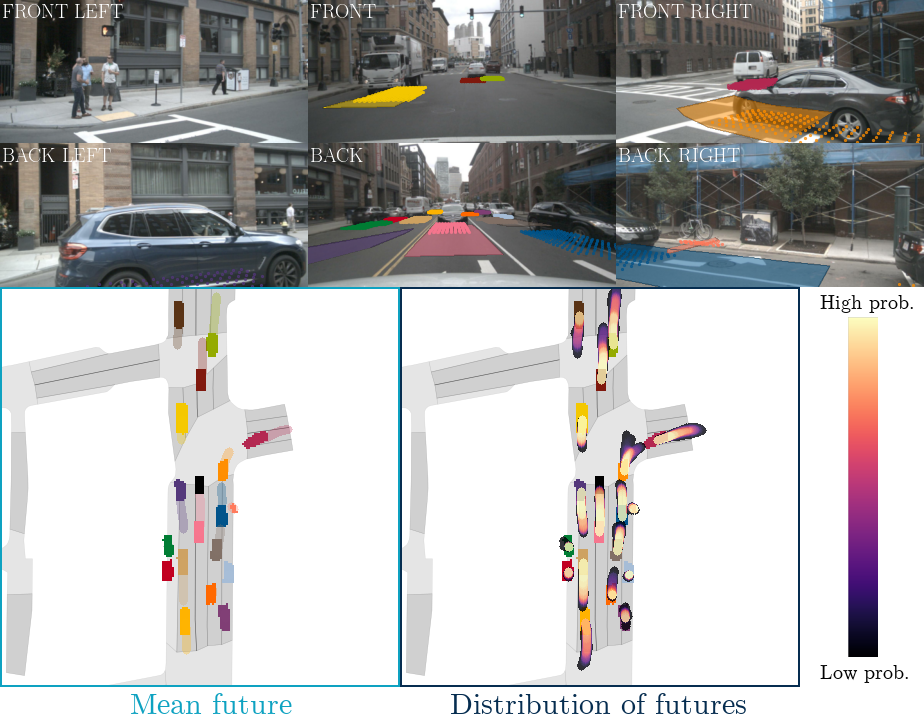}%
    \end{subfigure}
    \par\smallskip
    \begin{subfigure}[t]{0.55\textwidth}
    \includegraphics[width=\linewidth]{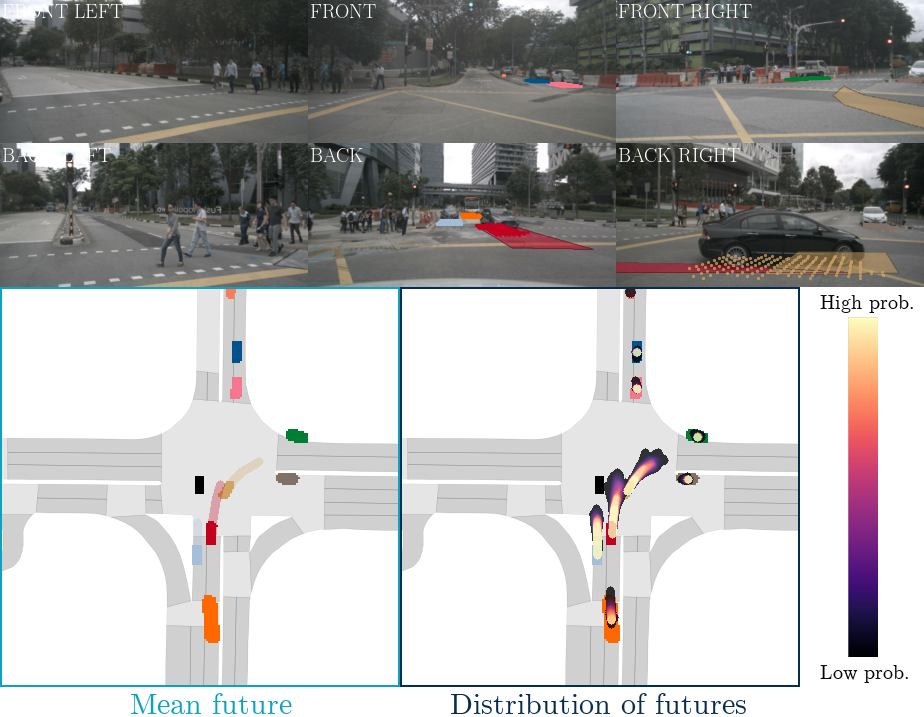}%
    \end{subfigure}
    \par\smallskip
    \begin{subfigure}[t]{0.55\textwidth}
    \includegraphics[width=\linewidth]{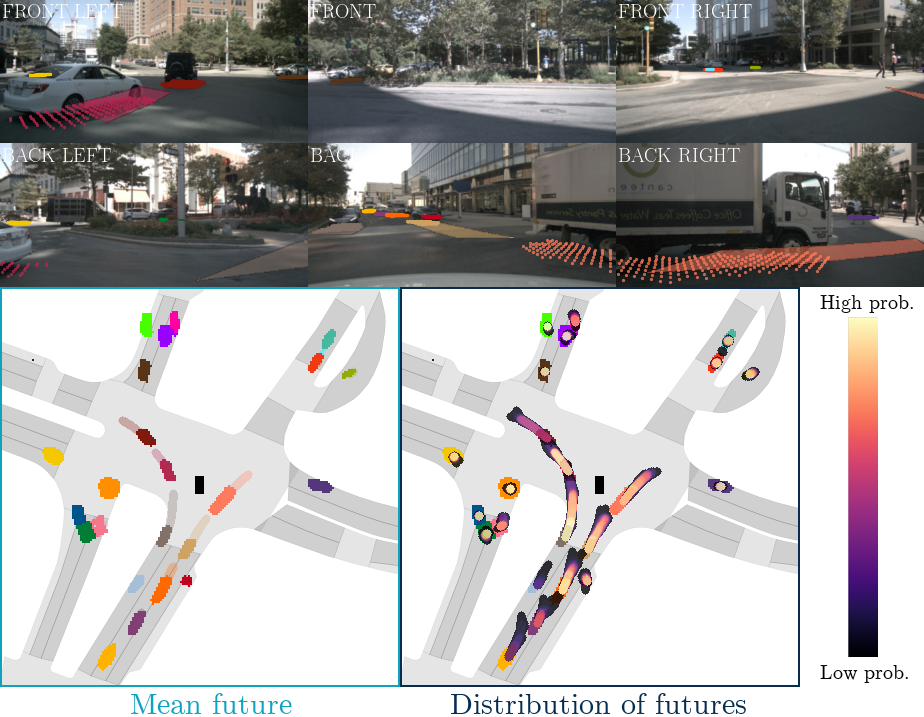}%
    \end{subfigure}
\caption[Multimodal future predictions by our bird's-eye view network.]{Multimodal future predictions by our bird's-eye view network.
Top two rows: RGB camera inputs. The predicted instance segmentations are projected to the ground plane in the images. We also visualise the mean future trajectory of dynamic agents as transparent paths.
Bottom row: future instance prediction in bird's-eye view in a $100\mathrm{m}\times100\mathrm{m}$ capture size around the ego-vehicle, which is indicated by a black rectangle in the center.}
\label{fiery-fig:teaser}
\end{figure}

Prediction of future states is a key challenge in many autonomous decision making systems. This is particularly true for motion planning in highly dynamic environments: for example in autonomous driving where the motion of other road users and pedestrians has a substantial influence on the success of motion planning \citep{cui2019multimodal}. Estimating the motion and future poses of these road users enables motion planning algorithms to better resolve multimodal outcomes where the optimal action may be ambiguous knowing only the current state of the world.

Autonomous driving is inherently a geometric problem, where the goal is to navigate a vehicle safely and correctly through 3D space. As such, an orthographic \newterm{bird's-eye view} (BeV) representation is commonly used for motion planning and prediction based on LiDAR sensing \citep{precog19, wu2020motionnet}. Recent advances in camera-based perception have rivalled LiDAR-based perception \citep{wang2019pseudo}, and we anticipate that this will also be possible for wider monocular vision tasks, including prediction. Building a perception and prediction system based on cameras would enable a leaner, cheaper and higher resolution visual recognition system over LiDAR sensing.

Most of the work in camera-based prediction to date has either been performed directly in the perspective view coordinate frame \citep{alahi2016social, jin2017predicting}, or using simplified BeV raster representations of the scene \citep{desire17, djuric2020uncertainty, cui2019multimodal} generated by HD-mapping systems such as \citep{liang2018deep, gao2020vectornet}. We wish to build predictive models that operate in an orthographic bird's-eye view frame (due to the benefits for planning and control \citep{7490340}), though \emph{without} relying on auxiliary systems to generate a BeV raster representation of the scene.

A key theme in robust perception systems for autonomous vehicles has been the concept of early sensor fusion, generating 3D object detections directly from image and LiDAR data rather than seeking to merge the predicted outputs of independent object detectors on each sensor input. Learning a task jointly from multiple sources of sensory data as in \citep{xu2018pointfusion}, rather than a staged pipeline, has been demonstrated to offer improvement to perception performance in tasks such as object detection. We seek similar benefits in joining perception and sensor fusion to prediction by estimating bird's-eye view prediction directly from surround RGB monocular camera inputs, rather than a multi-stage discrete pipeline of tasks.

Lastly, traditional autonomous driving stacks \citep{ferguson08} tackle future prediction by extrapolating the current behaviour of dynamic agents, without taking into account possible interactions. They rely on HD maps and use road connectivity to generate a set of future trajectories. Instead, FIERY learns to predict future motion of road agents directly from camera driving data in an end-to-end manner, without relying on HD maps. It can reason about the probabilistic nature of the future, and predicts multimodal future trajectories (see \href{https://wayve.ai/blog/fiery-future-instance-prediction-birds-eye-view}{blog post} and \Cref{fiery-fig:teaser}).


To summarise the main contributions of this chapter:
\begin{enumerate}[itemsep=1mm, parsep=0pt]
    \item We present the first future prediction model in bird's-eye view from monocular camera videos. Our framework explicitly reasons about multi-agent dynamics by predicting future instance segmentation and motion in bird's-eye view.
    \item Our probabilistic model predicts plausible and multi-modal futures of the dynamic environment.
    \item We demonstrate quantitative benchmarks for future dynamic scene segmentation, and show that our learned prediction outperforms previous prediction baselines for autonomous driving on the NuScenes \citep{nuscenes19} and Lyft \citep{lyft2019} datasets.
    
\end{enumerate}

\clearpage
\section{Related Work}
\paragraph{Bird's-eye view representation from cameras.}
Many prior works \citep{zhu2019generative,wang2019monocular} have tackled the inherently ill-posed problem \citep{groenendijk20} of lifting 2D perspective images into a bird's-eye view representation. \citet{Pan_2020,ng2020bevseg} dealt specifically with the problem of generating semantic BeV maps directly from images and used a simulator to obtain the ground truth.

Recent multi-sensor datasets, such as NuScenes \citep{nuscenes19} or Lyft \citep{lyft2019}, made it possible to directly supervise models on real-world data by generating bird's-eye view semantic segmentation labels from 3D object detections. \citet{roddick20} proposed a Bayesian occupancy network to predict road elements and dynamic agents in BeV directly from monocular RGB images.  Most similar to our approach, Lift-Splat \citep{philion20} learned a depth distribution over pixels to lift camera images to a 3D point cloud, and project the latter into BeV using camera geometry. Fishing Net \citep{hendy20} tackled the problem of predicting deterministic future bird's-eye view semantic segmentation using camera, radar and LiDAR inputs.

\paragraph{Future prediction.} Classical methods for future prediction generally employ a multi-stage detect-track-predict paradigm for trajectory prediction \citep{chai2019multipath, hong2019rules, tang2019multiple}.  However, these methods are prone to cascading errors and high latency, and thus many have turned to an end-to-end approach for future prediction.  Most end-to-end approaches rely heavily on LiDAR data \citep{Luo_2018_CVPR, djuric2020multixnet}, showing improvements by incorporating HD maps \citep{intentnet18},  encoding constraints \citep{casas2019spatiallyaware}, and fusing radar and other sensors for robustness \citep{shah2020liranet}.  These end-to-end methods are faster and have higher performance as compared to the traditional multi-stage approaches.

The above methods attempt future prediction by producing a single deterministic trajectory \citep{intentnet18, hendy20}, or a single distribution to model the uncertainty of each waypoint of the trajectory \citep{casas2019spatiallyaware, djuric2020multixnet}.  However, in the case of autonomous driving, one must be able to anticipate a range of behaviors for actors in the scene, jointly.  From an observed past, there are many valid and probable futures that could occur \citep{hu2020probabilistic}.  Other work \citep{chai2019multipath, tang2019multiple, phanminh2020covernet} has been done on probabilistic multi-hypothesis trajectory prediction, however all assume access to top-down rasterised representations as inputs. Our approach is the first to predict diverse and plausible future vehicle trajectories directly from raw camera video inputs.

\clearpage

In this chapter, we consider a special case of the framework from \Cref{chapter:generative-models}. 
From $k$ past observations $\rvo_{1:k}$, we would like to infer the latent state $\rvs_k$ that is predictive of future computer vision representations (bird's-eye view instance segmentation and flow) $\rvy_{k+1:k+H}$ for $H$ timesteps in the future. We consider only a single latent state $\rvs_k$, and not a sequence of $T$ latent states $\rvs_{1:T}$ as in the general case. We also focus on the prediction problem and do not predict any action.

\section{Model Architecture}
\begin{figure}[h!]
    \captionsetup{singlelinecheck=off}
    \centering
    \includegraphics[width=\linewidth]{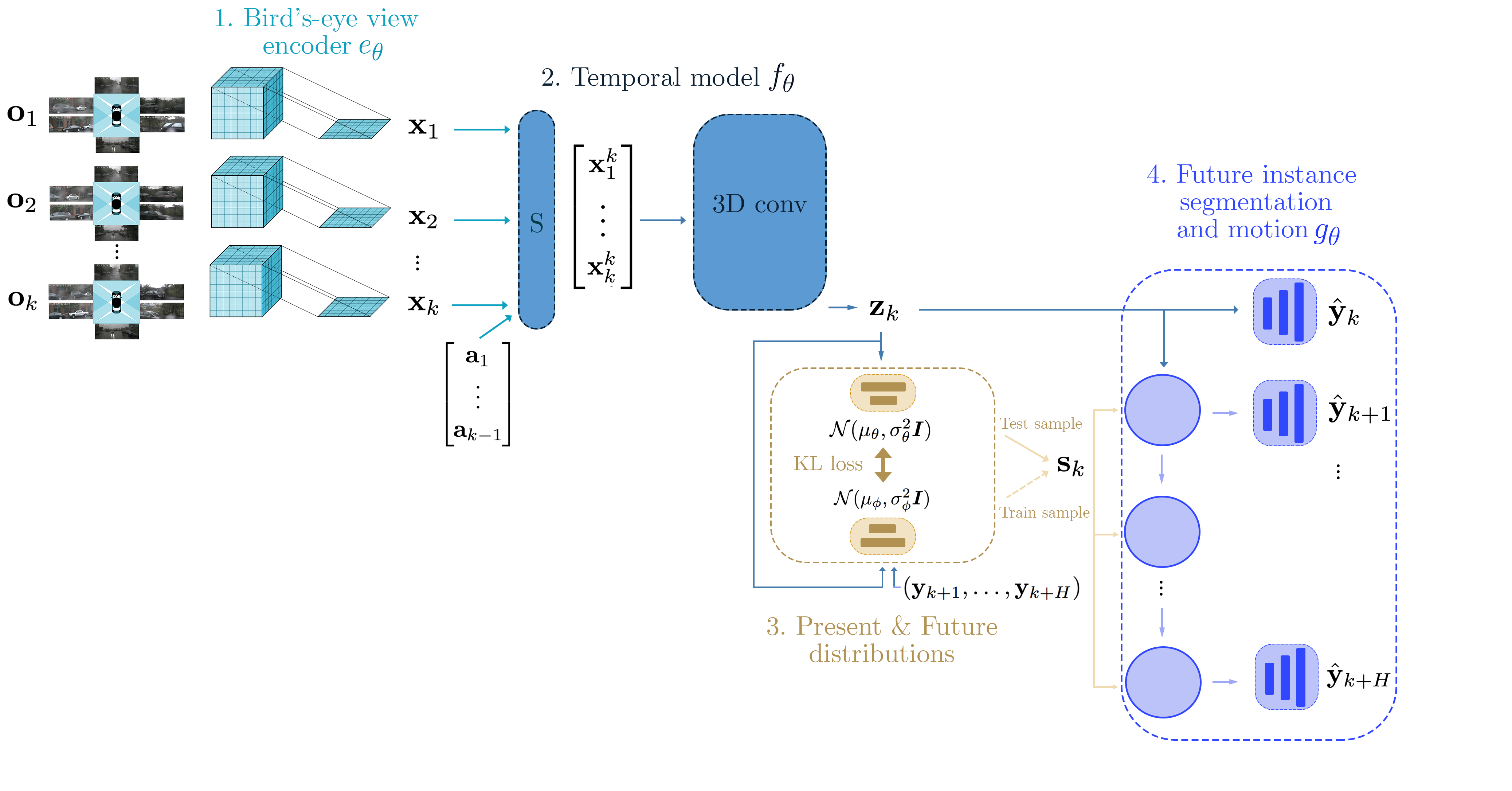}%
    \caption[The architecture of FIERY: a future instance and motion prediction model in bird's-eye view from camera inputs.]{The architecture of FIERY: a future instance and motion prediction model in bird's-eye view from camera inputs.
    \begin{enumerate}[itemsep=0.3mm, parsep=0pt]
        \item At each past timestep $\{1,...,k\}$, we lift camera inputs $(\rvo_1, \dots, \rvo_k)$ to 3D by predicting a depth probability distribution over pixels and using known camera intrinsics and extrinsics. These 3D features are pooled to bird's-eye view features $\rvx_{1:k}=e_{\theta}(\rvo_{1:k})$. 
        \item Using past ego-motion $(\rva_1, \dots, \rva_{k-1})$, we transform the bird's-eye view features into the present reference frame (time $k$) with a Spatial Transformer module $S$. These features are the input to a 3D convolutional temporal model that outputs a spatio-temporal feature $\rvz_k = f_{\theta}(\rvx_{1:k}, \rva_{1:k-1})$.
        \item We parametrise two probability distributions: the present and the future distribution. The present distribution $\mathcal{N}(\mu_{\theta}(\rvz_k), \sigma_{\theta}^2(\rvz_k)\mI)$ is conditioned on the current feature $\rvz_k$, and the future distribution $\mathcal{N}(\mu_{\phi}(\rvz_k, \rvy_{k+1:k+H}), \sigma_{\phi}^2(\rvz_k, \rvy_{k+1:k+H})\mI)$ is conditioned on both the current feature $\rvz_k$ and future labels $\rvy_{k+1:k+H}$ .
        \item We sample a latent code $\rvs_k$ from the future distribution during training, and from the present distribution during inference. The current feature $\rvz_k$ and the latent code $\rvs_k$ are the inputs to the future prediction model that recursively predicts future features that are decoded into future instance segmentation and future motion in bird's-eye view $(\hat{\rvy}_k, ..., \hat{\rvy}_{k+H})$.
    \end{enumerate}
    }
    \label{fiery-fig:model-diagram}
\end{figure}


Our model learns a latent state to predict future scene representation (semantic segmentation, instance segmentation, future motion) in bird's-eye view. \Cref{fiery-subsection:probabilistic-modelling} defines the probabilistic framework with the generative model and inference model, and specifies the evidence lower bound we want to maximise. Subsequently, the remaining sections details each part of the network that work in concert to optimise this lower bound. The model contains four components:
\begin{itemize}
    \item \textbf{Bird's-eye view encoder} (\Cref{fiery-subsection:perception}), a geometric bird's-eye view module;
    \item \textbf{Temporal model} (\Cref{fiery-subsection:temporal-model}), which learns a spatio-temporal representation from bird's-eye view features and past actions;
    \item \textbf{Present and future distributions} (\Cref{fiery-subsection:distributions}), the probabilistic prior and posterior distributions;
    \item \textbf{Future instance segmentation and motion} (\Cref{fiery-subsection:future-pred}), which predicts future instance segmentation and motion in bird's-eye view.
\end{itemize}

\Cref{fiery-fig:model-diagram} gives an overview of the model, and further details are described in \Cref{fiery-appendix:model}.

\subsection{Probabilistic Modelling}
\label{fiery-subsection:probabilistic-modelling}
Conditioned on $k$ frames of image observations $\rvo_{1:k}$ and $k-1$ past actions $\rva_{1:k-1}$, we want to infer the latent state $\rvs_k$, which is predictive of future bird's-eye view scene representation $\rvy_{k:k+H} = (\rvl_{k:k+H}, \rvi_{k:k+H}, \rvm_{k:k+H})$ (semantic segmentation, instance segmentation, and future motion) for $H$ timesteps in the future.

The observations $\rvo_{1:k}$ are first embedded with a bird's-eye view encoder $e_{\theta}$ (\Cref{fiery-subsection:perception}) as $\rvx_{1:k}=e_{\theta}(\rvo_{1:k})$, and then mapped to a spatio-temporal representation $\rvz_{k} = f_{\theta}(\rvx_{1:k}, \rva_{1:k-1})$ using past actions $\rva_{1:k-1}$ (\Cref{fiery-subsection:temporal-model}). We parametrise the latent state distribution as a diagonal Gaussian $p(\rvs_k|\rvo_{1:k}) \sim \mathcal{N}(\mu_{\theta}(\rvz_k), \sigma_{\theta}(\rvz_k)\mI)$. For $t=0,\dots,H$, let $\rvl_{k+t}$ be the future semantic segmentation, $\rvi_{k+t}$ be the future instance segmentation, and $\rvm_{k+t}$ be the future motion. The full generative model, parametrised by $\theta$, is defined as:
\begin{align}
    \label{fiery-eq:full-model}
        \rvz_k | \rvo_{1:k}, \rva_{1:k-1} &\sim \delta(f_{\theta}(e_{\theta}(\rvo_{1:k}), \rva_{1:k-1})) \\
        \rvs_k |\rvz_k &\sim \mathcal{N}(\mu_{\theta}(\rvz_k), \sigma_{\theta}(\rvz_k)\mI)\\
         \rvo_{k+t} | \rvz_k,\rvs_k&\sim \mathcal{N}(o_{t,\theta}(\rvz_k,\rvs_k),\mI) \\
        \rvl_{k+t} | \rvz_k,\rvs_k&\sim \mathrm{Categorical}(l_{t,\theta}(\rvz_k,\rvs_k)) \\
        \rvi_{k+t} |\rvz_k,\rvs_k &\sim \mathcal{N}(i_{t,\theta}(\rvz_k,\rvs_k),\mI) \\
        \rvm_{k+t}|\rvz_k,\rvs_k &\sim \mathcal{N}(m_{t,\theta}(\rvz_k,\rvs_k),\mI)
\end{align}
with $\delta$ the Dirac delta function, $(\mu_{\theta},\sigma_{\theta})$ a network parametrising the latent state distribution, and $\{o_{t,\theta},l_{t,\theta},i_{t,\theta},m_{t,\theta}\}_{t=0,\dots,H}$ parametrising the image, segmentation, instance and motion. Thus, the joint probability distribution is:
\begin{equation}
    \label{fiery-eq:generative-model}
    \begin{split}
    p(\rvz_k,\rvs_k, \rvo_{k:k+H},\rvy_{k:k+H}|\rvo_{1:k},\rva_{1:k-1}) =p(\rvz_k|\rvo_{1:k},\rva_{1:k-1})p(\rvs_k|\rvz_k) \prod_{t=0}^H p(\rvo_{k+t},\rvy_{k+t}|\rvz_k,\rvs_k)
    \end{split}
\end{equation}
with 
\begin{align}
p(\rvz_k|\rvo_{1:k},\rva_{1:k-1})&=\delta(\rvz_k - f_{\theta}(e_{\theta}(\rvo_{1:k}), \rva_{1:k-1})) \\ p(\rvo_{k+t},\rvy_{k+t}|\rvz_k,\rvs_k)&=p(\rvo_{k+t}|\rvz_k,\rvs_k)p(\rvy_{k+t}|\rvz_k,\rvs_k) \label{fiery-eq:generative-split}
\end{align}

In order to learn this distribution from observed data, we introduce a variational distribution $q_{Z,S}$ (parametrised by $\phi$) defined and factorised as:
\begin{equation}
    \label{fiery-eq:inference-model}
    q_{Z,S} \triangleq q(\rvz_k,\rvs_k|\rvo_{1:k},\rva_{1:k-1}, \rvy_{k+1:k+H}) = q(\rvz_k|\rvo_{1:k},\rva_{1:k-1}, \rvy_{k+1:k+H})q(\rvs_k|\rvz_k, \rvy_{k+1:k+H})
\end{equation}
where $q(\rvz_k|\rvo_{1:k},\rva_{1:k-1}, \rvy_{k+1:k+H})=p(\rvz_k|\rvo_{1:k},\rva_{1:k-1})$ the Dirac delta function defined above since $\rvz_k$ is deterministic. By applying Jensen's inequality, we obtain a lower bound on the log evidence:
\begin{alignat}{2}
&  && \log  p(\rvo_{k:k+H},\rvy_{k:k+H}|\rvo_{1:k},\rva_{1:k-1}) \notag \\
& \ge && ~\mathcal{L}(\rvo_{k:k+H},\rvy_{k:k+H}; \rvo_{1:k},\rva_{1:k-1},\theta, \phi) \notag \\
 &\triangleq&&  \E_{\rvz_k, \rvs_k \sim q_{Z,S}}\bigg[\sum_{t=0}^T\Big(\underbrace{\log p(\rvo_{k+t}| \rvz_k,\rvs_k)}_{\text{image prediction}} ~+~ \underbrace{\log p(\rvy_{k+t}|\rvz_k,\rvs_k)}_{\text{label prediction}}\Big)\bigg]  \notag \\
& && - \underbrace{\KL\Big( q(\rvs_k|\rvz_k, \rvy_{k+1:k+H}) ~||~ p(\rvs_k | \rvz_k) \Big)}_{\text{posterior and prior matching}}
\end{alignat}
Please refer to \Cref{fiery-appendix:lower-bound} for the full derivation. The probability distribution of $p(\rvo_{k+t}|\rvz_k,\rvs_k)$ being a diagonal Gaussian, the resulting loss is the mean-squared-error. The same logic applies to $\rvi_{k+t}$ and $\rvm_{k+t}$. The semantic segmentation $p(\rvl_{k+t}|\rvz_k,\rvs_k)$ being a categorical distribution, the resulting loss is the cross-entropy. During training, the expectation over the variational distribution is efficiently approximated using a single sample from the distribution and using the reparametrisation trick \citep{kingma14}.

\subsection{Bird's-Eye View Encoder}
\label{fiery-subsection:perception}
\paragraph{Lifting camera features to 3D} For every past timestep, we use the method of \citet{philion20} to extract image features from each camera and then lift and fuse them into a BeV feature map. In particular, each image is passed through a standard convolutional encoder $b_{\theta}$ (we use EfficientNet \citep{tan19} in our implementation. The backbone was pretrained on ImageNet) to obtain a set of features to be lifted and a set of discrete depth probabilities. Let $\rvo_i = \{\rvc_i^1, ..., \rvc_i^n\}$ be the set of $n=6$ camera images at time $i\in \{1,\dots,k\}$. We encode each image $\rvc_i^j$ with the encoder: $\rvb_i^j = b_{\theta}(\rvc_i^j) \in \R^{(C+D)\times H_e \times W_e}$, with $C$ the number of feature channels, $D$ the number of discrete depth values, and $(H_e, W_e)$ the feature spatial size. $D$ is equal to the number of equally spaced depth slices between $D_{\text{min}}$ (the minimum depth value) and $D_{\text{max}}$ (the maximum depth value) with size $D_{\mathrm{size}}=1.0\mathrm{m}$. Let us split this feature into two: $\rvb_i^j = (\rvb_{i,C}^j, \rvb_{i,D}^j)$ with $\rvb_{i,C}^j \in \R^{C\times H_e \times W_e}$ and $\rvb_{i,D}^j \in \R^{D\times H_e \times W_e}$.  A tensor $\rvu_i^j \in \R^{C \times D \times H_e \times W_e}$ is formed by taking the outer product of the features to be lifted with the depth probabilities:
\begin{equation}
    \rvu_i^j = \rvb_{i,C}^j \otimes \rvb_{i,D}^j
\end{equation}

The depth probabilities act as a form of self-attention, modulating the features according to which depth plane they are predicted to belong to.
Using the known camera intrinsics and extrinsics (position of the cameras with respect to the center of gravity of the vehicle), these tensors from each camera $(\rvu_i^1, ..., \rvu_i^n)$ are lifted to 3D in a common reference frame (the inertial center of the ego-vehicle at time $i$). 

\paragraph{Pooling to bird's-eye view} In our experiments, to obtain a bird's-eye view feature, we discretise the space in $0.50\mathrm{m}\times 0.50\mathrm{m}$ columns in a $100\mathrm{m}\times100\mathrm{m}$ capture size around the ego-vehicle. The 3D features are sum pooled along the vertical dimension to form bird's-eye view feature maps $\rvx_{1:k}=e_{\theta}(\rvo_{1:k}) \in \R^{C\times H \times W}$, with $(H, W) = (200, 200)$ the spatial extent of the BeV feature.

\subsection{Learning a Temporal Representation}
\label{fiery-subsection:temporal-model}
The past bird's-eye view features $\rvx_{1:k}$ are transformed to the present's reference frame (time $k$) using known past ego-motion $\rva_{1:k-1}$. $\rva_{k-1} \in SE(3)$ corresponds to the ego-motion from $k-1$ to $k$, i.e. the translation and rotation of the ego-vehicle. Using a Spatial Transformer \citep{jaderberg15} module $S$, we warp past features $\rvx_i$ to the present reference frame for $i \in \{1,\dots,k-1\}$:

\begin{equation}
    \rvx_i^k = S(\rvx_i, \rva_{k-1}\cdot \rva_{k-2} \cdot ... \cdot \rva_i)
\end{equation}

Since we lose the past ego-motion information with this operation, we concatenate spatially-broadcast actions to the warped past features $\rvx_i^k$. These features are then the input to a temporal model which outputs a spatio-temporal state $\rvz_k$:
\begin{equation}
    \rvz_k = f_{\theta}(\rvx_{1:k}, \rva_{1:k-1})
\end{equation}

$f_{\theta}$ is the composition of the spatial transformer module followed by the temporal model. The temporal model is a 3D convolutional network with local spatio-temporal convolutions, global 3D pooling layers, and skip connections. For more details on this module, refer to \Cref{fiery-appendix:model}.

\subsection{Present and Future Distributions}
\label{fiery-subsection:distributions}
Following \Cref{chapter:video-scene-understanding} we adopt a conditional variational approach to model the inherent stochasticity of future prediction. We introduce two distributions: a \emph{present distribution} $p(\rvs_k | \rvz_k)$ (or prior) which only has access to the current spatio-temporal state $\rvz_k$, and a \emph{future distribution} $q(\rvs_k|\rvz_k, \rvy_{k+1:k+H})$ (or posterior) that additionally has access to the observed future labels $\rvy_{k+1:k+H}$, with $H$ the future prediction horizon. The labels correspond to future centerness, offset, segmentation, and flow (see \Cref{fiery-subsection:future-pred}).

We parametrise both distributions as diagonal Gaussians of dimension $L$. During training, we use samples $\rvs_k \sim \mathcal{N}(\mu_{\phi}(\rvz_k, \rvy_{k+1:k+H}), \sigma_{\phi}^2(\rvz_k, \rvy_{k+1:k+H})\mI)$ from the future distribution to enforce predictions consistent with the observed future, and Kullback-Leibler divergence loss to encourage the present distribution to cover the observed futures.

During inference, we sample $\rvs_k \sim \mathcal{N}(\mu_{\theta}(\rvz_k), \sigma_{\theta}^2(\rvz_k)\mI)$ from the present distribution where each sample encodes a possible future.

\subsection{Future Prediction in Bird's-Eye View}
\label{fiery-subsection:future-pred}

The future prediction model $g_{\theta}=\{o_{t,\theta},l_{t,\theta},i_{t,\theta},m_{t,\theta}\}_{t=0,\dots,H}$ is the succession of (i) a convolutional gated recurrent unit network taking as input the embedding $\rvz_k$ and the latent code $\rvs_k$ sampled from the future distribution during training, or the present distribution for inference. And (ii) a bird's-eye view decoder which has multiple output heads: semantic segmentation, instance centerness and instance offset (similarily to \citet{cheng20}), and future instance flow.

\begin{equation}
    \hat{\rvy}_{k:k+H} = g_{\theta}(\rvz_k, \rvs_k)
\end{equation}

\begin{figure}
\captionsetup[subfigure]{justification=centering}
\centering
    \begin{subfigure}[b]{\textwidth}
    \centering
    \includegraphics[width=0.6\linewidth]{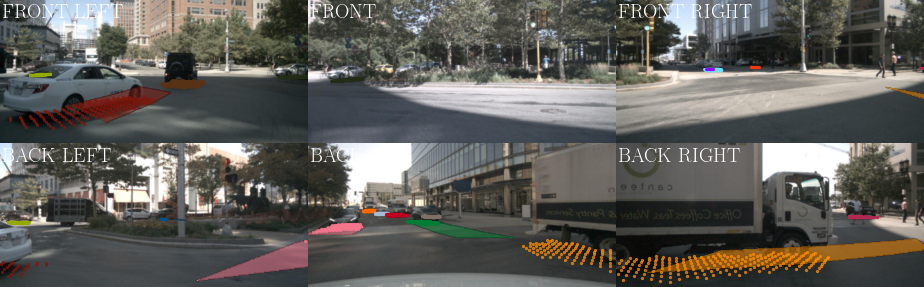}%
    \caption{Camera inputs.}
    \label{fiery-subfig-camera}
    \end{subfigure}
    \begin{subfigure}[b]{0.19\textwidth}
    \includegraphics[width=\linewidth]{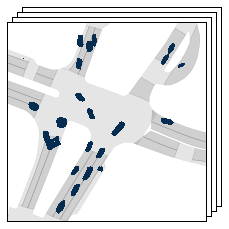}%
    \caption{Seg.}
    \label{fiery-subfig-segmentation}
    \end{subfigure}
    \begin{subfigure}[b]{0.19\textwidth}
    \includegraphics[width=\linewidth]{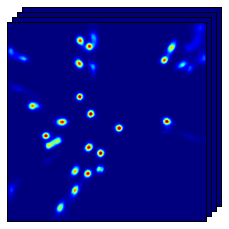}%
    \caption{Centerness.}
    \label{fiery-subfig-centerness}
    \end{subfigure}
    \begin{subfigure}[b]{0.19\textwidth}
    \includegraphics[width=\linewidth]{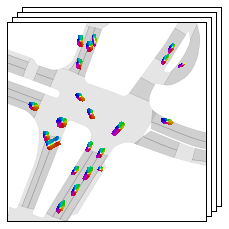}%
    \caption{Offset.}
    \label{fiery-subfig-offset}
    \end{subfigure}
    \begin{subfigure}[b]{0.19\textwidth}
    \includegraphics[width=\linewidth]{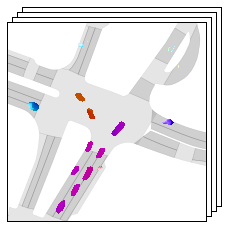}%
    \caption{Future flow.}
    \label{fiery-subfig-flow}
    \end{subfigure}
    \begin{subfigure}[b]{0.19\textwidth}
    \includegraphics[width=\linewidth]{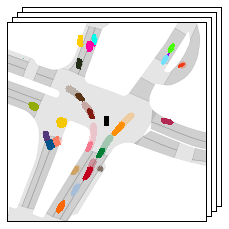}%
    \caption{Instance seg.}
    \label{fiery-subfig-instance}
    \end{subfigure}
\caption[Outputs from FIERY: segmentation, centerness, offset, future motion.]{Outputs from FIERY. (b) represents the vehicles segmentation. (c) shows a heatmap of instance centerness and indicates the probability of finding an instance center (from blue to red).  (d) shows a vector field indicating the direction to the instance center. (e) corresponds to future motion  -- notice how consistent the flow is for a given instance, since it is a rigid-body motion. (f) shows the final output of our model: a sequence of temporally consistent future instance segmentation in bird's-eye view where: (i) Instance centers are obtained by non-maximum suppression. (ii) The pixels are then grouped to their closest instance center using the offset vector. (iii) Future flow allows for consistent instance identification by comparing the warped centers using future flow from $k+t$ to $k+t+1$, and the centers at time $k+t+1$. The ego-vehicle is indicated by a black rectangle. We represent the predictions from a single sample (the mean).}
\label{fiery-fig:model-outputs}
\end{figure}

For each future timestep $k+t$, the instance centerness indicates the probability of finding an instance center (see \Cref{fiery-subfig-centerness}). By running non-maximum suppression, we get a set of instance centers. The offset is a vector pointing to the center of the instance (\Cref{fiery-subfig-offset}), and can be used jointly with the segmentation map (\Cref{fiery-subfig-segmentation}) to assign neighbouring pixels to its nearest instance center and form the bird's-eye view instance segmentation (\Cref{fiery-subfig-instance}). The future flow (\Cref{fiery-subfig-flow}) is a displacement vector field of the dynamic agents. It is used to consistently track instances over time by comparing the flow-warped instance centers at time $k+t$ and the detected instance centers at time $k+t+1$ and running a Hungarian matching algorithm \citep{hungarian55}. We do not reconstruct future RGB observations for efficiency.

A full description of our model is given in \Cref{fiery-appendix:model}.

\subsection{Losses}
We detail here a couple of practical modifications of the losses for improved performance. For semantic segmentation, we use a top-$k$ cross-entropy loss \citep{wang2019pseudo}. As the bird's-eye view image is largely dominated by the background, we only backpropagate the top-$k$ hardest pixels. In our experiments, we set $k=0.25$. The centerness loss is a $\normltwo$ distance, and both offset and flow losses are $\normlone$ distances. We exponentially discount future timesteps with a parameter $\gamma=0.95$.

\clearpage
\section{Experimental Setting}
\subsection{Dataset}
We evaluate our approach on the NuScenes \citep{nuscenes19} and Lyft \citep{lyft2019} datasets. NuScenes contains $1000$ scenes, each $20$ seconds in length, annotated at $2$Hz. The Lyft dataset contains $180$ scenes, each $25-45$ seconds in length, annotated at $5$Hz. In both datasets, the camera rig covers the full \ang{360} field of view around the ego-vehicle, and is comprised of $6$ cameras with a small overlap in field of view. Camera intrinsics and extrinsics are available for each camera in every scene.

The labels $\rvy_{k:k+H}$ are generated by projecting the provided 3D bounding boxes of vehicles into the bird's-eye view plane to create a bird's-eye view occupancy grid. See \Cref{fiery-appendix:dataset} for more details. All the labels $\rvy_{k:k+H}$ are in the present's reference frame and are obtained by transforming the labels with the ground truth future ego-motion.

\subsection{Metrics}
\paragraph{Future Video Panoptic Quality.} We want to measure the performance of our system in both:
\begin{enumerate}[label=(\roman*), itemsep=0.5mm, parsep=0pt]
    \item Recognition quality: how consistently the instances are detected over time.
    \item Segmentation quality: how accurate the instance segmentations are.
\end{enumerate}

We use the \emph{Video Panoptic Quality} (VQP) \citep{kim2020vps} metric defined as:

\begin{equation}
    \text{VPQ} = \sum_{t=0}^H \frac{\sum_{(p_t,q_t) \in TP_t} \text{IoU}(p_t,q_t)}{|TP_t| + \frac{1}{2}|FP_t| + \frac{1}{2}|FN_t|}
\end{equation}

with $TP_t$ the set of true positives at timestep $t$ (correctly detected ground truth instances), $FP_t$ the set of false positives at timestep $t$ (predicted instances that do not match any ground truth instance), and $FN_t$ the set of false negatives at timestep $t$ (ground truth instances that were not detected). A true positive corresponds to a predicted instance segmentation that has: (i) an intersection-over-union (IoU) over 0.5 with the ground truth, and (ii) an instance id that is consistent with the ground truth over time (correctly tracked).











\subsection{Training}
Our model takes $1.0$s of past context and predicts $2.0$s in the future. In NuScenes, this corresponds to $3$ frames of past temporal context and $4$ frames into the future at 2Hz. In the Lyft dataset, this corresponds to $6$ frames of past context and $10$ frames in the future at 5Hz.

For each past timestep, our model processes $6$ camera images at resolution $224\times480$. It outputs a sequence of $100\mathrm{m}\times100\mathrm{m}$ BeV predictions at $50\mathrm{cm}$ pixel resolution in both the $x$ and $y$ directions resulting in a bird's-eye view video with spatial dimension $200\times200$. We use the Adam optimiser with a constant learning rate of $3\times 10^{-4}$. We train our model on $4$ Tesla V100 GPUs with a batch size of $12$ for $20$ epochs with mixed precision.

\clearpage
\section{Results}
\subsection{Comparison with the Literature}
Since predicting future instance segmentation in bird's-eye view is a new task, we begin by comparing our model to previous published methods on bird's-eye view semantic segmentation from monocular cameras.

\begin{figure}
\captionsetup[subfigure]{justification=centering,font=scriptsize}
\centering
    \begin{subfigure}[c]{0.24\textwidth}
    \centering
    \includegraphics[width=\linewidth]{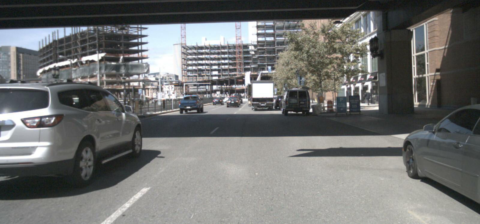}%
    \end{subfigure}
    \begin{subfigure}[c]{0.115\textwidth}
    \includegraphics[width=\linewidth]{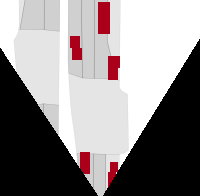}%
    \end{subfigure}
    \begin{subfigure}[c]{0.115\textwidth}
    \includegraphics[width=\linewidth]{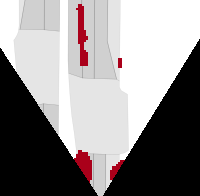}%
    \end{subfigure}
    \begin{subfigure}[c]{0.115\textwidth}
    \includegraphics[width=\linewidth]{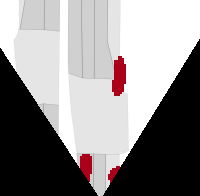}%
    \end{subfigure}
    \begin{subfigure}[c]{0.115\textwidth}
    \includegraphics[width=\linewidth]{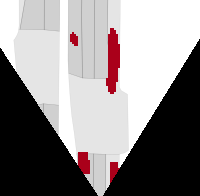}%
    \end{subfigure}
    \begin{subfigure}[c]{0.115\textwidth}
    \includegraphics[width=\linewidth]{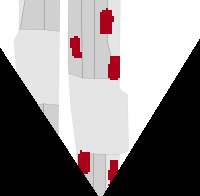}%
    \end{subfigure}
    \begin{subfigure}[c]{0.115\textwidth}
    \includegraphics[width=\linewidth]{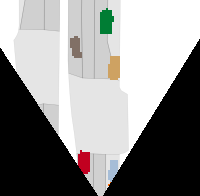}%
    \end{subfigure}
    \par\smallskip
    \begin{subfigure}[c]{0.24\textwidth}
    \centering
    \includegraphics[width=\linewidth]{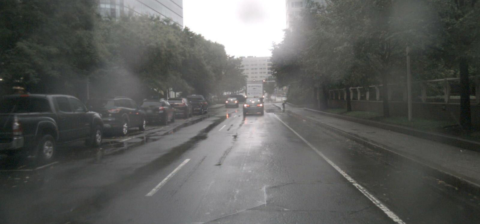}%
    \end{subfigure}
    \begin{subfigure}[c]{0.115\textwidth}
    \includegraphics[width=\linewidth]{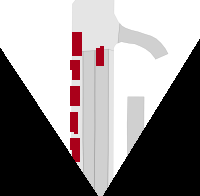}%
    \end{subfigure}
    \begin{subfigure}[c]{0.115\textwidth}
    \includegraphics[width=\linewidth]{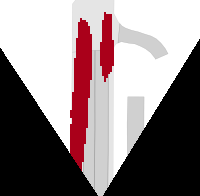}%
    \end{subfigure}
    \begin{subfigure}[c]{0.115\textwidth}
    \includegraphics[width=\linewidth]{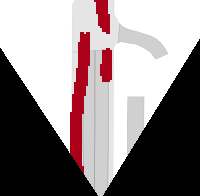}%
    \end{subfigure}
    \begin{subfigure}[c]{0.115\textwidth}
    \includegraphics[width=\linewidth]{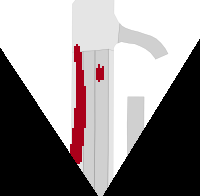}%
    \end{subfigure}
    \begin{subfigure}[c]{0.115\textwidth}
    \includegraphics[width=\linewidth]{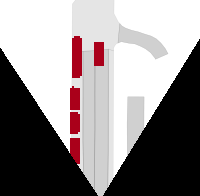}%
    \end{subfigure}
    \begin{subfigure}[c]{0.115\textwidth}
    \includegraphics[width=\linewidth]{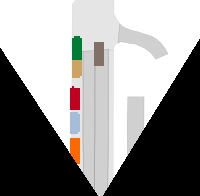}%
    \end{subfigure}

    \par\smallskip
    \begin{subfigure}[c]{0.24\textwidth}
    \centering
    \includegraphics[width=\linewidth]{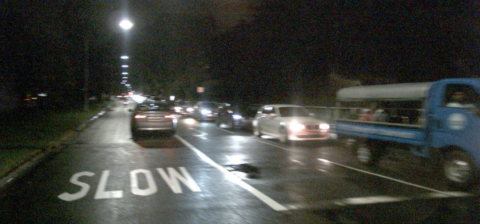}%
    \end{subfigure}
    \begin{subfigure}[c]{0.115\textwidth}
    \includegraphics[width=\linewidth]{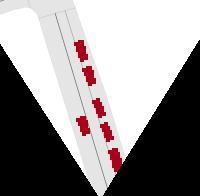}%
    \end{subfigure}
    \begin{subfigure}[c]{0.115\textwidth}
    \includegraphics[width=\linewidth]{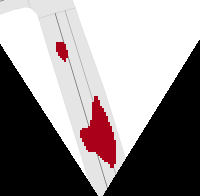}%
    \end{subfigure}
    \begin{subfigure}[c]{0.115\textwidth}
    \includegraphics[width=\linewidth]{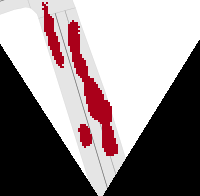}%
    \end{subfigure}
    \begin{subfigure}[c]{0.115\textwidth}
    \includegraphics[width=\linewidth]{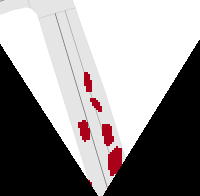}%
    \end{subfigure}
    \begin{subfigure}[c]{0.115\textwidth}
    \includegraphics[width=\linewidth]{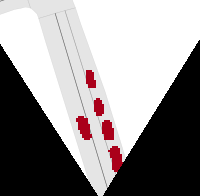}%
    \end{subfigure}
    \begin{subfigure}[c]{0.115\textwidth}
    \includegraphics[width=\linewidth]{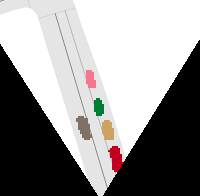}%
    \end{subfigure}

    \par\smallskip
    \begin{subfigure}[c]{0.24\textwidth}
    \centering
    \includegraphics[width=\linewidth]{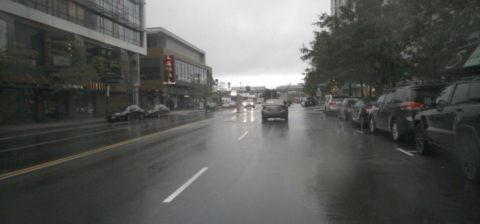}%
    \end{subfigure}
    \begin{subfigure}[c]{0.115\textwidth}
    \includegraphics[width=\linewidth]{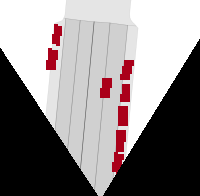}%
    \end{subfigure}
    \begin{subfigure}[c]{0.115\textwidth}
    \includegraphics[width=\linewidth]{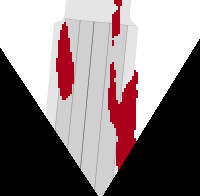}%
    \end{subfigure}
    \begin{subfigure}[c]{0.115\textwidth}
    \includegraphics[width=\linewidth]{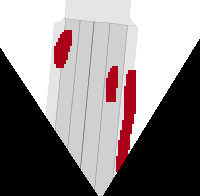}%
    \end{subfigure}
    \begin{subfigure}[c]{0.115\textwidth}
    \includegraphics[width=\linewidth]{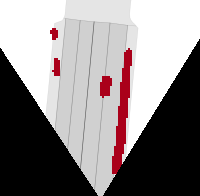}%
    \end{subfigure}
    \begin{subfigure}[c]{0.115\textwidth}
    \includegraphics[width=\linewidth]{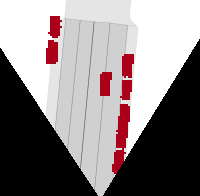}%
    \end{subfigure}
    \begin{subfigure}[c]{0.115\textwidth}
    \includegraphics[width=\linewidth]{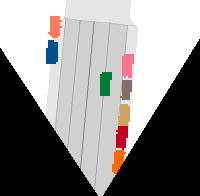}%
    \end{subfigure}

    \par\smallskip
    \begin{subfigure}[c]{0.24\textwidth}
    \centering
    \includegraphics[width=\linewidth]{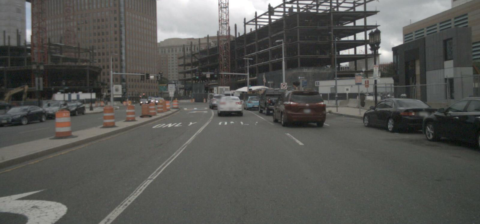}%
    \end{subfigure}
    \begin{subfigure}[c]{0.115\textwidth}
    \includegraphics[width=\linewidth]{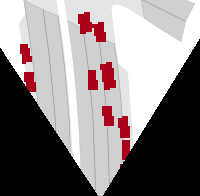}%
    \end{subfigure}
    \begin{subfigure}[c]{0.115\textwidth}
    \includegraphics[width=\linewidth]{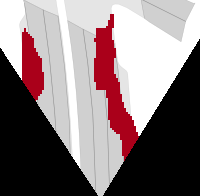}%
    \end{subfigure}
    \begin{subfigure}[c]{0.115\textwidth}
    \includegraphics[width=\linewidth]{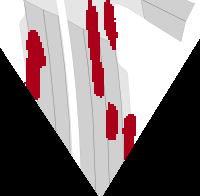}%
    \end{subfigure}
    \begin{subfigure}[c]{0.115\textwidth}
    \includegraphics[width=\linewidth]{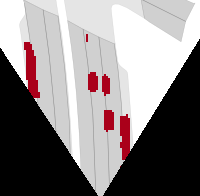}%
    \end{subfigure}
    \begin{subfigure}[c]{0.115\textwidth}
    \includegraphics[width=\linewidth]{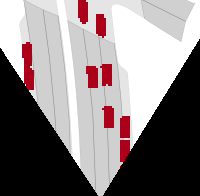}%
    \end{subfigure}
    \begin{subfigure}[c]{0.115\textwidth}
    \includegraphics[width=\linewidth]{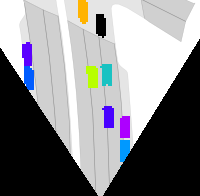}%
    \end{subfigure}

    \par\smallskip
    \begin{subfigure}[c]{0.24\textwidth}
    \centering
    \includegraphics[width=\linewidth]{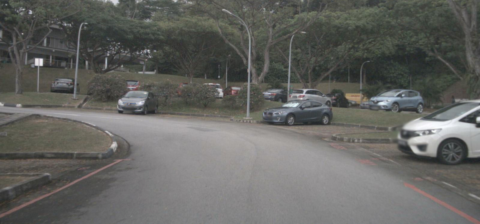}%
    \caption*{Camera input}
    \end{subfigure}
    \begin{subfigure}[c]{0.115\textwidth}
    \includegraphics[width=\linewidth]{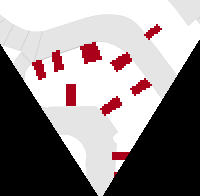}%
    \caption*{Ground truth}
    \end{subfigure}
    \begin{subfigure}[c]{0.115\textwidth}
    \includegraphics[width=\linewidth]{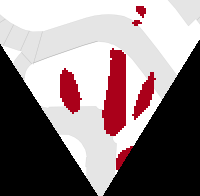}%
    \caption*{VPN}
    \end{subfigure}
    \begin{subfigure}[c]{0.115\textwidth}
    \includegraphics[width=\linewidth]{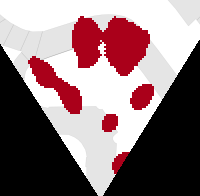}%
    \caption*{PON}
    \end{subfigure}
    \begin{subfigure}[c]{0.115\textwidth}
    \includegraphics[width=\linewidth]{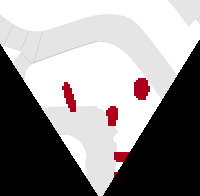}%
    \caption*{Lift-Splat}
    \end{subfigure}
    \begin{subfigure}[c]{0.115\textwidth}
    \includegraphics[width=\linewidth]{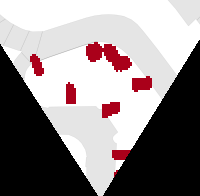}%
     \caption*{\textbf{Ours Seg.}}
    \end{subfigure}
    \begin{subfigure}[c]{0.115\textwidth}
    \includegraphics[width=\linewidth]{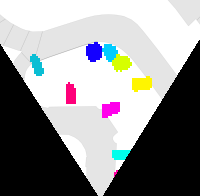}%
    \caption*{\textbf{Ours Inst.}}
    \end{subfigure}
\caption[Qualitative comparison of bird's-eye view prediction with published methods on NuScenes.]{Qualitative comparison of bird's-eye view prediction with published methods on NuScenes. The predictions of our model are much sharper and more accurate. Contrary to previous methods, FIERY can separate closely parked cars and correctly predict distant vehicles (near the top of the bird's-eye view image).}
\label{fiery-fig:qual-literature}
\end{figure}

Many previous works \citep{lu19-icra-ral,Pan_2020,roddick20,philion20,saha21} have proposed a model to output the dynamic scene bird's-eye view segmentation from multiview camera images of a single timeframe. For comparison, we adapt our model so that the past context is reduced to a single observation, and we set the future horizon $H=0$ (to only predict the present's segmentation). We call this model \emph{FIERY Static} and report the results in \Cref{fiery-table:single-timestep}. We observe that FIERY Static outperforms all previous baselines. Additionally, \Cref{fiery-fig:qual-literature} shows a qualitative comparison of the predictions from our model with previous published methods, on the task of present-frame bird's-eye view semantic segmentation. Our predictions are much sharper and more accurate.

\begin{table}
\centering
\caption[Bird's-eye view semantic segmentation on NuScenes in the settings of the respective published methods.]{Bird's-eye view semantic segmentation on NuScenes in the settings of the respective published methods. \\Setting 1: $100\mathrm{m}\times50\mathrm{m}$ at $25$cm resolution. Prediction of the present timeframe. \\Setting 2: $100\mathrm{m}\times100\mathrm{m}$ at $50$cm resolution. Prediction of the present timeframe. \\Setting 3: $32.0\mathrm{m}\times19.2\mathrm{m}$ at $10$cm resolution. Prediction $2.0$s in the future. In this last setting we compare our model to two variants of Fishing Net \citep{hendy20}: one using camera inputs, and one using LiDAR inputs.}
\begin{tabularx}{0.9\textwidth}{lYYY}
\toprule
\multicolumn{1}{c}{} & \multicolumn{3}{c}{\color{darkgray}{\textbf{Intersection-over-Union (IoU)}}}\\
& Setting 1 & Setting 2 & Setting 3\\
\midrule
VED \citep{lu19-icra-ral} & 8.8 & - & -\\
PON \citep{roddick20} & 24.7 & - & -\\
VPN \citep{Pan_2020} & 25.5 & - & -\\
STA \citep{saha21} & 36.0 & - & -\\
Lift-Splat \citep{philion20} & - & 32.1 & - \\
Fishing Camera \citep{hendy20} & - & - & 30.0\\
Fishing Lidar \citep{hendy20} & - & - & 44.3\\
FIERY Static & 37.7& 35.8& - \\
\textbf{FIERY} & \textbf{39.9}& \textbf{38.2}& \textbf{57.6}\\
\bottomrule
\end{tabularx}

\label{fiery-table:single-timestep}
\end{table}

\begin{figure}
\captionsetup[subfigure]{justification=centering}
\centering
    \begin{subfigure}[c]{0.95\textwidth}
    \centering
    \includegraphics[width=\linewidth]{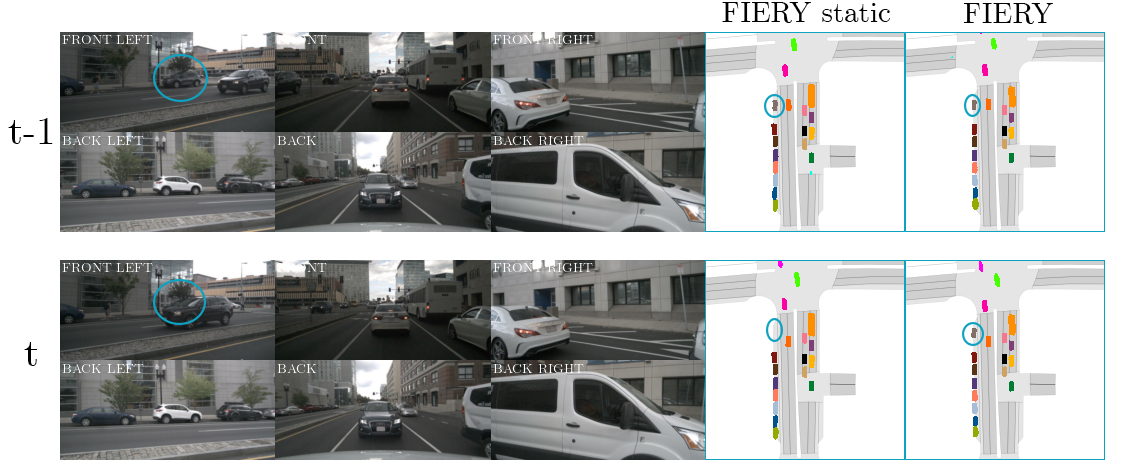}%
    \caption{The vehicle parked on the left-hand side is correctly predicted even through the occlusion.}
    \end{subfigure}
    \begin{subfigure}[c]{0.95\textwidth}
    \includegraphics[width=\linewidth]{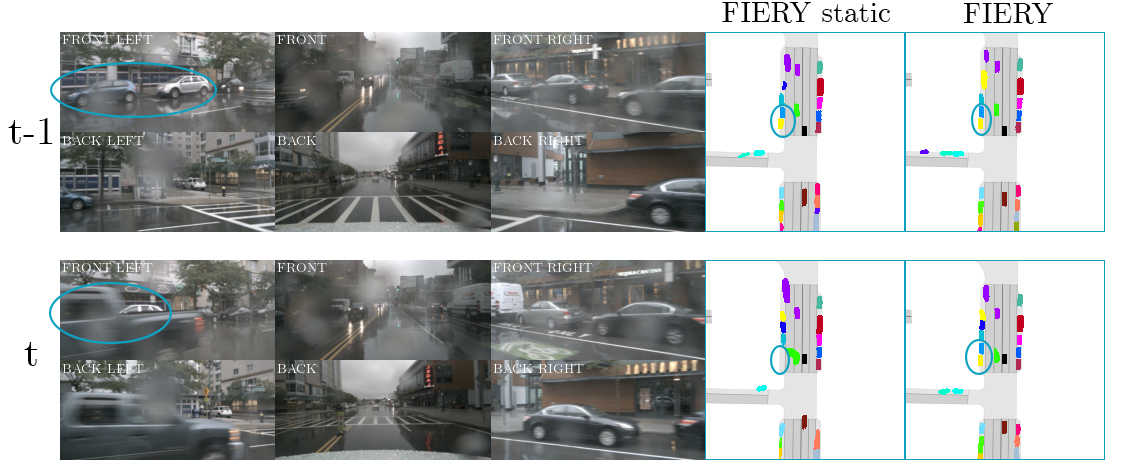}%
    \caption{The two vehicles parked on the left are heavily occluded by the 4x4 driving on the opposite lane, however by fusing past temporal information, the model is able to predict their positions accurately.}
    \end{subfigure}
\caption[Comparison of FIERY Static (no temporal context) and FIERY (1.0s of past context) on the task of present-frame bird's-eye view instance segmentation on NuScenes.]{Comparison of FIERY Static (no temporal context) and FIERY (1.0s of past context) on the task of present-frame bird's-eye view instance segmentation on NuScenes. FIERY can predict partially observable and occluded elements, as highlighted by the blue ellipses.}
\label{fiery-fig:temporal-fusion}
\end{figure}

We also train a model that takes 1.0s of past observations as context (\emph{FIERY}) and note that it achieves an even higher intersection-over-union over its single-timeframe counterpart that has no past context. This is due to our model's ability to accumulate information over time and better handle partial observability and occlusions (see \Cref{fiery-fig:temporal-fusion}).

Finally, we compare our model to Fishing Net \citep{hendy20}, where the authors predicts bird's-eye view semantic segmentation $2.0\mathrm{s}$ in the future. Fishing Net proposes two variants of their model: one using camera as inputs, and one using LiDAR as inputs. FIERY performs much better than both the camera and LiDAR models, hinting that computer vision networks are starting to become competitive with LiDAR sensing for the prediction task.

\subsection{Future Instance Prediction}
In order to compare the performance of our model in future instance segmentation and motion prediction, we introduce the following baselines:

\paragraph{Static model.} The most simple approach to model dynamic objects is to assume that they will not move and remain static. We use FIERY Static to predict the instance segmentation of the present timestep (time $t$), and repeat this prediction in the future. We call this baseline the \emph{Static model} as it should correctly detect all static vehicles, since the future labels are in the present's reference frame.
\paragraph{Extrapolation model.} Classical prediction methods \citep{fiorini98,fraichard03} extrapolate the current behaviour of dynamic agents in the future. We run FIERY Static on every past timesteps to obtain a sequence of past instance segmentations. We re-identify past instances by comparing the instance centers and running a Hungarian matching algorithm. We then obtain past trajectories of detected vehicles, which we extrapolate in the future and transform the present segmentation accordingly.
\paragraph{}
We also report the results of various ablations of our proposed architecture:
\begin{itemize}[itemsep=0.5mm, parsep=0pt]
\item \textbf{No temporal context.} This model only uses the features $\rvx_k$ from the present timestep to predict the future (i.e. we set the 3D convolutional temporal model to the identity function).
\item \textbf{No transformation.} Past bird's-eye view features $\rvx_{1:k}$ are not warped to the present's reference frame.

\item \textbf{No unrolling.} Instead of recursively predicting future features and decoding the corresponding instance information, this variant directly predicts all future instance centerness, offset, segmentation and flow without a recurrent module.

\item \textbf{No future flow.} This model does not predict future flow.

\begin{table}
\caption[Future instance segmentation in bird's-eye view for $2.0\mathrm{s}$ in the future on NuScenes.]{Future instance segmentation in bird's-eye view for $2.0\mathrm{s}$ in the future on NuScenes. We report future Intersection-over-Union (IoU) and Video Panoptic Quality (VPQ), evaluated at different ranges: $30\mathrm{m}\times30\mathrm{m}$ (Short) and $100\mathrm{m}\times100\mathrm{m}$ (Long) around the ego-vehicle. Results are reported as percentages.}
\centering
\begin{tabularx}{\textwidth}{lYYYY}
\toprule
    \multicolumn{1}{c}{} 
& \multicolumn{2}{c}{\textbf{Intersection-over-Union}}
& \multicolumn{2}{c}{\textbf{Video Panoptic Quality}}\\ 

& Short & Long & Short & Long \\
\midrule
Static model & 47.9 & 30.3 & 43.1 & 24.5 \\
Extrapolation model  & 49.2 & 30.8  & 43.8 & 24.9 \\
\cmidrule{1-5}
No temporal context & 51.7 & 32.6 & 40.3 & 24.1 \\
No transformation  & 53.0 & 33.8 & 41.7 & 24.6\\
No unrolling & 55.4 & 34.9 & 44.2 & 26.2 \\
No future flow  & 58.0 & 36.7  & 44.6 & 26.9\\
Uniform depth  & 57.1 & 36.2 & 46.8 & 27.8\\
Deterministic & 58.2 & 36.6  & 48.3 & 28.5\\
\textbf{FIERY}  & \textbf{59.4} & \textbf{36.7} & \textbf{50.2} & \textbf{29.9} \\
\bottomrule
\end{tabularx}
\label{fiery-table:nuscenes}
\end{table}

\item \textbf{Uniform depth.} We lift the features from the encoder $(\rvb_i^1, ..., \rvb_i^n)$ with the Orthographic Feature Transform \citep{roddick19} module. This corresponds to setting the depth probability distribution to a uniform distribution.

\item \textbf{Deterministic.} This variant does not include a prior and posterior distributions. Therefore there is no KL loss, and the prediction $\hat{\rvy}_{k:k+H}$ are conditioned on $\rvz_k$ only.
\end{itemize}

We report the results in \Cref{fiery-table:nuscenes} (on NuScenes) and \Cref{fiery-table:lyft} (on Lyft) of the mean prediction of our probabilistic model (i.e. we set the latent code $\rvs_k$ to the mean of the present distribution: $\rvs_k = \mu_{\theta}(\rvz_k)$).

\begin{table}[h]
\centering
\caption[Future instance prediction in bird's-eye view for $2.0$s in the future on the Lyft dataset.]{Future instance prediction in bird's-eye view for $2.0$s in the future on the Lyft dataset. We report future Intersection-over-Union and Video Panoptic Quality.}
\begin{tabularx}{0.6\textwidth}{lYY}
\toprule
    \multicolumn{1}{c}{} 
& \multicolumn{2}{c}{\textbf{IoU$\mid$VPQ}}\\ 

& Short & Long \\
\midrule
Static model &35.3$\mid$36.4& 24.1$\mid$20.7\\
Extrapolation model & 37.4$\mid$37.5 & 24.8$\mid$21.2\\
\textbf{FIERY} &  \textbf{57.8$\mid$50.0}& \textbf{36.3$\mid$29.2}\\
\bottomrule
\end{tabularx}

\label{fiery-table:lyft}
\end{table}

\subsection{Analysis}

\definecolor{grad1}{HTML}{AF2624}
\definecolor{grad2}{HTML}{912E3A}
\definecolor{grad3}{HTML}{753752}
\definecolor{grad4}{HTML}{5A406A}
\definecolor{grad5}{HTML}{434881}
\definecolor{grad6}{HTML}{2F4F97}
\definecolor{grad7}{HTML}{155AB6}
\begin{figure}[h]
\centering
\begin{tikzpicture}[scale=1.0]
\begin{axis}[
    xbar=0pt,
    /pgf/bar shift=0pt,
    legend style={
    legend columns=4,
        at={(xticklabel cs:0.5)},
        anchor=north,
        draw=none
    },
    ytick={0,...,6},
    axis y line*=left, 
    axis x line*=bottom, 
    xtick={20, 22,24,26,28,30},
    width=.6\textwidth,
    bar width=5mm,
    xmin=20,
    xmax=31,
    y=7mm, 
    enlarge y limits={abs=0.625},
    yticklabels={{\textbf{FIERY}}, 
    {Deterministic}, 
    {Uniform depth},
    {No future flow}, 
    {No unrolling}, 
    {No transformation},
    {No temporal}},
    nodes near coords,
    nodes near coords style={text=black},
    every axis plot/.append style={fill}
]
\addplot[grad1] coordinates {(29.9,0)};
\addplot[grad2] coordinates {(28.5,1)};
\addplot[grad3] coordinates {(27.8,2)};
\addplot[grad4] coordinates {(26.9,3)};
\addplot[grad5] coordinates {(26.2,4)};
\addplot[grad6] coordinates {(24.6,5)};
\addplot[grad7] coordinates {(24.1,6)};
\end{axis}  
\end{tikzpicture}
\caption[Performance comparison of various ablations of our model.]{Performance comparison of various ablations of our model. We measure future Video Panoptic Quality $2.0\mathrm{s}$ in the future on NuScenes.}
\label{fiery-fig:barplot-results}
\end{figure}
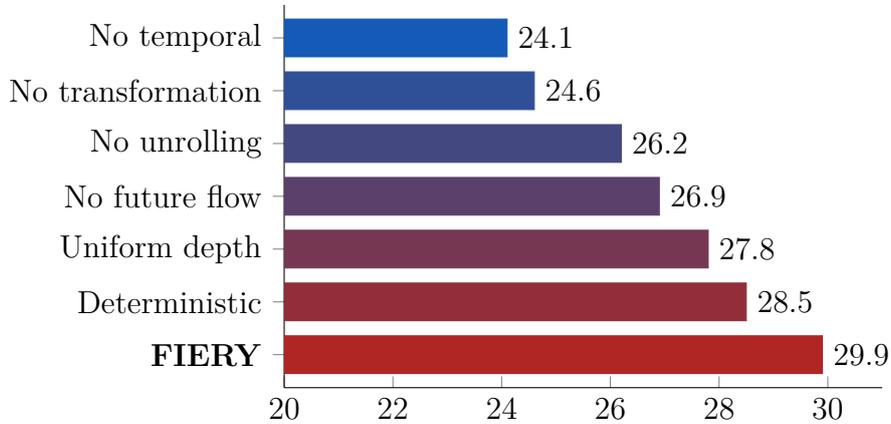

FIERY largely outperforms the Static and Extrapolation baselines for the task of future prediction. 
\Cref{fiery-fig:barplot-results} shows the performance boost our model gains from different parts of the model.

\paragraph{Temporal model.} The \emph{No temporal context} variant performs similarly to the static model. That is to be expected as this model does not have any information from the past, and cannot infer much about the motion of road agents.

\paragraph{Transformation to the present's reference frame.} There is a large performance drop when we do not transform past features to the present's reference frame. This can be explained by how much easier it is for the temporal model to learn correspondences between dynamic vehicles when the ego-motion is factored out.

Past prediction models either naively fed past images to a temporal model \citep{ballas16,hu2020probabilistic}, or did not use a temporal model altogether and simply concatenated past features \citep{luc17, hendy20}. We believe that in order to learn temporal correspondences, past features have to be mapped to a common reference frame and fed into a high capacity temporal model, such as our proposed 3D convolutional architecture.

\paragraph{Recurrent modelling.} When predicting the future, it is important to model its sequential nature, i.e. the prediction at time $k+t+1$ should be conditioned on the prediction at time $k+t$.

The \emph{No unrolling} variant which directly predicts all future instance segmentations and motions from the current state $(\rvz_k,\rvs_k)$, results in a large performance drop. This is because the sequential constraint is no longer enforced, contrarily to our approach that predicts future features in a recursive way.

\paragraph{Future motion.} Learning to predict future motion allows our model to re-identify instances using the predicted flow and comparing instance centers. Our model is the first to produce temporally consistent future instance segmentation in bird's-eye view of dynamic agents. Without future flow, the predictions are no longer temporally consistent explaining the sharp decrease in performance. 

\paragraph{Lifting the features to 3D} Using a perfect depth model we could directly lift each pixel to its correct location in 3D space. Since our depth prediction is uncertain, we instead lift the features at different possible depth locations and assign a probability mass at each location, similar to \citep{philion20}. The \emph{Uniform depth} baseline uses the Orthographic Feature Transform to lift features in 3D, by setting a uniform distribution on all depth positions. We observe that such a naive lifting performs worse compared to a learned weighting over depth.

\paragraph{Present and future distributions.} A deterministic model has a hard task at hand. It has to output with full confidence which future will happen, even though the said future is uncertain. In our probabilistic setting, the model is guided during training with the future distribution outputting a latent code that indicates the correct future. It also encourages the present distribution to cover the modes of the future distribution. This paradigm allows FIERY to predict both accurate and diverse futures.

\subsection{Visualisation of the Learned Latent Space}
\begin{figure}[h]
\captionsetup[subfigure]{justification=centering}
\centering
    \begin{subfigure}[c]{0.8\textwidth}
    \centering
    \includegraphics[width=\linewidth]{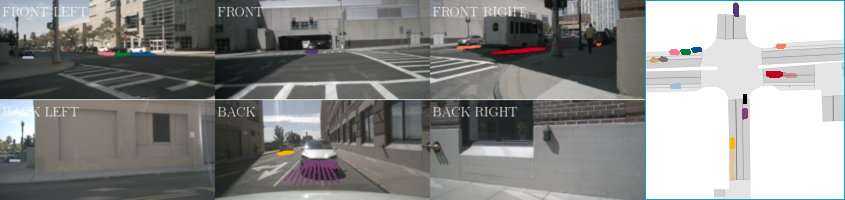}%
    \end{subfigure}
    \begin{subfigure}[c]{0.8\textwidth}
    \includegraphics[width=\linewidth]{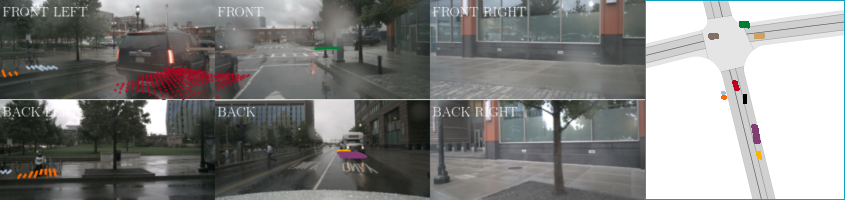}%
    \end{subfigure}
    \begin{subfigure}[c]{0.8\textwidth}
    \includegraphics[width=\linewidth]{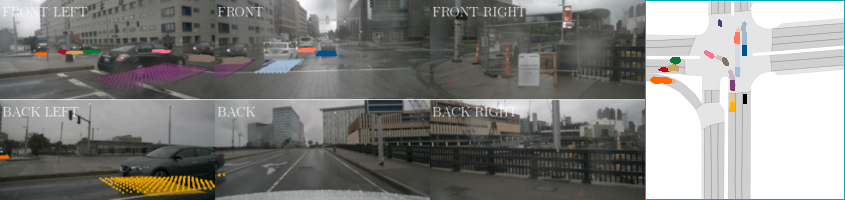}%
    \caption{Approaching an intersection.}
    \end{subfigure}
\caption[An example of cluster obtained from the spatio-temporal states $\rvz_k$ by running a Gaussian Mixture algorithm on the NuScenes validation set.]{An example of cluster obtained from the spatio-temporal states $\rvz_k$ by running a Gaussian Mixture algorithm on the NuScenes validation set. Our model learns to map similar situations to similar states. Even though the training dataset is deterministic, after mapping the RGB inputs to the state $\rvz_k$, different futures can be observed from the same starting state. This explains why our probabilistic paradigm can learn to predict diverse and plausible futures.}
\label{fiery-fig:cluster1}
\end{figure}

\begin{figure}
\captionsetup[subfigure]{justification=centering}
\centering
    \begin{subfigure}[c]{0.8\textwidth}
    \centering
    \includegraphics[width=\linewidth]{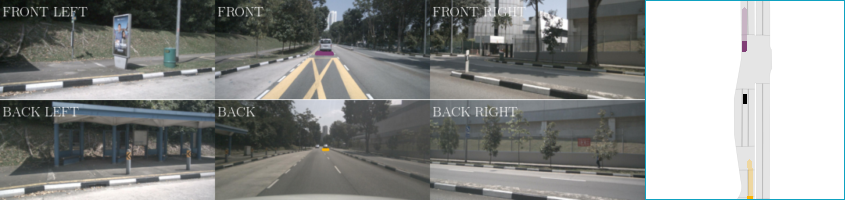}%
    \end{subfigure}
    \begin{subfigure}[c]{0.8\textwidth}
    \includegraphics[width=\linewidth]{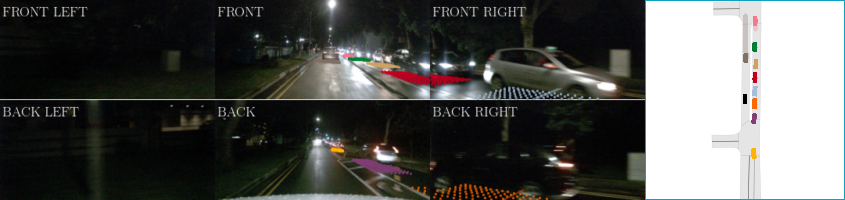}%
    \end{subfigure}
    \begin{subfigure}[c]{0.8\textwidth}
    \includegraphics[width=\linewidth]{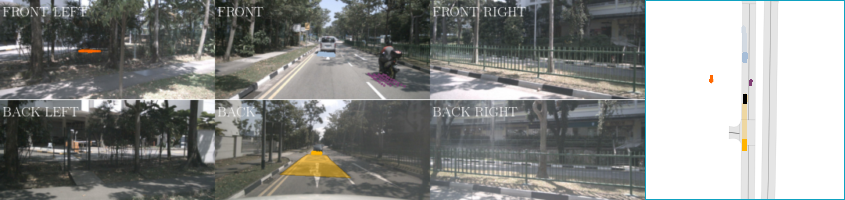}%
    \caption{Cruising behind a vehicle.}
    \end{subfigure}
    \par\smallskip
    \begin{subfigure}[c]{0.8\textwidth}
    \centering
    \includegraphics[width=\linewidth]{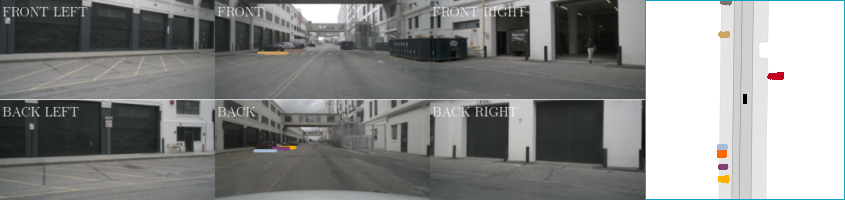}%
    \end{subfigure}
    \begin{subfigure}[c]{0.8\textwidth}
    \includegraphics[width=\linewidth]{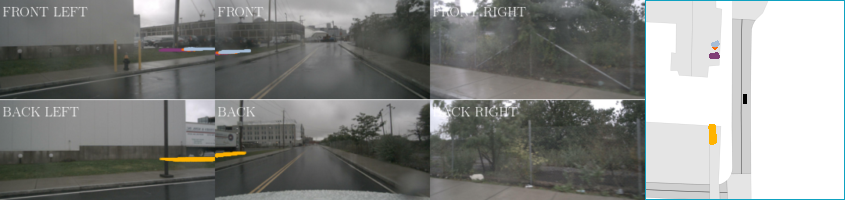}%
    \end{subfigure}
    \begin{subfigure}[c]{0.8\textwidth}
    \includegraphics[width=\linewidth]{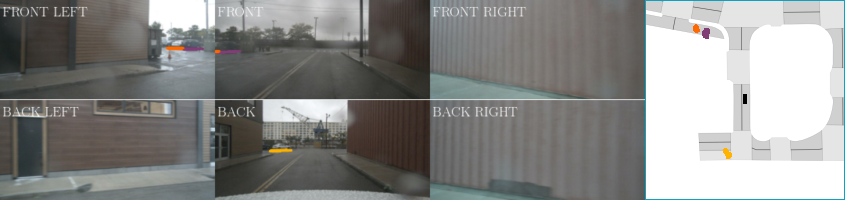}%
    \caption{Driving on open road.}
    \end{subfigure}
\caption{More example of clusters (1).}
\label{fiery-fig:cluster2-3}
\end{figure}

\begin{figure}
\captionsetup[subfigure]{justification=centering}
\centering
    \begin{subfigure}[c]{0.8\textwidth}
    \centering
    \includegraphics[width=\linewidth]{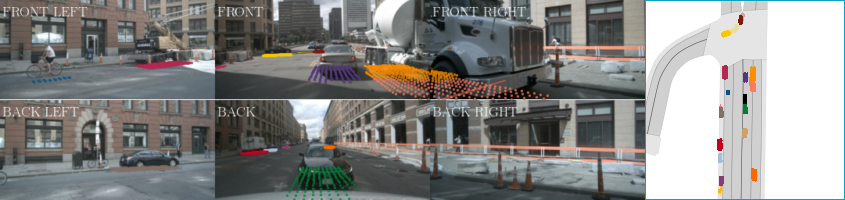}%
    \end{subfigure}
    \begin{subfigure}[c]{0.8\textwidth}
    \includegraphics[width=\linewidth]{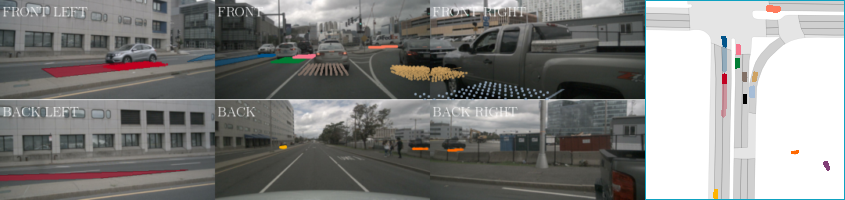}%
    \caption{Stuck in traffic.}
    \end{subfigure}
    \par\smallskip
    \begin{subfigure}[c]{0.8\textwidth}
    \centering
    \includegraphics[width=\linewidth]{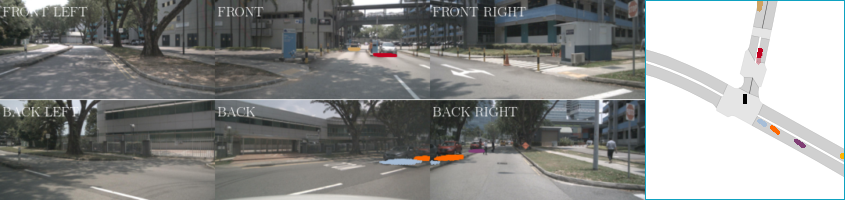}%
    \end{subfigure}
    \begin{subfigure}[c]{0.8\textwidth}
    \includegraphics[width=\linewidth]{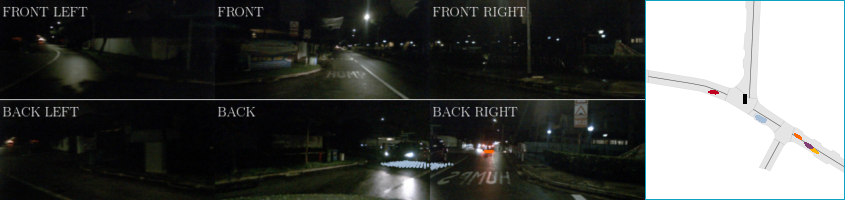}%
    \end{subfigure}
    \begin{subfigure}[c]{0.8\textwidth}
    \includegraphics[width=\linewidth]{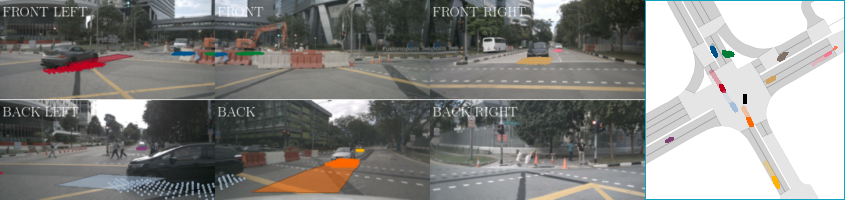}%
    \caption{Turning right at an intersection.}
    \end{subfigure}
\caption{More example of clusters (2).}
\label{fiery-fig:cluster4-5}
\end{figure}

We run a Principal Component Analysis on the latent feature $\rvz_k$ and a Gaussian Mixture algorithm on the projected features in order to obtain clusters. We then visualise the inputs and predictions of the clusters in \Cref{fiery-fig:cluster1,fiery-fig:cluster2-3,fiery-fig:cluster4-5}. We observe that examples in a given cluster correspond to similar scenarios. Therefore, we better understand why our model is able to learn diverse and multimodal futures from a deterministic training dataset. Since similar scenes are mapped to the same feature $\rvz_k$, our model will effectively observe different futures starting from the same initial state. The present distribution will thus be able to learn to capture the different modes in the future.

\paragraph{Limitations.} The clusters obtained in \Cref{fiery-fig:cluster1}, \Cref{fiery-fig:cluster2-3} and \Cref{fiery-fig:cluster4-5} were all interpretable, however the method could also yield a few clusters in which the examples were hard to link to each other from visual inspection. 

\clearpage

\clearpage
\section{Summary}
Autonomous driving requires decision making in multimodal scenarios, where the present state of the world is not always sufficient to reason correctly alone. Predictive models estimating the future state of the world -- particularly other dynamic agents -- are therefore a key component to robust driving. We presented the first prediction model of dynamic agents for autonomous driving in bird's-eye view from surround RGB videos. We posed this as an end-to-end learning problem in which our network models future stochasticity with a variational distribution. We demonstrated that FIERY predicts temporally consistent future instance segmentations and motion and is able to model diverse futures accurately.

The previous chapter (\Cref{chapter:video-scene-understanding}) was about modelling static and dynamic scene in the perspective image space. This chapter (\Cref{chapter:instance-prediction}) showed it was possible for the world model to reason in the bird's-eye view space, which is better suited for autonomous driving. In the next chapter (\Cref{chapter:imitation-learning}), we are going to connect the dots and jointly predict static scene, dynamic scene, and ego-behaviour in the bird's-eye view space. 

\chapter{Learning a World Model and a Driving Policy}
\label{chapter:imitation-learning}

\graphicspath{{Chapter6/Figures/}}

An accurate model of the environment and the dynamic agents acting in it offers great potential for improving motion planning. In this chapter, we present MILE: a Model-based Imitation LEarning approach to jointly learn a model of the world and a policy for autonomous driving. Our method leverages 3D geometry as an inductive bias and learns a highly compact latent space directly from high-resolution videos of expert demonstrations. Our model is trained on an offline corpus of urban driving data, without any online interaction with the environment. MILE improves upon prior state-of-the-art by 31\% in driving score on the CARLA simulator when deployed in a completely new town and new weather conditions. Our model can predict diverse and plausible states and actions, that can be interpretably decoded to bird's-eye view semantic segmentation. Further, we demonstrate that it can execute complex driving manoeuvres from plans entirely predicted in imagination. Our approach is the first camera-only method that models static scene, dynamic scene, and ego-behaviour in an urban driving environment.  The content of this chapter was published in the Advances in Neural Information Processing Systems, NeurIPS 2022 as \citet{hu2022mile}.

\section{Introduction}

From an early age we start building internal representations of the world through observation and interaction \citep{barlow1989unsupervised}. 
Our ability to estimate scene geometry and dynamics is paramount to generating complex and adaptable movements. 
This accumulated knowledge of the world, part of what we often refer to as common sense, allows us to navigate effectively in unfamiliar situations \citep{wolpert1998multiple}. 

In this chapter, we present MILE, a Model-based Imitation LEarning approach to jointly learn a model of the world and a driving policy. 
We demonstrate the effectiveness of our approach in the autonomous driving domain, operating on complex visual inputs labelled only with expert action and semantic segmentation. Unlike prior work on world models~\citep{ha18,hafner2019planet,hafner2021dreamerv2}, our method does not assume access to a ground truth reward, nor does it need any online interaction with the environment. Further, previous environments in OpenAI~Gym~\citep{ha18}, MuJoCo~\citep{hafner2019planet}, and Atari~\citep{hafner2021dreamerv2} were characterised by simplified visual inputs as small as $64\times64$ images. In contrast, MILE operates on high-resolution camera observations of urban driving scenes.

Driving inherently requires a geometric understanding of the environment, and MILE exploits 3D geometry as an important inductive bias by first lifting image features to 3D and pooling them into a bird's-eye view (BeV) representation.
The evolution of the world is modelled by a latent dynamics model that infers compact latent states from observations and expert actions.
The learned latent state is the input to a driving policy that outputs vehicle control, and can additionally be decoded to BeV segmentation for visualisation and as a supervision signal.

Our method also relaxes the assumption made in some recent work~\citep{chen2021learning,sobal2022separating} that neither the agent nor its actions influence the environment.
This assumption rarely holds in urban driving, and therefore MILE is action-conditioned, allowing us to model how other agents respond to ego-actions.
We show that our model can predict plausible and diverse futures from latent states and actions over long time horizons. It can even predict entire driving plans in imagination to successfully execute complex driving manoeuvres, such as negotiating a roundabout, or swerving to avoid a motorcyclist (see videos in the supplementary material).

We showcase the performance of our model on the driving simulator CARLA~\citep{dosovitskiy17b}, and demonstrate a new state-of-the-art. 
MILE achieves a 31\% improvement in driving score with respect to previous  methods~\citep{zhang2021end,chen2022learning} when tested in a new town and new weather conditions. 
Finally, during inference, because we model time with a recurrent neural network, we can maintain a single state that summarises all the past observations and then efficiently update the state when a new observation is available. We demonstrate that this design decision has important benefits for deployment in terms of latency, with negligible impact on the driving performance.

To summarise the main contributions of this chapter:
\begin{itemize}
    \item We introduce a novel model-based imitation learning architecture that scales to the visual complexity of autonomous driving in urban environments by leveraging 3D geometry as an inductive bias. Our method is trained solely using an offline corpus of expert driving data, and does not require any interaction with an online environment or access to a reward, offering strong potential for real-world application.
    \item Our camera-only model sets a new state-of-the-art on the CARLA simulator, surpassing other approaches, including those requiring LiDAR inputs.
    \item Our model predicts a distribution of diverse and plausible futures states and actions. We demonstrate that it can execute complex driving manoeuvres from plans entirely predicted in imagination (see the accompanying \href{https://wayve.ai/blog/learning-a-world-model-and-a-driving-policy/}{blog post}).

\end{itemize}
\clearpage
\section{Related Work}
MILE is at the intersection of imitation learning, 3D scene representation, and world modelling.

\paragraph{Imitation learning.}
Despite that the first end-to-end method for autonomous driving was envisioned more than 30 years ago~\citep{pomerleau1988alvinn}, early autonomous driving approaches were dominated by modular frameworks, where each module solves a specific task~\citep{bacha2008odin,dolgov2008practical,leonard2008perception}. 
Recent years have seen the development of several end-to-end self-driving systems that show strong potential to improve driving performance by predicting driving commands from high-dimensional observations alone. Conditional imitation learning has proven to be one successful method to learn end-to-end driving policies that can be deployed in simulation~\citep{codevilla2018end} and real-world urban driving scenarios~\citep{hawke2020urban}. Nevertheless, difficulties of learning end-to-end policies from high-dimensional visual observations and expert trajectories alone have been highlighted~\citep{codevilla2019exploring}.

Several works have attempted to overcome such difficulties by moving past pure imitation learning. DAgger~\citep{ross2011reduction} proposes iterative dataset aggregation to collect data from trajectories that are likely to be experienced by the policy during deployment.   NEAT~\citep{Chitta2021ICCV} additionally supervises the model with BeV semantic segmentation. ChauffeurNet~\citep{bansal2018chauffeurnet} exposes the learner to synthesised perturbations of the expert data in order to produce more robust driving policies.
Learning from All Vehicles (LAV)~\citep{chen2022learning} boosts sample efficiency by learning behaviours from not only the ego vehicle, but from all the vehicles in the scene. Roach~\citep{zhang2021end} presents an agent trained with supervision from a reinforcement learning coach that was trained on-policy and with access to privileged information.

\paragraph{3D scene representation.}
Successful planning for autonomous driving requires being able to understand and reason about the 3D scene, and this can be challenging from monocular cameras. One common solution is to condense the information from multiple cameras to a single bird's-eye representation of the scene. This can be achieved by lifting each image in 3D (by learning a depth distribution of features) and then splatting all frustums into a common rasterised BeV grid~\citep{philion20,saha21,hu2021fiery}.
An alternative approach is to rely on transformers to learn the direct mapping from image to bird's-eye view \citep{peng2022bevsegformer, gosala22bev,li22}, without explicitly modelling depth.

\paragraph{World models.}
Model-based methods have mostly been explored in a reinforcement learning setting and have been shown to be extremely successful~\citep{ha18,schrittwieser2020mastering,hafner2021dreamerv2}. These methods assume access to a reward, and online interaction with the environment, although progress has been made on fully offline reinforcement learning \citep{PLAS_corl2020,yu2020mopo}. Model-based imitation learning has emerged as an alternative to reinforcement learning in robotic manipulation~\citep{englert2013probabilistic} and OpenAI~Gym~\citep{kidambi2021mobile}. Even though these methods do not require access to a reward, they still require online interaction with the environment to achieve good performance.


Learning the latent dynamics of a world model from image observations was first introduced in video prediction \citep{babaeizadeh18,denton18,franceschi20}. Most similar to our approach, \citep{hafner2019planet,hafner2021dreamerv2} additionally modelled the reward function and optimised a policy inside their world model. Contrarily to prior work, our method does not assume access to a reward function, and directly learns a policy from an offline dataset. Additionally, previous methods operate on simple visual inputs, mostly of size $64\times 64$. In contrast, MILE is able to learn the latent dynamics of complex urban driving scenes from high resolution $600\times960$ input observations, which is important to ensure small details such as traffic lights can be perceived reliably.

\paragraph{Occupancy grid maps.} Occupancy grid maps (OGMs) are bird's-eye view representation of a scene. \citet{itkina19,toyungyernsub21,lange21,mohajerin19}
predicted future static and dynamic OGMs from past OGMs in urban driving scenes. They framed environment prediction as a video prediction problem. The static and dynamic OGMs are obtained by processing LiDAR measurements and 3D object detections. 

Although the output space between their method and our proposed approach is similar (BeV representation), there are three key differences. (i) Both the inputs and outputs of their model are OGMs. This means the static scene can be perfectly predicted by estimating ego-motion from past inputs. The motion prediction of dynamic agents is also made easier as the vehicles are already represented in a metric BeV space. In contrast, our proposed model operates on high-dimensional camera images and has to do all the heavy lifting to reason about 3D geometry and semantics from images in order to predict BeV outputs. (ii) Their methods are deterministic, whereas our approach is probabilistic and can predict multimodal futures. (iii) They do not model ego-behaviour. We jointly predict future states and actions, and demonstrate the efficacy of the learned policy in a driving simulator.  

\paragraph{Trajectory forecasting.} The goal of trajectory forecasting is to estimate the future trajectories of dynamic agents using past physical states (e.g. position, velocity), and scene context (e.g. as an offline HD map) \citep{desire17,precog19,zhao-tnt20,salzmann-trajectron20}. World models build a latent representation of the environment that explains the observations from the sensory inputs of the ego-agent (e.g. camera images) conditioned on their actions. While trajectory forecasting methods only model the dynamic scene, world models jointly reason on static and dynamic scenes. The future trajectories of moving agents is implicitly encoded in the learned latent representation of the world model, and could be explicitly decoded given we have access to future trajectory labels.

\citep{desire17,zhao-tnt20,salzmann-trajectron20} forecast the future trajectory of moving agents, but did not control the ego-agent. They focused on the prediction problem and not on learning expert behaviour from demonstrations. \citep{rhinehart20} inferred future trajectories of the ego-agent from expert demonstrations, and conditioned on some specified goal to perform new tasks. \citep{precog19} extended their work to jointly model the future trajectories of moving agents as well as of the ego-agent.

Our proposed model jointly models the motion of other dynamics agents, the behaviour of the ego-agent, as well as the static scene. Contrary to prior work, we do not assume access to ground truth physical states (position, velocity) or to an offline HD map for scene context. Our approach is the first camera-only method that models static scene, dynamic scene, and ego-behaviour in an urban driving environment.
\clearpage
\section{MILE: Model-based Imitation LEarning}
\label{mile-section:world-model}

This chapter is almost identical to the framework described in \Cref{chapter:generative-models}, except that it also assumes the existence of labels $\rvy_{1:T}$ in addition to the observations $\rvo_{1:T}$ in order to inject domain knowledge from computer vision in the latent states $\rvs_{1:T}$. The labels are bird's-eye view semantic segmentation.

\begin{figure}[h!]
    \centering
    \includegraphics[width=\linewidth]{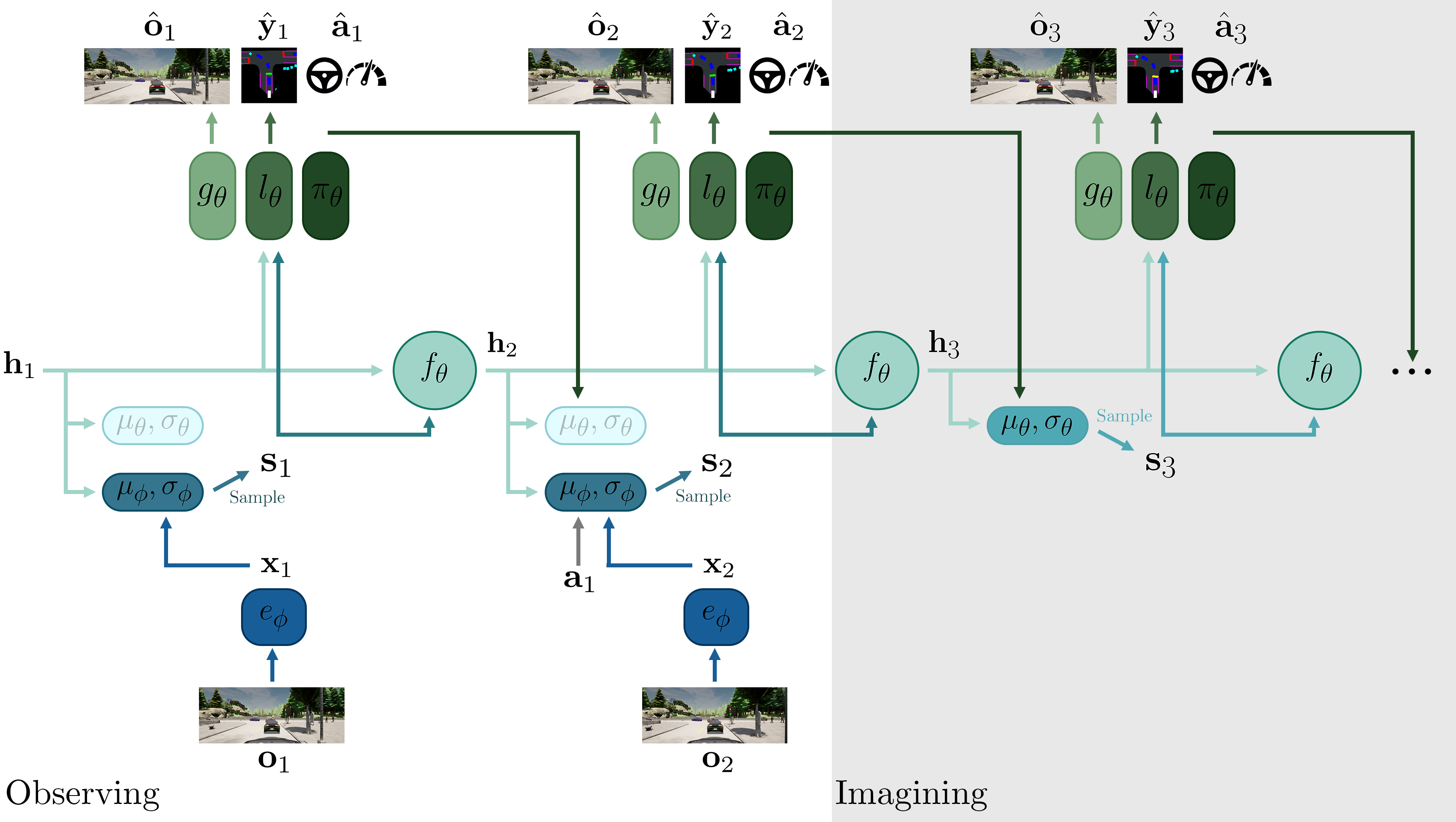}%
    \captionsetup{singlelinecheck=off}
    \caption[Architecture of MILE.]{Architecture of MILE.
    \begin{enumerate}[label=(\roman*), itemsep=0.5mm, parsep=0pt]
        \item The goal is to infer the \textbf{latent dynamics} $(\rvh_{1:T}, \rvs_{1:T})$ that generated the observations $\rvo_{1:T}$, the expert actions $\rva_{1:T}$ and the bird's-eye view labels $\rvy_{1:T}$. The latent dynamics contains a deterministic history $\rvh_t$ and a stochastic state $\rvs_t$.
        \item The \textbf{inference model}, with parameters $\phi$, estimates the posterior distribution of the stochastic state $q(\rvs_t | \rvo_{\le t}, \rva_{<t}) \sim \mathcal{N}(\mu_{\phi}(\rvh_{t}, \rva_{t-1},\rvx_t), \sigma_{\phi}(\rvh_{t}, \rva_{t-1}, \rvx_t)\bm{I})$ with $\rvx_t = e_{\phi}(\rvo_t)$. $e_{\phi}$ is the observation encoder that lifts image features to 3D, pools them to bird's-eye view, and compresses to a 1D vector.
        \item The \textbf{generative model}, with parameters $\theta$, estimates the prior distribution of the stochastic state $p(\rvs_t|\rvh_{t-1}, \rvs_{t-1}) \sim \mathcal{N}(\mu_{\theta}(\rvh_{t}, \hat{\rva}_{t-1}), \sigma_{\theta}(\rvh_t, \hat{\rva}_{t-1})\bm{I})$, with $\rvh_{t} = f_{\theta}(\rvh_{t-1}, \rvs_{t-1})$ the deterministic transition, and $\hat{\rva}_{t-1} = \pi_{\theta}(\rvh_{t-1}, \rvs_{t-1})$ the predicted action. It additionally estimates the distributions of the observation $p(\rvo_t|\rvh_t, \rvs_t) \sim \mathcal{N}(g_{\theta}(\rvh_t, \rvs_t), \bm{I})$, the bird's-eye view segmentation $p(\rvy_t| \rvh_t, \rvs_t) \sim \mathrm{Categorical}(l_{\theta}(\rvh_t, \rvs_t))$, and the action 
        $p(\rva_t| \rvh_t, \rvs_t) \sim \mathrm{Laplace}(\pi_{\theta}(\rvh_t, \rvs_t), \mathbf{1})$.
        \item In the diagram, we represented our model observing inputs for $T=2$ timesteps, and then imagining future latent states and actions for one step.
    \end{enumerate}}
    \label{mile-fig:architecture}
\end{figure}

In this section, we present MILE: our method that learns to jointly control an autonomous vehicle and model the world and its dynamics. An overview of the architecture is presented in \Cref{mile-fig:architecture} and the full description of the network can be found in \Cref{mile-appendix:model-description}. We begin by defining the generative model (\Cref{mile-subsection:probabilistic_model}), and then derive the inference model (\Cref{mile-subsection:variational_inference}). \Cref{mile-subsection:inference_model} and \Cref{mile-subsection:generative_model} describe the neural networks that parametrise the inference and generative models respectively. Finally, in \Cref{mile-subsection:imagine} we show how our model can predict future states and actions to drive in imagination.

\subsection{Probabilistic Generative Model}
\label{mile-subsection:probabilistic_model}

Let $\rvo_{1:T}$ be a sequence of $T$ video frames with associated expert actions $\rva_{1:T}$ and ground truth BeV semantic segmentation labels $\rvy_{1:T}$. We model their evolution by introducing latent variables $\rvs_{1:T}$ that govern the temporal dynamics. The initial distribution is parameterised as $\rvs_1 \sim \mathcal{N}(\rvzero, \bm{I})$, and we additionally introduce a variable $\rvh_1 \sim \delta(\rvzero)$ that serves as a deterministic history. The transition consists of (i) a deterministic update $\rvh_{t+1} = f_{\theta}(\rvh_t, \rvs_t)$ that depends on the past history $\rvh_t$ and past state $\rvs_t$, followed by (ii) a stochastic update $\rvs_{t+1}  \sim \mathcal{N}(\mu_{\theta}(\rvh_{t+1}, \rva_t), \sigma_{\theta}(\rvh_{t+1}, \rva_t)\bm{I})$, where we parameterised $\rvs_t$ as a normal distribution with diagonal covariance.
We model these transitions with neural networks: $f_{\theta}$ is a gated recurrent cell, and $(\mu_{\theta}, \sigma_{\theta})$ are  multi-layer perceptrons. The full probabilistic model is given by \Cref{equation:full-model}.

\begin{align}
    \label{equation:full-model}
    \begin{cases}
        \rvh_1 &\sim \delta(\rvzero)\\
        \rvs_1 &\sim \mathcal{N}(\rvzero, \bm{I})\\
        \rvh_{t+1} |\rvh_t, \rvs_t &= f_{\theta}(\rvh_t, \rvs_t) \\
        \rvs_{t+1} | \rvh_{t+1}, \rva_t& \sim \mathcal{N}(\mu_{\theta}(\rvh_{t+1}, \rva_t), \sigma_{\theta}(\rvh_{t+1}, \rva_t)\bm{I}) \\
        \rvo_t |\rvh_t, \rvs_t&\sim \mathcal{N}(g_{\theta}(\rvh_t, \rvs_t), \bm{I}) \\
        \rvy_t | \rvh_t, \rvs_t&\sim \mathrm{Categorical}(l_{\theta}(\rvh_t, \rvs_t))  \\
        \rva_t | \rvh_t, \rvs_t&\sim \mathrm{Laplace}(\pi_{\theta}(\rvh_t, \rvs_t), \mathbf{1})
    \end{cases}
\end{align}
with $\delta$ the Dirac delta function, $g_{\theta}$ the image decoder, $l_{\theta}$ the BeV decoder, and $\pi_{\theta}$ the policy, which will be described in \Cref{mile-subsection:generative_model}.

\subsection{Variational Inference}
\label{mile-subsection:variational_inference}
Following the generative model described in \Cref{equation:full-model}, we can factorise the joint probability as:
\begin{equation}
    \label{eq:generative-model}
    \begin{split}
    p(\rvo_{1:T},\rvy_{1:T}, \rva_{1:T}, \rvh_{1:T}, \rvs_{1:T}) =\prod_{t=1}^T p(\rvh_t, \rvs_t|\rvh_{t-1},\rvs_{t-1},\rva_{t-1})p(\rvo_t,\rvy_t,\rva_t|\rvh_t, \rvs_t)
    \end{split}
\end{equation}

with
\begin{align}
    p(\rvh_t, \rvs_t|\rvh_{t-1},\rvs_{t-1},\rva_{t-1}) &= p(\rvh_t|\rvh_{t-1},\rvs_{t-1})p(\rvs_t|\rvh_t,\rva_{t-1}) \label{eq:generative-factorisation}\\
    p(\rvo_t,\rvy_t,\rva_t|\rvh_t, \rvs_t) &= p(\rvo_t|\rvh_t, 
    \rvs_t)p(\rvy_t|\rvh_t, \rvs_t)p(\rva_t|\rvh_t, \rvs_t)
\end{align}

Given that $\rvh_t$ is deterministic according to \Cref{equation:full-model}, we have $p(\rvh_t|\rvh_{t-1},\rvs_{t-1})=\delta(\rvh_t -f_{\theta}(\rvh_{t-1},\rvs_{t-1}))$. Therefore, in order to maximise the marginal likelihood of the observed data $p(\rvo_{1:T},\rvy_{1:T}, \rva_{1:T})$, we need to infer the latent variables $\rvs_{1:T}$. 
We do this through deep variational inference by introducing a variational distribution $q_{H,S}$ defined and factorised as follows:

\begin{equation}
    \label{eq:inference-model}
    q_{H,S} \triangleq q(\rvh_{1:T},\rvs_{1:T}|\rvo_{1:T}, \rva_{1:T-1}) = \prod_{t=1}^T q(\rvh_t|\rvh_{t-1},\rvs_{t-1})q(\rvs_t|\rvo_{\le t},\rva_{<t})
\end{equation}

with $q(\rvh_t|\rvh_{t-1},\rvs_{t-1})=p(\rvh_t|\rvh_{t-1},\rvs_{t-1})$, the Delta dirac function defined above, and $q(\rvh_1)=\delta(\rvzero)$. We parameterise this variational distribution with a neural network with weights $\phi$. By applying Jensen's inequality, we can obtain a variational lower bound on the log evidence:

\begin{alignat}{2}
\log p(&\rvo_{1:T} &&, \rvy_{1:T}, \rva_{1:T}) \ge ~\mathcal{L}(\rvo_{1:T}, \rvy_{1:T}, \rva_{1:T} ; \theta, \phi) \notag \\
 &\triangleq&& \sum_{t=1}^T \E_{q(\rvh_{1:t},\rvs_{1:t}|\rvo_{\le t}, \rva_{<t})}\bigg[\underbrace{\log p(\rvo_t|\rvh_t, \rvs_t)}_{\text{image reconstruction}} +~ \underbrace{\log p(\rvy_t|\rvh_t, \rvs_t)}_{\text{bird's-eye segmentation}} ~+~ \underbrace{\log p(\rva_t|\rvh_t,\rvs_t)}_{\text{action}}\bigg]  \notag \\
& && - \sum_{t=1}^T \E_{q(\rvh_{1:t-1},\rvs_{1:t-1}|\rvo_{\le t-1}, \rva_{<t-1})}\bigg[\underbrace{\KL\Big( q(\rvs_t | \rvo_{\le t}, \rva_{<t}) ~||~ p(\rvs_t | \rvh_{t-1}, \rvs_{t-1}) \Big)}_{\text{posterior and prior matching}}\bigg]
\end{alignat}

Please refer to \Cref{mile-appendix:lower-bound} for the full derivation. We model $q(\rvs_t | \rvo_{\le t}, \rva_{<t})$ as a Gaussian distribution so that the Kullback-Leibler (KL) divergence can be computed in closed-form. Given that the image observations $\rvo_t$ are modelled as Gaussian distributions with unit variance, the resulting loss is the mean-squared error. Similarly, the action being modelled as a Laplace distribution and the BeV labels as a categorical distribution, the resulting losses are, respectively, $L_1$ and cross-entropy. The expectations over the variational distribution can be efficiently approximated with a single sequence sample from $q_{H,S}$, and backpropagating gradients with the reparametrisation trick \citep{kingma14}.

\subsection{Inference Network \texorpdfstring{$\phi$}{}}
\label{mile-subsection:inference_model}
The inference network, parameterised by $\phi$, models $q(\rvs_t | \rvo_{\le t}, \rva_{<t})$, which approximates the true posterior $p(\rvs_t | \rvo_{\le t}, \rva_{<t})$. It is constituted of two elements: the observation  encoder $e_{\phi}$, that embeds input images, route map and vehicle control sensor data to a low-dimensional vector, and the posterior network $(\mu_{\phi},\sigma_{\phi})$, that estimates the probability distribution of the Gaussian posterior.

\subsubsection{Observation Encoder}
The state of our model should be compact and low-dimensional in order to effectively learn dynamics. Therefore, we need to embed the high resolution input images to a low-dimensional vector. Naively encoding this image to a 1D vector similarly to an image classification task results in poor performance as shown in \Cref{mile-section:ablation-studies}. Instead, we explicitly encode 3D geometric inductive biases in the model.

\textbf{Lifting image features to 3D.}
Since autonomous driving is a geometric problem where it is necessary to reason on the static scene and dynamic agents in 3D, we first lift the image features to 3D. More precisely, we encode the image inputs $\rvo_t \in \mathbb{R}^{3\times H \times W}$ with an image encoder to extract features $\rvu_t \in \mathbb{R}^{C_e\times H_e \times W_e}$. Then similarly to \citet{philion20}, we predict a depth probability distribution for each image feature along a predefined grid of depth bins $\rvd_t \in \mathbb{R}^{D\times H_e, \times W_e}$. Using the depth probability distribution, the camera intrinsics $K$ and extrinsics $M$, we can lift the image features to 3D: $\mathrm{Lift}(\rvu_t, \rvd_t, K^{-1}, M)) \in \mathbb{R}^{C_e\times D \times H_e \times D_e \times 3}$. 

\textbf{Pooling to BeV.}
The 3D feature voxels are then sum-pooled to BeV space using a predefined grid with spatial extent $H_b\times W_b$ and spatial resolution $b_{\mathrm{res}}$. The resulting feature is $\rvb_t \in \mathbb{R}^{C_e\times H_b \times W_b}$.

\textbf{Mapping to a 1D vector.}
In traditional computer vision tasks (e.g. semantic segmentation \citep{chen17}, depth prediction \citep{godard19}), the bottleneck feature is usually a spatial tensor, in the order of $10^5-10^6$ features. Such high dimensionality is prohibitive for a world model that has to match the distribution of the priors (what it thinks will happen given the executed action) to the posteriors (what actually happened by observing the image input). Therefore, using a convolutional backbone, we compress the BeV feature $\rvb_t$ to a single vector $\rvx'_t \in \mathbb{R}^{C'}$.  As shown in \Cref{mile-section:ablation-studies}, we found it critical to compress in BeV space rather than directly in image space.

\textbf{Route map and speed.}
We provide the agent with a goal in the form of a route map \citep{zhang2021end}, which is a small grayscale image indicating to the agent where to navigate at intersections. The route map is encoded using a convolutional module resulting in a 1D feature $\rvr_t$. The current speed is encoded with fully connected layers as $\rvm_t$. At each timestep $t$, the observation embedding $\rvx_t$ is the concatenation of the image feature, route map feature and speed feature: $\rvx_t = [\rvx'_t, \rvr_t, \rvm_t] \in \mathbb{R}^C$, with $C=512$

\subsubsection{Posterior Network}
The posterior network $(\mu_{\phi},\sigma_{\phi})$ estimates the parameters of the variational distribution $q(\rvs_t | \rvo_{\le t},\rva_{<t}) \sim \mathcal{N}\left( \mu_{\phi}(\rvh_t, \rva_{t-1}, e_{\phi}(\rvo_t)), \sigma_{\phi}(\rvh_t, \rva_{t-1}, e_{\phi}(\rvo_t))\mI \right)$ with $\rvh_{t} = f_{\theta}(\rvh_{t-1}, \rvs_{t-1})$. Note that $\rvh_t$ was inferred using $f_{\theta}$ because we have assumed that $\rvh_t$ is deterministic, meaning that $q(\rvh_t|\rvh_{t-1},\rvs_{t-1})=p(\rvh_t|\rvh_{t-1},\rvs_{t-1})=\delta(\rvh_t -f_{\theta}(\rvh_{t-1},\rvs_{t-1}))$. The dimension of the Gaussian distribution is equal to $512$.

\subsection{Generative Network \texorpdfstring{$\theta$}{}}
\label{mile-subsection:generative_model}
The generative network, parameterised by $\theta$, models the latent dynamics $(\rvh_{1:T},\rvs_{1:T})$ as well as the generative process of $(\rvo_{1:T},\rvy_{1:T},\rva_{1:T})$. It comprises a gated recurrent cell $f_{\theta}$, a prior network $(\mu_{\theta}, \sigma_{\theta})$, an image decoder $g_{\theta}$, a BeV decoder $l_{\theta}$, and a policy $\pi_{\theta}$.

The prior network estimates the parameters of the Gaussian distribution\\ $p(\rvs_t|\rvh_{t-1}, \rvs_{t-1}) \sim \mathcal{N}\left( \mu_{\theta}(\rvh_t,\hat{\rva}_{t-1}), \sigma_{\theta}(\rvh_t,\hat{\rva}_{t-1})\mI \right)$ with $\rvh_{t} = f_{\theta}(\rvh_{t-1}, \rvs_{t-1})$ and $\hat{\rva}_{t-1}=\pi_{\theta}(\rvh_{t-1}, \rvs_{t-1})$. Since the prior does not have access to the ground truth action $\rva_{t-1}$, the latter is estimated with the learned policy $\hat{\rva}_{t-1} = \pi_{\theta}(\rvh_{t-1}, \rvs_{t-1})$. 

The Kullback-Leibler divergence loss between the prior and posterior distributions can be interpreted as follows. Given the past state $(\rvh_{t-1}, \rvs_{t-1})$, the objective is to predict the distribution of the next state $\rvs_t$. As we model an active agent, this transition is decomposed into (i) action prediction and (ii) next state prediction. This transition estimation is compared to the posterior distribution that has access to the ground truth action $\rva_{t-1}$, and the image observation $\rvo_t$. The prior distribution tries to match the posterior distribution. This divergence matching framework ensures the model predicts actions and future states that explain the observed data. The divergence of the posterior from the prior measures how many nats of information were missing from the prior when observing the posterior. At training convergence, the prior distribution should be able to model all action-state transitions from the expert dataset.

The image and BeV decoders have an architecture similar to StyleGAN \citep{karras19}. The prediction starts as a learned constant tensor, and is progressively upsampled to the final resolution. At each resolution, the latent state is injected in the network with adaptive instance normalisation. This allows the latent states to modulate the predictions at different resolutions. The policy is a multi-layer perceptron. Please refer to \Cref{mile-appendix:model-description} for a full description of the neural networks.

\subsection{Imagining Future States and Actions}
\label{mile-subsection:imagine}
Our model can imagine future latent states by using the learned policy to infer actions $\hat{\rva}_{T+i} = \pi_{\theta}(\rvh_{T+i}, \rvs_{T+i})$, predicting the next deterministic state $\rvh_{{T+i+1}} = f_{\theta}(\rvh_{T+i}, \rvs_{T+i})$ and sampling from the prior distribution $\rvs_{T+i+1} \sim \mathcal{N}(\mu_{\theta}(\rvh_{T+i+1},\hat{\rva}_{T+i}), \sigma_{\theta}(\rvh_{T+i+1},\hat{\rva}_{T+i})\mI)$, for $i\geq0$. This process can be iteratively applied to generate sequences of longer futures in latent space, and the predicted futures can be visualised through the decoders. 

\clearpage
\section{Experimental Setting}
\label{mile-section:experimental_setting}
\paragraph{Dataset.} The training data was collected in the CARLA simulator with an expert reinforcement learning (RL) agent \citep{zhang2021end} that was trained using privileged information as input (BeV semantic segmentations and vehicle measurements). This RL agent generates more diverse runs and has greater driving performance than CARLA's in-built autopilot \citep{zhang2021end}.

We collect data at $25\mathrm{Hz}$ in four different training towns (Town01, Town03, Town04, Town06) and four weather conditions (ClearNoon, WetNoon, HardRainNoon, ClearSunset) for a total of $2.9\mathrm{M}$ frames, or $32$ hours of driving data. At each timestep, we save a tuple $(\rvo_t, \mathbf{route}_t, \mathbf{speed}_t, \rva_t, \rvy_t)$, with $\rvo_t \in \mathbb{R}^{3\times600\times960}$ the forward camera RGB image, $\mathbf{route}_t \in \mathbb{R}^{1\times64\times64}$ the route map (visualized as an inset on the top right of the RGB images in Figure \ref{mile-fig:multimodal-predictions}), $\mathbf{speed}_t \in \mathbb{R}$ the current velocity of the vehicle, $\rva_t \in \mathbb{R}^2$ the action executed by the expert (acceleration and steering), and $\rvy_t \in \mathbb{R}^{C_b \times 192 \times 192}$ the BeV semantic segmentation. There are $C_b = 8$ semantic classes: background, road, lane marking, vehicles, pedestrians, and traffic light states (red, yellow, green). In urban driving environments, the dynamics of the scene do not contain high frequency components, which allows us to subsample frames at $5\mathrm{Hz}$ in our sequence model.

\paragraph{Training.} 
Our model was trained for $50,000$ iterations on a batch size of $64$ on $8$ V100 GPUs, with training sequence length $T=12$. We used the AdamW optimiser \citep{loshchilov19} with learning rate $10^{-4}$ and weight decay $0.01$.

\paragraph{Metrics.} We report metrics from the CARLA challenge \citep{carla-challenge} to measure on-road performance: route completion, infraction penalty, and driving score. These metrics are however very coarse, as they only give a sense of how well the agent performs with hard penalties (such as hitting virtual pedestrians). Core driving competencies such as lane keeping and driving at an appropriate speed are obscured. Therefore we also report the cumulative reward of the agent. At each timestep the reward \citep{toromanoff20} penalises the agent for deviating from the lane center, for driving too slowly/fast, or for causing infractions. It measures how well the agent drives at the timestep level. In order to account for the length of the simulation (due to various stochastic events, it can be longer or shorter), we also report the normalised cumulative reward. More details on the experimental setting is given in \Cref{mile-appendix:experimental-setting}.
\clearpage
\section{Results}
\label{mile-section:result}

\subsection{Driving Performance}
We evaluate our model inside the CARLA simulator on a town and weather conditions never seen during training. We picked Town05 as it is the most complex testing town, and use the 10 routes of Town05 as specified in the CARLA challenge \citep{carla-challenge}, in four different weather conditions. \Cref{mile-table:carla-challenge} shows the comparison against prior state-of-the-art methods: CILRS \citep{codevilla2019exploring}, LBC \citep{chen2020learning}, TransFus. \citep{prakash2021multi}, Roach \citep{zhang2021end}, and LAV \citep{chen2022learning}. We evaluate these methods using their publicly available pre-trained weights.

\begin{table}[t]
\caption[Driving performance on a new town and new weather conditions in CARLA.]{Driving performance on a new town and new weather conditions in CARLA. Metrics are averaged across three runs. We include reward signals from past work where available.}
\label{mile-table:carla-challenge}
\centering
\begin{small}
\begin{tabular}{lccccc}
\toprule
& \textbf{Driving Score}  & \textbf{Route} & \textbf{Infraction}& \textbf{Reward} & \textbf{Norm. Reward} \\
\midrule
CILRS & 7.8 $\pm~$0.3 & 10.3 $\pm~$0.0 & 76.2 $\pm~$0.5& -&- \\
LBC & 12.3 $\pm~$2.0 & 31.9 $\pm~$2.2 & 66.0 $\pm~$1.7 & -&- \\
TransFus. & 31.0 $\pm~$ 3.6 & 47.5 $\pm~$ 5.3& \textbf{76.8 $\pm~$3.9} & -&- \\
Roach & 41.6 $\pm~$1.8 & 96.4 $\pm~$2.1& 43.3 $\pm~$2.8&	4236 $\pm~$468&0.34 $\pm~$0.05 \\
LAV & 46.5 $\pm~$3.0 & 69.8 $\pm~$2.3& 73.4 $\pm~$2.2  & - & - \\
\textbf{MILE} & \textbf{61.1 $\pm~$3.2} & \textbf{97.4 $\pm~$0.8}	 & 63.0 $\pm~$3.0& \textbf{7621 $\pm~$460} &	\textbf{0.67 $\pm~$0.02}\\
\cmidrule{1-6}
Expert & 88.4 $\pm~$0.9& 97.6 $\pm~$1.2	& 90.5 $\pm~$1.2& 8694 $\pm~$88& 0.70 $\pm~$0.01\\
\bottomrule
\end{tabular}
\end{small}
\end{table}

MILE outperforms previous works on all metrics, with a 31\% relative improvement in driving score with respect to LAV. Even though some methods have access to additional sensor information such as LiDAR (TransFus.~\citep{prakash2021multi}, LAV~\citep{chen2022learning}), our approach demonstates superior performance while only using RGB images from the front camera. Moreover, we observe that our method almost doubles the cumulative reward of Roach (which was trained on the same dataset) and approaches the performance of the privileged expert.

\subsection{Ablation Studies}
\label{mile-section:ablation-studies}

We next examine the effect of various design decisions in our approach.

\paragraph{3D geometry.} We compare our model to the following baselines. \emph{Single frame} that predicts the action and BeV segmentation from a single image observation. \emph{Single frame, no 3D} which is the same model but without the 3D lifting step. And finally, \emph{No 3D} which is MILE without 3D lifting. As shown in \Cref{mile-table:ablations}, in both cases, there is a significant drop in performance when not modelling 3D geometry. For the single frame model, the cumulative reward drops from 6084 to 1878. For MILE, the reward goes from 7621 to 4564. These results highlights the importance of the 3D geometry inductive bias.

\paragraph{Probabilistic modelling.} At any given time while driving, there exist multiple possible valid behaviours. For example, the driver can slightly adjust its speed, decide to change lane, or decide what is a safe distance to follow behind a vehicle. A deterministic driving policy cannot model these subtleties. In ambiguous situations where multiple choices are possible, it will often learn the mean behaviour, which is valid in certain situations (e.g.~the mean safety distance and mean cruising speed are reasonable choices), but unsafe in others (e.g.~in lane changing: the expert can change lane early, or late; the mean behaviour is to drive on the lane marking). We compare MILE with a \emph{No prior/post. matching} baseline that does not have a Kullback-Leibler divergence loss between the prior and posterior distributions, and observe this results in a drop in cumulative reward from 7621 to 6084.

\begin{table}[t]
\begin{footnotesize}
\caption[Ablation studies.]{Ablation studies. We report driving performance on a new town and new weather conditions in CARLA. Results are averaged across three runs.}
\label{mile-table:ablations}
\centering
\begin{tabular}{lccccc}
\toprule
& \textbf{Driv. Score}  & \textbf{Route} & \textbf{Infraction}& \textbf{Reward} & \textbf{N. Reward} \\
\midrule
Single frame, no 3D& 51.8	$\pm~$3.0&78.3 $\pm~$3.0&68.3 $\pm~$2.8& 1878 $\pm~$296&0.20 $\pm~$0.04\\
Single frame &59.6 $\pm~$3.6&94.5 $\pm~$0.6&64.7 $\pm~$3.3& 6630 $\pm~$168& 0.60 $\pm~$0.01\\
\cmidrule{1-6}
No 3D &63.0 $\pm~$1.5&91.5 $\pm~$5.5&\textbf{69.1 $\pm~$2.8}&4564 $\pm~$1791&0.40  $\pm~$0.15\\
No prior/post matching &\textbf{63.3 $\pm~$2.2}&91.5 $\pm~$5.0&68.7 $\pm~$1.8&6084 $\pm~$1429& 0.55 $\pm~$0.07\\
No segmentation &55.0 $\pm~$3.3&92.5 $\pm~$2.4&60.9 $\pm~$3.9&7183 $\pm~$107&0.64  $\pm~$0.02\\
\textbf{MILE} & 61.1 $\pm~$3.2 & \textbf{97.4 $\pm~$0.8}	 & 63.0 $\pm~$3.0& \textbf{7621 $\pm~$460} &	\textbf{0.67 $\pm~$0.02}	 \\
\cmidrule{1-6}
Expert & 88.4 $\pm~$0.9& 97.6 $\pm~$1.2	& 90.5 $\pm~$1.2& 8694 $\pm~$88& 0.70 $\pm~$0.01\\
\bottomrule
\end{tabular}
\end{footnotesize}
\end{table}

\subsection{Fully Recurrent Inference in Closed-Loop Driving}
We compare the closed-loop performance of our model with two different strategies:

\begin{enumerate}[label=(\roman*), itemsep=0.5mm, parsep=0pt]
    \item \textbf{Reset state}: for every new observation, we re-initialise the latent state and recompute the new state $[h_T, s_T]$, with $T$ matching the training sequence length.
    \item \textbf{Fully recurrent}: the latent state is initialised at the beginning of the evaluation, and is recursively updated with new observations. It is never reset, and instead, the model must have learned a representation that generalises to integrating information for orders of magnitude more steps than the $T$ used during training.
\end{enumerate}

 \Cref{mile-table:online-deployment} shows that our model can be deployed with recurrent updates, matching the performance of the \emph{Reset state} approach, while being much more computationally efficient ($7\times$ faster from $6.2\mathrm{Hz}$ with $T=12$ of fixed context to $43.0\mathrm{Hz}$ with a fully recurrent approach). A hypothesis that could explain why the \emph{Fully recurrent} deployment method works well is because the world model has learned to always discard all past information and rely solely on the present input. To test this hypothesis, we add Gaussian noise to the past latent state during deployment. If the recurrent network is simply discarding all past information, its performance should not be affected. However in \Cref{mile-table:online-deployment}, we see that the cumulative reward significantly decreases, showing our model does not simply discard all past context, but actively makes use of it.

\begin{table}[h]
\caption[Comparison of two deployment methods: Reset state and Fully recurrent.]{Comparison of two deployment methods. (i) \emph{Reset state}: for each new observation a fresh state is computed from a zero-initialised latent state using the last $T$ observations, and (ii) \emph{Fully recurrent}: the latent state is recurrently updated with new observations. We report driving performance on an unseen town and unseen weather conditions in CARLA. Frequency is in Hertz.}
\label{mile-table:online-deployment}
\centering
\begin{footnotesize}
\begin{tabular}{lcccccc}
\toprule
 & \textbf{Driv. Score}   & \textbf{Route} & \textbf{Infraction} & \textbf{Reward} & \textbf{N. Reward} & \textbf{Freq.}\\
\midrule
Reset state& 61.1 $\pm~$3.2 & \textbf{97.4 $\pm~$0.8}	 & 63.0 $\pm~$3.0& \textbf{7621 $\pm~$460} &	\textbf{0.67 $\pm~$0.02}  &6.2\\
Fully recurrent  &  \textbf{62.1 $\pm~$0.5}& 93.5 $\pm~$4.8& \textbf{66.6 $\pm~$3.4} & 7532 $\pm~$1122 & 0.67 $\pm~$0.04 &\textbf{43.0}\\
\cmidrule{1-7}
Recur.+noise&48.8 $\pm~$1.8&81.1 $\pm~$7.0  & 61.5 $\pm~$6.4& 3603 $\pm~$780&0.35 $\pm~$0.07 &\textbf{43.0}\\
\bottomrule
\end{tabular}
\end{footnotesize}
\vspace{-10pt}
\end{table}

\subsection{Long Horizon, Diverse Future Predictions}

\begin{figure}[h!]
    \centering
    \begin{subfigure}[b]{\textwidth}
    \centering
    \includegraphics[width=\linewidth]{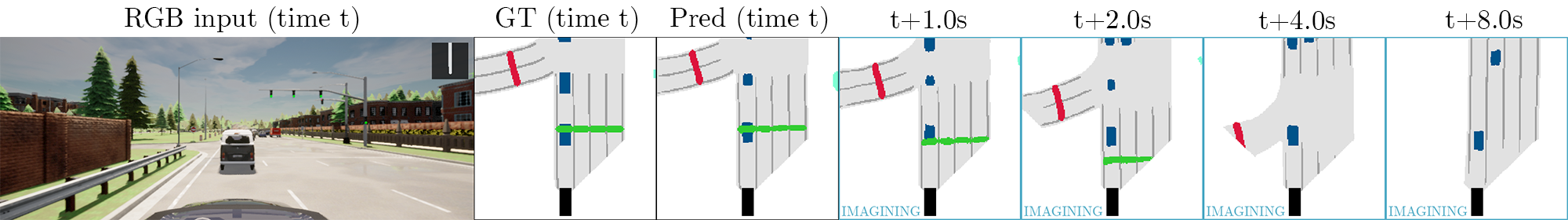}%
    \end{subfigure}
    \begin{subfigure}[t]{\textwidth}
    \includegraphics[width=\linewidth]{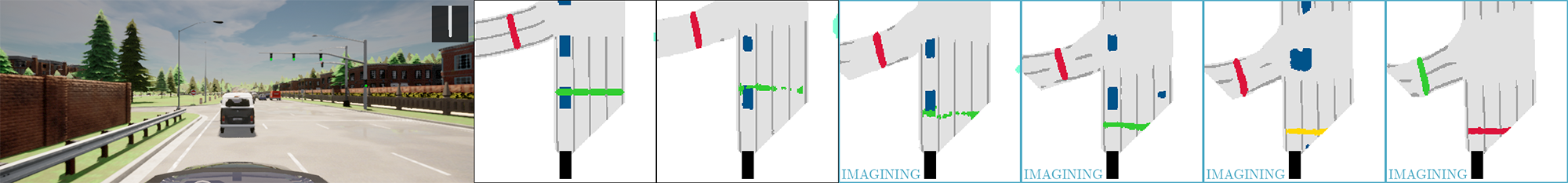}%
    \end{subfigure}
    \caption[Qualitative example of multi-modal predictions, for 8 seconds in the future.]{Qualitative example of multi-modal predictions, for 8 seconds in the future. BeV segmentation legend: black = ego-vehicle, white = background, gray =  road, dark gray=lane marking, blue = vehicles, cyan = pedestrians, green/yellow/red = traffic lights. Ground truth labels (GT) outside the field-of-view of the front camera are masked out. In this example, we visualise two distinct futures predicted by the model: 1) (top row) driving through the green light, 2) (bottom row) stopping because the model imagines the traffic light turning red. Note the light transition from green, to yellow, to red, and also at the last frame $t+8.0\mathrm{s}$ how the traffic light in the left lane turns green.}
    \label{mile-fig:multimodal-predictions}
    \vspace{-10pt}
\end{figure}

Our model can imagine diverse futures in the latent space, which can be decoded to BeV semantic segmentation for interpretability. \Cref{mile-fig:multimodal-predictions} shows examples of multi-modal futures predicted by MILE.

\clearpage
\section{Insights from the World Model}
\label{mile-section:insights}

\definecolor{blue_recurrent}{RGB}{14,162,193}
\definecolor{blue_reset}{RGB}{21,65,103}

\subsection{Latent State Dimension}
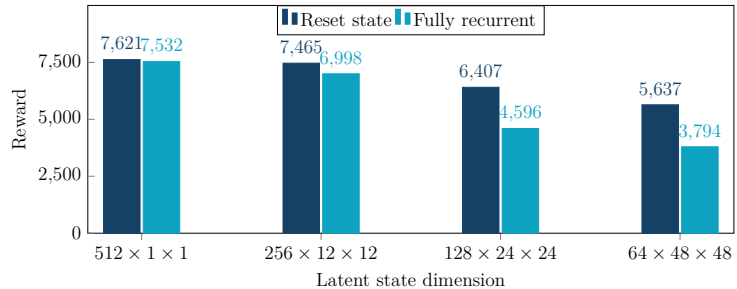
\begin{figure}[h]
\centering
\begin{tikzpicture}[scale=0.6]
    \begin{axis}[
            ybar,
            bar width=.8cm,
            width=\textwidth,
            height=.42\textwidth,
            legend style={at={(0.5,1)},
                anchor=north,legend columns=-1},
            symbolic x coords={512x1x1,256x12x12,128x24x24,64x48x48},
            xticklabels={$512\times1\times1$,$256\times12\times12$,$128\times24\times24$,$64\times48\times48$},
            extra y ticks={0,2500, 5000, 7500},
            ytick=\empty,
            xtick=data,
            xtick pos=left,
            ytick pos=left,
            nodes near coords,
            nodes near coords align={vertical},
            ymin=0,ymax=10000,
            ylabel={Reward},
            xlabel={Latent state dimension}
        ]
        \addplot[blue_reset,fill=blue_reset] table[x=dimension,y=reset]{\latentstatedata};
        \addplot[blue_recurrent,fill=blue_recurrent] table[x=dimension,y=recurrent]{\latentstatedata};
        \legend{Reset state,Fully recurrent}
    \end{axis}
\end{tikzpicture}
\caption{Analysis on the latent state dimension. We report closed-loop driving performance in a new town and new weather in CARLA.}
\label{mile-fig:latent-state-dimensionality}
\vspace{-10pt}
\end{figure}

In our model, we have set the latent state to be a low-dimensional 1D vector of size $512$. In dense image reconstruction however, the bottleneck feature is often a 3D spatial tensor of dimension (channel, height, width). We test whether it is possible to have a 3D tensor as a latent probabilistic state instead of a 1D vector. We change the latent state to have dimension $256\times 12 \times 12$ (40k distributions), $128\times 24 \times 24$ (80k distributions),  and $64\times 48 \times 48$ (160k distributions, which is the typical bottleneck size in dense image prediction). Since the latent state is now a spatial tensor, we adapt the recurrent network to be convolutional by switching the fully-connected operations with convolutions. We evaluate the model in the reset state and fully recurrent setting and report the results in \Cref{mile-fig:latent-state-dimensionality}. 

\pgfplotstableread[row sep=\\,col sep=&]{
    dimension & reset & recurrent \\
    512x1x1     & 7621  & 7532   \\
    256x12x12    & 7465 & 6998 \\
    128x24x24   & 6407& 4596 \\
    64x48x48   & 5637 & 3794\\
    }\latentstatedata
    
In the reset state setting, performance decreases as the dimensionality of the latent state increases. Surprisingly, even though the latent space is larger and has more capacity, driving performance is negatively impacted. This seems to indicate that optimising the prior and posterior distributions in the latent space is difficult, and especially more so as dimensionality increases. The prior, which is a multivariate Gaussian distribution needs to match the posterior, another multivariate Gaussian distribution. What makes this optimisation tricky is that the two distributions are non-stationary and change over time during the course of training. The posterior needs to extract the relevant information from the high-resolution images and incorporate it in the latent state in order to reconstruct BeV segmentation and regress the expert action. The prior has to predict the transition that matches the distribution of the posterior. 

Even more intriguing is when we look at the results in the fully recurrent deployment setting. When deployed in a fully recurrent manner in the simulator, without resetting the latent state, the model needs to discard information that is no longer relevant and continuously update its internal state with new knowledge coming from image observations. In our original latent state dimension of $512$, there is almost no different in driving performance between the two deployment modes. The picture is dramatically different when using a higher dimensional spatial latent state. For all the tested dimensions, there is a large gap between the two deployment settings. This result seems to indicate that the world model operating on high-dimensional spatial states has not optimally learned this behaviour, contrarily to the one operating on low-dimensional vector states.

\subsection{Driving in Imagination}
Humans are believed to build an internal model of the world in order to navigate in it \citep{madl15,epstein17,park18}. Since the stream of information they perceive is often incomplete and noisy, their brains fill missing information through imagination. This explains why it is possible for them to continue driving when blinded by sunlight for example. Even if no visual observations are available for a brief moment, they can still reliably predict their next states and actions to exhibit a safe driving behaviour.
We demonstrate that similarly, MILE can execute accurate driving plans entirely predicted from imagination, without having access to image observations. We qualitatively show that it can perform complex driving maneuvers such as navigating a roundabout, marking a pause a stop sign, or swerving to avoid a motorcyclist, using an imagined plan from the model (see supplementary material).

Quantitatively, we measure how accurate the predicted plans are by operating in the fully recurrent setting. We alternate between the \emph{observing mode} where the model can see image observations, and the \emph{imagining mode} where the model has to imagine the next states and actions, similarly to a driver that temporarily loses sight due to sun glare. In \Cref{mile-appendix:driving-imagination} we show that our model can retain the same driving performance with up to 30\% of the drive in imagining mode. This demonstrates that the model can imagine driving plans that are accurate enough for closed loop driving. Further, it shows that the latent state of the world model can seamlessly switch between the observing and imagining modes. The evolution of the latent state is predicted from imagination when observations are not available, and updated  with image observations when they become accessible.

\clearpage
\section{Summary}
\label{mile-section:conclusion}

In this chapter, we presented MILE: a Model-based Imitation LEarning approach for urban driving, that jointly learns a driving policy and a world model from offline expert demonstrations alone. Our approach exploits geometric inductive biases, operates on high-dimensional visual inputs, and sets a new state-of-the-art on the CARLA simulator. MILE can predict diverse and plausible future states and actions, allowing the model to drive from a plan entirely predicted from imagination. 

An open problem is how to infer the driving reward function from expert data, as this would enable explicit planning in the world model. Another exciting avenue is self-supervision in order to relax the dependency on the bird's-eye view segmentation labels. Self-supervision could fully unlock the potential of world models for real-world driving and other robotics tasks.


\chapter{Conclusions and Future Directions}

\section{Conclusions}
In this thesis, we presented a general framework to train a world model and a policy with deep neural networks using camera observations and expert demonstrations. Our proposed model could jointly predict future static scene, dynamic scene, and ego-behaviour in complex urban driving scenes from high-resolution camera observations only.

We derived the probabilistic model of this active inference setting where the goal is to infer the latent dynamics that explain the observations and actions of the active agent. We optimised the lower bound of the log evidence, which contained a reconstruction term and a divergence term. The reconstruction term ensured that the learned latent states contained the information necessary to reconstruct the observations and predict the ego-actions. The divergence term ensured that the latent states were predictive of the future. To achieve this, it matched the prior distribution, which only had access to the current state, to the posterior distribution that additionally had access to the ground truth action and ground truth future observation. Therefore, in order to minimise this divergence loss, the model had to predict both plausible actions and plausible transitions.

First, we proposed in \Cref{chapter:video-scene-understanding} a model that predicted important quantities in computer vision, such as semantic segmentation, depth and optical flow. The main objective of this work was to figure out which temporal architecture was suitable to infer a compact latent representation from high-dimensional input images. Our model could predict diverse and plausible futures for a couple of seconds in the future.

Then, we leveraged 3D geometry as an inductive bias to learn a bird's-eye view state representation from surround monocular cameras. Our learned latent state could predict the future trajectories of dynamic agents in a probabilistic manner. Our approach predicted depth from image observations and used camera parameters to lift image features to 3D in a common reference frame. The 3D features were subsequently pooled to bird's-eye view. The bird's-eye view representation was a well-suited space to learn dynamics as it is an Euclidean metric space, as opposed to image observations which are in perspective projection space. One key benefit was that the equivariance property of convolutions could be fully leveraged as the network no longer needed to learn to predict trajectories for each object at a different distance from the camera. We showed for the first time an end-to-end model that could predict multimodal future trajectories of moving agents from surround camera observations.

Finally, we presented a model that could jointly predict future static scene, dynamic scene and ego-behaviour in urban driving scenes from image observations and expert demonstrations. We demonstrated the effectiveness of our approach in closed-loop simulation and set a new state-of-the-art on CARLA. Our model could predict diverse and plausible future states and actions, where the states could be decoded to bird's-eye view semantic segmentation for interpretability. The learned world model and driving policy were stable enough to generate predictions of over 1 hour ($2000\times$ longer than the training sequence size). When deploying the model in the simulator in closed-loop, we found the latent state of the world model could be recursively updated with each new observation, without ever being reset. The world model has therefore learned over the course of training to forget information that are no longer relevant, and update its internal state with new information. Lastly, we showed that the predicted states and actions from our model were accurate enough to be executed in the simulator without decreasing driving performance. The model could control the vehicle using the predicted actions it had imagined using its world model. This behaviour is similar to that of a human that would be able to predict a reasonable sequence of actions even when closing their eyes for a brief moment when driving.

\clearpage
\section{Future Directions}
To conclude this thesis, we discuss future research directions we find to be highly promising. These include self-supervision with 3D geometry, leveraging simulation, and training foundation vision models.
\subsection{Self-Supervision with 3D Geometry}
The world models presented in this thesis were all supervised with labels, either generated by a teacher (\Cref{chapter:video-scene-understanding}) or ground truth labels (\Cref{chapter:instance-prediction} and \Cref{chapter:imitation-learning}). Labels accelerate training, but even though they are easy to acquire in simulation, they are hard and expensive to obtain in the real world.

Self-supervision could remove the need for labels, by using image reconstruction as a loss and enforcing 3D geometry with depth and scene flow \citep{vedula_three-dimensional_1999,hur20}. From the current latent state at time $t$, if we could infer the depth of the image observation, the ego-pose from $t-1$ to $t$ and the residual 3D scene flow from $t-1$ to $t$, then we could have an additional reprojection loss enforcing correct depth prediction, correct ego-motion and correct motion of dynamic objects. Scale ambiguity could be addressed by supervising the ego-pose with vehicle velocity similarly to \citep{guizilini20}.




\subsection{Leveraging Simulation}
Simulation has driven massive breakthroughs in artificial intelligence. AlphaGo \citep{SilverHuangEtAl16nature}, OpenAI Five \citep{openaifive19}, AlphaFold \citep{alphafold-jumper21} all have something in common: they maximise a reward function using millions or billions of simulated data. How can we achieve that for autonomous driving? One way would be to: 1) craft a generic reward function that obeys traffic rules. 2) Train a league of agents that all try to improve this reward function. 3) Self-improving and diverse behaviour should emerge. 

Another difficult arises however -- how could we deploy this learned policy in the real-world? There are several ways to bridge the reality gap:

\begin{itemize}
    \item Use a shared high-level representation (e.g. depth, semantic segmentation) \citep{muller18,loquercio21}. However this approach might be brittle because real-world depth and segmentation will not be as accurate as in simulation.
    \item Domain adaptation by sharing features between real world images and simulated images \citep{zhu17,wulfmeier17,bewley19}.
    \item Domain randomisation by randomising the appearance and behaviour of the simulation so much that the real world becomes a subset of all the diversity encountered during training and transfers naturally \citep{tobin17,sadeghi17,openai-rubiks19}.
\end{itemize}

    

\subsection{Foundation Vision Models}
Foundation models are large-scale deep neural networks trained on a massive amount of unsupervised data that can be fine-tuned for a broad range of downstream tasks. A prominent example is in language modelling with GPT-3 \citep{brown20}, which was trained on a large corpus of internet text data to predict the next word. GPT-3 could then be applied to solve other tasks, such as question answering, code writing, or language translation. It could do so even in a zero-shot manner, without being explicitly trained for the new task. Instead, a prompt would be given to the model as textual input, and GPT-3 would predict the relevant text in order to complete the task.

One long-standing goal of computer vision is to train such a foundation model. A foundation vision model is a large-scale neural network pre-trained on a vast amount of visual data \citep{yuan21}. Such a model could be fine-tuned to solve important computer vision tasks such as 3D object detection, video panoptic segmentation, depth and 3D scene flow prediction, or future video prediction. This behaviour could emerge from training a large world model on millions of videos.

A world model infers the latent dynamics that explain the observations of an active agent conditioned on their actions. It needs to understand how the static and dynamic environments evolve over time, as well as how the action of the active agent affects the environment around them. If the world model is able to faithfully reconstruct the sensory inputs of the active agent from its neural representation, it necessarily means the world model has some understanding of 3D geometry, semantics, and motion. The learned neural representation (latent states) could then be fine-tuned to downstream computer vision tasks. Nonetheless, there are many challenges in training such a large-scale world model, including data, model architecture, and compute. 

One challenge is in the data -- what videos should the model be trained on? Humans learn the physics of the world through both passive observation and active interaction. The vast majority of online videos are passive observations where the camera is static. They constitute a solid basis to learn about the world. However, it is important to understand the consequences of the action of the ego-agent on the environment in order to reason about interactions. Examples include ego-centric videos or driving videos.

Training a foundation vision model will imply a massive scaling in terms of model size, similarly to GPT-3 with language modelling. A vision model with hundreds of billions of parameters will require progress in neural network architecture and optimisation. Transformers have a flexible architecture that only requires framing the inputs as a sequence of tokens. They have become ubiquitous in natural language processing and are slowly replacing convolutional neural networks in image understanding on most vision benchmarks. It might only be a question of time before they also become prevalent in video modelling, especially since transformers are remarkably computationally efficient and scale extremely well. With models and datasets getting larger and larger in natural language processing, there are still no signs of performance saturation.

However, video data takes much more memory and is much higher dimensional than text data from natural language processing. The compute requirements will therefore be massive. Progress in hardware and clever engineering to compute the forward pass of the network and store the gradients of the parameters for the backward pass will be necessary. 

Foundation vision models have the potential to advance computer vision as a field by understanding geometry, semantics, and motion in a diverse range of environments. They would be a huge leap forward in embodied intelligence.


\begin{spacing}{0.9}

\bibliographystyle{natbib}  
\cleardoublepage
\bibliography{z_references} 



\end{spacing}


\begin{appendices} 

\chapter{Video Scene Understanding} 
\section{Lower Bound Derivation}
\label[appendix]{video-appendix:lower-bound}

\begin{proof}
The Kullback-Leibler divergence between the variational distribution $q_{Z,S}$ and the posterior distribution $p(\rvz_k,\rvs_k|\rva_k,\rvo_{k+1:k+H},y_{k+1:k+H};\rvo_{1:k})$ is:

\begin{align}
   &\KL\big(q(\rvz_k,\rvs_k|\rvo_{1:k+H}) ~||~ p(\rvz_k,\rvs_k|\rva_k,\rvo_{k+1:k+H},\rvy_{k+1:k+H};\rvo_{1:k}) \big) \notag\\
    =& \E_{\rvz_k,\rvs_k \sim q_{Z,S}} \left[ \log \frac{q(\rvz_k,\rvs_k|\rvo_{1:k+H})}{p(\rvz_k,\rvs_k|\rva_k,\rvo_{k+1:k+H},y_{k+1:k+H};\rvo_{1:k})}\right]\\
    =& \E_{\rvz_k,\rvs_k \sim q_{Z,S}} \left[ \log \frac{q(\rvz_k,\rvs_k|\rvo_{1:k+H})p(\rva_k, \rvo_{k+1:k+H},\rvy_{k+1:k+H}|\rvo_{1:k})}{p(\rvz_k,\rvs_k|\rvo_{1:k})p(\rva_k, \rvo_{k+1:k+H},\rvy_{k+1:k+H}|\rvz_k,\rvs_k)} \right] \\
    =& \log p(\rva_k, \rvo_{k+1:k+H},\rvy_{k+1:k+H}|\rvo_{1:k}) - \E_{\rvz_k,\rvs_k \sim q_{Z,S}} \big[ \log p(\rva_k, \rvo_{k+1:k+H},\rvy_{k+1:k+H}|\rvz_k,\rvs_k) \big] \notag \\
    &+ \KL \big(q(\rvz_k,\rvs_k|\rvo_{1:k+H})~||~p(\rvz_k,\rvs_k|\rvo_{1:k})\big) \notag
\end{align}

Since $\KL\big(q(\rvz_k,\rvs_k|\rvo_{1:k+H}) ~||~ p(\rvz_k,\rvs_k|\rva_k,\rvo_{k+1:k+H},\rvy_{k+1:k+H};\rvo_{1:k}) \big) \geq 0$, we obtain the following evidence lower bound:
\begin{alignat}{2}
   & &&\log p(\rva_k, \rvo_{k+1:k+H},\rvy_{k+1:k+H}|\rvo_{1:k}) \notag\\
   &\geq && \E_{\rvz_k,\rvs_k \sim q_{Z,S}} \big[ \log p(\rva_k, \rvo_{k+1:k+H},\rvy_{k+1:k+H}|\rvz_k,\rvs_k) \big] \notag \\
    & &&- \KL \big(q(\rvz_k,\rvs_k|\rvo_{1:k+H})~||~p(\rvz_k,\rvs_k|\rvo_{1:k})\big) \label{video-eq:lower-bound}
\end{alignat}

Let us now calculate the two terms of this lower bound separately. On the one hand:
\begin{align}
   &\E_{\rvz_k,\rvs_k \sim q_{Z,S}} \Big[ \log p(\rva_k, \rvo_{k+1:k+H},\rvy_{k+1:k+H}|\rvz_k,\rvs_k) \Big] \notag\\
    =&~ \E_{\rvz_k,\rvs_k \sim q_{Z,S}} \Big[ \log \big(p(\rva_k|\rvz_k,\rvs_k)\prod_{t=1}^H p( \rvo_{k+t}|\rvz_k,\rvs_k) p(\rvy_{k+t}|\rvz_k,\rvs_k) \big)\Big] \label{video-eq:expectation1}\\
    =&~\E_{\rvz_k,\rvs_k \sim q_{Z,S}} \Big[ \log p(\rva_k|\rvz_k,\rvs_k) + \sum_{t=1}^H \big(\log p( \rvo_{k+t}|\rvz_k,\rvs_k) + \log p(\rvy_{k+t}|\rvz_k,\rvs_k)\big) \Big] \label{video-eq:last-expectation}
\end{align}
where \Cref{video-eq:expectation1} follows from the factorisation defined in \Cref{video-eq:generative-model}, and \Cref{video-eq:last-expectation} was obtained by splitting the log of products.

On the other hand:
\begin{align}
    &\KL \big(q(\rvz_k,\rvs_k|\rvo_{1:k+H})~||~p(\rvz_k,\rvs_k|\rvo_{1:k})\big) \notag\\
    =&~\E_{\rvz_k,\rvs_k \sim q_{Z,S}} \left[\log \frac{q(\rvz_k,\rvs_k|\rvo_{1:k+H})}{p(\rvz_k,\rvs_k|\rvo_{1:k})} \right] \notag\\
    =&~\int_{\rvz_k,\rvs_k } q(\rvz_k,\rvs_k|\rvo_{1:k+H}) \log \frac{q(\rvz_k,\rvs_k|\rvo_{1:k+H})}{p(\rvz_k,\rvs_k|\rvo_{1:k})} \diff \rvz_k \diff \rvs_k \notag\\
    =&~\int_{\rvz_k,\rvs_k } q(\rvz_k,\rvs_k|\rvo_{1:k+H}) \log \frac{q(\rvz_k|\rvo_{1:k+H})q(\rvs_k | \rvz_k,\rvo_{k+1:k+H})}{p(\rvz_k|\rvo_{1:k})p(\rvs_k|\rvz_k)} \diff \rvz_k \diff \rvs_k  \label{video-eq:kl-step-1}
\end{align}
where \Cref{video-eq:kl-step-1} results from the factorisations defined in \Cref{video-eq:generative-model} and \Cref{video-eq:inference-model}. 

Thus:
\begin{align}
    &\KL \big(q(\rvz_k,\rvs_k|\rvo_{1:k+H})~||~p(\rvz_k,\rvs_k|\rvo_{1:k})\big) \notag\\
    =&~\int_{\rvz_k,\rvs_k } q(\rvz_k,\rvs_k|\rvo_{1:k+H}) \log \frac{q(\rvs_k | \rvz_k,\rvo_{k+1:k+H})}{p(\rvs_k|\rvz_k)} \diff \rvz_k \diff \rvs_k  \label{video-eq:kl-first-term} \\
    =&~\int_{\rvs_k } q(\rvs_k | \rvz_k,\rvo_{k+1:k+H}) \log \frac{q(\rvs_k | \rvz_k,\rvo_{k+1:k+H})}{p(\rvs_k|\rvz_k)} \diff \rvs_k  \label{video-eq:kl-second-term}\\
    =&~\KL \big(q(\rvs_k | \rvz_k,\rvo_{k+1:k+H})~||~p(\rvs_k|\rvz_k)\big) \label{video-eq:kl-final}
\end{align}
where \Cref{video-eq:kl-first-term} follows from $q(\rvz_k|\rvo_{1:k+H})=p(\rvz_k|\rvo_{1:k})$ and \Cref{video-eq:kl-second-term} was obtained by integrating over the deterministic variable $\rvz_k$.

Finally, we inject \Cref{video-eq:last-expectation} and \Cref{video-eq:kl-final} in \Cref{video-eq:lower-bound} to obtain the desired lower bound:

\begin{align*}
    &\log p(\rva_k, \rvo_{k+1:k+H},\rvy_{k+1:k+H}|\rvo_{1:k}) \\
    \geq&~\E_{\rvz_k,\rvs_k \sim q_{Z,S}} \Big[ \log p(\rva_k|\rvz_k,\rvs_k) + \sum_{t=1}^H \big(\log p( \rvo_{k+t}|\rvz_k,\rvs_k) + \log p(\rvy_{k+t}|\rvz_k,\rvs_k)\big) \Big]\\
    & - \KL \big(q(\rvs_k | \rvz_k,\rvo_{k+1:k+H})~||~p(\rvs_k|\rvz_k)\big) \qedhere
\end{align*}

\end{proof}

\clearpage
\section{Model}
\label[appendix]{video-appendix:architecture}

In total, our architecture has $30.4$M parameters, comprising of modules:
\begin{itemize}
    \item Perception $e_{\theta}$, $25.3$M parameters ;
    \item Temporal module $f_{\theta}$ and present/future distributions $(\mu_{\theta},\sigma_{\theta})$ and $(\mu_{\phi},\sigma_{\phi})$, $0.8$M parameters ;
    \item Future prediction $g_{\theta}=\{d_{t,\theta},l_{t,\theta},m_{t,\theta}\}_{t=1,\dots,H}$, $3.5$M parameters ;
    \item Policy $\pi_{\theta}$, $0.7$M parameters.
\end{itemize}

\subsection{Perception}

\paragraph{Semantics and Geometry.}
Our model is an encoder-decoder model with five encoder blocks and three decoder blocks, followed by an atrous spatial pyramid pooling (ASPP) module \cite{chen17}. The encoders contain 2, 4, 8, 8, 8 layers respectively, downsampling by a factor of two each time with a strided convolution. The decoders contain 3 layers each, upsampling each time by a factor of two with a sub-strided convolution. All layers have residual connections and many are low rank, with varying kernel and dilation sizes. Furthermore, we employ skip connections from the encoder to decoder at each spatial scale.

We pretrain the scene understanding encoder on a number of heterogeneous datasets to predict depth and semantic segmentation: Cityscapes \citep{cityscapes16}, Mapillary Vistas \citep{neuhold_mapillary_2017}, ApolloScape \citep{huang2018apolloscape} and Berkeley Deep Drive \citep{yu2018bdd100k}. We collapse the classes to $14$ semantic segmentation classes shared across these datasets and sample each dataset equally during training. We train for 200,000 gradient steps with a batch size of $32$ using SGD with an initial learning rate of $0.1$ with momentum $0.9$. We use cross entropy for segmentation and the scale-invariant loss \citep{li_megadepth_2018} to learn depth with a weight of $1.0$ and $0.1$, respectively.

\paragraph{Motion.}
In addition to this geometry and semantics encoder, we also use a pretrained optical flow network, PWCNet~\citep{sun_pwc-net_2018}. We use the pretrained authors' implementation.

\paragraph{Perception.}
To form our perception encoder we concatenate these two feature representations (from the perception encoder and optical flow net) concatenated together. 
We use the features two layers before the output optical flow regression as the feature representation.
The decoders of these networks are used for generating pseudo-ground truth segmentation and depth labels to train our dynamics and future prediction modules.

\subsection{Training}
Our model was trained on 8 GPUs, each with a batch size of 4, for 200,000 steps using an Adam optimiser with learning rate $3\mathrm{e}{-4}$. The input of our model is a sequence of 15 frames at resolution $224\times480$ and a frame rate of 5Hz ($256\times512$ and 17Hz for Cityscapes). The first 5 frames correspond to the past and present context (1s), and the following 10 frames to the future we want to predict (2s). All layers in the network use batch normalisation and a ReLU activation function. We now describe each module of our architecture in more detail.

\paragraph{Dynamics} four temporal blocks with kernel size $k=(2,3,3)$, stride $s=1$ and output channels $c=[80, 88, 96, 104]$. In between every temporal block, four 2D residual convolutions ($k=(3,3)$, $s=1$) are inserted.

\paragraph{Present and Future Distribution} two downsampling 2D residual convolutions ($k=(3,3)$, $s=2$, $c=[52, 52]$). An average pooling layer flattens the feature spatially, and a final dense layer maps it to a vector of size $2L$ ($L=16$).

\paragraph{Future Prediction} the main structure is a convolutional GRU ($k=(3,3)$, $s=1$). Each convolutional GRU is followed by three 2D residual convolutions ($k=(3,3)$, $s=1$). This structure is stacked five times.
The decoders: two upsampling convolutions ($k=(3,3)$, $s=1$, $c=32$), a convolution ($k=(3,3)$, $s=1$, $c=16$), and finally a convolution without activation followed by a bilinear interpolation to the original resolution $224\times480$.

\paragraph{Control} two downsampling convolutions ($k=(3,3)$, $s=2$, $c=[64, 32]$), followed by dense layers ($c=[1024, 512, 256, 128, 64, 32, 16, 4]$).

\clearpage
\section{Nomenclature}

We detail the symbols used to describe our model in this paper.

\begin{table}[h]
\centering
\begin{tabular}{ll}
    \toprule
    \textbf{Networks} &\\
    \midrule
    Perception encoder &  $e_{\theta}$ \\
    Temporal module & $f_{\theta}$ \\
    Present network & $(\mu_{\theta}, \sigma_{\theta})$ \\
    Future network & $(\mu_{\phi}, \sigma_{\phi})$ \\
    Future prediction module & $g_{\theta}=\{d_{t,\theta},l_{t,\theta},m_{t,\theta}\}_{t=1,\dots,H}$ \\
    Control module & $\pi_{\theta}$ \\
    \cmidrule{1-2}
    
    \textbf{Tensors} &\\
    \midrule
    Past temporal context & $k$ \\
    Future prediction horizon & $H$ \\
    Future control horizon & $N_c$ \\
    Images & $\rvo_{1:k+H}$ \\
    Perception features & $\rvx_i = e_{\theta}(\rvo_i)$ \\
    Temporal feature & $\rvz_k = f_{\theta}(\rvx_{1:k})$ \\
    Present distribution & $(\mu_{\theta}(\rvz_k), \sigma_{\theta}(\rvz_k))$ \\
    Future distribution & $(\mu_{\phi}(\{\rvz_{k+j}\}_{j\in J}), \sigma_{\phi}(\{\rvz_{k+j}\}_{j\in J})$ \\
    Sample (train) & $\rvs_k \sim \mathcal{N}(\mu_{\phi}(\{\rvz_{k+j}\}_{j\in J}), \sigma^2_{\phi}(\{\rvz_{k+j}\}_{j\in J})\mI)$ \\
    Sample (test) & $\rvs_k \sim \mathcal{N}(\mu_{\theta}(\rvz_k), \sigma^2_{\theta}(\rvz_k)\mI)$ \\
    $\begin{aligned} \text{Future perception outputs}  \\ \end{aligned}$ & $\begin{aligned}\hat{\rvy}_{k+t} &= (\hat{\rvd}_{k+t},\hat{\rvl}_{k+t},  \hat{\rvf}_{k+t} )  \\
    & = (d_{t,\theta}(\rvz_k, \rvs_k),l_{t,\theta}(\rvz_k, \rvs_k), m_{t,\theta}(\rvz_k, \rvs_k)) \end{aligned}$\\
    Control outputs  &  $\rva_k =\pi_{\theta}(\rvz_k)  = (\hat{\rv_k}, \hat{\dot{\rv_k}}, \hat{\rw_k}, \hat{\dot{\rw_k}})$\\
    \bottomrule
\end{tabular}
\end{table}


\chapter{Bird's-Eye View Future Prediction}
\section{Lower Bound Derivation}
\label[appendix]{fiery-appendix:lower-bound}

\begin{proof}
The Kullback-Leibler divergence between the variational distribution $q_{Z,S}$ and the posterior distribution $p(\rvz_k,\rvs_k|\rvo_{k:k+H},\rvy_{k:k+H};\rvo_{1:k},\rva_{1:k-1})$ is:

\begin{align}
   &\KL\big(q(\rvz_k,\rvs_k|\rvo_{1:k}, \rva_{1:k-1},\rvy_{k+1:k+H}) ~||~ p(\rvz_k,\rvs_k|\rvo_{k:k+H},\rvy_{k:k+H};\rvo_{1:k},\rva_{1:k-1}) \big) \notag\\
    =& \E_{\rvz_k,\rvs_k \sim q_{Z,S}} \left[ \log \frac{q(\rvz_k,\rvs_k|\rvo_{1:k}, \rva_{1:k-1},\rvy_{k+1:k+H})}{p(\rvz_k,\rvs_k|\rvo_{k:k+H},\rvy_{k:k+H};\rvo_{1:k},\rva_{1:k-1})}\right]\\
    =& \E_{\rvz_k,\rvs_k \sim q_{Z,S}} \left[ \log \frac{q(\rvz_k,\rvs_k|\rvo_{1:k}, \rva_{1:k-1},\rvy_{k+1:k+H})p( \rvo_{k:k+H},\rvy_{k:k+H}|\rvo_{1:k},\rva_{1:k-1})}{p(\rvz_k,\rvs_k|\rvo_{1:k},\rva_{1:k-1})p( \rvo_{k:k+H},\rvy_{k:k+H}|\rvz_k,\rvs_k)} \right] \\
    =& \log p(\rvo_{k:k+H},\rvy_{k:k+H}|\rvo_{1:k},\rva_{1:k-1}) - \E_{\rvz_k,\rvs_k \sim q_{Z,S}} \big[ \log p(\rvo_{k:k+H},\rvy_{k:k+H}|\rvz_k,\rvs_k) \big] \notag \\
    &+ \KL \big(q(\rvz_k,\rvs_k|\rvo_{1:k}, \rva_{1:k-1},\rvy_{k+1:k+H})~||~p(\rvz_k,\rvs_k|\rvo_{1:k},\rva_{1:k-1})\big) \notag
\end{align}

Since $\KL\big(q(\rvz_k,\rvs_k|\rvo_{1:k}, \rva_{1:k-1},\rvy_{k+1:k+H}) ~||~ p(\rvz_k,\rvs_k|\rvo_{k:k+H},\rvy_{k:k+H};\rvo_{1:k},\rva_{1:k-1}) \big) \geq 0$, we obtain the following evidence lower bound:
\begin{alignat}{2}
   & &&\log p(\rvo_{k:k+H},\rvy_{k:k+H}|\rvo_{1:k},\rva_{1:k-1}) \notag\\
   &\geq && \E_{\rvz_k,\rvs_k \sim q_{Z,S}} \big[ \log p(\rvo_{k:k+H},\rvy_{k:k+H}|\rvz_k,\rvs_k) \big] \notag \\
    & &&- \KL \big(q(\rvz_k,\rvs_k|\rvo_{1:k}, \rva_{1:k-1},\rvy_{k+1:k+H})~||~p(\rvz_k,\rvs_k|\rvo_{1:k},\rva_{1:k-1})\big) \label{fiery-eq:lower-bound}
\end{alignat}

Let us now calculate the two terms of this lower bound separately. On the one hand:
\begin{align}
   &\E_{\rvz_k,\rvs_k \sim q_{Z,S}} \big[ \log p(\rvo_{k:k+H},\rvy_{k:k+H}|\rvz_k,\rvs_k) \big] \notag\\
    =&~ \E_{\rvz_k,\rvs_k \sim q_{Z,S}} \Big[ \log \big(\prod_{t=0}^H p( \rvo_{k+t}|\rvz_k,\rvs_k) p(\rvy_{k+t}|\rvz_k,\rvs_k) \big)\Big] \label{fiery-eq:expectation1}\\
    =&~\E_{\rvz_k,\rvs_k \sim q_{Z,S}} \Big[\sum_{t=0}^H \big(\log p( \rvo_{k+t}|\rvz_k,\rvs_k) + \log p(\rvy_{k+t}|\rvz_k,\rvs_k)\big) \Big] \label{fiery-eq:last-expectation}
\end{align}
where \Cref{fiery-eq:expectation1} follows from the factorisation defined in \Cref{fiery-eq:generative-model} and \Cref{fiery-eq:generative-split}, and \Cref{fiery-eq:last-expectation} was obtained by splitting the log of products.

On the other hand:
\begin{align}
    &\KL \big(q(\rvz_k,\rvs_k|\rvo_{1:k}, \rva_{1:k-1},\rvy_{k+1:k+H})~||~p(\rvz_k,\rvs_k|\rvo_{1:k},\rva_{1:k-1})\big) \notag\\
    =&~\E_{\rvz_k,\rvs_k \sim q_{Z,S}} \left[\log \frac{q(\rvz_k,\rvs_k|\rvo_{1:k}, \rva_{1:k-1},\rvy_{k+1:k+H})}{p(\rvz_k,\rvs_k|\rvo_{1:k},\rva_{1:k-1})} \right] \notag\\
    =&~\int_{\rvz_k,\rvs_k } q(\rvz_k,\rvs_k|\rvo_{1:k}, \rva_{1:k-1},\rvy_{k+1:k+H}) \log \frac{q(\rvz_k,\rvs_k|\rvo_{1:k}, \rva_{1:k-1},\rvy_{k+1:k+H})}{p(\rvz_k,\rvs_k|\rvo_{1:k},\rva_{1:k-1})} \diff \rvz_k \diff \rvs_k \notag\\
    =&~\int_{\rvz_k,\rvs_k } q(\rvz_k,\rvs_k|\rvo_{1:k}, \rva_{1:k-1},\rvy_{k+1:k+H})  \notag\\
    &\quad\quad\ \ \;\log \frac{q(\rvz_k|\rvo_{1:k}, \rva_{1:k-1},\rvy_{k+1:k+H}) q(\rvs_k|\rvz_k,\rvy_{k+1:k+H})}{p(\rvz_k|\rvo_{1:k},\rva_{1:k-1}) p(\rvs_k|\rvz_k)} \diff \rvz_k \diff \rvs_k  \label{fiery-eq:kl-step-1}
\end{align}
where \Cref{fiery-eq:kl-step-1} results from the factorisations defined in \Cref{fiery-eq:generative-model} and \Cref{fiery-eq:inference-model}. 

Thus:
\begin{align}
    &\KL \big(q(\rvz_k,\rvs_k|\rvo_{1:k}, \rva_{1:k-1},\rvy_{k+1:k+H})~||~p(\rvz_k,\rvs_k|\rvo_{1:k},\rva_{1:k-1})\big) \notag\\
    =&~\int_{\rvz_k,\rvs_k } q(\rvz_k,\rvs_k|\rvo_{1:k}, \rva_{1:k-1},\rvy_{k+1:k+H}) \log \frac{q(\rvs_k|\rvz_k,\rvy_{k+1:k+H})}{p(\rvs_k|\rvz_k)} \diff \rvz_k \diff \rvs_k  \label{fiery-eq:kl-first-term} \\
    =&~\int_{\rvs_k } q(\rvs_k|\rvz_k,\rvy_{k+1:k+H}) \log \frac{q(\rvs_k|\rvz_k,\rvy_{k+1:k+H})}{p(\rvs_k|\rvz_k)} \diff \rvs_k  \label{fiery-eq:kl-second-term}\\
    =&~\KL \big(q(\rvs_k|\rvz_k,\rvy_{k+1:k+H})~||~p(\rvs_k|\rvz_k)\big) \label{fiery-eq:kl-final}
\end{align}
where \Cref{fiery-eq:kl-first-term} follows from $q(\rvz_k|\rvo_{1:k}, \rva_{1:k-1},\rvy_{k+1:k+H})=p(\rvz_k|\rvo_{1:k},\rva_{1:k-1})$ and \Cref{fiery-eq:kl-second-term} was obtained by integrating over the deterministic variable $\rvz_k$.

Finally, we inject \Cref{fiery-eq:last-expectation} and \Cref{fiery-eq:kl-final} in \Cref{fiery-eq:lower-bound} to obtain the desired lower bound:

\begin{align*}
    &\log p(\rvo_{k:k+H},\rvy_{k:k+H}|\rvo_{1:k},\rva_{1:k-1})  \\
    \geq&~\E_{\rvz_k,\rvs_k \sim q_{Z,S}} \Big[\sum_{t=0}^H \big(\log p( \rvo_{k+t}|\rvz_k,\rvs_k) + \log p(\rvy_{k+t}|\rvz_k,\rvs_k)\big) \Big]\\
    & - \KL \big(q(\rvs_k|\rvz_k,\rvy_{k+1:k+H})~||~p(\rvs_k|\rvz_k)\big) \qedhere
\end{align*}

\end{proof}

\clearpage
\section{Model and Dataset}
\label[appendix]{fiery-appendix:model}

\subsection{Model Description}
Our model processes $k=3$ past observations each with $n=6$ cameras images at resolution $(H_{\mathrm{in}}, W_{\mathrm{in}}) = (224\times480)$, i.e. $18$ images. 
The minimum depth value we consider is $D_{\text{min}} = 2.0\mathrm{m}$, which corresponds to the spatial extent of the ego-car. The maximum depth value is $D_{\text{max}}=50.0\mathrm{m}$, and the size of each depth slice is set to $D_{\mathrm{size}}=1.0\mathrm{m}$.

 We use uncertainty \citep{kendall18} to weight the segmentation, centerness, offset and flow losses. The probabilistic loss is weighted by $\lambda_{\text{probabilistic}} = 100$. All our layers use batch normalisation and a ReLU activation function. The bird's-eye view encoder and decoder layers are initialised with the weights of a pretrained \emph{Static model}.

Our model contains a total of $8.1$M parameters and trains in a day on $4$ Tesla V100 GPUs with $32$GB of memory.

\paragraph{Bird's-eye view encoder.} 
For every past timestep $i\in \{1,\dots,k \}$, each image in the observation $\rvo_i = \{\rvc_i^1, ..., \rvc_i^n\}$ is encoded: $\rvb_i^j = b_{\theta}(\rvc_i^j) \in \mathbb{R}^{(C+D)\times H_e \times W_e}$. We use the EfficientNet-B4 \citep{tan19} backbone with an output stride of $8$ in our implementation, so $(H_e, W_e) = (\frac{H_{\mathrm{in}}}{8}, 
\frac{W_{\mathrm{in}}}{8}) = (28,60)$. The number of channel is $C=64$ and the number of depth slices is $D = \frac{D_{\mathrm{max}} - D_{\mathrm{min}}}{D_{\mathrm{size}}}=48$.

These features are then lifted and projected to bird's-eye view to obtain a tensor $\rvx_i \in \mathbb{R}^{C\times H \times W}$ with $(H, W) = (200, 200)$. Using past ego-motion and a spatial transformer module, we transform the bird's-eye view features to the present's reference frame.

\paragraph{Temporal model.} The 3D convolutional temporal model is composed of \emph{Temporal Blocks}. Let $C_{\mathrm{in}}$ be the number of input channels and $C_{\mathrm{out}}$ the number of output channels. A single Temporal block is composed of:
\begin{itemize}[itemsep=0.3mm, parsep=0pt]
    \item a 3D convolution, with kernel size $(k_t, k_s, k_s) = (2, 3, 3)$. $k_t$ is the temporal kernel size, and $k_s$ the spatial kernel size.
    \item a 3D convolution with kernel size $(1, 3, 3)$.
    \item a 3D global average pooling layer with kernel size $(2, H, W)$.
\end{itemize}

Each of these operations is preceded by a feature compression layer, which is a $(1,1,1)$ 3D convolution with output channels $\frac{C_{\mathrm{in}}}{2}$.

All the resulting features are concatenated and fed through a $(1,1,1)$ 3D convolution with output channel $C_{\mathrm{out}}$. The temporal block module also has a skip connection. The final feature $\rvz_k$ is in $\mathbb{R}^{64\times 200 \times 200}$. 

\paragraph{Present and future distributions.}
The architecture of the present and future distributions are identical, except for the number of input channels. The present distribution takes as input $\rvz_k$, and the future distribution takes as input the concatenation of $(\rvz_k, \rvy_{k+1:k+H})$. Let $C_p=64$ be the number of input channel of the present distribution and $C_f=64 + C_y\cdot H = 88$ the number of input channels of the future distribution (since $C_y=6$ and $H=4$). The module contains four residual block layers \citep{he16} each with spatial downsampling 2. These four layers divide the number of input channels by 2. A spatial average pooling layer then decimates the spatial dimension, and a final (1,1) 2D convolution regress the mean and log standard deviation of the distribution in $\R^L \times \R^L$ with $L=32$.

\paragraph{Future prediction.}
The future prediction module is made of the following structure repeated three times: a convolutional Gated Recurrent Unit \citep{ballas16} followed by $3$ residual blocks with kernel size $(3,3)$.

\paragraph{Future instance segmentation and motion decoder.}
The decoder has a shared backbone and multiple output heads to predict centerness, offset, segmentation and flow. The shared backbone contains:

\begin{itemize}[itemsep=0.3mm, parsep=0pt]
    \item a 2D convolution with output channel 64 and stride 2.
    \item the following block repeated three times: four 2D residual convolutions with kernel size (3, 3). The respective output channels are [64, 128, 256] and strides [1, 2, 2].
    \item three upsampling layers of factor 2, with skip connections and output channel 64.
\end{itemize}

Each head is then the succession two 2D convolutions outputting the required number of channels.

\subsection{Labels Generation}
\label[appendix]{fiery-appendix:dataset}
We compute instance center labels as a 2D Gaussian centered at each instance center of mass with standard deviation $\sigma_x = \sigma_y = 3$. The centerness label indicates the likelihood of a pixel to be the center of an instance and is a $\mathbb{R}^{1\times H\times W}$ tensor.
For all pixels belonging to a given instance, we calculate the offset labels as the vector pointing to the instance center of mass (a $\mathbb{R}^{2\times H\times W}$ tensor). Finally, we obtain future flow labels (a $\mathbb{R}^{2\times H\times W}$ tensor) by comparing the position of the instance centers of gravity between two consecutive timesteps.

We use the \emph{vehicles} category to obtain 3D bounding boxes of road agents, and filter the vehicles that are not visible from the cameras.

We report results on the official NuScenes validation split. Since the Lyft dataset does not provide a validation set, we create one by selecting random scenes from the dataset so that it contains roughly the same number of samples (6,174) as NuScenes (6,019).




\chapter{World Models and Active Inference}

\section{Additional Results}

\subsection{Driving in Imagination}
\label[appendix]{mile-appendix:driving-imagination}
We deploy the model in the fully recurrent setting, and at fixed intervals: (i) we let the model imagine future states and actions, without observing new images, and execute those actions in the simulator. (ii) We then let the model update its knowledge of the world by observing new image frames. More precisely, we set the fixed interval to a two-second window, and set a ratio of imagining vs. observing. If for example that ratio is set to 0.5, we make the model imagine by sampling from the prior distribution for 1.0s, then sample from the posterior distribution for 1.0s, and alternate between these two settings during the whole evaluation run.

We make the ratio of imagining vs. observing vary from $0$ (always observing each image frame, which is the default behaviour) to $0.6$ (imagining for $60\%$ of the time). We report both the driving performance and perception accuracy in \Cref{mile-fig:dreaming-driving}. The driving performance is measured with the driving score, and the perception accuracy using the intersection-over-union with the ground truth BeV semantic segmentation. We compare MILE with a one-frame baseline which has no memory and only uses a single image frame for inference.

\Cref{mile-fig:dreaming-driving-performance} shows that our model can imagine for up to $30\%$ of the time without any significant drop in driving performance. After this point, the driving score starts decreasing but remains much higher than its one-frame counterpart. In \Cref{mile-fig:dreaming-driving-perception}, we see that the predicted states remain fairly accurate (by decoding to BeV segmentation), even with an important amount of imagining.
These results demonstrate that our model can predict plausible future states and actions, accurate enough to control a vehicle in closed-loop.

\begin{figure}[h]
    \centering
    \includegraphics[width=0.7\linewidth]{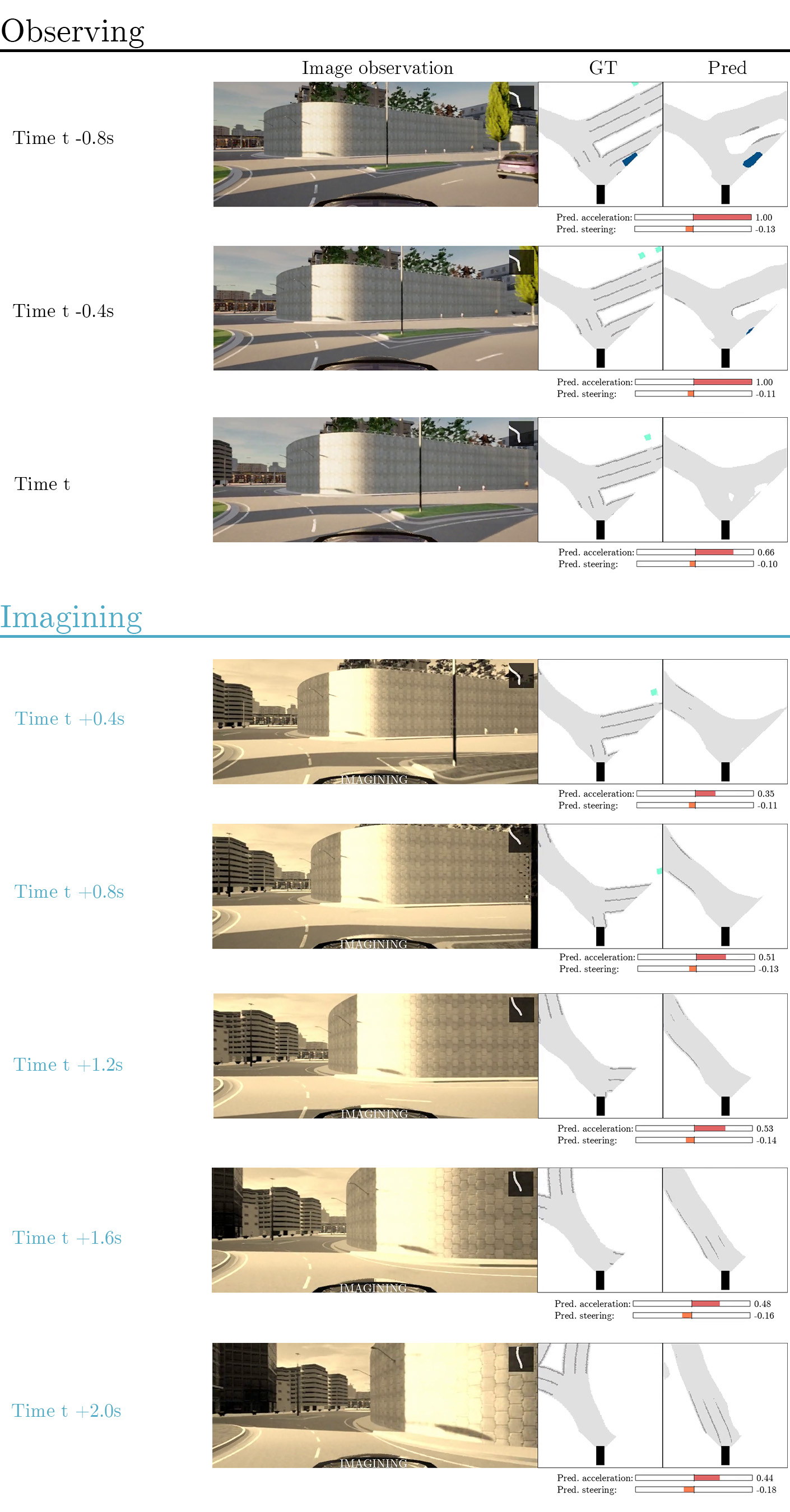}%
    \caption{An example of the model imagining and accurately predicting future states and actions to negotiate a roundabout. When imagining, the model does not observe the image frames, but predicts the future states and actions from its current latent state.}
    \label{fig:dreaming-driving-traffic}
\end{figure}

\Cref{fig:dreaming-driving-traffic} illustrates an example of the model driving in imagination and successfully negotiating a roundabout.

\definecolor{gray1}{RGB}{125,125,125}
\definecolor{blue2}{RGB}{14,162,193}

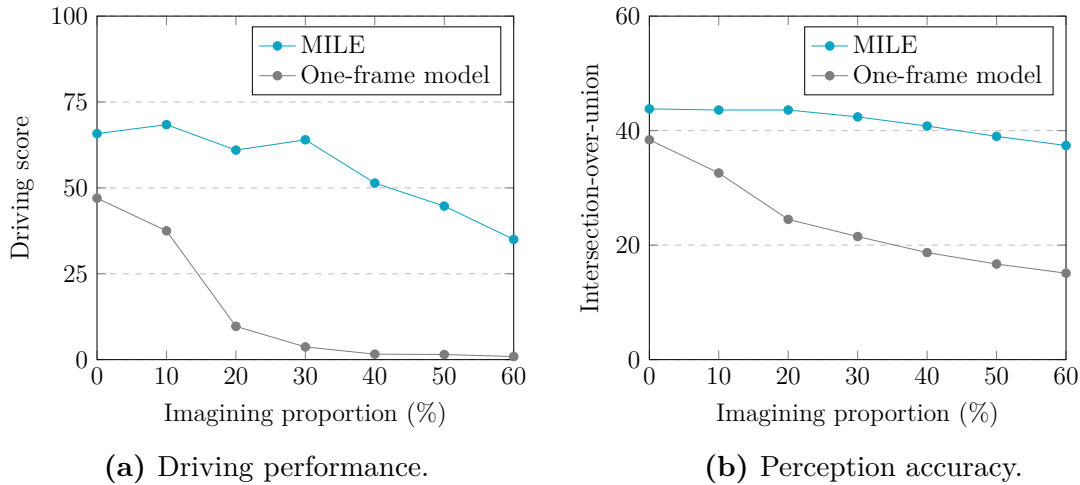
\begin{figure}[h]
\centering
    \begin{subfigure}[b]{0.49\textwidth}
    \centering
    \begin{tikzpicture}[scale=0.8]
    \begin{axis}[
        xlabel={Imagining proportion (\%)},
        ylabel={Driving score},
        xmin=0.0, xmax=60.0,
        ymin=0, ymax=100,
        xtick={0, 10, 20, 30, 40, 50, 60},
        ytick={0, 25, 50, 75, 100},
        legend pos=north east,
        legend cell align={left},
        ymajorgrids=true,
        grid style=dashed,
    ]
    
    \addplot[
        color=blue2,
        mark=*,
        ]
        coordinates {
        (0, 65.8)(10,68.4)(20,61.0)(30,64.0)(40,51.4)(50,44.7)(60,35.0)
        };
    \addplot[
        color=gray1,
        mark=*,
        ]
        coordinates {
        (0,47.0)(10,37.5)(20,9.7)(30,3.7)(40,1.6)(50,1.5)(60,0.9)
        };
    \addlegendentry{MILE}
    \addlegendentry{One-frame model}
        
    \end{axis}
    \end{tikzpicture}
    \caption{Driving performance.}
    \label{mile-fig:dreaming-driving-performance}
    \end{subfigure}
    \begin{subfigure}[b]{0.49\textwidth}
    \begin{tikzpicture}[scale=0.8]
    \begin{axis}[
        xlabel={Imagining proportion (\%)},
        ylabel={Intersection-over-union},
        xmin=0.0, xmax=60.0,
        ymin=0, ymax=60,
        xtick={0, 10, 20, 30, 40, 50, 60},
        ytick={0, 20, 40, 60},
        legend pos=north east,
        legend cell align={left},
        ymajorgrids=true,
        grid style=dashed,
    ]
    
    \addplot[
        color=blue2,
        mark=*,
        ]
        coordinates {
        (0,43.8)(10,43.6)(20,43.6)(30,42.4)(40,40.8)(50,39.0)(60,37.4)
        };
    \addplot[
        color=gray1,
        mark=*,
        ]
        coordinates {
        (0,38.4)(10,32.6)(20,24.5)(30,21.5)(40,18.7)(50,16.7)(60,15.1)
        };
    \addlegendentry{MILE}
    \addlegendentry{One-frame model}
        
    \end{axis}
    \end{tikzpicture}
    \caption{Perception accuracy.}
    \label{mile-fig:dreaming-driving-perception}
    \end{subfigure}
\caption{Driving in imagination. We report the closed-loop driving performance and perception accuracy in CARLA when the model imagines future states and actions and does not observe a proportion of the images.}
\label{mile-fig:dreaming-driving}
\end{figure}

\clearpage
\subsection{Image Resolution}
\begin{figure}[t]
    \centering
    \begin{subfigure}[b]{\textwidth}
    \centering
    \includegraphics[width=\linewidth]{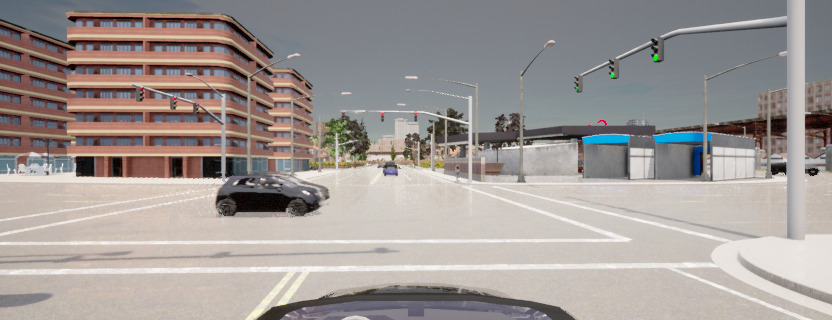}%
    \caption{Resolution $600\times960$.}
    \end{subfigure}
    \par\smallskip
    \begin{subfigure}[t]{0.5\textwidth}
    \includegraphics[width=\linewidth]{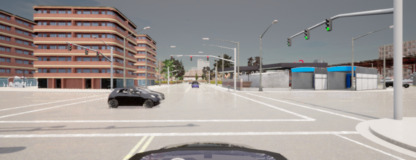}%
    \caption{Resolution $300\times480$.}
    \end{subfigure}
    \begin{subfigure}[t]{0.24\textwidth}
    \includegraphics[width=\linewidth]{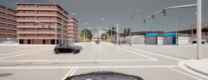}%
    \caption{Resolution $150\times240$.}
    \end{subfigure}
    \begin{subfigure}[t]{0.24\textwidth}
    \includegraphics[width=\linewidth]{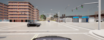}%
    \caption{Resolution $75\times120$.}
    \end{subfigure}
    \caption{Input image observation at different resolutions. The red traffic light becomes almost indistinguishable in lower resolutions.}
    \label{fig:image-resolution-traffic-light}
\end{figure}

\begin{figure}[t]
    \centering
    \begin{subfigure}[b]{\textwidth}
    \centering
    \includegraphics[width=\linewidth]{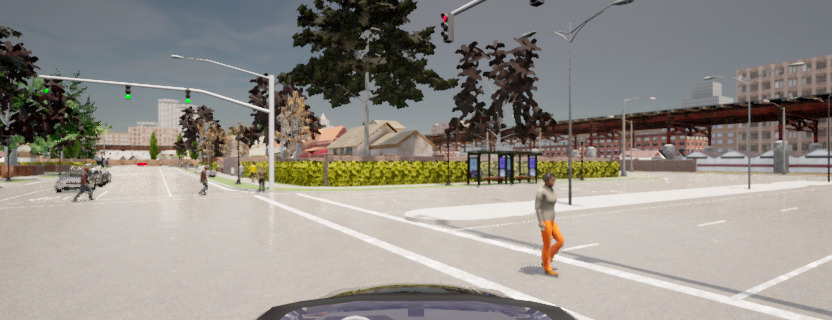}%
    \caption{Resolution $600\times960$.}
    \end{subfigure}
    \par\smallskip
    \begin{subfigure}[t]{0.5\textwidth}
    \includegraphics[width=\linewidth]{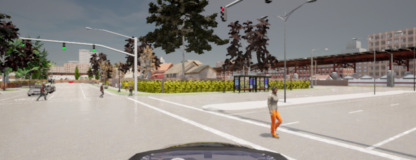}%
    \caption{Resolution $300\times480$.}
    \end{subfigure}
    \begin{subfigure}[t]{0.24\textwidth}
    \includegraphics[width=\linewidth]{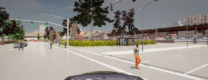}%
    \caption{Resolution $150\times240$.}
    \end{subfigure}
    \begin{subfigure}[t]{0.24\textwidth}
    \includegraphics[width=\linewidth]{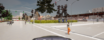}%
    \caption{Resolution $75\times120$.}
    \end{subfigure}
    \caption{Input image observation at different resolutions. It becomes increasingly harder to see the pedestrian as the resolution decreases.}
    \label{fig:image-resolution-pedestrian}
\end{figure}

In urban driving, small elements in the scene can have an important role  in decision making. One typical example is traffic lights, which only occupy a small portion of the image, but dictate whether a vehicle can continue driving forward or needs to stop at a red light. \Cref{fig:image-resolution-traffic-light} and \Cref{fig:image-resolution-pedestrian} illustrate how traffic lights and pedestrians become much harder to distinguish in lower image resolutions.

We evaluate the importance of image resolution by training MILE at different resolutions: $75\times120$, $150\times240$, $300\times480$, and $600\times960$ (our proposed resolution). We report the results in \Cref{mile-table:resolution} and observe a significant decrease in both driving score and cumulative reward. The performance drop is most severe in the infraction penalty metric. To get a better understanding of what is happening, we detail in \Cref{mile-table:resolution-breakdown} the breakdown of the infractions. We report the number of red lights run, the number of vehicle collisions, and the number of pedestrian collisions, all per kilometre driven. As the resolution of the image lowers, the number of infractions increases across all modalities (red lights, vehicles, and pedestrians). These results highlight the importance of high resolution images to reliably detect traffic lights, vehicles, and pedestrians.

\begin{table}[h]
\begin{footnotesize}
\caption[Analysis on the image resolution.]{Analysis on the image resolution. We report driving performance on a new town and new weather conditions in CARLA.}
\label{mile-table:resolution}
\centering
\begin{tabular}{lccccc}
\toprule
\textbf{Image resolution}& \textbf{Driving Score}  & \textbf{Route} & \textbf{Infraction}& \textbf{Reward} & \textbf{Norm. Reward} \\
\midrule
$75\times120$ &20.9 $\pm~$0.0&87.5 $\pm~$0.0&25.3 $\pm~$0.0&5674 $\pm~$0.0&0.65  $\pm~$0.0\\
$150\times240$ &27.9 $\pm~$0.0&81.8 $\pm~$0.0&40.4 $\pm~$0.0&5017 $\pm~$0.0&0.65  $\pm~$0.0\\
$300\times480$ &43.3 $\pm~$0.0&96.1 $\pm~$0.0&44.4 $\pm~$0.0&5814 $\pm~$0.0&0.55  $\pm~$0.0\\
$600\times960$ & \textbf{61.1 $\pm~$3.2} & \textbf{97.4 $\pm~$0.8}	 & \textbf{63.0 $\pm~$3.0}& \textbf{7621 $\pm~$460} &	\textbf{0.67 $\pm~$0.02}	 \\
\cmidrule{1-6}
Expert & 88.4 $\pm~$0.9& 97.6 $\pm~$1.2	& 90.5 $\pm~$1.2& 8694 $\pm~$88& 0.70 $\pm~$0.01\\
\bottomrule
\end{tabular}
\end{footnotesize}
\end{table}

\begin{table}[h]
\caption[Analysis on the image resolution. We report the breakdown of infraction penalties on a new town and new weather conditions in CARLA.]{Analysis on the image resolution. We report the breakdown of infraction penalties on a new town and new weather conditions in CARLA. The metrics are: number of red lights run, number of vehicle collisions, and number of pedestrian collisions. They are normalised per kilometre driven. Lower is better.}
\label{mile-table:resolution-breakdown}
\centering
\begin{tabular}{lccc}
\toprule
\textbf{Image resolution}& \textbf{Red lights ($\downarrow$)}  & \textbf{Vehicles ($\downarrow$)}& \textbf{Pedestrians ($\downarrow$)} \\
\midrule
$75\times120$ & 3.07 $\pm~$0.0& 0.77 $\pm~$0.0&0.07  $\pm~$0.0\\
$150\times240$ & 2.39 $\pm~$0.0& 0.35 $\pm~$0.0&0.03  $\pm~$0.0\\
$300\times480$ &  0.99 $\pm~$0.0& 0.31 $\pm~$0.0&0.05  $\pm~$0.0\\
$600\times960$ & \textbf{0.13 $\pm~$0.04}& \textbf{0.24 $\pm~$0.05} & \textbf{0.01 $\pm~$0.01}\\
\cmidrule{1-4}
Expert & 0.04 $\pm~$0.01& 0.15 $\pm~$0.01& 0.02 $\pm~$0.00\\
\bottomrule
\end{tabular}
\end{table}

\subsection{Training Town Evaluation}
We also evaluate our method on towns and weather conditions seen during training.  As reported in \Cref{mile-table:carla-challenge-train}, our model shows a 21\% relative improvement in driving score with respect to Roach. Note that the RL expert has a lower performance than in test town Town05, because Town03 was designed as the most complex town \citep{carla-maps}.

\begin{table}[h]
\caption[Driving performance in CARLA on a town and weather conditions seen during training.]{Driving performance in CARLA on a town and weather conditions seen during training. Metrics are averaged across three runs.}
\label{mile-table:carla-challenge-train}
\centering
\begin{small}
\begin{tabular}{lccccc}
\toprule
 & \textbf{Driving Score}   & \textbf{Route} & \textbf{Infraction}& \textbf{Reward} & \textbf{Norm. Reward}\\
\midrule
Roach & 50.6 $\pm~$1.9& \textbf{91.0 $\pm~$0.7} & 56.9 $\pm~$1.2& 4419 $\pm~$487 & 0.38 $\pm~$0.04 \\
\textbf{MILE} & \textbf{61.4 $\pm~$0.3}& 89.3 $\pm~$2.5&\textbf{69.4 $\pm~$1.3}&\textbf{7627 $\pm~$190}&\textbf{0.71 $\pm~$0.01}\\
\cmidrule{1-6}
Expert & 81.5 $\pm~$2.8& 95.1 $\pm~$1.2	& 85.6 $\pm~$1.7&  7740 $\pm~$220& 0.69 $\pm~$0.03 \\
\bottomrule
\end{tabular}
\end{small}
\end{table}

\subsection{Evaluation in the Settings of Past Works}

We also evaluated our model in the evaluation settings of: 

\begin{itemize}
    \item TransFus. \citep{prakash2021multi}: the full 10 test routes of Town05 in ClearNoon weather and no scenarios (\Cref{mile-table:transfuser-setting});
    \item LAV \citep{chen2022learning}: 2 test routes from Town02 and 2 test routes in Town05 in weathers [SoftRainSunset, WetSunset, CloudyNoon, MidRainSunset] and no scenarios  (\Cref{mile-table:lav-setting}).
\end{itemize}

\begin{table}[h]
\caption{Driving performance in CARLA in the TransFus. \citep{prakash2021multi} evaluation setting.}
\label{mile-table:transfuser-setting}
\centering
\begin{small}
\begin{tabular}{lccccc}
\toprule
 & \textbf{Driving Score}   & \textbf{Route} & \textbf{Infraction}& \textbf{Reward} & \textbf{Norm. Reward}\\
\midrule
TransFus. &43.7  $\pm~2.4$& 79.6 $\pm~$8.5 &  -&  - &  - \\
\textbf{MILE} & \textbf{69.9 $\pm~$7.0}&\textbf{98.3  $\pm~$2.1}&70.9 $\pm~$6.8&7792 $\pm~$663&0.69 $\pm~0.03$\\
\bottomrule
\end{tabular}
\end{small}
\end{table}

\begin{table}[h]
\caption{Driving performance in CARLA in the LAV \citep{chen2022learning} evaluation setting.}
\label{mile-table:lav-setting}
\centering
\begin{small}
\begin{tabular}{lccccc}
\toprule
 & \textbf{Driving Score}   & \textbf{Route} & \textbf{Infraction}& \textbf{Reward} & \textbf{Norm. Reward}\\
\midrule
LAV & 54.2 $\pm~$8.0& 78.7 $\pm~$5.8 & \textbf{73.0 $\pm~$4.9}&  - &  - \\
\textbf{MILE} & \textbf{64.3 $\pm~$5.2}&  \textbf{99.1 $\pm~$1.5}&64.6 $\pm~$5.4&9631 $\pm~$341&0.72 $\pm~$0.01\\
\bottomrule
\end{tabular}
\end{small}
\end{table}

\clearpage
\section{Lower Bound Derivation}
\label[appendix]{mile-appendix:lower-bound}
\begin{proof}
Let $q_{H,S} \triangleq q(\rvh_{1:T},\rvs_{1:T}|\rvo_{1:T},\rvy_{1:T},\rva_{1:T}) = q(\rvh_{1:T},\rvs_{1:T}|\rvo_{1:T},\rva_{1:T-1})$ be the variational distribution (where we have assumed independence of $(\rvy_{1:T}, \rva_T)$ given $(\rvo_{1:T}, \rva_{1:T-1}$)), and $p(\rvh_{1:T},\rvs_{1:T}|\rvo_{1:T},\rvy_{1:T},\rva_{1:T})$ be the posterior distribution. The Kullback-Leibler divergence between these two distributions writes as:

\begin{align}
   &\KL\left(q(\rvh_{1:T},\rvs_{1:T}|\rvo_{1:T},\rvy_{1:T},\rva_{1:T}) ~||~ p(\rvh_{1:T},\rvs_{1:T}|\rvo_{1:T},\rvy_{1:T},\rva_{1:T}) \right) \notag\\
    =& \E_{\rvh_{1:T}, \rvs_{1:T} \sim q_{H,S}} \left[ \log \frac{q(\rvh_{1:T},\rvs_{1:T}|\rvo_{1:T},\rvy_{1:T},\rva_{1:T})}{p(\rvh_{1:T},\rvs_{1:T}|\rvo_{1:T},\rvy_{1:T},\rva_{1:T})}\right] \notag\\
    =& \E_{\rvh_{1:T}, \rvs_{1:T} \sim q_{H,S}} \left[ \log \frac{q(\rvh_{1:T},\rvs_{1:T}|\rvo_{1:T},\rvy_{1:T},\rva_{1:T})p(\rvo_{1:T},\rvy_{1:T},\rva_{1:T})}{p(\rvh_{1:T},\rvs_{1:T}) p(\rvo_{1:T},\rvy_{1:T},\rva_{1:T}|\rvh_{1:T},\rvs_{1:T})} \right] \notag\\
    =& \log p(\rvo_{1:T},\rvy_{1:T},\rva_{1:T}) - \E_{\rvh_{1:T}, \rvs_{1:T} \sim q_{H,S}} \left[ \log p(\rvo_{1:T},\rvy_{1:T},\rva_{1:T}|\rvh_{1:T},\rvs_{1:T}) \right] \notag \\
    &+ \KL (q(\rvh_{1:T},\rvs_{1:T}|\rvo_{1:T},\rvy_{1:T},\rva_{1:T})~||~p(\rvh_{1:T},\rvs_{1:T})) \notag
\end{align}

Since $\KL\left(q(\rvh_{1:T},\rvs_{1:T}|\rvo_{1:T},\rvy_{1:T},\rva_{1:T}) ~||~ p(\rvh_{1:T},\rvs_{1:T}|\rvo_{1:T},\rvy_{1:T},\rva_{1:T}) \right) \geq 0$, we obtain the following evidence lower bound:
\begin{align}
    \log p(\rvo_{1:T},\rvy_{1:T},\rva_{1:T}) \geq &\E_{\rvh_{1:T}, \rvs_{1:T} \sim q_{H,S}} \left[ \log p(\rvo_{1:T},\rvy_{1:T},\rva_{1:T}|\rvh_{1:T},\rvs_{1:T}) \right] \notag \\
    &- \KL (q(\rvh_{1:T},\rvs_{1:T}|\rvo_{1:T},\rva_{1:T-1})~||~p(\rvh_{1:T},\rvs_{1:T})) \label{eq:lower-bound}
\end{align}
\\
Let us now calculate the two terms of this lower bound separately. On the one hand:
\begin{align}
   & \E_{\rvh_{1:T}, \rvs_{1:T} \sim q_{H,S}} \left[ \log p(\rvo_{1:T},\rvy_{1:T},\rva_{1:T}|\rvh_{1:T},\rvs_{1:T}) \right] \notag\\
    =&~ \E_{\rvh_{1:T}, \rvs_{1:T} \sim q_{H,S}} \left[ \log \prod_{t=1}^T p(\rvo_t|\rvh_t,\rvs_t) p(\rvy_t|\rvh_t,\rvs_t) p(\rva_t|\rvh_t,\rvs_t) \right] \label{eq:expectation1}\\
    =&~\sum_{t=1}^T \E_{\rvh_{1:t}, \rvs_{1:t} \sim q(\rvh_{1:t}, \rvs_{1:t}| \rvo_{\le t}, \rva_{<t})} \left[ \log p(\rvo_t|\rvh_t,\rvs_t) + \log p(\rvy_t|\rvh_t,\rvs_t) + \log p(\rva_t|\rvh_t,\rvs_t) \right] \label{eq:last-expectation}
\end{align}
where \Cref{eq:expectation1} follows from \Cref{eq:generative-model}, and \Cref{eq:last-expectation} was obtained by integrating over remaining latent variables $(\rvh_{t+1:T}, \rvs_{t+1:T})$.
\\~\\
On the other hand:
\begin{align}
    &\KL (q(\rvh_{1:T},\rvs_{1:T}|\rvo_{1:T},\rva_{1:T-1})~||~p(\rvh_{1:T},\rvs_{1:T})) \notag\\
    =&~\E_{\rvh_{1:T}, \rvs_{1:T} \sim q_{H,S}} \left[\log \frac{q(\rvh_{1:T}, \rvs_{1:T}| \rvo_{1:T}, \rva_{1:T-1})}{p(\rvh_{1:T}, \rvs_{1:T})} \right] \notag\\
    =&~\int_{\rvh_{1:T}, \rvs_{1:T}} q(\rvh_{1:T}, \rvs_{1:T}| \rvo_{1:T}, \rva_{1:T-1}) \log \frac{q(\rvh_{1:T}, \rvs_{1:T}| \rvo_{1:T}, \rva_{1:T-1})}{p(\rvh_{1:T}, \rvs_{1:T})} \diff \rvh_{1:T} \diff \rvs_{1:T} \notag\\
    =&~\int_{\rvh_{1:T}, \rvs_{1:T}} q(\rvh_{1:T}, \rvs_{1:T}| \rvo_{1:T}, \rva_{1:T-1}) \log\left[ \prod_{t=1}^T \frac{q(\rvh_t| \rvh_{t-1}, \rvs_{t-1})q(\rvs_t| \rvo_{\le t}, \rva_{<t})}{p(\rvh_{t} | \rvh_{t-1}, \rvs_{t-1}) p(\rvs_t | \rvh_{t-1}, \rvs_{t-1})} \right] \diff \rvh_{1:T} \diff \rvs_{1:T} \label{eq:kl-step-1}\\
    =&~\int_{\rvh_{1:T}, \rvs_{1:T}} q(\rvh_{1:T}, \rvs_{1:T}| \rvo_{1:T}, \rva_{1:T-1}) \log\left[ \prod_{t=1}^T \frac{q(\rvs_t| \rvo_{\le t}, \rva_{<t})}{p(\rvs_t | \rvh_{t-1}, \rvs_{t-1})} \right] \diff \rvh_{1:T} \diff \rvs_{1:T} \label{eq:kl-step-2}\\
\end{align}
where:
\begin{itemize}
\item \Cref{eq:kl-step-1} follows from the factorisations defined in \Cref{eq:generative-model} and \Cref{eq:inference-model}.
\item The simplification in \Cref{eq:kl-step-2} results of $q(\rvh_t| \rvh_{t-1}, \rvs_{t-1}) = p(\rvh_{t} | \rvh_{t-1}, \rvs_{t-1})$ .
\end{itemize}

Thus:
\begin{align}
    &\KL (q(\rvh_{1:T},\rvs_{1:T}|\rvo_{1:T},\rva_{1:T-1})~||~p(\rvh_{1:T},\rvs_{1:T})) \notag\\
    =&~\int_{\rvh_{1:T}, \rvs_{1:T}} \left(\prod_{t=1}^T q(\rvh_{t}| \rvh_{t-1}, \rvs_{t-1})q(\rvs_{t}| \rvo_{\le t}, \rva_{<t}) \right)\left(  \sum_{t=1}^T \log \frac{q(\rvs_t| \rvo_{\le t}, \rva_{<t})}{p(\rvs_t | \rvh_{t-1}, \rvs_{t-1})} \right) \diff \rvh_{1:T} \diff \rvs_{1:T}\notag \\
    =&~\int_{\rvh_{1:T}, \rvs_{1:T}} \left(\prod_{t=1}^T q(\rvh_{t}| \rvh_{t-1}, \rvs_{t-1})q(\rvs_{t}| \rvo_{\le t}, \rva_{<t}) \right)\Bigg( \log \frac{q(\rvs_1| \rvo_1)}{p(\rvs_1)}\notag \\ 
    & \qquad\qquad\qquad\qquad\qquad\qquad\qquad\qquad\qquad\qquad\quad+ \sum_{t=2}^T \log \frac{q(\rvs_t| \rvo_{\le t}, \rva_{<t})}{p(\rvs_t | \rvh_{t-1}, \rvs_{t-1})} \Bigg) \diff \rvh_{1:T} \diff \rvs_{1:T} \notag \\
    =&~\E_{\rvs_1 \sim q(\rvs_1|\rvo_1)} \left[ \log \frac{q(\rvs_1|\rvo_1)}{p(\rvs_1)}\right] \notag \\
    &+~\int_{\rvh_{1:T}, \rvs_{1:T}} \left( \prod_{t=1}^T q(\rvh_{t}| \rvh_{t-1}, \rvs_{t-1})q(\rvs_{t}| \rvo_{\le t}, \rva_{<t}) \right) \left(\sum_{t=2}^T \log \frac{q(\rvs_t| \rvo_{\le t}, \rva_{<t})}{p(\rvs_t | \rvh_{t-1}, \rvs_{t-1})} \right) \diff \rvh_{1:T} \diff \rvs_{1:T} \label{eq:kl-first-term} \\
    =&~\KL ( q(\rvs_1|\rvo_1)~||~p(\rvs_1)) \notag \\
    &+~\int_{\rvh_{1:T}, \rvs_{1:T}} \left( \prod_{t=1}^T q(\rvh_{t}| \rvh_{t-1}, \rvs_{t-1})q(\rvs_{t}| \rvo_{\le t}, \rva_{<t}) \right) \Bigg(\log \frac{q(\rvs_2| \rvo_{1:2}, \rva_1)}{p(\rvs_2 | \rvh_1, \rvs_1)}\notag\\
    &\qquad\qquad\qquad\qquad\qquad\qquad\qquad\qquad\qquad\qquad\enspace\  +\sum_{t=3}^T \log \frac{q(\rvs_t| \rvo_{\le t}, \rva_{<t})}{p(\rvs_t | \rvh_{t-1}, \rvs_{t-1})} \Bigg) \diff \rvh_{1:T} \diff \rvs_{1:T} \notag\\
    =&~\KL ( q(\rvs_1|\rvo_1)~||~p(\rvs_1)) + ~\E_{\rvh_1,\rvs_1 \sim q(\rvh_1,\rvs_1|\rvo_1)} \left[ \KL ( q(\rvs_2|\rvo_{1:2}, \rva_1)~||~p(\rvs_2|\rvh_1, \rvs_1)) \right]\notag \\
    &+~\int_{\rvh_{1:T}, \rvs_{1:T}} \left( \prod_{t=1}^T q(\rvh_{t}| \rvh_{t-1}, \rvs_{t-1})q(\rvs_{t}| \rvo_{\le t}, \rva_{<t}) \right) \left(\sum_{t=3}^T \log \frac{q(\rvs_t| \rvo_{\le t}, \rva_{<t})}{p(\rvs_t | \rvh_{t-1}, \rvs_{t-1})} \right) \diff \rvh_{1:T} \diff \rvs_{1:T} \label{eq:kl-second-term}
\end{align}
where \Cref{eq:kl-first-term} and \Cref{eq:kl-second-term} were obtained by splitting the integral in two and integrating over remaining latent variables.
By recursively applying this process on the sum of logarithms indexed by $t$, we get:
\begin{align}
    &\KL (q(\rvh_{1:T},\rvs_{1:T}|\rvo_{1:T},\rva_{1:T-1})~||~p(\rvh_{1:T},\rvs_{1:T})) \notag\\
    =&~ \sum_{t=1}^T \E_{\rvh_{1:t-1},\rvs_{1:t-1} \sim q(\rvh_{1:t-1},\rvs_{1:t-1}|\rvo_{\le t-1},\rva_{<t-1})} \left[ \KL ( q(\rvs_t|\rvo_{\le t}, \rva_{<t})~||~p(\rvs_t | \rvh_{t-1}, \rvs_{t-1})) \right] \label{eq:last-kl}
\end{align}

Finally, we inject \Cref{eq:last-expectation} and \Cref{eq:last-kl} in \Cref{eq:lower-bound} to obtain the desired lower bound:

\begin{align*}
    &\log p(\rvo_{1:T},\rvy_{1:T},\rva_{1:T})\\
    \geq&~\sum_{t=1}^T \E_{\rvh_{1:t}, \rvs_{1:t} \sim q(\rvh_{1:t}, \rvs_{1:t}| \rvo_{\leq t}, \rva_{<t})} \left[ \log p(\rvo_t|\rvh_t,\rvs_t) + \log p(\rvy_t|\rvh_t,\rvs_t) + \log p(\rva_t|\rvh_t,\rvs_t) \right]\\
    & - \sum_{t=1}^T \E_{\rvh_{1:t-1},\rvs_{1:t-1} \sim q(\rvh_{1:t-1},\rvs_{1:t-1}|\rvo_{\le t-1},\rva_{<t-1})} \left[ \KL ( q(\rvs_t|\rvo_{\le t}, \rva_{<t})~||~p(\rvs_t | \rvh_{t-1}, \rvs_{t-1})) \right] \qedhere
\end{align*}

\end{proof}

\clearpage
\section{Model Description}
\begin{figure}[h] 
\begin{subfigure}[b]{0.5\textwidth} 
\centering 
\resizebox{0.9\linewidth}{!}{ 
\begin{tikzpicture} 

    \node[draw,inner sep=0pt,minimum size =10mm] (h1) {\Large$\rvh_1$};
    \node[draw=black,right=2cm of h1,inner sep=0pt,minimum size =10mm] (h2) {\Large$\rvh_2$};
    \node[draw=black,right=2cm of h2,inner sep=0pt,minimum size =10mm] (h3) {\Large$\rvh_3$};
    \node[draw=black,circle,below=1.5cm of h1,inner sep=0pt,minimum size =10mm] (s1) {\Large$\rvs_1$};
    \node[draw=black,circle,below=1cm of s1,inner sep=0pt,minimum size =10mm,fill=lightgray] (o1y1) {$\rvo_1,\rvy_1$};
    \node[draw=black,circle,below right=0.5cm and 0.6cm of h1,inner sep=0pt,minimum size =10mm,fill=lightgray] (a1) {\Large$\rva_1$};
    \node[draw=black,circle,below=1.5cm of h2,inner sep=0pt,minimum size =10mm] (s2) {\Large$\rvs_2$};
    \node[draw=black,circle,below=1cm of s2,inner sep=0pt,minimum size =10mm,fill=lightgray] (o2y2) {$\rvo_2,\rvy_2$};  
    \node[draw=black,circle,below right=0.5cm and 0.6cm of h2,inner sep=0pt,minimum size =10mm,fill=lightgray] (a2) {\Large$\rva_2$};  
    \coordinate[below =1.5cm of h3] (s3);
    \node[draw=white,circle,right=1cm of h3] (h4){\Huge $...$};
    \path[->,thick]
        (h1) edge node[pos=0,below] {} (s1)
        (h2) edge node[pos=0,below] {} (s2)
        (s1) edge (a1)
        (s2) edge (a2)
        (h1) edge (h2)
        (h2) edge (h3)
        (s1) edge (o1y1)
        (s2) edge (o2y2)
        (s1) edge[bend left=35] node[pos=0.0,above] {} (h2)
        (s2) edge[bend left=35] node[pos=0.0,above] {} (h3)
        (h1) edge[bend right=30] node[pos=0.5,above] {} (o1y1)
        (h1) edge (a1)
        (a1) edge (s2)
        (h2) edge[bend right=30] node[pos=0.5,above] {} (o2y2)
        (h2) edge (a2)
        (h3) edge[path fading = south] (s3)
        (a2) edge[path fading = south] (s3)
        (h3) edge[path fading = east] (h4);
\end{tikzpicture}
}
\caption{Generative model.} 
\label{mile-fig:generative-model} 
\end{subfigure} 
\begin{subfigure}[b]{0.5\textwidth} 
\centering \resizebox{0.9\linewidth}{!}{ 
\begin{tikzpicture} 
   \node[draw,inner sep=0pt,minimum size =10mm] (h1) {\Large$\rvh_1$};
    \node[draw=black,right=2cm of h1,inner sep=0pt,minimum size =10mm] (h2) {\Large$\rvh_2$};
    \node[draw=black,right=2cm of h2,inner sep=0pt,minimum size =10mm] (h3) {\Large$\rvh_3$};
    \node[draw=black,circle,below=1.5cm of h1,inner sep=0pt,minimum size =10mm] (s1) {\Large$\rvs_1$};
    \node[draw=black,circle,below=1cm of s1,inner sep=0pt,minimum size =10mm,fill=lightgray] (o1y1) {$\rvo_1,\rvy_1$};
    \node[draw=black,circle,below right=0.5cm and 0.6cm of h1,inner sep=0pt,minimum size =10mm,fill=lightgray] (a1) {\Large$\rva_1$};
    \node[draw=black,circle,below=1.5cm of h2,inner sep=0pt,minimum size =10mm] (s2) {\Large$\rvs_2$};
    \node[draw=black,circle,below=1cm of s2,inner sep=0pt,minimum size =10mm,fill=lightgray] (o2y2) {$\rvo_2,\rvy_2$};  
    \node[draw=black,circle,below right=0.5cm and 0.6cm of h2,inner sep=0pt,minimum size =10mm,fill=lightgray] (a2) {\Large$\rva_2$};  
    \coordinate[below =1.5cm of h3] (s3);
    \node[draw=white,circle,right=1cm of h3] (h4){\Huge $...$};
    \path[->,thick]
        (o1y1) edge (s1)
        (o2y2) edge (s2)
        (o1y1) edge (s2)
        (a1) edge (s2)
        (h1) edge[bend right=23,color=white] node[pos=0.5,above] {} (o1y1)
        (h1) edge[bend right=35,color=white] node[pos=1.0,above] {} (a1)
        (a2) edge[path fading=east] (s3)
        (o2y2) edge[path fading=east] (s3)
        (o1y1) edge[path fading=east] (s3)
        (a1) edge[bend right = 10, path fading=east] (s3);
\end{tikzpicture} } 
\caption{Inference model.} 
\label{mile-fig:inference-model} 
\end{subfigure} 
\caption[Graphical models representing the conditional dependence between states.]{Graphical models representing the conditional dependence between states. Deterministic and stochastic states are represented by, respectively, squares and circles. Observed states are in gray.} \label{mile-fig:probabilistic-model} 
\end{figure}
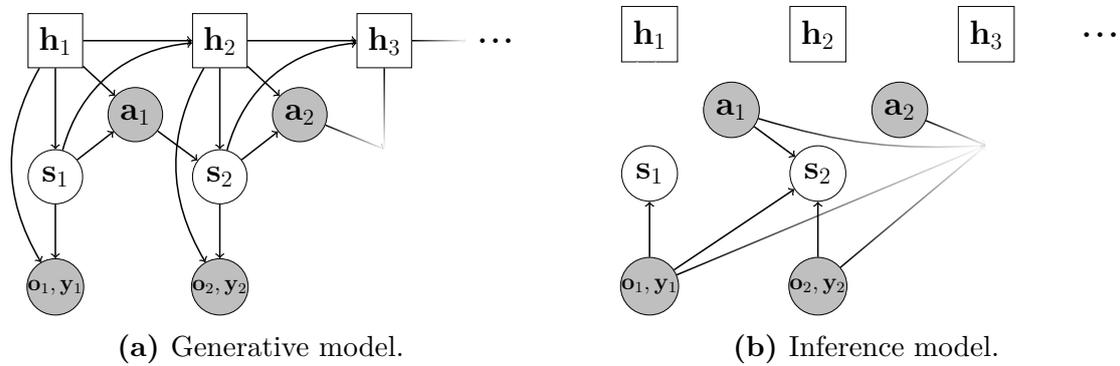

\label[appendix]{mile-appendix:model-description}
We give a full description of MILE. The graphical models of the generative and inference models are depicted in \Cref{mile-fig:probabilistic-model}. \Cref{mile-appendix:parameters} shows the number of parameters of each component of the model, and \Cref{mile-appendix:hyperparameters} contains all the hyperparameters used during training. \Cref{mile-appendix:inference-model} describes the inference network, and \Cref{mile-appendix:generative-model} the generative network. 

\subsection{Graphical Models}
\label[appendix]{mile-appendix:graphical-models}

\subsection{Network Description}
\begin{table}[h]
\caption{Parameters of the model.}
\centering
\begin{tabular}{ll r}
\toprule
& \textbf{Name}& \textbf{Parameters}\\
\midrule
\multirow{2}{*}{\textbf{{Inference model $\phi$}}} &Observation encoder $e_{\phi}$ &34.9M \\
&Posterior network $(\mu_{\phi}, \sigma_{\phi})$& 3.9M\\
\cmidrule{1-3}
\multirow{4}{*}{\textbf{{Generative model $\theta$}}} &Prior network $(\mu_{\theta}, \sigma_{\theta})$& 2.1M\\
&Recurrent cell $f_{\theta}$& 6.9M\\
&BeV decoder $l_{\theta}$& 34.2M\\
&Policy $\pi_{\theta}$& 5.9M\\
\bottomrule
\end{tabular}
\label{mile-appendix:parameters}
\end{table}

\begin{table}[h]
\caption{Hyperparameters.}
\centering
\begin{small}
\begin{tabular}{ll l}
\toprule
\textbf{Category} & \textbf{Name}& \textbf{Value}\\
\midrule
\multirow{4}{*}{\textbf{Training}} &GPUs& \num{8} Tesla V100\\
&batch size& \num{64}\\
&precision& Mixed precision (16-bit)\\
& iterations & \num{5e4} \\
\cmidrule{1-3}
\multirow{8}{*}{\textbf{Optimiser}} &name& AdamW\\
&learning rate& \num{1e-4}\\
&weight decay& \num{1e-2}\\
& $\beta_1$& \num{0.9}\\
& $\beta_2$& \num{0.999}\\
& $\varepsilon$& \num{1e-8}\\
& scheduler& OneCycleLR\\
& pct start & \num{0.2}\\
\cmidrule{1-3}
\multirow{5}{*}{\textbf{Input image}} &size& $600\times 960$\\
 &crop& $[64, 138, 896, 458]$ (left, top, right, bottom)\\
 &field of view & \ang{100}\\
 & camera position & $[-1.5, 0.0, 2.0]$ (forward, right, up) \\
& camera rotation & $[0.0, 0.0, 0.0]$ (pitch, yaw, roll)\\
\cmidrule{1-3}
\multirow{2}{*}{\textbf{BeV label}} &size $H_b\times W_b$ & $192\times 192$\\
& resolution $b_{\mathrm{res}}$& $0.2\text{m/pixel}$\\
\cmidrule{1-3}
\multirow{3}{*}{\textbf{Sequence}} &length $T$& $12$\\
& frequency & \SI{5}{\Hz}\\
& observation dropout $p_{\text{drop}}$ & 0.25\\
\cmidrule{1-3}
\multirow{10}{*}{\textbf{Loss}} &action weight& $1.0$\\
& image weight & $0.0$\\
& segmentation weight & $0.1$\\
& segmentation top-k & $0.25$\\
& instance weight & $0.1$\\
& instance center weight & $200.0$\\
& instance offset weight & $0.1$\\
& image weight & $0.0$\\
& kl weight & \num{1e-3}\\
& kl balancing & \num{0.75}\\
\bottomrule
\end{tabular}
\end{small}
\label{mile-appendix:hyperparameters}
\end{table}

\subsection{Details on the Network and on Training.} 

\paragraph{Lifting to 3D.}
The $\mathrm{Lift}$ operation can be detailed as follows: (i) Using the inverse intrinsics $K^{-1}$ and predicted depth, the features in the pixel image space are lifted to 3D in camera coordinates with a pinhole camera model, (ii) the rigid body motion $M$ transforms the 3D camera coordinates to 3D vehicle coordinates (center of inertia of the ego-vehicle).

\paragraph{Observation dropout.} At training time the priors are trained to match posteriors through the KL divergence, however they are not necessarily optimised for robust long term future prediction. \citet{hafner2019planet} optimised states for robust multi-step predictions by iteratively applying the transition model and integrating out intermediate states. In our case, we  supervise priors unrolled with random temporal horizons  (i.e. predict states at $t+k$ with $k \geq 1$). More precisely, during training, with probability $p_{\text{drop}}$ we sample the stochastic state $\rvs_t$ from the prior instead of the posterior. We call this observation dropout. If we denote $X$ the random variable representing the $k$ number of times a prior is unrolled, $X$ follows a geometric distribution with probability of success $(1 - p_{\text{drop}})$. 
Observation dropout resembles $z$-dropout from \citet{HenaffCL19}, where the posterior distribution is modelled as a mixture of two Gaussians, one of which comes from the prior. During training, some posterior variables are randomly dropped out, forcing other posterior variables to maximise their information extraction from input images. Observation dropout can be seen as a global variant of $z$-dropout since it drops out all posterior variables together.

\paragraph{Additional details} The action space is in $\mathbb{R}^2$ with the first component being the acceleration in $[-1,1]$. Negative values correspond to braking, and positive values to throttle. The second component is steering in $[-1, 1]$, with negative values corresponding to turning left, and positive values to turning right. For simplicity, we have set the weight parameter of the image reconstruction to $0$. In order to improve reconstruction of the bird's-eye view vehicles and pedestrians, we also include an instance segmentation loss \citep{cheng20}. We use the KL balancing technique from \citep{hafner2021dreamerv2}. Finally, the encoder and decoder are initialised with the weights of a pre-trained \emph{Single frame} model.

\begin{table}
\caption{Inference model $\phi$.}
\centering
\begin{tiny}
\begin{tabular}{ll ll}
\toprule
\textbf{Category} & \textbf{Layer}& \textbf{Output size} & \text{Parameters}\\
\midrule
\multirow{6}{*}{\textbf{Image encoder} $e_{\phi}$} &Input& $3\times320\times832 =\rvo_t$ & 0\\
&ResNet18 \citep{he16}& $[128\times40\times104$,
$256\times20\times52$,
$512\times10\times26]$ &11.2M \\
& Feature aggregation & $64\times40\times104 =\rvu_t$ & 0.5M\\
& Depth & $37\times40\times104 =\rvd_t$ & 0.5M\\
& Lifting to 3D & $64\times37\times40\times104$ & 0 \\
& Pooling to BeV & $64\times48\times48 =\rvb_t$ & 0\\
\cmidrule{1-4}
\multirow{2}{*}{\textbf{Route map encoder} $e_{\phi}$} &Input& $3\times64\times64=\mathbf{route}_t$ & 0\\
&ResNet18 \citep{he16}& $16=\rvr_t$ & 11.2M\\
\cmidrule{1-4}
\multirow{2}{*}{\textbf{Speed encoder} $e_{\phi}$} &Input& $1=\mathbf{speed}_t$ & 0\\
&Dense layers& $16 =\rvm_t$ & 304\\
\cmidrule{1-4}
\multirow{2}{*}{\textbf{Compressing to 1D} $e_{\phi}$} & Input & $[64\times48\times48, 16, 16]=[\rvb_t,\rvr_t,\rvm_t]$ & 0 \\ 
&ResNet18 \citep{he16}& $512=\rvx_t$ & 11.5M \\ 
\cmidrule{1-4}
\multirow{2}{*}{\textbf{Posterior network} $(\mu_{\phi}, \sigma_{\phi})$} &Input& $[1024, 512] = [\rvh_t, \rvx_t]$ & 0\\
&Dense layers& $[512, 512]$ & 3.9M\\
\bottomrule
\end{tabular}
\end{tiny}
\label{mile-appendix:inference-model}
\end{table}

\begin{table}
\caption{Generative model $\theta$.}
\centering
\begin{tiny}
\begin{tabular}{ll ll}
\toprule
\textbf{Category} & \textbf{Layer}& \textbf{Output size} & \text{Parameters}\\
\midrule
\multirow{2}{*}{\textbf{Prior network} $(\mu_{\theta}, \sigma_{\theta})$} &Input& $1024 =\rvh_t$ & 0\\
&Dense layers& $[512, 512]$ & 2.1M\\
\cmidrule{1-4}
\multirow{2}{*}{\textbf{Recurrent cell} $f_{\theta}$} &Input& $[1024, 512, 2] = [\rvh_t, \rvs_t, \rva_t]$ & 0\\
& Action layer & $64$ & 192\\
& Pre GRU layer & $1024$ & 0.6M\\
&GRU cell& $1024 = \rvh_{t+1}$ & 6.3M\\
\cmidrule{1-4}
\multirow{10}{*}{\textbf{BeV decoder} $l_{\theta}$} &Input& $[512\times3\times3, 1024, 512] = [\text{constant}, \rvh_t, \rvs_t]$ & 0\\
&Adaptive instance norm& $512\times3\times3$ & 1.6M\\
&Conv. instance norm& $512\times3\times3$ & 3.9M\\
& Upsample conv. instance norm &$512\times6\times6$ & 7.9M\\
& Upsample conv. instance norm &$512\times12\times12$ & 7.9M\\
& Upsample conv. instance norm &$512\times24\times24$ & 7.9M\\
& Upsample conv. instance norm &$256\times48\times48$ & 3.3M\\
& Upsample conv. instance norm &$128\times96\times96$ & 1.2M\\
& Upsample conv. instance norm &$64\times192\times192$ & 0.5M\\
& Output layer &$[8\times192\times192,1\times192\times192,2\times192\times192]$ & 715\\
\cmidrule{1-4}
\multirow{2}{*}{\textbf{Policy} $\pi_{\theta}$} &Input& $[1024, 512] = [\rvh_t, \rvs_t]$ & 0\\
&Dense layers& $2$ & 5.9M\\
\bottomrule
\end{tabular}
\end{tiny}
\label{mile-appendix:generative-model}
\end{table}

\clearpage
\section{Experimental Setting}
\label[appendix]{mile-appendix:experimental-setting}
\subsection{Dataset} Each run was randomised with a different start and end position, as well as with traffic agents \citep{zhang2021end}. A random number of vehicles and pedestrians were spawned in the environment as specified in \Cref{mile-table:data-collection}.

\begin{table}[h]
\caption{Uniform sampling intervals of spawned vehicles and pedestrians in each town during training.}
\centering
\begin{tabular}{lcc}
\toprule
\textbf{Town}& \textbf{Number of vehicles}& \textbf{Number of pedestrians}\\
\midrule
Town01&$[80,160]$& $[80,160]$\\
Town03&$[40,100]$& $[40,100]$\\
Town04&$[100,200]$& $[40,120]$\\
Town06&$[80,160]$& $[40,120]$\\
\bottomrule
\end{tabular}
\label{mile-table:data-collection}
\end{table}

\subsection{Metrics} 
We report metrics from the CARLA challenge \citep{carla-challenge} to measure on-road performance: route completion, infraction penalty, and driving score.

\begin{itemize}[itemsep=0.3mm, parsep=0pt]
    \item \textbf{Route completion} \textbf{$R_{\mathrm{completion}} \in [0, 1]$}: for a given simulation scenario, the percentage of route completed by the driving agent. The simulation can end early if the agent deviates from the desired route by more than $30\mathrm{m}$, or does not take any action for $180\mathrm{s}$.
    \item \textbf{Infraction penalty} \textbf{$I_{\mathrm{penalty}}$}: multiplicative penalty due to various infractions from the agent (collision with pedestrians/vehicles/static objects, running red lights etc.). $I_{\mathrm{penalty}} \in [0, 1]$, with $I_{\mathrm{penalty}}=1$ meaning no infraction was observed.
    \item \textbf{Driving score} \textbf{$D$}: measures both how far the agent drives on the given route, but also how well it drives. $D$ is defined as $D = R_{\mathrm{completion}}\times I_{\mathrm{penalty}} \in [0,1]$, with $D=1$ corresponding perfect driving. For a full description of these metrics, please refer to \citep{carla-challenge}.
\end{itemize}

We now define how the normalised cumulative reward is defined. At every timestep, the environment computes a reward $r\in[R_{\mathrm{min}}, 1]$ \citep{toromanoff20} for the driving agent.
 If $N$ is the number of timesteps the agent was deployed for without hitting a termination criteria, then the \textbf{cumulative reward} $R \in[N\times R_{\mathrm{min}}, N]$. In order to account for the length of the simulation (due to various stochastic events, it can be longer or shorter), we also report the \textbf{normalised cumulative reward} $\overline{R} =R/N$.

We also wanted to highlight the limitations of the driving score as it is obtained by multiplying the route completion with the infraction penalty. The route completion (in $[0,1]$) can be understood as the recall: how far the agent has travelled along the specified route. The infraction penalty (also in $[0, 1]$) starts at $1.0$ and decreases with each infraction with multiplicative penalties. It can be understood as the precision: how many infractions has the agent successfully avoided. Therefore, two models are only comparable at a given recall (or route completion), as the more miles are driven, the more likely the agent risks causing infractions. We instead suggest reporting the cumulative reward in future, that overcomes the limitations of the driving score by being measured at the timestep level. The more route is driven, the more rewards are accumulated along the way. This reward is however modulated by the driving abilities of the model (and can be negative when encountering hard penalties).

\subsection{Evaluation Settings} 
\label{mile-subsection:evaluation}

We measure the performance of our model on two settings. Each evaluation is repeated three times.

\begin{itemize}
    \item \textbf{New town, new weathers}: the 10 test scenarios in Town05 \citep{carla-challenge}, on 4 unseen weather conditions: SoftRainSunset, WetSunset, CloudyNoon, MidRainSunset.
    \item \textbf{Train town, train weathers}: the 20 train scenarios in Town03 \citep{carla-challenge}, on 4 train weather conditions: ClearNoon, WetNoon, HardRainNoon, ClearSunset.
\end{itemize}

\end{appendices}


\end{document}